\documentclass[11pt, a4paper, twoside]{./AllPackages/Thesis} 
\usepackage{./AllPackages/slashbox} 

\usepackage{algorithm,algpseudocode}
\usepackage{wrapfig}
\usepackage{listings}
\usepackage{float}
\usepackage{lscape}
\usepackage{rotating}
\usepackage{graphicx}
\usepackage{caption}
\usepackage{amsmath}
\usepackage{enumerate}
\usepackage{multirow}
\usepackage{tabularx,ragged2e} 
\usepackage[section]{placeins} 
\usepackage{longtable,pdflscape,booktabs}

\usepackage{setspace}
\usepackage{cite}
\usepackage[english]{babel}
\usepackage[square, numbers, comma, sort&compress]{natbib}

\usepackage[final]{pdfpages}

\newcommand{\cmmnt}[1]{\ignorespaces}

\newcolumntype{M}[1]{>{\centering\arraybackslash}m{#1}}
\newcolumntype{L}[1]{>{\raggedright\let\newline\\\arraybackslash\hspace{0pt}}m{#1}}
\newcolumntype{C}[1]{>{\centering\let\newline\\\arraybackslash\hspace{0pt}}m{#1}}
\newcolumntype{R}[1]{>{\raggedleft\let\newline\\\arraybackslash\hspace{0pt}}m{#1}}

\newcolumntype{A}[1]{>{\raggedright}m{#1}}

\newcommand{\expnumber}[2]{{#1}\mathrm{e}{#2}}

\graphicspath{{./Pictures/}} 

\hypersetup{urlcolor=blue, linkcolor=blue, colorlinks=true, citecolor=blue, anchorcolor=green, menucolor=red, runcolor=cyan, filecolor=cyan} 

\title{\ttitle} 

\begin{document}
\makeatletter
\renewcommand*{\NAT@nmfmt}[1]{\textsc{#1}}
\makeatother

\setstretch{1.6} 

\fancyhead{} 
\rhead{\thepage} 
\lhead{} 

\pagestyle{fancy} 

\newcommand{\HRule}{\rule{\linewidth}{0.5mm}} 

\hypersetup{pdftitle={\ttitle}}
\hypersetup{pdfsubject=\subjectname}
\hypersetup{pdfauthor=\authornames}
\hypersetup{pdfkeywords=\keywordnames}


\begin{titlepage}
\begin{center}

\HRule \\[0.1cm] 
{\huge \bfseries \ttitle}\\[0.1cm] 
\HRule \\[0.4cm] 

\large \textit{A Thesis Submitted} \\
	In Partial Fulfilment of the Requirements for the Degree of\\ \textbf{\large Doctor of Philosophy}\\[0.1cm] 

\vfill

\begin{figure}[hb]
  \centering
  \includegraphics[width=0.3\linewidth]{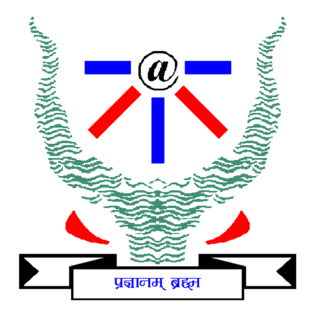}
\end{figure}
\textit{Submitted by}\\
\href{https://www.researchgate.net/profile/Santosh_Barnwal2}{\textbf{\authornames}}\\
\textbf{(RS128)}\\[0.4cm]
Under the supervision of \\ 
\textbf{\facname}
\\[0.4cm]
\textit{to the}\\[0.4cm]
\textbf{\DEPTNAME}\\ 
\textbf{\textsc{\UNIVNAME}}\\ 
\textbf{\textsc{DEOGHAT, JHALWA, PRAYAGRAJ-211015 (U.P.) INDIA}}\\[1.5cm] 
\large \textbf{August, 2021}\\[2cm] 

\end{center}

\end{titlepage}

\clearpage 





\frontmatter 
\setlength{\parindent}{4em}
\setlength{\parskip}{1em}



\abstract{\addtocontents{toc}{\vspace{1em}} 
Reading comprehension, which has been defined as gaining an understanding of written text through a process of translating grapheme into meaning, is an important academic skill. Other language learning skills - writing, speaking and listening, all are connected to reading comprehension. There have been several measures proposed by researchers to automate the assessment of comprehension skills for second language (L2) learners, especially English as Second Language (ESL) and English as Foreign Language (EFL) learners. However, current methods measure particular skills without analysing the impact of reading frequency on comprehension skills. In this dissertation, we show how different skills could be measured and scored automatically. We also demonstrate, using example experiments on multiple forms of learners' responses, how frequent reading practices could impact on the variables of multimodal skills (reading pattern, writing, and oral fluency).

This thesis comprises of five studies. The first and second studies are based on eye-tracking data collected from EFL readers in repeated reading (RR) sessions. The first eye-movement study presents readers' in-depth reading efforts analysis in the second chapter of the thesis. In this study, many state-of-the-art machine learning techniques, viz. Support Vector Machine (SVM), Decision Tree (DT), and Multi-Layer Perceptron (MLP) are used to predict the label of reading efforts and the results are validated by classification accuracy (C. rate), Unweighted Average Recall (UAR) and mean Cross Validation (CV) accuracy. At last in the study, the impact of repeated reading practice on eye-movement features is analysed using a statistical model, Linear Mixed-Effects Regression (LMER). The ANOVA tests of the LMER models are also used to find out if the LMER results on reading efforts are statistically significant. The second eye-movement study presents a psycholinguistics analysis of readers' in-depth reading patterns in the third chapter of the thesis. In this study, the correlation between psycholinguistics factors and the gaze features extracted from eye movements is explained. At last in the second study, the results of LMER, to analyse the effects of these factors at gaze features, are discussed. The ANOVA tests of the LMER models are discussed to find out if the LMER results show the influence of psycholinguistics factors on gaze features in repeated reading approach. 

The third and fourth studies are to evaluate free-text summary written by EFL readers in repeated reading sessions. The third one, described in the fourth chapter of the thesis, examines the summary text using different linguistics and psycholinguistics measures for calculating L2 summary writing proficiency. In this study, many machine learning classification techniques, viz. SVM, DT, and MLP are used to predict the label of summary text on writing proficiency and the results are validated by C. rate, UAR and mean CV accuracy. In addition, a machine learning regression method, Support Vector Regression (SVR) is also used to predict the score of summary text on writing proficiency and the results are validated by the Pearson correlation coefficient ($r$), Root Mean Square Error (RMSE), Quadratic Weighted Kappa (QWK) agreement, and mean Cross Validation RMSE (CV-RMSE). At last in the study, the results of LMER, to analyse the significant impact of repeated reading on these measures, are discussed. The ANOVA tests of the LMER models are also discussed to find out if the LMER results show the influence of repeated reading practice on linguistics and psycholinguistics measures. 

The fourth study, described in the fifth chapter of the thesis, examines the content enrichment in summaries by determining the semantic similarity between the summary text and the referred (read) text. The overall semantic similarity is calculated by measuring lexical, syntactic and concept similarity between summaries and referred text. For measuring lexical similarity, four external knowledge sources viz. a lexical database- WordNet and three pre-trained word embedding models-- Word2Vec, GloVe, and SpaCy are used. The similarity scores are validated by the Pearson correlation coefficient ($r$), Spearman's rank correlation coefficient ($\rho$) and QWK agreement. At last in the study, ANOVA tests of the LMER models are discussed to find out if the LMER results show the impact of knowledge sources on the predicted similarity scores and writing proficiency labels. 

The fifth and last study, described in the sixth chapter of the thesis, is to evaluate recorded oral summaries recited by EFL readers in repeated reading sessions. In this study, the summary speech audio are labelled as high or low based on oral fluency features. Two types of feature sets viz. acoustics-prosody set and temporal sets are extracted from the audio files. Many machine learning classification techniques, viz. SVM, DT, and MLP are used to predict the label of the summary audio on oral fluency and the results are validated by C. rate, UAR and mean CV accuracy. At last in the study, ANOVA tests of LMER models are discussed to find out if the LMER results show the influence of repeated reading practice on oral fluency labels. 

In a nutshell, through this dissertation, we show that multimodal skills of learners could be assessed to measure their comprehension skills as well as to measure the effect of repeated readings on these skills in the course of time, by finding significant features and by applying machine learning techniques with a combination of statistical models such as LMER.
}
\\[0.5cm]
\clearpage 


\setstretch{1.3} 
\acknowledgements{\addtocontents{toc}{\vspace{1em}} 
This dissertation is a result of the inspirations, efforts, and contributions of many people, who I have worked with and to whom I owe my deepest gratitude. First of all, I would like to thank my supervisor cum advisor, \supname. He always encourages his research students to be inquisitive, mature and self-reliant by providing learning conditions in  SILP\footnote{\url{https://silp.iiita.ac.in}} lab as well as by sharing his ideas and suggestions regarding academia, research, and administration. Prof. Albert Einstein quoted a similar approach ``I never teach my pupils. I only attempt to provide the conditions in which they can learn''. His continuous guidance kept me motivated to work hard during the research journey. His constructive critique helped me expanding my knowledge and limits. Without his scientific advice and ideas, this research work would not have reached this state of maturity.

I want to pay my sincere regards to my former Ph.D. supervisor Prof. R. C. Tripathi, who wanted to see me as an entrepreneur. He had encouraged me to think research as a process of an invention of a product and had also taught about the protection of intellectual properties through patents.

Research is never done in a vacuum. I owe this in great measure to my colleagues Shrikant Malviya, and Rohit Mishra who inspired me to learn Python, \LaTeX~ and PyDial dialogue system; helped in operating servers several times; as well as supported me mastering in Ubuntu System. I thank Punit Singh for starting some discussions on various issues, mostly were nonsense; however, these had really helped to relax my mind from academic stress for a moment. I am also grateful to Sudhakar, Varsha, Sumit and other members for making SILP lab a great working place for my research.

I would like to pay my sincere gratitude to Dr Archita Rai, who supported me as an elder sister during the entire period.

I am also indebted to all the ``anonymous" participants of the studies, for their sincere, pretensionless, and genuine participation; even though I was forcing them to sit straight with their chins held up by an uncomfortable ophthalmologic rest.

I pay my sincere gratitude to the members of my review committee for recommending the extension of my Ph.D. duration. I would also like to thank MHRD India for awarding me the Ph.D. fellowship. 

Finally, I offer my deepest thanks to my brothers-- Sheo, Ashutosh and Ashish-- whose love and support has sustained me; and to my mother-- Shakuntala Devi, for her love, blessing, and care. They are and will always be a motivating factor for me to be able to make them proud. At the end, I would like to thank the partner of my life, Nibha. She was always there for me. I really cannot thank her enough for coping up with my crazy work schedule, completely changing her life for being close to me, her unfailing love, and unconditional support throughout this journey. I dedicate my thesis to her.
\\[0.5cm]
\place\\
\fxdate 
\hfill
\authornames.

}
\clearpage 



\lhead{\emph{Contents}} 
\tableofcontents 
\lhead{\emph{List of Figures}} 
\listoffigures 
\lhead{\emph{List of Tables}} 
\listoftables 


\clearpage 

\setstretch{1.4} 

\lhead{\emph{Abbreviations}} 
\listofsymbols{ll} 
{
	\textbf{AoI} & \textbf{A}rea \textbf{o}f \textbf{I}nterest \\
	\textbf{ANOVA} & \textbf{AN}alysis \textbf{O}f \textbf{VA}riance \\
	\textbf{C. rate} & \textbf{C}lassification accuracy \textbf{rate} \\
	\textbf{CAA} & \textbf{C}omputer-\textbf{A}ssisted \textbf{A}ssessment \\
	\textbf{CV} & \textbf{C}ross \textbf{V}alidation \\
	\textbf{CV-RMSE} & \textbf{C}ross \textbf{V}alidation-\textbf{R}oot \textbf{M}ean \textbf{S}quare \textbf{E}rror \\
	\textbf{DT} & \textbf{D}ecision \textbf{T}ree \\
	\textbf{EFL} & \textbf{E}nglish as \textbf{F}oreign \textbf{L}anguage \\
	\textbf{ESL} & \textbf{E}nglish as \textbf{S}econd \textbf{L}anguage \\
	\textbf{ER} & \textbf{E}xtensive \textbf{R}eading \\
	\textbf{IR} & \textbf{I}ntensive \textbf{R}eading \\
	\textbf{L1} & First \textbf{L}anguage \\
	\textbf{L2} & Second \textbf{L}anguage \\
	\textbf{LMER} & \textbf{L}inear \textbf{M}ixed-\textbf{E}ffect \textbf{R}egression \\
	\textbf{MI} & \textbf{M}utual \textbf{I}nformation \\
	\textbf{MLP} & \textbf{M}ulti-\textbf{L}ayer \textbf{P}erceptron \\
	\textbf{NLP} & \textbf{N}atural \textbf{L}anguage \textbf{P}rocessing \\
	\textbf{QWK} & \textbf{Q}uadratic \textbf{W}eighted \textbf{K}appa \\
	\textbf{RMSE} & \textbf{R}oot \textbf{M}ean \textbf{S}quare \textbf{E}rror \\
	\textbf{RR} & \textbf{R}epeated \textbf{R}eading \\
	\textbf{STEM} & \textbf{S}cience \textbf{T}echnology \textbf{E}ngineering \textbf{M}athematics \\
	\textbf{SVM} & \textbf{S}upport \textbf{V}ector \textbf{M}achine \\
	\textbf{SVR} & \textbf{S}upport \textbf{V}ector \textbf{R}egression \\
	\textbf{UAR} & \textbf{U}nweighted \textbf{A}verage \textbf{R}ecall \\
	
}

\mainmatter 

\pagestyle{fancy} 


\chapter{Introduction} 

\label{Chapter1} 

\lhead{Chapter 1. \emph{Introduction}} 

\section{Introduction and Motivation}

Learning is a continuous process and it takes time to process the knowledge. Multiple reading is a strategy that improves reading fluency and comprehension by reading and rereading a text multiple times \cite{roundy2009effect}. The purpose of the first-attempt of reading is to understand the main ideas of a text –what it is about, whose point of view represented, who the characters are, where the text is set –and to become familiar with the language and structure of the text. In second- and third-attempt of reading, learners can overcome any initial confusion, work through the unfamiliarity of the text, and move beyond the literal meaning of the text. It is through multiple reading that learners can understand how they made inferences and developed their opinions about the text as well as make connections within and between texts. One version of multiple reading is repeated reading, which is popular in an academic practice that aims to increase oral reading fluency. Repeated reading \cite{meyer1999repeated, therrien2006developing, o2007repeated, webb2012vocabulary, lee2017effects} can be used with learners who have developed initial word reading skills but demonstrate inadequate reading fluency for their grade level. Multiple reading and repeated reading techniques are studied by mostly the researchers of teaching and learning domains. The researchers conclude that repeated reading is more effective at increasing learners' reading fluency than question generation intervention \cite{therrien2006effect}. Additionally, the learners in the repeated reading perform significantly better than the question generation learners on factual comprehension measures \cite{therrien2008comparison}. In recent decades, to assess learners' performance skills, multimodal responses including reading, text production, writing and speaking are addressed.

Over the past decades, researchers of psychology, language acquisition, teaching \& learning domains study eye-movements to analyse the reading and learning in learners (mostly school-students) \cite{spichtig2017comparison, shin2013analysis, jian2016fourth, scheiter2018self, jian2017eye, yun2020comparing}. They studied various types of reading patterns including skimming, scanning, reading, and searching. The researchers of psycholinguistics domain proposed that during reading several factors affect learners' reading and the degree of impact depends on their skills and the complexity of reading text. Most studied psycholinguistics factors are word frequency, meaningful, imagery, familiarity, emotion, and age-of-acquisition \cite{joseph2013using, gruhn2016english, ljubevsic2018predicting, desrochers2009subjective, dirix2017eye, dye2013children, juhasz2018experience, juhasz2003investigating, juhasz2006role, juhasz2015database, khwaileh2018imageability, rofes2018imageability, warriner2013norms}. The impacts of these factors on first-language and second-language writing skills were also studied and were reported about their significance. Other important linguistic measures are lexicons, syntactic complexity, cohesion, readability index –to assess language proficiency in free-text writing or oral speech production. To assess content enrichment in written text, different Natural Language Processing (NLP) tools and techniques are used to measure lexical-similarity, syntactical-similarity, and semantic-similarity; these all were reported in linguistic domain prior arts. One important part of comprehension assessment is to measure oral fluency in a spontaneous speech; for that several acoustics-prosody as well as temporal variables were reported in oral proficiency studies.



The prior arts of related domains provide sufficient supports to develop computer-assisted assessment (CAA) systems to assess learners' learning, acquisition and comprehension skills using multimodal responses such as eye movements, writings, and oral speeches. Several methods, tools and techniques had been used to develop computer-assisted assessment systems and were reported in \cite{bahari2020computer, berzak2018assessing, liu2016automated, roy2015perspective, d2012gaze, kakkonen2011essayaid}. However, there are some limitations of these systems listed below, which are addressed in this thesis: 
\begin{itemize}
	\item Prior art systems evaluate learners' performance based on only one mode of their response.
	\item These systems can not measure the changes happened in learners' comprehension skills due to multiple reading or repeated reading intervention.
\end{itemize}
To address these limitations, there was a requirement to develop a holistic framework to assess learners' multimodal responses for measuring the level of comprehension skills and became the main motivation behind this PhD work.


\section{Objectives}

The first objective of this thesis is to study the changes occurred in different modalities measures, that show readers' comprehension in terms of reading, writing and spontaneous speaking, due to repeated reading practice. Since readers' comprehension depends on several factors including text complexity. Therefore, the second objective of the thesis is to analyse readers' skills performance on two levels of text complexity-- high (university level) and low (middle school level). Several measures have been already proposed by prior researchers, but all of them would not be applicable in all the cases. Therefore, the third objective of the thesis is to find significant measures which can improve machine learning outputs. The fourth and last objective of the thesis is to develop a machine-learning-based framework to determine as well as to track the changes that occurred in readers' comprehension skills due to repeated reading practice.


\section{Research Context}
This thesis proposes analytic learning\footnote{``Analytic Learning is an analytical approach to learning that uses prior knowledge as a base from which concepts can be described, hypotheses can be developed, and concepts can be rationally generalized by analysing the components and the structure of the concepts. In the fields of artificial intelligence and computer science, analytical learning is a form of machine learning which concerns the design and development of algorithms in computer science, or related sciences.'' \cite{seel2011encyclopedia}} approach to understand the impact of repeated reading practice on learners' multimodal response. The data collection is a major part of this approach. We use data extracted from different responses such as eye movement (gaze), written text and oral speech as the main learning measures in our experiments. As shown in figure \ref{fig:Chapter-1fig1} our work lies at the mixed of several domains:
\begin{figure}[!h]
	\centering
	\includegraphics[scale=0.4]{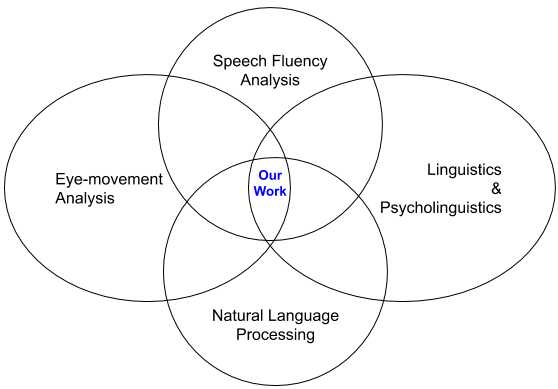}
	\captionsetup{font={small}, labelfont=bf, skip=4pt}
	\caption{The placement of this dissertation work within the relevant research domains}
	\label{fig:Chapter-1fig1}
\end{figure}
\begin{enumerate}[(i)]
	\item \textbf{Eye-movement Analysis:} It is used to understand various cognitive processes called by ongoing comprehension process during the reading of a text. It can establish a relationship between reading skills and text properties. Therefore, it is vital for explaining the differences among learners of different levels of comprehension skills. 

	\item \textbf{Linguistics and Psycholinguistics:} Linguistic features such as lexicon, syntax, morphology, semantics and discourse analysis are extracted from referred text as well as from learners' produced text to analyse their comprehension level. Psycholinguistics features such as cohesion, readability etc. indicate the mental representation of text in learners' mind. 

	\item \textbf{Natural Language Processing:} NLP facilitates to perform lexical, syntactic, semantic and discourse analysis of given text by applying tools and techniques such as parsing, stemming, named entity recognition, semantic relation among words, knowledge representation using concept network, electronic thesaurus etc. NLP is hugely used for calculating a similarity score between two text.

	\item \textbf{Oral Fluency Analysis:} The fluency in oral speech production is calculated using two methods- i) acoustics-prosody (low-level descriptors) analysis, and ii) temporal features analysis. In the former method, audio is treated as signals and various signal properties are extracted using a sliding window analysis. Whereas, in the latter method, the audio is transcribed and separated into segments corresponding to the speech text. Oral disfluencies such as brief pause, filled pause etc. in an audio are explicitly marked at the time domain.
\end{enumerate}
\section{Problem Statement}
Since first language (L1) learners have learned their mother tongue orally before learning to read, and they are enough exposed to language, they are in general proficient in L1 comprehension; but second language (L2) or foreign language (FL) learners oral language and reading development occur simultaneously, and their contact with language data is so limited. That is why reading in L2 or FL is a laborious process and learners need go through hardships to develop their own reading fluency and comprehension.

Repeated reading or multiple reading (henceforth, both are used as synonyms) is common in learners' learning practice, especially for low comprehend learners. Repeated reading can help L2 learners to improve their reading comprehension, spontaneous text production, oral proficiency, as well as linguistic and psycholinguistics processing.

Through this dissertation, we try to answer the following research questions:\\
\textbf{How to classify L2 reading effort in terms of eye movements into high and low automatically in repeated reading practice?}
During reading of a text, eye movements depend upon readers' prior knowledge, the complexity of the text as well as the degree of familiarity with the text. So, readers' efforts in terms of eye movements can be varied during a reading. Predicting the reading efforts of a reader during reading can provide a facility to compare that of with others' reading efforts and thus helps to determine the level of comprehension skills. Also, during repeated reading practice, readers' efforts can provide a clue about the improvements in their reading skills. In this thesis, we try to classify learners' efforts during repeated reading practice.\\
\textbf{How to measure the influence of L2 word ratings on eye movements and learners' text production in repeated reading?}
Every content word in a text carries meaning as well as is given some ratings on different psycholinguistics measures, such as word frequency, age-of-acquisition, familiarity, meaningfulness, imagery, emotion and concreteness. In the prior studies, the influence of words on reading eye movements as well as text production has been related to their ratings. These studies describe the rating effect on a single reading situation. However, the role of the word rating effects in repeated reading has not explored yet. Therefore, we try to explore it in this thesis.\\
\textbf{How to measure the improvements in L2 linguistics in repeated reading?}
The research in the domain of L2 study the effect of repeated reading on reading fluency, pronunciation, and reading comprehension development. Repeated reading can also help to improve other linguistic features especially in EFL/ESL learners. These linguistic features can include lexical richness, syntactic complexity, cohesion, readability and psycholinguistics influence. We try to measure the improvement in linguistics due to repeated reading.\\
\textbf{How to measure the improvement in L2 text production in repeated reading?}
The importance of repeated reading practice on the improvement of spontaneous text production has not been studied in the previous experiments. However, learners use this reading approach to improve their spontaneous text production ability by adding new concepts into their knowledge base. We try to measure the improvement in knowledge (or concept) regarding the reading text during repeated reading practice.\\
\textbf{How to measure the improvement in L2 oral proficiency in repeated reading?}
In a repeated reading practice, a text exposes several times; therefore, the familiarity of words is increased. And so, the oral fluency in related text production would be improved. We try to measure the improvement in oral proficiency during repeated reading.\\   
\textbf{How to measure the impact of the complexity of reading text on comprehension skills improvement in repeated reading?}
Prior studies examine the impact of text difficulty on learners' different skills such as oral reading prosody, text comprehension, reading patterns, text production etc. In this thesis, we try to measure the impact of text difficulty on improvements in comprehension skills during repeated reading practice. 
\section{Thesis Roadmap}
This dissertation is organised as follows:\\
In the next chapter-\ref{Chapter2}, we explain the experiment for collecting eye-movement data in two repeated reading sessions, one for intensive reading and the other for extensive reading. The data, recorded using an eye-tracking device and representing learners' reading efforts, was labelled as high and low in terms of eye-movements. Then we explain a machine learning procedure to automatically classify the data. In this chapter, we propose several hypotheses to explain the significance of eye movement features for machine learning in the triumvirate of reading approaches: repeated reading, intensive reading, and extensive reading.\\
Chapter-\ref{Chapter3} present an eye-movement study to find the relationship between gaze patterns, and the words ratings on various psycholinguistics factors-- word frequency, age-of-acquisition, familiarity (meaningfulness), imagery (imageability), concreteness, and emotion. In this chapter, we propose several hypotheses to explain the relation between fixation features and the rating of words in the triumvirate of reading approaches: repeated reading, intensive reading, and extensive reading. We use a linear mixed-effects regression method for measuring the significant effect of repeated reading on eye-movements under psycholinguistics factors.\\
In chapter-\ref{Chapter4}, we present linguistic proficiency analysis of learners' text production (summary writing). In this chapter, we propose several hypotheses-- a) to explain machine learning on linguistic features to categorise learners' writings proficiency automatically, and b) to study the effect of repeated reading on the linguistic features to show changes in learners' writing skills.\\
Chapter-\ref{Chapter5} present NLP methods to measure content enrichment in learners' text production (summary writing). In this chapter, we propose a hypothesis that content similarity between the referred text and learners' summary text could be automatically scored by machine learning methods like as human raters do. We also propose another hypothesis that repeated reading has an effect on similarity scores and shows improvement in learners' writing in terms of content enrichment.\\
In chapter-\ref{Chapter6}, we present acoustics-prosody and temporal features to measure the oral fluency label of learners' summary speech audio. In this chapter, we propose a hypothesis that using significant features of spontaneous audio speech, learners' summary speech fluency could be automatically classified by applying machine learning methods. We also propose another hypothesis that the repeated reading has an effect on oral fluency features.\\
Finally in chapter-\ref{Chapter7}, we conceptualise a computational framework for learning assessment based on multimodal responses, and conclude with a summary and general discussion about the contributions of this work. This chapter also explain the limitations and the implication of our work for future research.

In social experiments, measurements involve assigning scores to individuals so that they represent some characteristic of the individuals \cite{jhangiani2015research}. In our framework for evaluating a measurement, we consider two different dimensions-- reliability and repeatability. Reliability refers to the consistency of a measure. We use Pearson correlation coefficient ($r$), Quadratic Weighted Kappa (QWK) agreement and Spearman’s rank correlation coefficient ($\rho$) for calculating inter-rater reliability \cite{wiki:Inter-rater_reliability}. The repeatability describes the relative partitioning of variance into within-group and between-group sources of variance. LMER model framework is considered as a powerful tool for estimating repeatability. In this model, random-effect predictors are used to estimate variances at different hierarchical levels \cite{stoffel2019introduction}.

\chapter{Classification of In-depth Reading Patterns}
\label{Chapter2} 
\lhead{Chapter 2. \emph{Classification of In-depth Reading Patterns}} 

\section{Introduction}

In this chapter, we present the classification of readers' in-depth eye movements recorded during repeated reading of two texts. Readers, such as students generally read the same text more than once for improving their comprehension and also for memorizing more details. Studying eye movements during repeated reading of a text provides a unique opportunity to observe changes in reading patterns. The changes in these patterns that occurred from one reading attempt to another attempt can be a witness to a change in readers' reading comprehension level. Various reading patterns hint different reading circumstances and purposes. Readers read for enjoyment (extensive reading), while their eye movement trajectory is fast and leaping \cite{herman2019relationship}. Readers read also for learning new knowledge (intensive reading), by moving eyes slowly and stagnated.

Through this study, we infer reading patterns from eye movement tracking data. We also develop a machine learning model based on different classifiers trained to identify in-depth reading behaviour. The chapter first describes the context using prior studies-- eye movements in reading, gaze patterns in classification, English as a foreign language (EFL), extensive and intensive approaches to language learning in academics, and repeated reading (RR). Once the context is established, we provide the details of a repeated reading experiment, which is followed by the definition of different variables. These variables are used to analyse the relationship among the triumvirate and then the data is classified to label users reading efforts as `high' and `low'. Finally, we present the results of the study and discussion. For this chapter, we conceptualise our domain of investigation as a triumvirate that consists of eye-movement pattern, repeated reading, and extensive \& intensive reading approaches in EFL reference.


\section{Context}

We explain here some established theories, concepts and the findings of previous researches in related fields.

\subsection{Eye Movements in Reading}

When a reader reads a text, the eyes move back and forth horizontally across the sentences from one word to another. The reader makes one or more brief pauses at some words to take in information. These pauses, called fixations, generally end between 50—1500 ms. The frequency and duration of a fixation usually depend on the length and difficulty of the word. The rapid movements in between the fixations that change the point of the reader's gaze focus are called saccades. Generally, a saccade takes 20—50 ms to complete, depending upon the length of the movement, and no visual information is extracted during such jerky eye movements. In a reading, typically, more than 80\% of all saccadic eye movements are forward and the remaining 20\% are backward eye movements occurring to reread previous words. These backward movements, called regressions, are mostly caused by difficulties in linguistic processing, i.e., regressions are induced with difficult sentences, which often lead to incorrect syntactic analyses and a reader often makes regressions back to the point of difficulty to reinterpret the sentence.

The process of measuring where one looks, i.e., the point of gaze is called eye-tracking. From the last several decades, eye tracking devices are used to objectively measure eye movements by applying non-invasive methods including video recording sensors.

Researchers of various fields including psychology, linguistics, affective domain, psycholinguistics etc. have been working on reading in relation to eye movements \cite{lohr1998efficacy, radach2008role, monty2017eye, ward2007linguistic, coco2014classification, sharmin2016reading, stolicyn2020prediction, rauthmann2012eyes, moed2013psycholinguistic}. They have proposed different eye-tracking research models to analyse three levels of readings \cite{jarodzka2017tracking}: first level- reading words and sentences, second level- reading a whole-text and third level- reading multiple text. During a reading, a reader's decision about where to fixate eyes next, is determined mainly by low-level visual cues in the text such as word length and the spaces between words. The linguistics properties of a word influence the decision about when to move the eyes. For example, a reader's gaze can be longer at low-frequency words than at high-frequency words. Similarly, longer foveal vision is required to read longer (having four or more characters such as `institution') and more complex words, whereas short (having one-three characters such as `and' and `or') words are often skipped. Fixation count, fixation duration, first-pass, second-pass, saccade-length, and total reading time represent various cognitive and psycholinguistics processes that occurred during reading.

Usually, in a reading, five types of eye movements appear \cite{liao2017classification}:
\begin{enumerate}[(i)]
	\item \textbf{Speed reading:} To achieve speed reading, readers spend little time (\textless200ms) on words and the fixation is not word by word but sentence by sentence.
	\item \textbf{Slow reading:} It reflects a slower way of reading. Readers read word by word smoothly.
	\item \textbf{Skim-and-skip:} In skim-and-skip mode, readers focus on the part they found important or interesting and skip unimportant content to get necessary information quickly.
	\item \textbf{Keyword spotting:} During a reading, readers may find some keywords in a text, so their fixation points are moving between some words.
	\item \textbf{In-depth reading:} When readers focus on reading for learning, it shows the in-depth reading pattern. They mostly read word by word over and over again, until they understand the content.
\end{enumerate}	

\subsection{Gaze Patterns in Classification} 
Several researchers have used eye movements data to extract significant and relevant features to implement machine learning models. \\
Liao et al. \cite{liao2017classification} proposed a classification based model to predict the type of reading eye movements using SVM classifier. Their average classification accuracies were 78.24\%, 74.19\%, 93.75\%, 87.96\%, and 96.20\% for speed reading, slow reading, in-depth reading, skim-and-skip, and keyword spotting respectively. \\
Miyahira et al. \cite{miyahira2000gender} proposed a methodology to examine gender differences regarding four eye movement parameters, mean gazing time, total number of gaze points, mean eye scanning (i.e., saccade) length and total eye scanning length.\\
Landsmann et al. \cite{landsmann2019classification} developed a web-based tool called `Vocabulometer' to classify short sequences of fixations from eye gaze data into actual reading and actual not reading.\\
Mozaffari et al. \cite{mozaffari2018reading} developed machine learning models using SVM and BLSTM to classify eye movements into reading, skimming, and scanning with more than 95\% classification accuracy.\\
Fraser et al. \cite{fraser2017analysis} presented a machine learning analysis of eye-tracking data for the detection of mild cognitive impairment using two methods-- concatenation and merging of combining information from (aloud and silently) reading trials. They distinguished between participants with and without cognitive impairment with up to 86\% accuracy.\\
Parikh and Kalva \cite{parikh2018eye} proposed a Feature Weighted Linguistics Classifier tool (FWLC) to predict difficult words and concepts with 85\% accuracy using eye responses.

\subsection{English as a Foreign Language}

Due to emergence as a global language, English as a foreign language (EFL) has been established as the primary medium of instruction in higher education in several developing countries including India. Reading in a second language (L2) differs from reading in a first language (L1) in distinct ways. L1 learners have well-developed oral proficiency, vocabulary knowledge, and tacit grammar knowledge at the time they start learning to read, which leads to fluent processing of text information. Whereas, L2 learners have limited oral proficiency, vocabulary and underdeveloped grammar knowledge. Therefore, compared to L1 learners, L2 learners are invariably slower and less accurate in processing text. In this thesis, L2, EFL and ESL are interchangeable and represent the same meaning in language learning context.

Extensive reading, intensive reading, and repeated reading are approaches of reading and teaching programs that have been using in EFL settings as effective means of developing reading speed, fluency, and comprehension.

\subsection{Extensive Reading}

Extensive Reading (ER, henceforth) can be defined as the independent reading of a large quantity of materials for information or pleasure. The primary aim of ER program is ``to get students reading in the second language and liking it'' \cite{day1998extensive}. Readers read self-selected materials from books, which have reduced vocabulary range and simplified grammatical structures, to achieve the goal of reaching specified target times of sustained reading. ER is thought to increase L2 learners' fluency, i.e., their ability to automatically recognize more number of words and phrases, an essential step to the comprehension of L2 text. ER encourages L2 readers: a) to read for pleasure and information both inside and outside the classroom, b) to read for meaning, and c) to engage in sustained silent reading developing \cite{taguchi2004developing}. Research, investigating the benefits of ER in L2 contexts, has shown that ER improves L2 readers' comprehension, promotes their vocabulary knowledge development, and enhances their writing skills and oral proficiency \cite{iwahori2008developing, soltani2011extensive}. ER has also been reported to be effective in facilitating the growth of readers' positive attitudes toward reading and increasing their motivation to read \cite{endris2018effects}. To date, several empirical studies have supported the positive impact of the ER approach in promoting the reading rate of L2 learners compared to the traditional intensive reading approach \cite{park2017comparison, park2020comparison}.

\subsection{Intensive Reading}

Intensive Reading (IR, henceforth) approach is a conventional reading approach that aims to support L2 learners in constructing detailed meaning from a reading text through close analysis and translation led by teachers in order to develop their linguistic knowledge. In IR, learners usually read text that are more difficult in terms of content and language than those used for ER. Therefore, to help learners make sense of text that may present a significant challenge in terms of vocabulary, grammar and concepts, teachers focus on reading skills, such as identifying main ideas and guessing the meaning of unfamiliar words from context \cite{lang2009exploring, segueni2019investigating}.

\subsection{Repeated Reading}

When beginning to learn new knowledge, most L2 learners hope to achieve proficiency. To help learners reach this goal, it is essential to apply some reading strategies with a focus on the development of L2 fluency. One such method which is highly recommended in academics is repeated reading (RR, henceforth). In the RR approach, L2 learners read specified passages repeatedly in order to increase their sight recognition of words and phrases, resulting in increased reading fluency and comprehension. The RR approach has been extensively studied in English as a first language (L1), English as a second language (ESL) and English as a foreign language (EFL) contexts and overall it has been shown to be effective in developing reading fluency of monolingual and bilingual readers of English \cite{gorsuch2008repeated, taguchi2012assisted, taguchi2004developing}. There are two forms of repeated reading: unassisted and assisted \cite{young1996effects}. With unassisted repeated reading, learners are given reading passages that contain recognizable words at their independent reading levels. Learners silently or orally read given passage several times until they reach the predetermined level of fluency. Assisted repeated reading, on the other hand, involves repeated reading whilst or after listening to either a teacher reading the same text or a recorded version \cite{liu2016implementation}. Foster et al. \cite{foster2013underlying} stated that ``RR facilitates improved fluency for beginning readers by decreasing the amount of time they spend reading individual words and by reducing their need to reconsider previously read content. This suggests that RR not only improves word recognition and sight word acquisition, but also potentially influences higher level comprehension processes."


\section{Problem}

In this chapter, we propose three expectations as listed below:
\begin{enumerate}
	\item Using significant features of eye movements recorded during repeated readings, readers' in-depth reading efforts can be automatically classified.
	\item Repeated readings have a significant effect on eye-movement features.
	\item There is a relation between eye-movement features and learners' objective answers.
\end{enumerate}

In order to test these expectations, we conducted an eye-tracking experiment, in which participants attended two repeated reading sessions for reading two texts. In a session, they read one text once in a day for consecutive three days. The reading of text-1 and text-2 had given the experience of ER and IR respectively. After finishing the reading of a text, they answered some objective type questions, which were related to the text. 

The data collected in the repeated reading experiment are studied in the current chapter as well as in the next chapter with a different perspective.


\section{Experiment}
A brief description of the participants, materials, apparatus, and procedure that we used in this study are described here.
\subsection{Participants}
Thirty EFL bilingual Indian Institute of Information Technology Allahabad students (10 females and 20 males, age-range = 19-22 years, mean age = 20.8 years) had participated in the experiment. All participants were enrolled in a bachelor program of STEM discipline. To encourage active involvement, they were awarded some course credits for their participation in the experiment. They had a normal or corrected-to-normal vision. None of the participants reported having any language or reading impairments. They performed regular academic activities (e.g., listening to class lectures, writing assignments, watching lecture videos) in English (L2) only; whereas their primary language (L1) might be different from each other. Also, they reported that they could carry on a conversation, read and comprehend instructions, read articles, books, as well as watch TV shows and movies in English (L2) as well.

\subsection{Materials}

To fulfil the purpose of the experiment, participants needed to read completely strange text. Therefore, a separate group of five Indian Institute of Information Technology Allahabad undergraduates had selected two texts from the outside of their academics for the experiment with a full majority. The text-1 (\ref{AppendixA1}) is a simple narrative story, whereas the text-2 (\ref{AppendixA2}) is a relatively more complex informative article. The text-1 was used for simulating enjoyable ER and the text-2 was for IR. These texts were unread in the participants’ lifespan until the experiment begin.

\subsection{Apparatus}

\begin{figure}[!h]
	\centering
	\includegraphics[scale=0.5]{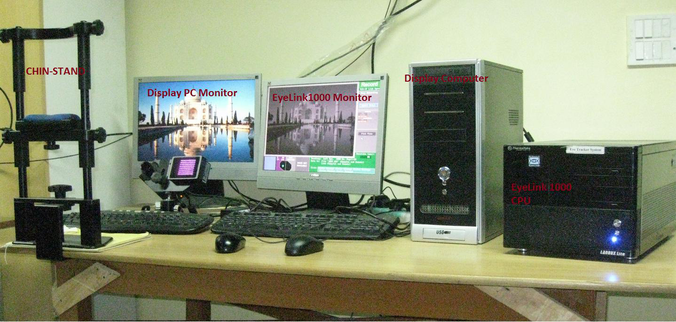}
	\captionsetup{font={small}, labelfont=bf, skip=4pt}
	\caption{Eye Tracker (SR Research Eyelink 1000) setup at SILP lab}
	\label{fig:Chapter-2fig1}
\end{figure}

The reading eye movements were recorded from a 1 KHz remote video binocular eye tracker (EyeLink 1000; SR Research, Canada) at a viewing distance of 70 cm as recommended in the EyeLink 1000 User Manual \cite{eyelink1000manual}. The setup of the eye-tracker system, available at SILP research laboratory, is shown in figure \ref{fig:Chapter-2fig1}. A chin-stand was used to reduce participants' head movements during recording their eye movements. Reading eye movements data was acquired with the system having a desktop mounted video binocular camera, but was recorded only from the participant's one eye, that was selected based on individual's nine-point calibration accuracy. 

The text appeared in paragraphs on the screen of a 22 inch LCD monitor. Each text was presented in black 14 point Courier New font on a light grey background and the lines were single-spaced. Title, heading, bullets, images etc. of the original text were excluded and only core sentences were displayed, which made the text anonymous to the participants. All paragraphs of a text were presented in five slides. The participants were able to change the slides forth and back by pressing left and right arrow keys on a keyboard, without looking at them. During the recording of eye movements, the room was quiet and dimly illuminated.
\subsection{Procedure}

Upon their first arrival in SILP research laboratory, the participants read and signed a consent form prior to starting the experimental procedure. They were informed about the three-day experiment and its procedure. They were aware that, they had to read two texts per day for consecutive three days, but they had not been given a hint that both texts, reading on first day (day-1), would be repeated on the next two days. The design of data collection experiment was inspired by Sukhram et al. \cite{sukhram2017effects} repeated reading experiment. 

Thus, in the experiment, participants attended two reading sessions in a day for three consecutive days. In a session, they individually sat in front of a monitor on which after the success of calibration process, the slide of text was displayed, and the individual started to read. There was no time limit given for finishing their reading to provide them a natural reading condition. So, participants were able to read at their own pace. They were instructed to read the text silently while the eye tracker recorded their eye movements, and to move their head and body as little as possible during their reading. After completion of reading, they were given some puzzles (Appendix \ref{pdf:puzzle}) to solve on a paper sheet in two minutes. After two minutes, the puzzle-sheet was collected and they were given another paper sheet having ten objective type questions (five fill in the blanks and five multiple-choice; the maximum mark was 10) (Appendix \ref{pdf:questionnaire}) all related to the read text. In the experiment, puzzles were used to clear their rote/working/short-term memory to ensure that answers would come from participants' long-term memory, where text related knowledge (comprehension) stored. Also, it had been ensured that they had no prior information that the questions in both sessions would be the same across the three-day trial of the experiment. The intention behind introducing questions was that the participants would pay full attention to text during reading, as well as to find any significant relationship between participants' eye movements and their obtained scores. Thus, in session-1, participants read text-1 and gave answers to its objective questions. Similarly, in session-2, they read text-2 and gave answers to its questions. Their scores were announced when the experiment ended on the third day. The participants took 15—25 minutes to finish a session. Between the two sessions, they were given a break of 15-minute for refreshment.

The distribution of participants' obtained scores, given in table \ref{tab:Chapter-2tab1}, shows that in session-1 highest seven participants achieved the maximum score on the second day, whereas in session-2 only three participants achieved the maximum score on the third day.

\begin{table}[!h]
	\captionsetup{font={small}, labelfont=bf, skip=4pt}
	\caption{Participants' score distribution} 
	\centering 
	\small	
	\renewcommand{\arraystretch}{1.2}
	\begin{tabular}
		{C{2cm} R{0.8cm} C{0.8cm} C{0.8cm} C{0.8cm} C{0.8cm} C{0.8cm} C{0.8cm} C{0.8cm} C{0.8cm} C{1cm}}
		\hline 
		\textbf{\backslashbox{Day}{Score}} & \textbf{1} & \textbf{2} & \textbf{3} & \textbf{4} & \textbf{5} & \textbf{6} & \textbf{7} & \textbf{8} & \textbf{9} & \textbf{10} \\[0.5ex]
		\hline 
		\multicolumn{11}{c}{\textbf{Session 1}} \\
		\hline 
		
		1 & 0 & 0 & 2 & 4 & 3 & 16 & 3 & 1 & 0 & 1 \\
		2 & 0 & 0 & 0 & 0 & 0 & 4 & 2 & 8 & 9 & 7 \\
		3 & 0 & 0 & 0 & 0 & 0 & 1 & 4 & 10 & 11 & 4 \\[1ex] 
		\hline 
		
		\multicolumn{11}{c}{\textbf{Session 2}} \\
		\hline 
		1 & 0 & 3 & 1 & 9 & 5 & 11 & 0 & 1 & 0 & 0 \\
		2 & 0 & 1 & 0 & 2 & 1 & 12 & 4 & 8 & 2 & 0 \\
		3 & 0 & 0 & 0 & 0 & 0 & 9 & 3 & 8 & 7 & 3 \\[1ex] 
		\hline 
				
	\end{tabular}
	\label{tab:Chapter-2tab1}
\end{table}

Table \ref{tab:Chapter-2tab2} shows a summary of the properties of two texts. Text-1 has relatively more a) number of words, b) average words per sentence, c) average content words per sentence, d) average word frequency, and e) average content word frequency than those of text-2. Whereas, text-2 has relatively more a) number of unique words, b) number of unique content words, c) number of unique stop words, and d) number of sentences than those of text-1. Paired t-test yields significant difference between both texts concerning average word frequency and average content word frequency. Text-2 has more unique words as well as unique content words than those of text-1, which means text-2 is more complex than text-1 in the terms of comprehensibility.

\begin{table}[!h]
	\captionsetup{font={small}, labelfont=bf, skip=4pt}
	\caption{Summary of the properties of the text} 
	\centering 
	\small	
	\scalebox{0.92}{
		\renewcommand{\arraystretch}{1.4}
		\begin{tabular}{L{7.8cm} L{1.8cm} L{1.5cm} L{3.cm}}
			\hline 
			\textbf{Descriptive parameters} & \textbf{Text-1} & \textbf{Text-2} & \textbf{T-value (Welch’s t-test)} \\[0.5ex]
			\hline 
			
			Number of words & 718 & 662 & - \\
			Number of unique words & 199 & 307 & - \\
			Number of unique content words & 97 & 179 & - \\
			Number of unique function (stop) words & 102 & 128 & - \\
			Number of sentences & 30 & 35 & - \\
			Average (SD) of words per sentence & 24.5 (14.5) & 19 (7.4) & 1.9 ($p=0.06$) \\
			Average (SD) of content words per sentence & 9.1 (4.8) & 7.9 (3.3) & 1.09 ($p=0.27$) \\
			Average (SD) word frequency & 3.7 (6.5) & 2.1 (4.9) & 2.8\textsuperscript{**} ($p=0.004$) \\
			Average (SD) content word frequency & 2.8 (4.1) & 1.5 (1.5) & 2.9\textsuperscript{**} ($p=0.004$) \\[0.5ex]
			\hline 
		\end{tabular}}
		\label{tab:Chapter-2tab2}
	\end{table}

\section{Method}

To classify participants' eye movements showing in-depth reading efforts; the eye movement (gaze) dataset was categorised into two labels- high and low, based on fixation and (forward \& backward) saccade characteristics. The methodology, to develop an automatic learning and classification system, has following steps: a) gaze extraction from recorded eye movement data, b) gaze data labelling, c) noise removal from gaze data, d) feature extraction, e) feature normalisation, f) feature selection, and g) automatic classification. 
\subsection{Gaze Extraction from Eye Movement Data}

\begin{figure}[!h]
	\centering
		\captionsetup{font={small}, labelfont=bf, skip=4pt}
	\begin{minipage}[b]{0.51\textwidth}
		\includegraphics[width=\textwidth]{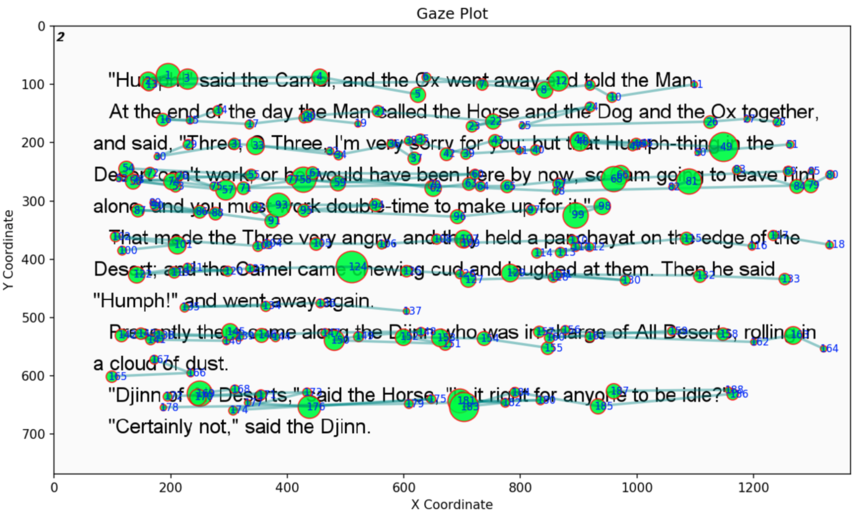}
		\caption{Fixation and saccade events on a slide (with text)}
		\label{fig:Chapter-2fig2}
	\end{minipage}
	\hfill
	\begin{minipage}[b]{0.48\textwidth}
		\includegraphics[width=\textwidth]{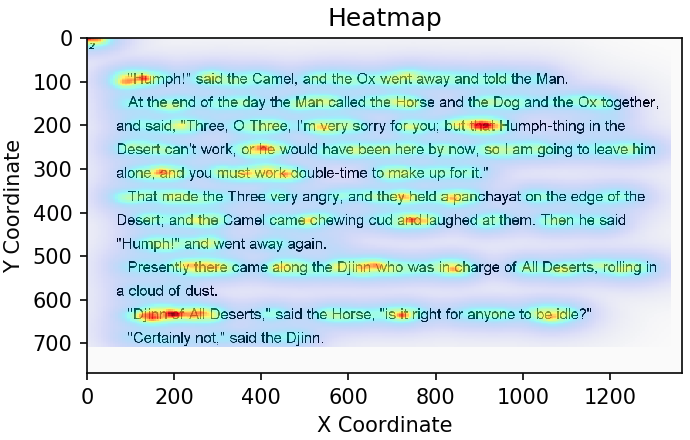}
		\caption{Heatmap of fixation events on a slide}
		\label{fig:Chapter-2fig3}
	\end{minipage}
\end{figure}

The eye tracker provides three types of eye events- fixation, saccade and blink. Both fixation and saccade events contain various information including event-name, eyes (left/right), event-start-timestamp, event-end-timestamp, event-duration (in milliseconds), average X and Y coordinates of eye gaze position, and average-pupil-size of the recorded eye. The saccade events contain some additional information such as average-velocity and peak-velocity. The fixation and saccade events generated from a participant's reading eye data are shown in figure \ref{fig:Chapter-2fig2}. In the figure, the circles represent fixations, and their diameters are in proportion to their fixation-durations. The fixation numbers show fixation sequence as well as the direction of eye movements during a reading. In the figure, two fixations are connected with a line representing a saccade-length. Figure \ref{fig:Chapter-2fig3} show a heatmap of fixation events generated from a participant's eye-movement data, the degree of redness shows that participant spent relatively more time (long fixation-duration) on some words. The blink events indicate at what time the eyes were not tracked by the eye-tracker. Since these have not any vital role in the analysis of reading; therefore, blink events were excluded from the gaze data.

\subsection{Gaze Data Labelling}

When the visuals of gaze data were inspected on text plot (fig. \ref{fig:Chapter-2fig2}), it had been observed that some participants put more reading efforts than others during their in-depth readings i.e., they had spent more time in reading words and thus more number of fixation and saccade events were generated. Since we had not conducted a pre-test for measuring participants' reading skills; therefore, at the moment, we can not establish a relationship between their reading efforts (in terms of eye movements) and their comprehension levels. 

The gaze data of participants were labelled by an expert having 8 years of experience in the field of eye-tracking data analysis. For labelling, all 180 gaze data (30 students $\times$ 3 days $\times$ 2 sessions) were separated into two groups based on their collecting session: the gaze data collected in session-1 formed one group and session-2 gaze data formed another group. Each gaze data of both groups was assigned trial day number (1, 2, \& 3), and was made anonymous by removing i) participants' personal information (name, age, gender etc.) and ii) the text. This dataset was provided to the expert for labelling them as `high' and `low' to quantify in-depth reading efforts. The expert compared each anonymous gaze data within its session-group and trial-day, and provided a label either `high' or `low', representing the participant's eye movements effort during in-depth reading in terms of fixations and saccades. 

\begin{figure}[!h]
	\centering
	\includegraphics[scale=0.35]{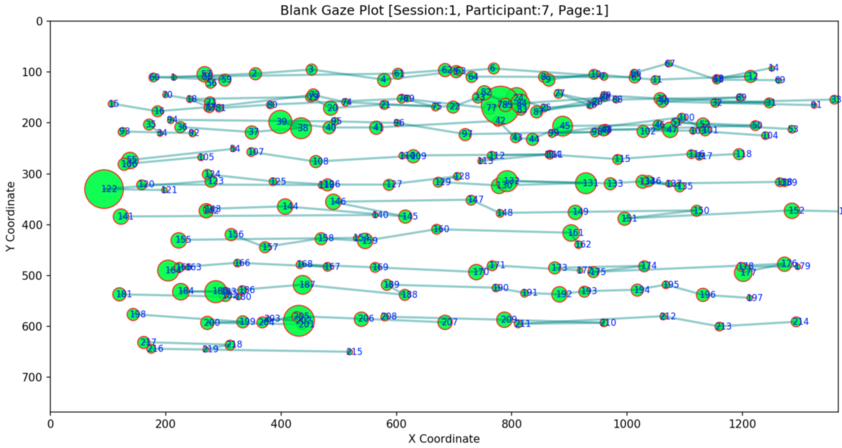}
	\captionsetup{font={small}, labelfont=bf, skip=4pt}
	\caption{Fixation and saccade events on a slide (without text)}
	\label{fig:Chapter-2fig4}
\end{figure}

For labelling the data, the expert used a visualization tool for plotting gaze on a blank surface as shown in figure \ref{fig:Chapter-2fig4}. The distribution of high/low labels of gaze data, given by the expert, is reported in table \ref{tab:Chapter-2tab3}. In session-1 gaze dataset, the number of high-label is gradually decreasing and thus, low-label is accordingly increasing as repeated reading trial performed from day-1 to day-3. However, in session-2 gaze dataset, changing in label distribution is not so clear from day-1 to day-3. The expert labelled the gaze data considering weightage of three events- fixation, regression, and saccade as roughly 40\%, 30\%, and 20\% respectively and 10\% for randomness (treated as noise).

\begin{table}[!h]
	\captionsetup{font={small}, labelfont=bf, skip=4pt}
	\caption{Details of label distribution in gaze dataset} 
	\centering 
	\small	
	\scalebox{1}{
	\renewcommand{\arraystretch}{1.4}
	\begin{tabular}
		{C{2.2cm} C{1.8cm} C{2.9cm} C{2.9cm} C{2.9cm}}
		\hline 
		\textbf{Session} & \textbf{Day} & \textbf{Total participants} & \textbf{Total High-label} & \textbf{Total Low-label} \\[0.5ex]
		\hline 
		
		\multirow{3}{*}{1} & 1 & 30 & 16 & 14 \\
		& 2 & 30 & 14 & 16 \\
		& 3 & 30 & 13 & 17 \\
		\hline 
		\multirow{3}{*}{2} & 1 & 30 & 20 & 10 \\
		& 2 & 30 & 17 & 13 \\
		& 3 & 30 & 18 & 12 \\[0.5ex]
		\hline 
	\end{tabular}}
	\label{tab:Chapter-2tab3}
\end{table}

\subsection{Noise Removal from Gaze Data}

Different types of noises can be present in a gaze data including-- isolated fixation, fixation having extreme low/high duration, calibration error- the distance between the actual eye location and predicted gaze location, random movements, etc. These noises were removed using a heuristic-based approach followed by manual correction approach. In heuristic-based approach \cite{mishra2012heuristic}, recorded gaze data was corrected in three phases. In phase-1, isolated fixations and fixations having extreme low duration (below  a threshold) and high duration (above another threshold) were removed from the gaze data. In our case the thresholds for extreme high and extreme low were-- 50 ms and 1000 ms respectively. In phase-2, the y-coordinate of each fixation was re-assigned the y-coordinate of the closest sentence line. In the last phase, continuous abnormalities in fixation sequence had been corrected. In the manual correction approach, the modified gaze data was examined manually to detect any remaining abnormalities and such disparities were corrected manually. Recently, Carr et al. \cite{carr2020algorithms} proposed additional algorithms for the automated correction of vertical drifts in eye movement data.

\subsection{Feature Extraction}

From rectified gaze data, following features were extracted: seven fixation features from the fixation events, seven forward movement (saccade) features from the saccade events and eight regression movement (re-reading) features from the saccade events, as explained in \cite{mishra2017harnessing, mishra2018scanpath, mishra2016predicting, mishra2018cognitively, fraser2017analysis, faber2018automated, foster2013underlying, cop2015eye, parikh2018eye, mozaffari2018reading, yang2017examining, mathias2018eyes, kuperman2018contributions}. The details of these features are given below:

\vspace{40mm}

\begin{enumerate}[(I)]
\item \textbf{Fixation Features:}
\begin{enumerate}
\item \textbf{total Fixation Duration (tFD):} the total of all fixation duration in millisecond on a word.
\item \textbf{First Fixation Duration (FFD):} the duration (in ms) of fixation during the first pass reading of a word.
\item \textbf{Second Fixation Duration (SFD):} the duration (in ms) of fixation during the second pass reading of a word.
\item \textbf{Last Fixation Duration (LFD):} the duration (in ms) of fixation during the third or more pass reading of a word.
\item \textbf{average Fixation Duration (aFD):} the average (in ms) of fixation duration on a word.
\item \textbf{total Fixation Count (tFC):} the total count (in number) of all fixations during the multiple pass reading of a word.
\item \textbf{average Fixation Count (aFC):} the average (in number) of all fixation counts during the multiple pass reading of a word.
\end{enumerate}

\item \textbf{Saccade Features:} 
\begin{enumerate}
\item \textbf{total Saccade Duration (SD):} the total of all saccade duration in millisecond in an area of interest (AoI).
\item \textbf{total Saccade Count (SC):} the total count (in number) of all saccade in an AoI.
\item \textbf{average Saccade Velocity (SV):} the average of saccade velocity in an AoI.
\item \textbf{average Saccade peak Velocity (SpV):} the average of saccade peak velocity in an AoI.
\item \textbf{total word read in a Saccade (rS):} the number of words that are viewed during a saccade in an AoI.
\item \textbf{total word skipped in a Saccade (sS):} the number of words that are skipped during a saccade in an AoI.
\item \textbf{average Saccade Amplitude (SA):} the average of saccade amplitude in an AoI.
\end{enumerate}

\item \textbf{Regression Features:}
\begin{enumerate}
\item \textbf{total Regression Duration (RD):} the total of all regression duration in millisecond in an AoI.
\item \textbf{total Regression Count (RC):} the total count (in number) of all regression in an AoI.
\item \textbf{average Regression Velocity (RV):} the average of regression velocity in an AoI.
\item \textbf{average Regression peak Velocity (RpV):} the average of regression peak velocity in an AoI.
\item \textbf{total word read in Regression count (rR):} the number of words that are viewed during a regression in an AoI.
\item \textbf{total word skipped in Regression count (sR):} the number of words that are skipped during a regression in an AoI.
\item \textbf{average Regression Amplitude (RA):} the average of regression amplitude in an AoI.
\item \textbf{ratio of Saccade and Regression counts (rSR):} the ratio of total-saccade-count and total-regression-count during the multiple pass reading of an AoI.
\end{enumerate}
\end{enumerate}

\noindent\textbf{Gaze features in different levels of textual areas of interest:} The statistical characteristics of above features were calculated at different levels of textual areas (Area of Interests) including word, sub-sentence, sentence, paragraph, slide and the whole-text (all slides). The total possible number of gaze features in each AoI extracted from sessions 1 and 2 gaze data are shown in table \ref{tab:Chapter-2tab4}. The interaction of gaze with words produce only fixation features, whereas gaze with other areas (sub-sentence, sentence, paragraph, slide and whole-text) provide all three types of gaze-feature patterns (fixation, saccade, and regression). 

\begin{table}[!h]
	\captionsetup{font={small}, labelfont=bf, skip=4pt}
	\caption{Possible gaze-text interaction in different levels of AoI} 
	\raggedright 
	\small	
	\scalebox{1}{
	\renewcommand{\arraystretch}{1.4}
	\begin{tabular}{L{2.6cm} C{1.8cm} C{1.8cm} C{1.8cm} C{2.3cm} C{1.8cm}}
		\hline 
		\textbf{AoI type} & \textbf{Total AoI} & \textbf{Fixation features (\#7)} & \textbf{Saccade features (\#7)} & \textbf{Regression features (\#8)}  & \textbf{Total-features}\\[0.5ex]
		\hline 
		\multicolumn{6}{l}{\textbf{Session-1 gaze features}} \\
		\hline 
		Word & 718 & 5026 & - & - & \textbf{5026} \\
		Sub-sentence & 69 & 483 & 483 & 552 & \textbf{1518} \\
		Sentence & 30 & 210 & 210 & 240 & \textbf{660} \\
		Paragraph & 7 & 49 & 49 & 56 & \textbf{154} \\
		Slide & 5 & 35 & 35 & 40 & \textbf{110} \\
		Whole-text & 1 & 7 & 7 & 8 & \textbf{22} \\[0.5ex]
		\hline 

		\textbf{Total-gaze-features} &  & \textbf{5810} & \textbf{784} & \textbf{896} & \textbf{7490} \\[0.5ex]
		\hline 
		\multicolumn{6}{l}{\textbf{Session-2 gaze features}} \\
		\hline 
		Word & 662 & 4634 & - & - & \textbf{4634} \\
		Sub-sentence & 66 & 462 & 462 & 528 & \textbf{1452} \\
		Sentence & 35 & 245 & 245 & 280 & \textbf{770} \\
		Paragraph & 6 & 42 & 42 & 48 & \textbf{132} \\
		Slide & 5 & 35 & 35 & 40 & \textbf{110} \\
		Whole-text & 1 & 7 & 7 & 8 & \textbf{22} \\
		\hline 

		\textbf{Total-gaze-features} &  & \textbf{5425} & \textbf{791} & \textbf{904} & \textbf{7120} \\[0.5ex]
		\hline 
	\end{tabular}}
	\label{tab:Chapter-2tab4}
\end{table}

\subsection{Feature Normalisation}

For applying machine learning techniques, it is required that the data should be normally distributed. Generally, most features of a dataset differ from each other in terms of the range of values, so the data is normalised using a normalisation method, such that the differences in the range of values do not affect the accuracy of classifiers. Normalization also improves the numerical stability of the model and gives equal considerations for each feature. There are several normalization methods proposed such as Z-score (zero mean and unit variance) normalization, min-max normalization, unit vector normalization etc. We applied the Z-score normalization method to normalise the gaze dataset.

\subsection{Feature Selection}

As shown in table \ref{tab:Chapter-2tab4}, thousands of features (gaze-text interaction) were extracted from the gaze dataset to characterize the eye movements of the participants. Too many features relative to the number of observations cause overfit problem. The overfitting in training phase should be avoided because the classifier adapts to the concrete set of inputs, and this adaptation can produce good classification results for this particular set, but can negatively affect the generalization capacity of the classifier. Since, most discriminating features provide better results, therefore the feature selection becomes an important step for improving classification result.

We employed the Welch's t-test technique \cite{welch1947generalization, delacre2017psychologists} to select discriminating features from all (possible gaze-text interaction) features obtained from the gaze data. This approach is based on the concept that whether the mean of two sample groups (here, high and low) of the population are similar or not. This test can be used to detect the significant differences between the features of high-labelled and low-labelled gaze data. Only the features with p-value lower than 0.01 were selected as statistically significant features for classification.

We also used Mutual Information (MI) and Chi-Square feature selection methods for comparative analysis against to the Welch's t-test method. The comparative results are presented in section- \ref{ssec:Chapter2ssec6.3}.\\
\textbf{Significant gaze features on text AoI:} The tables \ref{tab:Chapter-2tab5} and \ref{tab:Chapter-2tab6} show gaze features, having statistically significant differences (\textit{$p-value < 0.01$}), of gaze-patterns i.e., fixation, saccade, and regression; all were collected in different AoIs (words, sub-sentence, sentence, paragraph, slide, and the whole-text) of sessions 1 and 2 respectively. Both tables of the sessions report variations in the number of significant features across the trial days. Most of the features extracted from the gaze data were not found statistically significant; because several words (e.g., stop words) hardly got any fixation from most of the participants. Therefore, the feature vectors of such words carry insignificant values. Both tables show the count of statistically significant features under two broad groups-- AoI and gaze-pattern. In both tables, within AoI, `word' has the most number of statistically significant features; similarly, within gaze-pattern, `fixation' has the most number of statistically significant features. Both tables also report that, in gaze-pattern group, `regression' has more number of statistically significant features than of `saccade'; which shows the importance of the former feature over the latter one. 

In table \ref{tab:Chapter-2tab5}, two AoIs- `whole-text' and `slide' do not contain any statistically significant features for days - 1 \& 3, and 3 respectively. The sixth column of both tables contains the number of statistically significant features, which were extracted from all three days gaze data i.e., they were calculated from 90 samples (30 participants $\times$ 3 days) of the corresponding session. As seen, the column has the maximum number of features than those of three-days combined. Within both tables, the total number of features of both groups-- AoI and gaze-pattern are the same in all columns. Also, we had not found any relation between the number of statistically significant features and their corresponding days.

\begin{table}[!h]
	\captionsetup{font={small}, labelfont=bf, skip=4pt}
	\caption{Statistically significant gaze features of session-1 gaze data} 
	\centering 
	\small	
	\scalebox{0.95}{
	\renewcommand{\arraystretch}{1.4}
	\begin{tabular}{L{2.7cm} C{2.1cm} C{2.1cm} C{2.1cm} C{2.1cm} C{2.1cm}}
		\hline 
		\textbf{AoI group} & \textbf{\footnotesize All feature (from table \ref{tab:Chapter-2tab4})} & \textbf{Sig. feature (Day-1)} & \textbf{Sig. feature (Day-2)} & \textbf{Sig. feature (Day-3)}  & \textbf{Sig. feature (All days)}\\[0.5ex]
		\hline 
		
		Word & 5026 & 54 & 55 & 61 & 254 \\
		Sub-sentence & 1518 & 54 & 32 & 18 & 163 \\
		Sentence & 660 & 25 & 29 & 15 & 89 \\
		Paragraph & 154 & 6 & 13 & 14 & 37 \\
		Slide & 110 & 6 & 14 & 0 & 35 \\
		Whole-text & 22 & 0 & 6 & 0 & 11 \\
		\hline 
		Total feature in AoI & \textbf{7490} & \textbf{145} & \textbf{149} & \textbf{108} & \textbf{589} \\[0.5ex]
		\hline 
		
		\textbf{Gaze-pattern group} & \textbf{\footnotesize All feature (from table \ref{tab:Chapter-2tab4})} & \textbf{Sig. feature (Day-1)} & \textbf{Sig. feature (Day-2)} & \textbf{Sig. feature (Day-3)}  & \textbf{Sig. feature (All days)}\\[0.5ex]
		\hline 
		
		Fixation & 5810 & 91 & 85 & 73 & 145 \\
		Saccade & 784 & 15 & 20 & 12 & 22 \\
		Regression & 896 & 39 & 44 & 23 & 142 \\
		\hline 
		Total feature in Gaze-pattern & \textbf{7490} & \textbf{145} & \textbf{149} & \textbf{108} & \textbf{589} \\[0.5ex]
		\hline 
		
	\end{tabular}}
	\label{tab:Chapter-2tab5}
\end{table}

\begin{table}[!h]
	\captionsetup{font={small}, labelfont=bf, skip=4pt}
	\caption{Statistically significant gaze features of session-2 gaze data} 
	\centering 
	\small	
	\smallskip\noindent
	\scalebox{0.95}{
		\renewcommand{\arraystretch}{1.4}
		\begin{tabular}{L{2.7cm} C{2.1cm} C{2.cm} C{2.cm} C{2.cm} C{2.cm}}
			\hline 
			\textbf{AoI group} & \textbf{\footnotesize All feature (from table \ref{tab:Chapter-2tab4})} & \textbf{Sig. feature (Day-1)} & \textbf{Sig. feature (Day-2)} & \textbf{Sig. feature (Day-3)}  & \textbf{Sig. feature (All days)}\\[0.5ex]
			\hline 
			
			Word & 4634 & 59 & 89 & 120 & 500 \\
			Sub-sentence & 1452 & 24 & 58 & 66 & 255 \\
			Sentence & 770 & 26 & 35 & 32 & 176 \\
			Paragraph & 132 & 13 & 23 & 16 & 56 \\
			Slide & 110 & 5 & 18 & 15 & 48 \\
			Whole-text & 22 & 6 & 7 & 7 & 11 \\
			\hline 
			Total feature in AoI & \textbf{7120} & \textbf{133} & \textbf{230} & \textbf{256} & \textbf{1046} \\[0.5ex]
			\hline 
			
			\textbf{Gaze-pattern group} & \textbf{\footnotesize All feature (from table \ref{tab:Chapter-2tab4})} & \textbf{Sig. feature (Day-1)} & \textbf{Sig. feature (Day-2)} & \textbf{Sig. feature (Day-3)}  & \textbf{Sig. feature (All days)}\\[0.5ex]
			\hline 
			
			Fixation & 5425 & 75 & 137 & 174 & 809 \\
			Saccade & 791 & 12 & 13 & 14 & 32 \\
			Regression & 904 & 46 & 80 & 68 & 205 \\
			\hline 
			Total feature in Gaze-pattern & \textbf{7120} & \textbf{133} & \textbf{230} & \textbf{256} & \textbf{1046} \\[0.5ex]
			\hline 
			
		\end{tabular}}
		\label{tab:Chapter-2tab6}
	\end{table}
\subsection{Automatic Classification}
In order to make an automatic classification of the gaze dataset, the Scikit-learn machine learning library (Version 0.20.4, 2019) \cite{pedregosa2011scikit} for the Python programming language had been used. This library permits a collection of machine learning algorithms to be accessed for data mining tasks. Three different classifiers were used to compare their performance on the gaze dataset: Decision Tree (DT), MultiLayer Perceptron (MLP) and Support Vector Machine (SVM). The configuration of the classifiers and the regression are given here. For DT classifier, the quality of a split was set to Gini impurity and the strategy used to choose the split at each node = best. For MLP, the learning rate = 0.001, Number of epochs = 200, Activation function for the hidden layer = relu, and hidden layer size was set to 100. For SVM, the C parameter was set to 1.0 using RBF kernel and the tolerance parameter was set to 0.001.

In the experiment, the data set was randomly divided into two sets: 70\% as the training set and 30\% as the test set. In addition, the 5-fold cross-validation technique was also used to create another separate training and validation set. To analyse the performance of the classification, we used classification accuracy (C. rate), Unweighted Average Recall (UAR) and mean cross-validation (CV) accuracy. The classification accuracy is the number of correct predictions made divided by the total number of predictions made. The unweighted average recall is the mean of sensitivity (recall of positive instances) and specificity (recall of negative instances). UAR was chosen because it equally weights each class regardless of its number of samples, so it represents more precisely the accuracy of a classification test using unbalanced data. Since our dataset had a small size, cross-validation became a good alternative for measuring the classifiers' performance.


\section{Results and Discussion}

\subsection{Classification Results on All Extracted Features}

Tables \ref{tab:Chapter-2tab7} and \ref{tab:Chapter-2tab8} show the classification results in the task of identifying the label (High and Low) of participants using all normalised gaze features (i.e. all possible gaze-text interactions) extracted from the gaze data collected during sessions 1 and 2 respectively. The purpose of reporting these values is to represent them as the classifiers' baseline accuracy and both tables are compared with final accuracy reported in tables \ref{tab:Chapter-2tab9} and \ref{tab:Chapter-2tab10} respectively.

The tables \ref{tab:Chapter-2tab7} and \ref{tab:Chapter-2tab8} report that, the outputs of all three measures of the accuracy of three classifiers on both feature groups (AoI and gaze-pattern) vary across days (1, 2, 3 and all days).

In table \ref{tab:Chapter-2tab7}, the average accuracy for SVM, DT, and MLP are [0.60, 0.61, 0.62], [0.58, 0.57, 0.57], and [0.65, 0.63, 0.62] respectively in the sequence of mean CV accuracy, UAR and C. rate. Here, MLP shows comparatively slightly better average-performance in all three accuracy measures.

Within the AoI group, the classifiers jointly show slightly better performance for `sentence' (which is 0.64); whereas, their joint accuracy is lowest for `whole-text' (0.51). Within the gaze-pattern group, classifiers joint accuracy shows slightly better performance over all days on `regression' and `fixation' (0.64); whereas their joint accuracy is comparatively lowest for `saccade' (0.57) among the group. The day column reports that, the classifiers jointly shows better performance for day `2' (0.62) and shows worst performance for day `3' (0.44). 
In terms of accuracy measures, across classifiers, the minimum CV accuracy is 0.33, given seven times mostly by SVM in day `3'; whereas, the maximum CV accuracy is 0.89, given two times by DT and MLP in day `2'. The minimum UAR and C. rate across classifiers are 0.17, given two times by DT and MLP in day `3'. The maximum UAR across classifiers is 0.88, given nine times by mostly DT and MLP in days `1' \& `2'. The maximum C. rate across classifiers is 0.89, given four times by mostly DT in day `1'.

\begingroup
\setstretch{1.2}
{\small	
	\setlength{\tabcolsep}{3pt}
	\renewcommand{\arraystretch}{1.0}
	\begin{longtable}
		{>{\raggedright}m{2cm} C{0.8cm} C{1.5cm} C{0.8cm} C{0.8cm} C{1.5cm} C{0.8cm} C{1.cm} C{1.5cm} C{0.8cm} C{0.8cm}}
		\captionsetup{font={small}, labelfont=bf, skip=4pt}
		\caption{Classification results on all possible gaze-text interaction features of session-1}
		\label{tab:Chapter-2tab7}\\

		\toprule
		\midrule
		\textbf{Feature set} & \textbf{Day} & \multicolumn{3}{c}{\textbf{SVM}} & \multicolumn{3}{c}{\textbf{DT}} & \multicolumn{3}{c}{\textbf{MLP}} \\
		\midrule
		& & \textbf{\scriptsize CV accuracy} & \textbf{\scriptsize UAR} & \textbf{\scriptsize C. rate} & \textbf{\scriptsize CV accuracy} & \textbf{\scriptsize UAR} & \textbf{\scriptsize C. rate} & \textbf{\scriptsize CV accuracy} & \textbf{\scriptsize UAR} & \textbf{\scriptsize C. rate}\\
		\midrule
		
		\multirow{4}{1.8cm}{\raggedright \textbf{Word (5026)}} & 1 & 0.62 & 0.5 & 0.56 & 0.44 & 0.65 & 0.67 & 0.62 & 0.6 & 0.56 \\
		&  2 & 0.6 & 0.67 & 0.71 & 0.49 & 0.54 & 0.57 & 0.69 & 0.67 & 0.71 \\
		&  3 & 0.33 & 0.5 & 0.5 & 0.57 & 0.33 & 0.33 & 0.67 & 0.5 & 0.5 \\
		&  All  & 0.67 & 0.65 & 0.65 & 0.51 & 0.45 & 0.45 & 0.79 & 0.75 & 0.75 \\
		\midrule
		\multirow{4}{2.cm}{\raggedright \textbf{Sub-sentence (1518)}} & 1 & 0.58 & 0.5 & 0.56 & 0.64 & 0.65 & 0.67 & 0.6 & 0.45 & 0.44 \\
		&  2 & 0.86 & 0.8 & 0.78 & 0.43 & 0.51 & 0.51 & 0.77 & 0.88 & 0.86 \\
		&  3 & 0.33 & 0.5 & 0.5 & 0.53 & 0.33 & 0.33 & 0.53 & 0.5 & 0.5 \\
		&  All & 0.74 & 0.65 & 0.65 & 0.64 & 0.55 & 0.55 & 0.72 & 0.75 & 0.75 \\
		\midrule
		\multirow{4}{1.8cm}{\raggedright \textbf{Sentence (660)}} & 1 & 0.56 & 0.62 & 0.67 & 0.62 & 0.88 & 0.89 & 0.64 & 0.68 & 0.67 \\
		&  2 & 0.83 & 0.81 & 0.8 & 0.63 & 0.58 & 0.58 & 0.8 & 0.75 & 0.71 \\
		&  3 & 0.33 & 0.5 & 0.5 & 0.67 & 0.5 & 0.5 & 0.37 & 0.67 & 0.67 \\
		&  All & 0.74 & 0.65 & 0.65 & 0.55 & 0.5 & 0.5 & 0.75 & 0.6 & 0.6 \\
		\midrule
		\multirow{4}{1.8cm}{\raggedright \textbf{Paragraph (154)}} & 1 & 0.53 & 0.5 & 0.56 & 0.51 & 0.45 & 0.44 & 0.53 & 0.68 & 0.67 \\
		&  2 & 0.86 & 0.88 & 0.87 & 0.66 & 0.58 & 0.57 & 0.8 & 0.46 & 0.43 \\
		&  3 & 0.47 & 0.5 & 0.5 & 0.5 & 0.67 & 0.67 & 0.7 & 0.67 & 0.67 \\
		&  All & 0.63 & 0.65 & 0.65 & 0.71 & 0.65 & 0.65 & 0.72 & 0.65 & 0.65 \\
		\midrule
		\multirow{4}{1.8cm}{\raggedright \textbf{Slide (110)}} & 1 & 0.51 & 0.4 & 0.44 & 0.49 & 0.55 & 0.56 & 0.67 & 0.2 & 0.22 \\
		&  2 & 0.8 & 0.8 & 0.79 & 0.77 & 0.61 & 0.61 & 0.61 & 0.65 & 0.61 \\
		&  3 & 0.43 & 0.5 & 0.5 & 0.53 & 0.17 & 0.17 & 0.6 & 0.67 & 0.67 \\
		&  All & 0.63 & 0.65 & 0.65 & 0.69 & 0.7 & 0.7 & 0.69 & 0.65 & 0.65 \\
		\midrule
		\multirow{4}{1.8cm}{\raggedright \textbf{Whole-text (22)}} & 1 & 0.58 & 0.53 & 0.56 & 0.58 & 0.42 & 0.44 & 0.53 & 0.57 & 0.56 \\
		&  2 & 0.8 & 0.8 & 0.8 & 0.6 & 0.51 & 0.53 & 0.66 & 0.25 & 0.29 \\
		&  3 & 0.53 & 0.5 & 0.5 & 0.47 & 0.33 & 0.33 & 0.33 & 0.17 & 0.17 \\
		&  All & 0.63 & 0.55 & 0.55 & 0.56 & 0.65 & 0.65 & 0.5 & 0.45 & 0.45 \\
		\midrule
		\multirow{4}{1.8cm}{\raggedright \textbf{Fixation (5810)}} & 1 & 0.62 & 0.5 & 0.56 & 0.64 & 0.88 & 0.89 & 0.67 & 0.68 & 0.67 \\
		&  2 & 0.66 & 0.67 & 0.71 & 0.89 & 0.88 & 0.86 & 0.69 & 0.68 & 0.7 \\
		&  3 & 0.33 & 0.5 & 0.5 & 0.7 & 0.33 & 0.33 & 0.7 & 0.67 & 0.67 \\
		&  All & 0.68 & 0.65 & 0.65 & 0.7 & 0.45 & 0.45 & 0.74 & 0.65 & 0.65 \\
		\midrule
		\multirow{4}{1.8cm}{\raggedright \textbf{Saccade (784)}} & 1 & 0.53 & 0.4 & 0.44 & 0.62 & 0.78 & 0.78 & 0.49 & 0.75 & 0.78 \\
		&  2 & 0.6 & 0.67 & 0.71 & 0.46 & 0.67 & 0.71 & 0.66 & 0.75 & 0.71 \\
		&  3 & 0.4 & 0.5 & 0.5 & 0.4 & 0.4 & 0.4 & 0.43 & 0.4 & 0.4 \\
		&  All & 0.6 & 0.75 & 0.75 & 0.45 & 0.6 & 0.6 & 0.6 & 0.5 & 0.5 \\
		\midrule
		\multirow{4}{1.8cm}{\raggedright \textbf{Regression (896)}} & 1 & 0.6 & 0.75 & 0.78 & 0.53 & 0.35 & 0.33 & 0.73 & 0.65 & 0.67 \\
		&  2 & 0.8 & 0.81 & 0.8 & 0.51 & 0.58 & 0.57 & 0.89 & 0.88 & 0.86 \\
		&  3 & 0.33 & 0.33 & 0.33 & 0.5 & 0.67 & 0.67 & 0.53 & 0.5 & 0.5 \\
		&  All & 0.75 & 0.7 & 0.7 & 0.56 & 0.8 & 0.8 & 0.81 & 0.75 & 0.75  \\
		\midrule
		\multirow{4}{1.8cm}{\raggedright \textbf{All features (7490)}} & 1 & 0.6 & 0.5 & 0.56 & 0.71 & 0.88 & 0.89 & 0.57 & 0.88 & 0.89 \\
		&  2 & 0.71 & 0.83 & 0.86 & 0.86 & 0.58 & 0.57 & 0.77 & 0.88 & 0.86 \\
		&  3 & 0.33 & 0.5 & 0.5 & 0.4 & 0.5 & 0.5 & 0.67 & 0.83 & 0.83 \\
		&  All & 0.72 & 0.65 & 0.65 & 0.67 & 0.65 & 0.65 & 0.78 & 0.7 & 0.7 \\
		\bottomrule
				
\end{longtable}
}\endgroup

In table \ref{tab:Chapter-2tab8}, the average accuracy for SVM, DT, and MLP are [0.57, 0.70, 0.65], [0.54, 0.57, 0.56], and [0.63, 0.66, 0.65] respectively in the sequence of mean CV accuracy, UAR and C. rate. Here, in contrast to DT, SVM and MLP show better average performance in all three accuracy measures. 

\begingroup
\setstretch{1.3}
{\small	
	\setlength{\tabcolsep}{3pt}
	\renewcommand{\arraystretch}{1.0}
	\begin{longtable}
		{>{\raggedright}m{2.cm} C{0.8cm} C{1.5cm} C{0.8cm} C{0.8cm} C{1.5cm} C{0.8cm} C{1.cm} C{1.5cm} C{0.8cm} C{0.8cm}}
		\captionsetup{font={small}, labelfont=bf, skip=4pt}
		\caption{Classification results on all possible gaze-text interaction features of session-2}
		\label{tab:Chapter-2tab8}\\
		\toprule
		\midrule
		\textbf{Feature set} & \textbf{Day} & \multicolumn{3}{c}{\textbf{SVM}} & \multicolumn{3}{c}{\textbf{DT}} & \multicolumn{3}{c}{\textbf{MLP}} \\
		\midrule
		& & \textbf{\scriptsize CV accuracy} & \textbf{\scriptsize UAR} & \textbf{\scriptsize C. rate} & \textbf{\scriptsize CV accuracy} & \textbf{\scriptsize UAR} & \textbf{\scriptsize C. rate} & \textbf{\scriptsize CV accuracy} & \textbf{\scriptsize UAR} & \textbf{\scriptsize C. rate}\\
		\midrule
		
		\multirow{4}{1.8cm}{\raggedright \textbf{Word (4632)}} & 1 & 0.53 & 0.58 & 0.38 & 0.5 & 0.58 & 0.62 & 0.47 & 0.92 & 0.88  \\ 
		&  2 & 0.3 & 0.8 & 0.75 & 0.55 & 0.37 & 0.38 & 0.7 & 0.73 & 0.75  \\ 
		&  3 & 0.6 & 0.5 & 0.5 & 0.57 & 0.5 & 0.5 & 0.77 & 0.67 & 0.67  \\ 
		&  All  & 0.65 & 0.69 & 0.62 & 0.51 & 0.5 & 0.52 & 0.62 & 0.71 & 0.67 \\ 
		\midrule
		\multirow{4}{2.cm}{\raggedright \textbf{Sub-sentence (1452)}} & 1 & 0.53 & 0.67 & 0.5 & 0.5 & 0.58 & 0.62 & 0.58 & 0.52 & 0.58  \\ 
		&  2 & 0.4 & 0.8 & 0.75 & 0.4 & 0.63 & 0.62 & 0.7 & 0.8 & 0.88  \\ 
		&  3 & 0.6 & 0.83 & 0.83 & 0.8 & 0.67 & 0.67 & 0.83 & 0.81 & 0.82  \\ 
		&  All & 0.74 & 0.65 & 0.57 & 0.6 & 0.62 & 0.71 & 0.61 & 0.65 & 0.57 \\ 
		\midrule
		\multirow{4}{1.8cm}{\raggedright \textbf{Sentence (770)}} & 1 & 0.55 & 0.75 & 0.62 & 0.6 & 0.58 & 0.62 & 0.68 & 0.72 & 0.78  \\ 
		&  2 & 0.4 & 0.7 & 0.62 & 0.5 & 0.53 & 0.5 & 0.6 & 0.63 & 0.68  \\ 
		&  3 & 0.57 & 0.83 & 0.83 & 0.67 & 0.71 & 0.73 & 0.73 & 0.72 & 0.72  \\ 
		&  All & 0.69 & 0.69 & 0.62 & 0.52 & 0.49 & 0.43 & 0.62 & 0.77 & 0.71 \\ 
		\midrule
		\multirow{4}{1.8cm}{\raggedright \textbf{Paragraph (132)}} & 1 & 0.6 & 0.83 & 0.75 & 0.61 & 0.71 & 0.65 & 0.58 & 0.73 & 0.61  \\ 
		&  2 & 0.5 & 0.8 & 0.75 & 0.6 & 0.65 & 0.68 & 0.72 & 0.73 & 0.75  \\ 
		&  3 & 0.67 & 0.6 & 0.6 & 0.74 & 0.73 & 0.73 & 0.77 & 0.67 & 0.67  \\ 
		&  All & 0.68 & 0.73 & 0.67 & 0.63 & 0.68 & 0.67 & 0.68 & 0.68 & 0.67 \\ 
		\midrule
		\multirow{4}{1.8cm}{\raggedright \textbf{Slide (110)}} & 1 & 0.53 & 0.83 & 0.75 & 0.45 & 0.83 & 0.75 & 0.72 & 0.72 & 0.78  \\ 
		&  2 & 0.5 & 0.8 & 0.75 & 0.57 & 0.53 & 0.5 & 0.68 & 0.37 & 0.38  \\ 
		&  3 & 0.73 & 0.67 & 0.67 & 0.57 & 0.83 & 0.83 & 0.83 & 0.83 & 0.83  \\ 
		&  All & 0.67 & 0.65 & 0.57 & 0.49 & 0.43 & 0.47 & 0.45 & 0.45 & 0.48 \\ 
		\midrule
		\multirow{4}{1.8cm}{\raggedright \textbf{Whole-text (22)}} & 1 & 0.55 & 0.67 & 0.5 & 0.17 & 0.17 & 0.25 & 0.62 & 0.75 & 0.62  \\ 
		&  2 & 0.5 & 0.5 & 0.55 & 0.3 & 0.31 & 0.32 & 0.42 & 0.4 & 0.42  \\ 
		&  3 & 0.77 & 0.67 & 0.67 & 0.63 & 0.83 & 0.83 & 0.33 & 0.67 & 0.67  \\ 
		&  All & 0.68 & 0.73 & 0.67 & 0.66 & 0.61 & 0.57 & 0.68 & 0.5 & 0.38 \\ 
		\midrule
		\multirow{4}{1.8cm}{\raggedright \textbf{Fixation (5425)}} & 1 & 0.55 & 0.58 & 0.38 & 0.47 & 0.67 & 0.75 & 0.55 & 0.67 & 0.5  \\ 
		&  2 & 0.33 & 0.8 & 0.75 & 0.47 & 0.5 & 0.38 & 0.62 & 0.63 & 0.62  \\ 
		&  3 & 0.63 & 0.83 & 0.83 & 0.5 & 0.5 & 0.5 & 0.87 & 0.83 & 0.83  \\ 
		&  All & 0.68 & 0.73 & 0.67 & 0.57 & 0.53 & 0.48 & 0.65 & 0.59 & 0.52 \\ 
		\midrule
		\multirow{4}{1.8cm}{\raggedright \textbf{Saccade (791)}} & 1 & 0.53 & 0.58 & 0.62 & 0.47 & 0.42 & 0.38 & 0.5 & 0.5 & 0.5  \\ 
		&  2 & 0.35 & 0.73 & 0.75 & 0.47 & 0.47 & 0.5 & 0.57 & 0.73 & 0.75  \\ 
		&  3 & 0.37 & 0.5 & 0.5 & 0.43 & 0.3 & 0.3 & 0.43 & 0.33 & 0.33  \\ 
		&  All & 0.56 & 0.5 & 0.38 & 0.61 & 0.46 & 0.48 & 0.57 & 0.52 & 0.52 \\ 
		\midrule
		\multirow{4}{1.8cm}{\raggedright \textbf{Regression (904)}} & 1 & 0.55 & 0.83 & 0.75 & 0.65 & 0.42 & 0.38 & 0.55 & 0.72 & 0.73  \\ 
		&  2 & 0.5 & 0.8 & 0.75 & 0.57 & 0.53 & 0.5 & 0.72 & 0.7 & 0.71  \\ 
		&  3 & 0.73 & 0.7 & 0.7 & 0.57 & 0.53 & 0.53 & 0.53 & 0.53 & 0.53  \\ 
		&  All & 0.8 & 0.73 & 0.67 & 0.59 & 0.71 & 0.67 & 0.62 & 0.75 & 0.71  \\ 
		\midrule
		\multirow{4}{1.8cm}{\raggedright \textbf{All features (7120)}} & 1 & 0.53 & 0.58 & 0.38 & 0.45 & 0.67 & 0.5 & 0.7 & 0.7 & 0.7  \\ 
		&  2 & 0.33 & 0.8 & 0.75 & 0.45 & 0.63 & 0.62 & 0.6 & 0.63 & 0.62  \\ 
		&  3 & 0.63 & 0.83 & 0.83 & 0.67 & 0.63 & 0.63 & 0.73 & 0.73 & 0.73  \\ 
		&  All & 0.69 & 0.69 & 0.62 & 0.57 & 0.58 & 0.57 & 0.72 & 0.78 & 0.76 \\
		\bottomrule
		
\end{longtable}			
}\endgroup

The joint accuracy of three classifiers shows better performance for `paragraph' (0.68) and shows lowest performance for `saccade' (0.5). The day column reports that, the classifiers jointly shows better performance for day `3' (0.59) and worst performance for day `2' (0.53).

In terms of accuracy measures, the minimum CV accuracy and UAR across classifiers are 0.17, given two times by DT in day `1'. The maximum CV accuracy across classifiers is 0.87, given once by MLP in day `3'. The maximum UAR is 0.83, given twelve times by all three classifiers across days. The minimum C. rate is 0.25, given once by DT in day `1'. The maximum C. rate is 0.88, given two times by MLP in days `1' \& `2'.
\subsection{Classification Results on Statistically Significant Features}
\label{ssec:Chapter2ssec6.2}
Tables \ref{tab:Chapter-2tab9} and \ref{tab:Chapter-2tab10} show three classifications' final results to predict the given label (High or Low) of the participant using only the normalised features having statistically significant differences (between High and Low groups) of the gaze data of sessions 1 and 2 respectively.

As similar to last two tables, tables \ref{tab:Chapter-2tab9} and \ref{tab:Chapter-2tab10} report that the three measures of the accuracy of the classifiers on both feature groups (AoI and gaze-pattern) vary across days (1, 2, 3 and all days).

In table \ref{tab:Chapter-2tab9}, the average accuracy for SVM, DT, and MLP are [0.79, 0.84, 0.84], [0.65, 0.65, 0.65], and [0.77, 0.82, 0.82] respectively in the sequence of mean CV accuracy, UAR and C. rate. Among the three classifiers, SVM shows best average performance in the accuracy measures.

The joint accuracy of three classifiers shows better performance for `fixation' (which is 0.9) and shows the lowest performance for `saccade' (0.72). The day column reports that, the classifiers jointly shows better performance for day `2' (0.77) and shows the lowest performance for day `3' (0.6). 

In terms of accuracy measures, the minimum mean CV accuracy across classifiers is 0.56, given once by DT in day `1'. \textbf{The maximum CV accuracy across classifiers is 1.0, given five times by SVM in days- `2' \& `3'.} The minimum UAR is 0.33, given once by DT in day `3'. \textbf{The maximum UAR is 1.0, given thirty-one times mostly by SVM across days.} The minimum C. rate is 0.67, given nine times by all classifiers in days `1' \& `3'. \textbf{The maximum C. rate is 1.0, given thirty-nine times by all classifiers across days.}

\textit{As compared to the baseline accuracy table \ref{tab:Chapter-2tab7}, all classifiers in the present table perform better accuracy and among them, SVM outputs are the best. Also, by comparing both tables we observe that the joint accuracy of classifiers in the latter table shows a minimum performance improvement of 23.4\% for `regression'; whereas a maximum improvement of 50\% for `word' than the accuracies reported in the former table.}

\begingroup
\setstretch{1.3}
{\small	
	\setlength{\tabcolsep}{2pt}
	\renewcommand{\arraystretch}{1.0}
	\begin{longtable}
		{A{2.cm} C{0.5cm} C{1.5cm} C{1.5cm} C{0.8cm} C{0.8cm} C{1.5cm} C{0.8cm} C{1.cm} C{1.5cm} C{0.8cm} C{0.8cm}}
		\captionsetup{font={small}, labelfont=bf, skip=4pt}
		\caption{Classification results on significant gaze features of session-1}
		\label{tab:Chapter-2tab9}\\
		\toprule
		\midrule
		\textbf{\scriptsize Feature set} & \textbf{\scriptsize Day} & \textbf{\scriptsize \# Sig. feat. (from table \ref{tab:Chapter-2tab5})} & \multicolumn{3}{c}{\textbf{SVM}} & \multicolumn{3}{c}{\textbf{DT}} & \multicolumn{3}{c}{\textbf{MLP}} \\
		\midrule
		& & & \textbf{\scriptsize CV accuracy} & \textbf{\scriptsize UAR} & \textbf{\scriptsize C. rate} & \textbf{\scriptsize CV accuracy} & \textbf{\scriptsize UAR} & \textbf{\scriptsize C. rate} & \textbf{\scriptsize CV accuracy} & \textbf{\scriptsize UAR} & \textbf{\scriptsize C. rate}\\
		\midrule
		
		\multirow{4}{1.8cm}{\raggedright \textbf{\footnotesize Word (5026)}} & 1 & 54 & 0.98 & 1.0 & 1.0 & 0.71 & 0.68 & 0.67 & 0.98 & 0.9 & 0.89 \\ 
		& 2 & 55 & 1.0 & 1.0 & 1.0 & 0.74 & 0.67 & 0.71 & 1.0 & 1.0 & 1.0 \\ 
		& 3 & 61 & 1.0 & 1.0 & 1.0 & 0.6 & 0.67 & 0.67 & 0.93 & 1.0 & 1.0 \\ 
		& All  & 254 & 0.96 & 0.95 & 0.95 & 0.71 & 0.55 & 0.55 & 0.92 & 0.9 & 0.9 \\ 
		\midrule
		\multirow{4}{2.5cm}{\raggedright \textbf{\footnotesize Sub-sentence (1518)}} & 1 & 54 & 0.91 & 1.0 & 1.0 & 0.87 & 0.65 & 0.67 & 0.91 & 0.9 & 0.89 \\ 
		& 2 & 32 & 0.94 & 1.0 & 1.0 & 0.6 & 0.54 & 0.57 & 0.94 & 1.0 & 1.0 \\ 
		& 3 & 18 & 0.77 & 1.0 & 1.0 & 0.63 & 0.83 & 0.83 & 0.63 & 1.0 & 1.0 \\ 
		& All & 163 & 0.86 & 0.95 & 0.95 & 0.74 & 0.7 & 0.7 & 0.83 & 0.95 & 0.95 \\ 
		\midrule
		\multirow{4}{1.8cm}{\raggedright \textbf{\footnotesize Sentence (660)}} & 1 & 25 & 0.8 & 0.78 & 0.78 & 0.78 & 1.0 & 1.0 & 0.89 & 0.78 & 0.78 \\ 
		& 2 & 29 & 0.91 & 1.0 & 1.0 & 0.63 & 0.58 & 0.57 & 0.86 & 1.0 & 1.0 \\ 
		& 3 & 15 & 0.77 & 0.83 & 0.83 & 0.77 & 0.83 & 0.83 & 0.8 & 1.0 & 1.0 \\ 
		& All & 89 & 0.81 & 0.85 & 0.85 & 0.59 & 0.55 & 0.55 & 0.72 & 0.8 & 0.8 \\ 
		\midrule
		\multirow{4}{1.8cm}{\raggedright \textbf{\footnotesize Paragraph (154)}} & 1 & 6 & 0.76 & 0.9 & 0.89 & 0.58 & 0.78 & 0.78 & 0.67 & 0.78 & 0.78 \\ 
		& 2 & 13 & 0.91 & 1.0 & 1.0 & 0.71 & 0.58 & 0.57 & 0.91 & 1.0 & 1.0 \\ 
		& 3 & 14 & 0.77 & 1.0 & 1.0 & 0.8 & 1.0 & 1.0 & 0.83 & 0.83 & 0.83 \\ 
		& All & 37 & 0.71 & 0.7 & 0.7 & 0.71 & 0.7 & 0.7 & 0.69 & 0.75 & 0.75 \\ 
		\midrule
		\multirow{4}{1.8cm}{\raggedright \textbf{\footnotesize Slide (110)}} & 1 & 6 & 0.78 & 0.75 & 0.78 & 0.56 & 0.88 & 0.89 & 0.73 & 0.88 & 0.89 \\ 
		& 2 & 14 & 0.94 & 0.88 & 0.86 & 0.97 & 0.58 & 0.57 & 0.94 & 0.88 & 0.86 \\ 
		& 3 & 0 & - & - & - & - & - & - & - & - & - \\ 
		& All & 35 & 0.72 & 0.75 & 0.75 & 0.63 & 0.85 & 0.85 & 0.72 & 0.6 & 0.6 \\ 
		\midrule
		\multirow{4}{1.8cm}{\raggedright \textbf{\footnotesize Whole-text (22)}} & 1 & 0 & - & - & - & - & - & - & - & - & - \\ 
		& 2 & 6 & 0.8 & 1.0 & 1.0 & 0.71 & 0.62 & 0.57 & 0.71 & 1.0 & 1.0 \\ 
		& 3 & 0 & - & - & - & - & - & - & - & - & - \\ 
		& All & 11 & 0.7 & 0.7 & 0.7 & 0.57 & 0.55 & 0.55 & 0.68 & 0.75 & 0.75 \\ 
		\midrule
		\multirow{4}{1.8cm}{\raggedright \textbf{\footnotesize Fixation (5810)}} & 1 & 91 & 0.93 & 1.0 & 1.0 & 0.87 & 0.75 & 0.78 & 0.89 & 1.0 & 1.0 \\ 
		& 2 & 85 & 1.0 & 1.0 & 1.0 & 0.83 & 0.88 & 0.86 & 0.94 & 0.88 & 0.86 \\ 
		& 3 & 73 & 1.0 & 1.0 & 1.0 & 0.7 & 0.83 & 0.83 & 0.93 & 1.0 & 1.0 \\ 
		& All & 145 & 0.94 & 1.0 & 1.0 & 0.73 & 0.5 & 0.5 & 0.92 & 1.0 & 1.0 \\ 
		\midrule
		\multirow{4}{1.8cm}{\raggedright \textbf{\footnotesize Saccade (784)}} & 1 & 15 & 0.84 & 0.65 & 0.67 & 0.78 & 0.65 & 0.67 & 0.82 & 0.65 & 0.67 \\ 
		& 2 & 20 & 0.77 & 0.88 & 0.86 & 0.57 & 1.0 & 1.0 & 0.89 & 1.0 & 1.0 \\ 
		& 3 & 12 & 0.77 & 0.83 & 0.83 & 0.67 & 0.33 & 0.33 & 0.77 & 0.67 & 0.67 \\ 
		& All & 22 & 0.69 & 0.7 & 0.7 & 0.59 & 0.5 & 0.5 & 0.59 & 0.65 & 0.65 \\ 
		\midrule
		\multirow{4}{1.8cm}{\raggedright \textbf{\footnotesize Regression (896)}} & 1 & 39 & 0.82 & 0.9 & 0.89 & 0.71 & 0.68 & 0.67 & 0.78 & 0.9 & 0.89 \\ 
		& 2 & 44 & 0.86 & 1.0 & 1.0 & 0.69 & 0.58 & 0.57 & 0.91 & 1.0 & 1.0 \\ 
		& 3 & 23 & 0.83 & 0.83 & 0.83 & 0.83 & 0.83 & 0.83 & 0.63 & 0.83 & 0.83 \\ 
		& All & 142 & 0.79 & 0.8 & 0.8 & 0.63 & 0.6 & 0.6 & 0.7 & 0.75 & 0.75 \\ 
		\midrule
		\multirow{4}{1.8cm}{\raggedright \textbf{\footnotesize All features (7490)}} & 1 & 145 & 0.91 & 1.0 & 1.0 & 0.78 & 1.0 & 1.0 & 0.96 & 0.88 & 0.89 \\ 
		& 2 & 149 & 1.0 & 1.0 & 1.0 & 0.8 & 0.58 & 0.57 & 1.0 & 1.0 & 1.0 \\ 
		& 3 & 108 & 0.97 & 1.0 & 1.0 & 0.67 & 0.67 & 0.67 & 0.93 & 1.0 & 1.0 \\ 
		& All & 589 & 0.91 & 0.95 & 0.95 & 0.69 & 0.75 & 0.75 & 0.92 & 1.0 & 1.0 \\
		\bottomrule
			
\end{longtable}
}\endgroup

In table \ref{tab:Chapter-2tab10}, the average accuracy for SVM, DT, and MLP are [0.81, 0.85, 0.81], [0.65, 0.68, 0.65], and [0.76, 0.80, 0.78] respectively in the sequence of mean CV accuracy, UAR and C. rate. In contrast to DT and MLP, SVM shows better average performance in the three accuracy measures.

The joint accuracy of all classifiers shows better performance for `fixation' (0.84) and the lowest performance for `whole-text' (0.65). The day column reports that the classifiers jointly show better performance for day `3' (0.72) and the lowest performance for `all' days (0.62).

In terms of accuracy measures, the minimum mean CV accuracy across classifiers is 0.50, given two times by DT in day `1'. \textbf{The maximum mean CV accuracy across classifiers is 1.0, given seven times mostly by SVM in days- `2' \& `3'.} The minimum UAR is 0.20, given once by DT in day `2'. \textbf{The maximum UAR is 1.0, given ten times mostly by SVM across days.} The minimum C. rate is 0.25, given once by DT in day `2'. \textbf{The maximum C. rate is 1.0, given ten times mostly by SVM across days.}

\textit{As compared to the baseline accuracy table \ref{tab:Chapter-2tab8}, all classifiers in the present table perform better accuracy and among them, SVM outputs are the best. Also, by comparing both tables we observe that the joint accuracy of classifiers in the latter table shows the minimum performance improvement of 10\% for `paragraph'; as well as the maximum improvement of 38\% for `fixation' than the accuracies reported in the former table.}

\begingroup
\setstretch{1.2}
{\small	
	\setlength{\tabcolsep}{2pt}
	\renewcommand{\arraystretch}{1.0}
	\begin{longtable}
		{A{2cm} C{0.5cm} C{1.5cm} C{1.5cm} C{0.8cm} C{0.8cm} C{1.5cm} C{0.8cm} C{1.cm} C{1.5cm} C{0.8cm} C{0.8cm}}
		\captionsetup{font={small}, labelfont=bf, skip=4pt}
		\caption{Classification results on significant gaze features of session-2}
		\label{tab:Chapter-2tab10}\\
		\toprule	
		\midrule
		
		\textbf{\scriptsize Feature set} & \textbf{\scriptsize Day} & \textbf{\scriptsize \# Sig. feat. (from table \ref{tab:Chapter-2tab6})} & \multicolumn{3}{c}{\textbf{SVM}} & \multicolumn{3}{c}{\textbf{DT}} & \multicolumn{3}{c}{\textbf{MLP}} \\
		\midrule
		& & & \textbf{\scriptsize CV accuracy} & \textbf{\scriptsize UAR} & \textbf{\scriptsize C. rate} & \textbf{\scriptsize CV accuracy} & \textbf{\scriptsize UAR} & \textbf{\scriptsize C. rate} & \textbf{\scriptsize CV accuracy} & \textbf{\scriptsize UAR} & \textbf{\scriptsize C. rate}\\
		\midrule
			
		\multirow{4}{1.8cm}{\raggedright \textbf{\footnotesize Word (4634)}} & 1 & 59 & 0.97 & 0.92 & 0.88 & 0.7 & 0.83 & 0.75 & 0.93 & 1.0 & 1.0 \\ 
		& 2 & 89 & 1.0 & 1.0 & 1.0 & 0.72 & 0.9 & 0.88 & 0.93 & 0.9 & 0.88 \\ 
		& 3 & 120 & 1.0 & 0.83 & 0.83 & 0.57 & 0.5 & 0.5 & 0.93 & 0.83 & 0.83 \\ 
		& All  & 500 & 0.83 & 0.88 & 0.86 & 0.61 & 0.48 & 0.48 & 0.76 & 0.86 & 0.86 \\ 
		\midrule
		\multirow{4}{2.5cm}{\raggedright \textbf{\footnotesize Sub-sentence (1452)}} & 1 & 24 & 0.82 & 1.0 & 1.0 & 0.5 & 0.92 & 0.88 & 0.62 & 0.5 & 0.5 \\ 
		& 2 & 58 & 0.88 & 0.9 & 0.88 & 0.65 & 0.9 & 0.88 & 0.85 & 0.9 & 0.88 \\ 
		& 3 & 66 & 1.0 & 0.83 & 0.83 & 0.87 & 0.67 & 0.67 & 0.9 & 0.83 & 0.83 \\ 
		& All & 255 & 0.77 & 0.85 & 0.81 & 0.6 & 0.6 & 0.62 & 0.75 & 0.73 & 0.67 \\ 
		\midrule
		\multirow{4}{1.8cm}{\raggedright \textbf{\footnotesize Sentence (770)}} & 1 & 26 & 0.6 & 0.83 & 0.75 & 0.5 & 0.83 & 0.75 & 0.55 & 0.75 & 0.62 \\ 
		& 2 & 35 & 0.85 & 0.9 & 0.88 & 0.68 & 0.83 & 0.88 & 0.6 & 0.8 & 0.75 \\ 
		& 3 & 32 & 0.87 & 1.0 & 1.0 & 0.63 & 0.67 & 0.67 & 0.87 & 0.83 & 0.83 \\ 
		& All & 176 & 0.76 & 0.81 & 0.76 & 0.52 & 0.59 & 0.52 & 0.66 & 0.68 & 0.67 \\ 
		\midrule
		\multirow{4}{1.8cm}{\raggedright \textbf{\footnotesize Paragraph (132)}} & 1 & 13 & 0.78 & 0.83 & 0.75 & 0.82 & 0.75 & 0.62 & 0.75 & 0.67 & 0.75 \\ 
		& 2 & 23 & 0.75 & 0.9 & 0.88 & 0.62 & 0.57 & 0.62 & 0.78 & 0.9 & 0.88 \\ 
		& 3 & 16 & 0.77 & 0.83 & 0.83 & 0.8 & 0.67 & 0.67 & 0.83 & 0.83 & 0.83 \\ 
		& All & 56 & 0.75 & 0.77 & 0.71 & 0.66 & 0.66 & 0.67 & 0.68 & 0.67 & 0.62 \\ 
		\midrule
		\multirow{4}{1.8cm}{\raggedright \textbf{\footnotesize Slide (110)}} & 1 & 5 & 0.75 & 0.83 & 0.75 & 0.55 & 0.75 & 0.62 & 0.62 & 0.92 & 0.88 \\ 
		& 2 & 18 & 0.78 & 0.73 & 0.75 & 0.68 & 0.73 & 0.75 & 0.7 & 0.63 & 0.62 \\ 
		& 3 & 15 & 0.83 & 1.0 & 1.0 & 0.7 & 0.83 & 0.83 & 0.8 & 1.0 & 1.0 \\ 
		& All & 48 & 0.73 & 0.73 & 0.67 & 0.64 & 0.41 & 0.33 & 0.58 & 0.69 & 0.62 \\ 
		\midrule
		\multirow{4}{1.8cm}{\raggedright \textbf{\footnotesize Whole-text (22)}} & 1 & 6 & 0.68 & 0.67 & 0.5 & 0.53 & 0.5 & 0.5 & 0.57 & 0.75 & 0.62 \\ 
		& 2 & 7 & 0.68 & 0.53 & 0.5 & 0.68 & 0.2 & 0.25 & 0.65 & 0.63 & 0.62 \\ 
		& 3 & 7 & 0.8 & 1.0 & 1.0 & 0.57 & 0.83 & 0.83 & 0.6 & 0.83 & 0.83 \\ 
		& All & 11 & 0.72 & 0.77 & 0.71 & 0.59 & 0.67 & 0.62 & 0.7 & 0.71 & 0.67 \\ 
		\midrule
		\multirow{4}{1.8cm}{\raggedright \textbf{\footnotesize Fixation (5425)}} & 1 & 75 & 0.95 & 0.92 & 0.88 & 0.75 & 0.83 & 0.75 & 0.9 & 0.92 & 0.88 \\ 
		& 2 & 137 & 0.95 & 1.0 & 1.0 & 0.57 & 1.0 & 1.0 & 0.88 & 0.9 & 0.88 \\ 
		& 3 & 174 & 1.0 & 0.83 & 0.83 & 0.67 & 0.83 & 0.83 & 1.0 & 0.83 & 0.83 \\ 
		& All & 809 & 0.76 & 0.77 & 0.71 & 0.62 & 0.71 & 0.67 & 0.76 & 0.84 & 0.86 \\ 
		\midrule
		\multirow{4}{1.8cm}{\raggedright \textbf{\footnotesize Saccade (791)}} & 1 & 12 & 0.57 & 0.83 & 0.75 & 0.65 & 0.42 & 0.38 & 0.65 & 0.92 & 0.88 \\ 
		& 2 & 13 & 0.68 & 0.9 & 0.88 & 0.53 & 0.73 & 0.75 & 0.6 & 0.83 & 0.88 \\ 
		& 3 & 14 & 0.6 & 0.83 & 0.83 & 0.53 & 0.5 & 0.5 & 0.67 & 0.83 & 0.83 \\ 
		& All & 32 & 0.66 & 0.5 & 0.48 & 0.53 & 0.47 & 0.43 & 0.7 & 0.5 & 0.48 \\ 
		\midrule
		\multirow{4}{1.8cm}{\raggedright \textbf{\footnotesize Regression (904)}} & 1 & 46 & 0.7 & 0.83 & 0.75 & 0.7 & 0.67 & 0.5 & 0.7 & 0.75 & 0.62 \\ 
		& 2 & 80 & 0.82 & 0.9 & 0.88 & 0.72 & 0.53 & 0.5 & 0.85 & 0.73 & 0.75 \\ 
		& 3 & 68 & 1.0 & 0.83 & 0.83 & 0.67 & 0.5 & 0.5 & 0.97 & 0.83 & 0.83 \\ 
		& All & 205 & 0.85 & 0.92 & 0.9 & 0.59 & 0.71 & 0.67 & 0.78 & 0.85 & 0.81 \\ 
		\midrule
		\multirow{4}{1.8cm}{\raggedright \textbf{\footnotesize All features (7120)}} & 1 & 133 & 0.97 & 0.92 & 0.88 & 0.8 & 0.83 & 0.75 & 0.93 & 1.0 & 1.0 \\ 
		& 2 & 230 & 0.9 & 0.9 & 0.88 & 0.6 & 0.9 & 0.88 & 0.88 & 0.83 & 0.88 \\ 
		& 3 & 256 & 1.0 & 0.83 & 0.83 & 0.83 & 0.5 & 0.5 & 0.93 & 0.83 & 0.83 \\ 
		& All & 1046 & 0.82 & 0.88 & 0.86 & 0.7 & 0.62 & 0.62 & 0.74 & 0.88 & 0.86 \\

		\bottomrule
					
\end{longtable}
}\endgroup


\subsection{Analysis of Feature Selection Methods using Effect of the Classifiers}
\label{ssec:Chapter2ssec6.3}
\begin{figure}[!h] 
	\centerline{
		\includegraphics[scale=0.35]{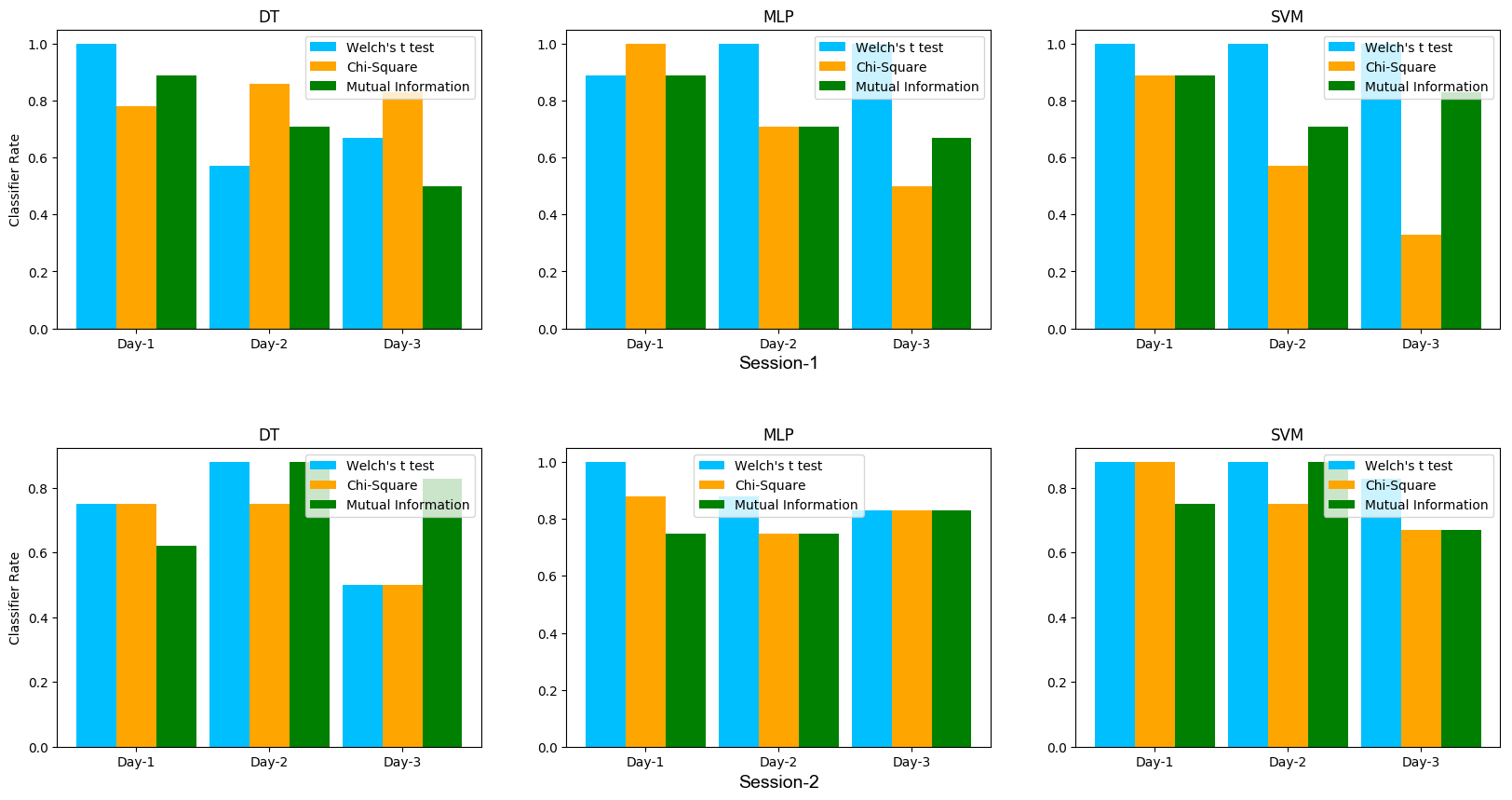}}
	\captionsetup{font={small}, labelfont=bf, skip=4pt}
	\caption{Classification rate by three classifiers on the best features of three feature selection methods}
	\label{fig:Chapter-2fig14}
\end{figure}

The classifiers' performance on the features selected by Welch's t-test has been presented in the previous section- \ref{ssec:Chapter2ssec6.2}. Here we present a comparative analysis of the classifiers' performance on the best-features selected by three feature selection methods-- Welch's t-test, Mutual Information and Chi-Square. The best performance of the three classifiers-- DT \& MLP, \& SVM (in terms of classification rate) on the best features (table-\ref{tab:No.ofFeatures}) of these methods are shown in figure- \ref{fig:Chapter-2fig14}. The figure shows that SVM classifier outperformed with Welch's t-test in both sessions.

\begin{table}[!h]
	\centering
	\captionsetup{font={small}, labelfont=bf, skip=4pt}
	\caption{Numbers of beast features selected by three feature selection methods}
	
	\small
	\renewcommand{\arraystretch}{1.4}
	\scalebox{1}{
	\begin{tabular}{C{2.cm} C{1cm} C{4cm} C{4cm} C{2.cm}}
	\hline
	\textbf{Session} & \textbf{Day} & \textbf{Welch's t-test} & \textbf{Chi-Square} & \textbf{MI} \\
	\midrule
	\multirow{3}{*}{1}&  1  & 145 & 20 & 50\\ 
	& 2 & 149  & 50 & 50\\
	& 3 & 108 & 20 & 150\\ \hline
	\multirow{3}{*}{2}& 1 & 133 & 40 & 50\\ 
	& 2 & 230 & 10 & 10\\ 
	& 3 & 256 & 250 & 20\\ 
	\bottomrule
\end{tabular}}
\label{tab:No.ofFeatures}
\end{table}

\subsection{Eye-movement Features Analysis on Text Structure}

A typical story is composed of four sections-- setting, plot, conflict and resolution. The previous study \cite{hyona1990eye} has shown that the eye-movement characteristics depend also on the structure of the text. 

\subsubsection{Eye-movement Characteristics on Text-1}

\begin{figure}[h!] 
	\centerline{
		\includegraphics[width=\columnwidth,height=100mm, scale=0.3]{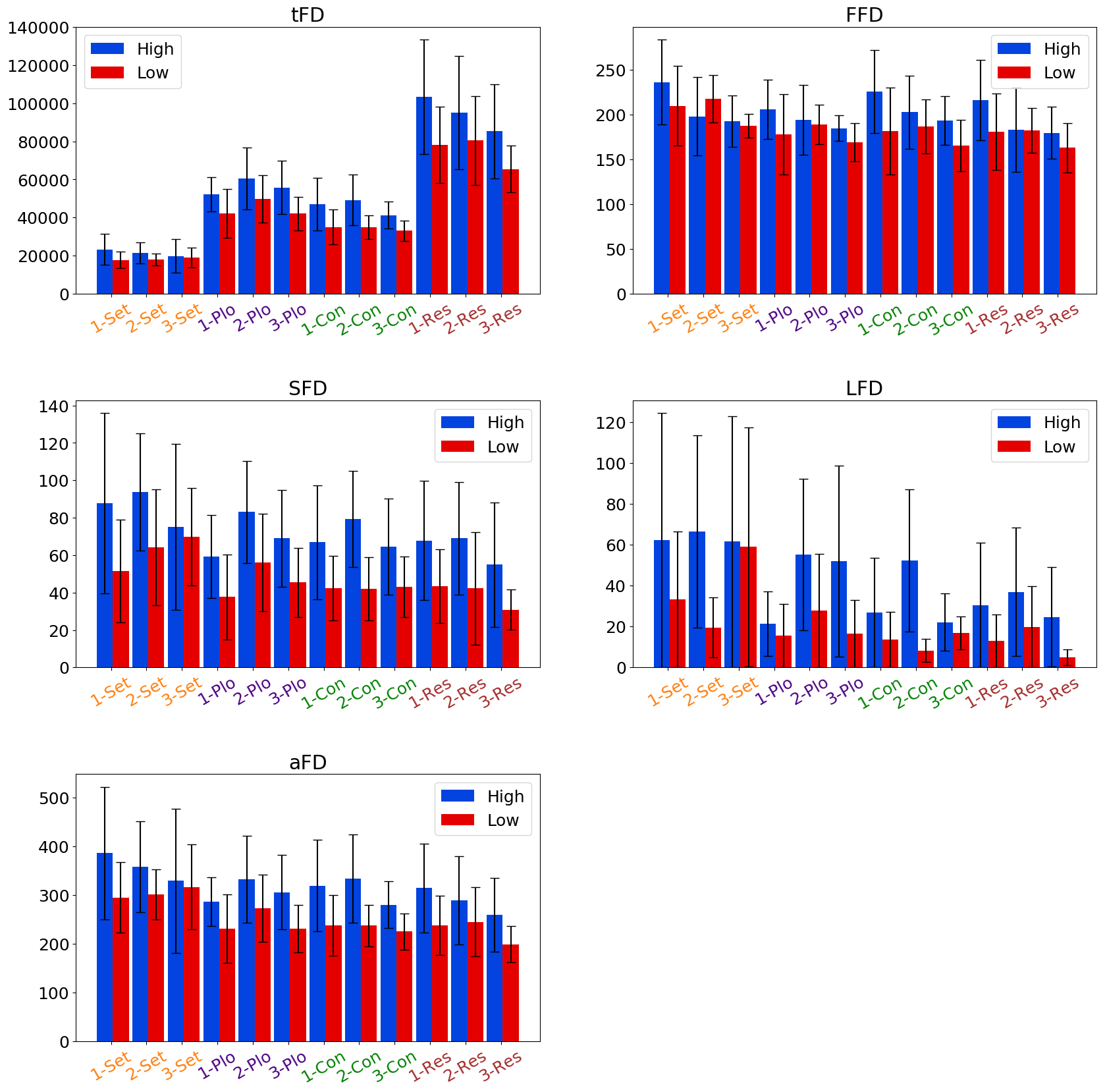}}
	\captionsetup{font={small}, labelfont=bf, skip=4pt}
	\caption{Fixation feature on text-1 structure}
	\label{fig:Chapter-2fig5}
\end{figure}

\begin{figure}[h!]
	\centerline{
		\includegraphics[width=\columnwidth,height=100mm, scale=0.3]{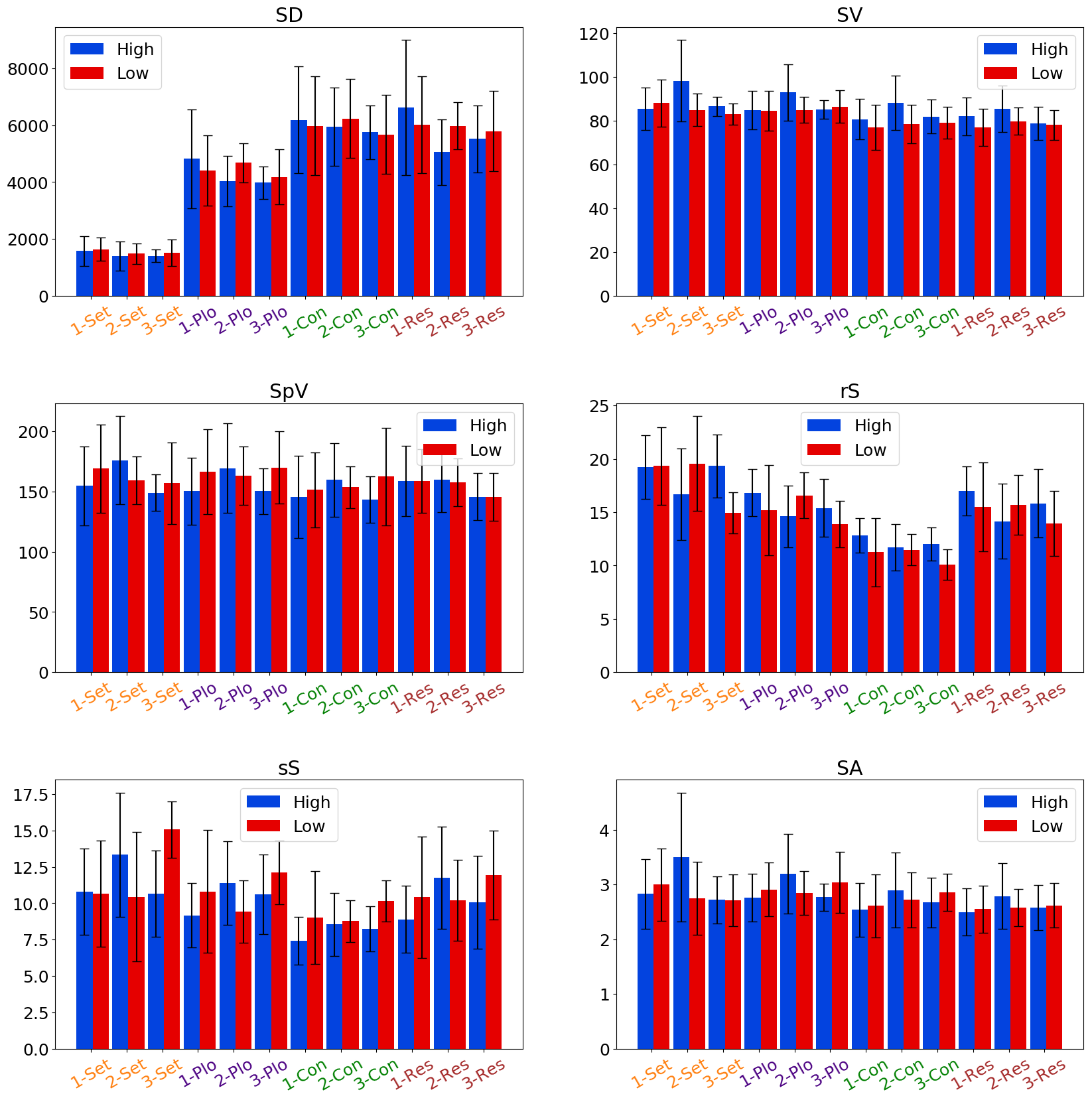}}
	\captionsetup{font={small}, labelfont=bf, skip=4pt}
	\caption{Saccade feature on text-1 structure}
	\label{fig:Chapter-2fig6}
\end{figure}

\begin{figure}[h!]
	\centerline{
		\includegraphics[width=\columnwidth,height=120mm, scale=0.3]{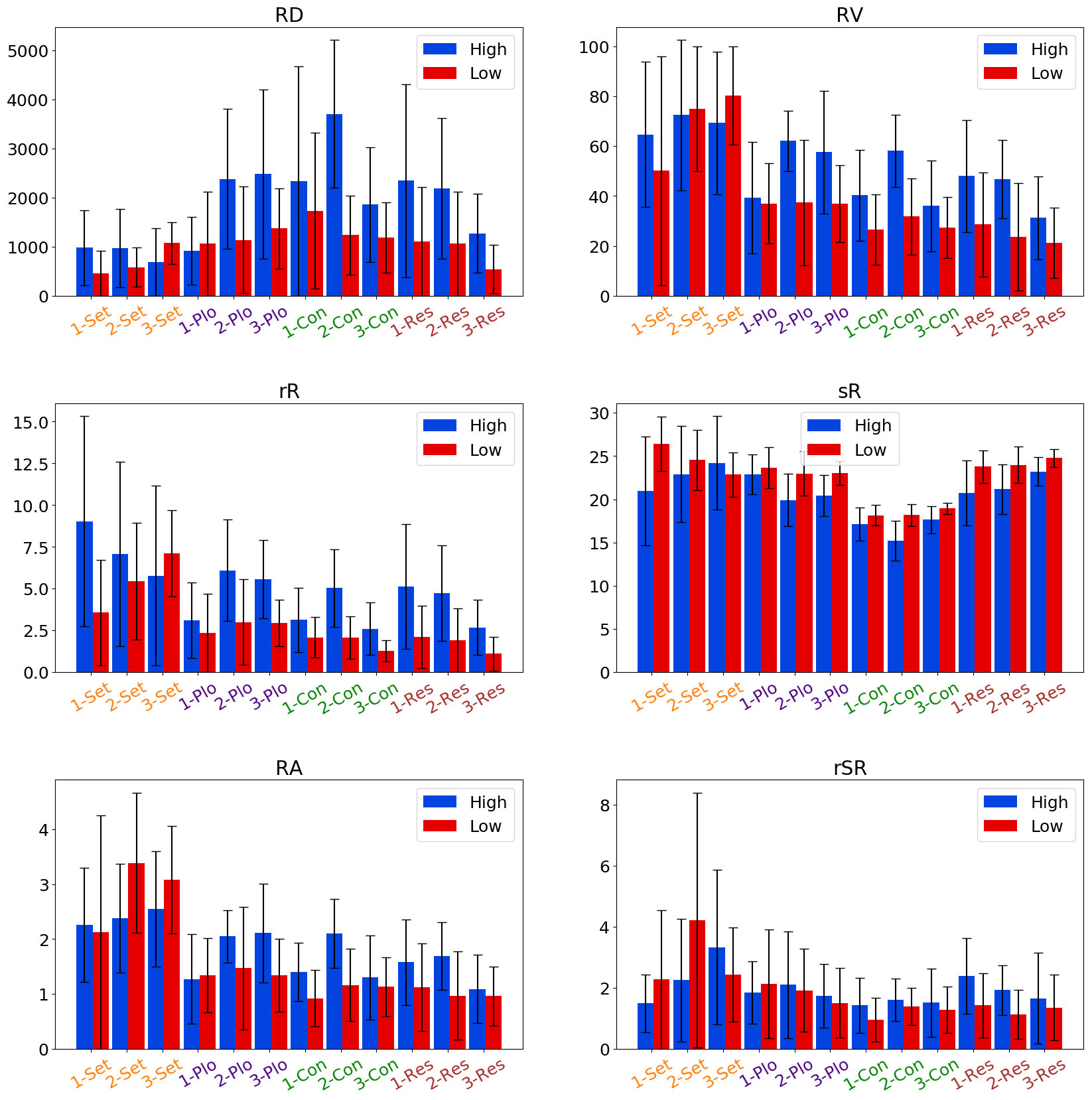}}
	\captionsetup{font={small}, labelfont=bf, skip=4pt}
	\caption{Regression feature on text-1 structure}
	\label{fig:Chapter-2fig7}
\end{figure}

Text-1 is divided into four sections– setting, plot, conflict, and resolution. Figure- \ref{fig:Chapter-2fig5} shows five bar charts, representing different fixation features (in ms), to analyse participants' day-wise repeated readings on text-1 sections; similarly next two figures-- \ref{fig:Chapter-2fig6} and \ref{fig:Chapter-2fig7} show six bar charts, representing saccade and regression features respectively on the sections. In a bar chart, blue and red bars represent the mean with the standard deviation of `high' and `low' groups respectively. In all charts, a pair of blue and red bars displays feature values from left to right in the sequence-- day-1 setting, day-2 setting, day-3 setting, day-1 plot, day-2 plot, day-3 plot, day-1 conflict, day-2 conflict, day-3 conflict, day-1 resolution, day-2 resolution, and day-3 resolution.

As shown in figure- \ref{fig:Chapter-2fig5}, in most cases of fixation features, the means are smaller in `low' group than those of `high' group. The variations in the mean of the groups are different across days on text-sections. tFD bar chart shows that participants spent maximum time on the resolution; however it has less words than the plot and the conflict. Therefore, aFD graph shows less variation in mean fixation duration across sections. The difference between first fixation means of both groups is smaller in most cases (shown in FFD bar graph); whereas, in case of second and last fixations, the differences between the means of both groups are higher (shown in SFD \& LFD graphs). These three bar graphs also show that the fixation features gradually decrease from first day to second day to third day. LFD graph has more standard deviation than the mean in several cases, because some participants spent more time but others spent little to no time in third or last pass. 

\begin{figure}[h!]
	\centerline{
		\includegraphics[width=\columnwidth,height=40mm, scale=0.3]{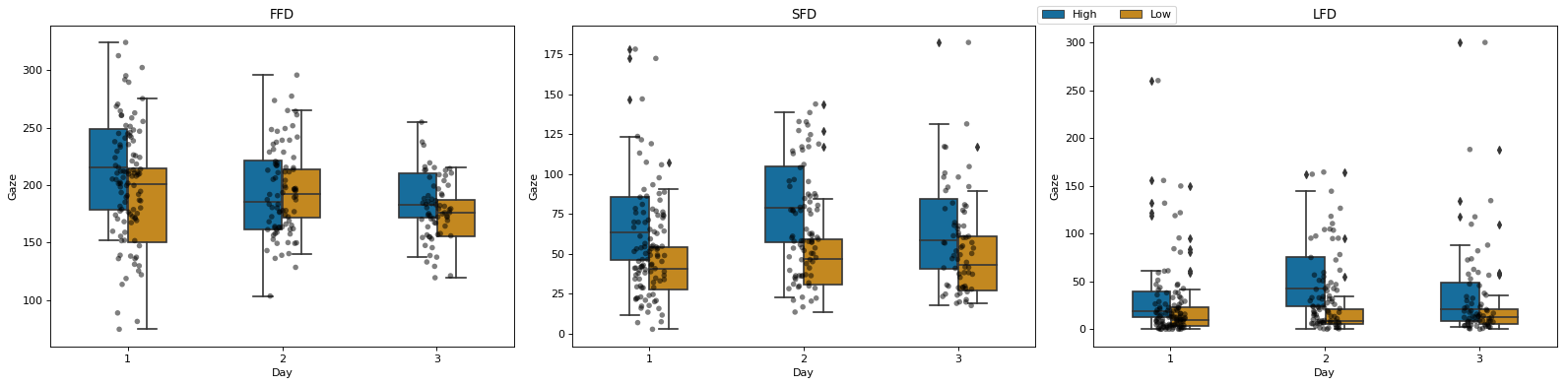}}
	\captionsetup{font={small}, labelfont=bf, skip=4pt}
	\caption{Gaze duration of the fixation features across trial days in session 1}
	\label{fig:Chapter-2fig51}
\end{figure}

The FFD, SFD and LFD gaze durations of the participants across trial days in session 1 are shown in figure- \ref{fig:Chapter-2fig51}. The figure clearly shows how the participants' gaze duration decreased from highest in the first fixation (FFD) to low in the second fixation (SFD) and then became lowest in the third fixation (LFD). We can see that some participants' gaze duration became outliers in SFD and LFD. 

As shown in figure- \ref{fig:Chapter-2fig6}, in SD graph, mean saccade duration is higher on conflict and resolution relative to the other two sections. SV graph shows that, saccade velocity is steady across all days on all sections, also there is no large difference in the means of both groups. Also, in saccade peak velocity (SpV) and saccade amplitude (SA) graphs, respective means are steady. 

The rS graph shows that, participants read lowest number of words in a first-pass saccade on conflict, whereas sS graph shows lowest number of words skipped during a saccade on this section. Thus, both graphs demonstrate that average length of first-pass saccade is smallest on conflict section than that on the other sections.

As shown in figure- \ref{fig:Chapter-2fig7}, in RD graph, mean regression duration is lowest on setting relative to the other sections. RV graph shows that, regression velocity is more in `high' group than `low' group in most cases. The rR graph shows that, participants read highest number of words in a regression on setting, whereas sR graph shows lowest number of words skipped on conflict section. The graph also shows that `low' group skipped more word than `high' group. The RA graph, the mean size of the regression, shows highest degree of eye-movement arc on setting. The last graph rSR shows, the ratio of read words in saccade vs those in regression is lowest on the conflict.


\subsubsection{Eye-movement Characteristics on Text-2}
\begin{figure}[h!]
	\centerline{
		\includegraphics[width=\columnwidth,height=100mm, scale=0.3]{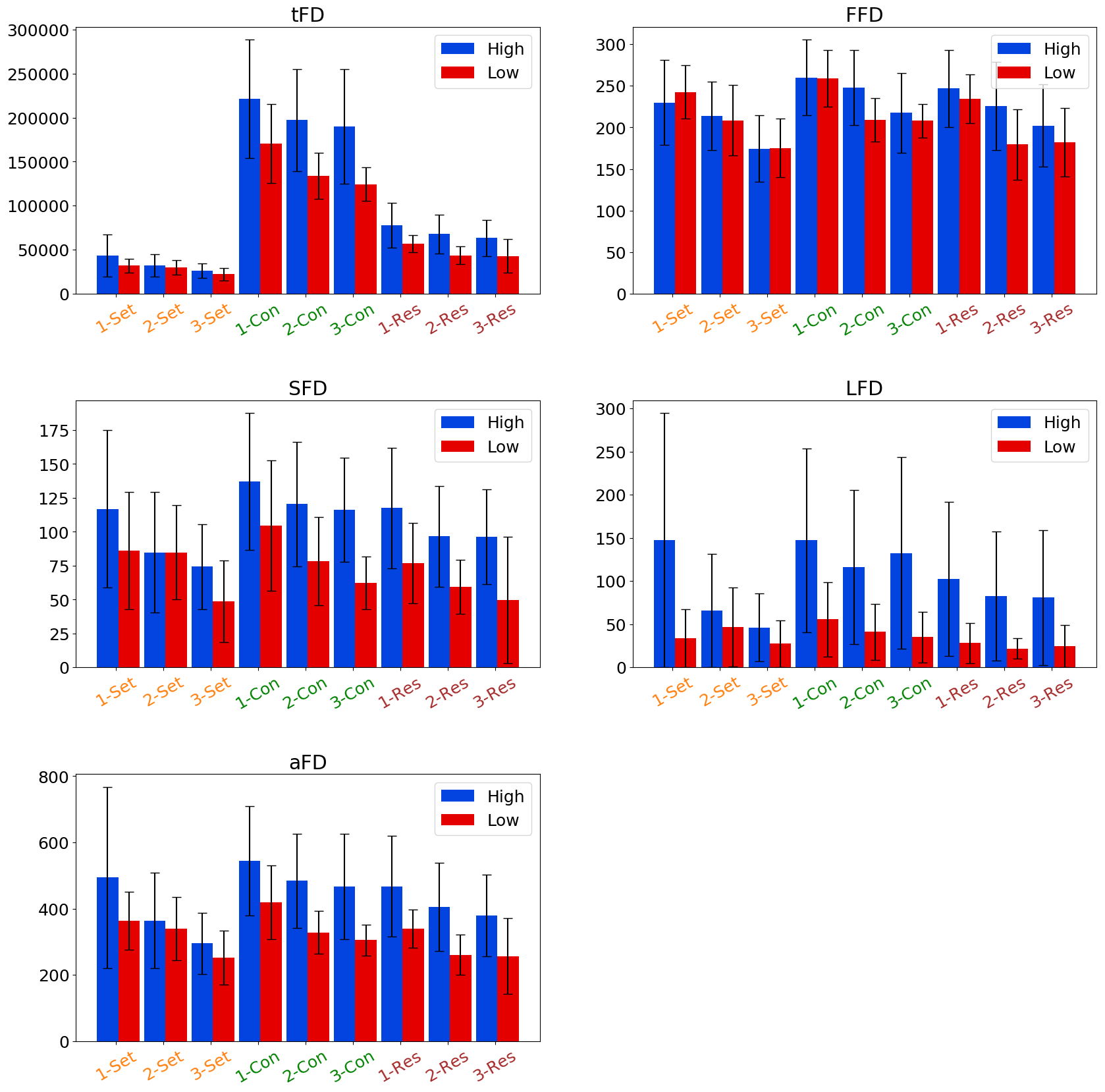}}
	\captionsetup{font={small}, labelfont=bf, skip=4pt}
	\caption{Fixation feature on text-2 structure}
	\label{fig:Chapter-2fig8}
\end{figure}

Text-2 is divided into three sections– setting, conflict, and resolution. As shown in figure- \ref{fig:Chapter-2fig8}, in most cases of fixation features, the means of both groups decrease across days. The tFD bar chart shows that participants spent time on the sections according to sections' size. The first fixation durations of both groups are very near as shown in FFD graph; however for second and last fixation durations, the difference between the means of both groups are more as shown in SFD and LFD graphs. Also, the average fixation durations of both groups decrease in the consecutive days across the sections.  

\begin{figure}[!h]
	\centerline{
		\includegraphics[width=\columnwidth,height=40mm, scale=0.3]{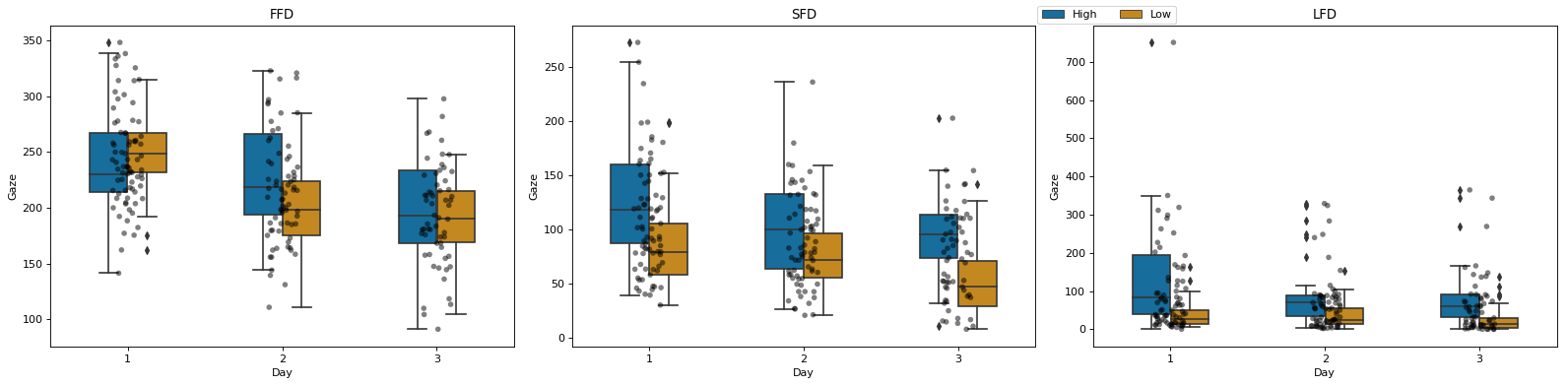}}
	\captionsetup{font={small}, labelfont=bf, skip=4pt}
	\caption{Gaze duration of the fixation features across trial days in session 2}
	\label{fig:Chapter-2fig81}
\end{figure}

The FFD, SFD and LFD gaze durations of the participants across trial days in session 2 are shown in figure- \ref{fig:Chapter-2fig81}. The figure clearly shows how the participants' gaze duration was decreased from highest in the first fixation (FFD) to low in the second fixation (SFD) and then became lowest in the third fixation (LFD). We can see that some participants' gaze duration became outliers in FFD, SFD and LFD. 

\begin{figure}[h!]
	\centerline{
		\includegraphics[width=\columnwidth,height=100mm, scale=0.3]{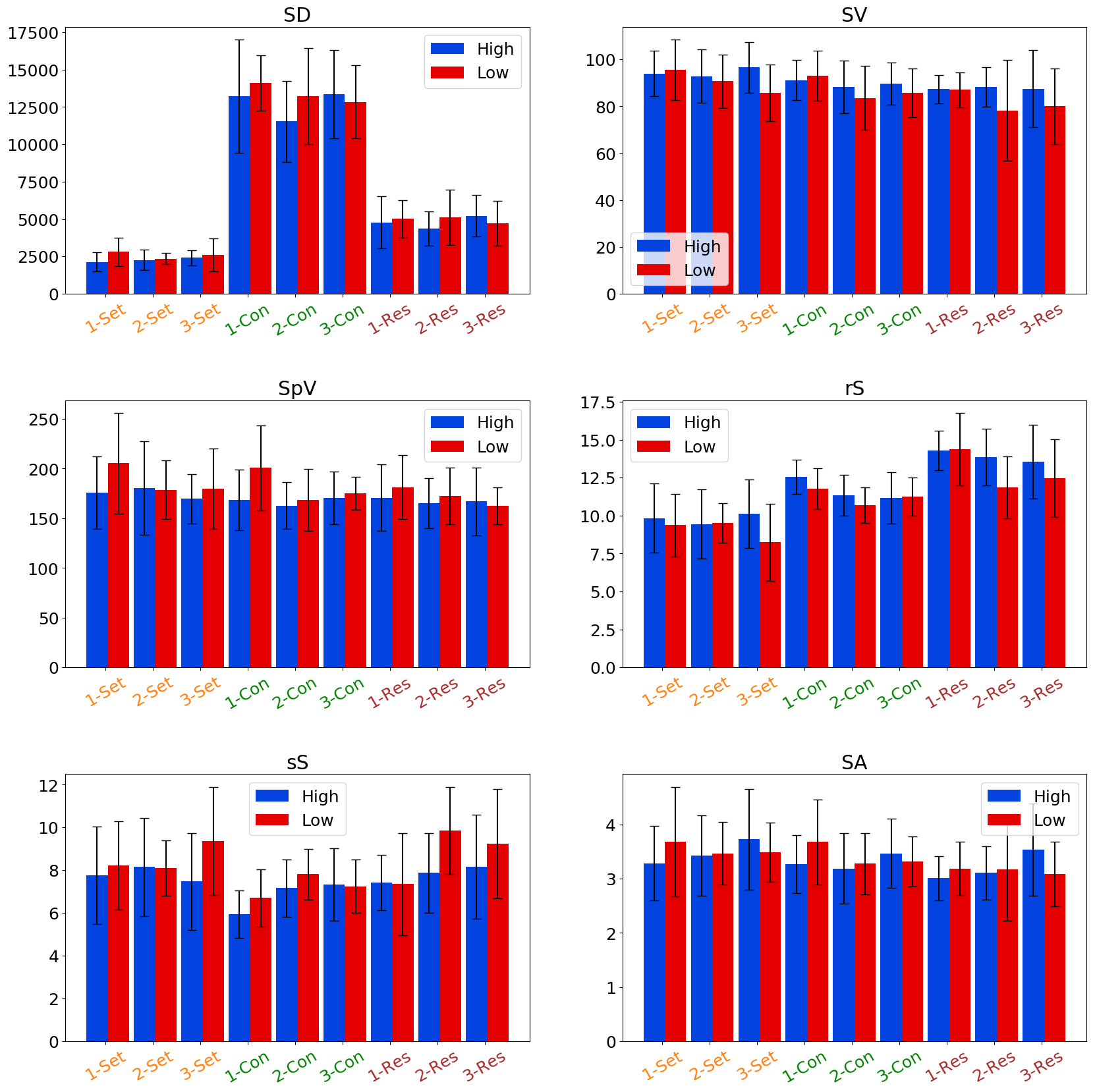}}
	\captionsetup{font={small}, labelfont=bf, skip=4pt}
	\caption{Saccade feature on text-2 structure}
	\label{fig:Chapter-2fig9}
\end{figure}

As shown in figure- \ref{fig:Chapter-2fig9}, the SD bar chart shows that both groups' mean saccade duration is according to the respective section's size; whereas, other graphs do not show more variation in the means of both groups across days on the sections. The rS bar chat shows that participants read fewer words in first-pass saccade on setting as compared to the other sections.  

\begin{figure}[h!]
	\centerline{
		\includegraphics[width=\columnwidth,height=100mm, scale=0.3]{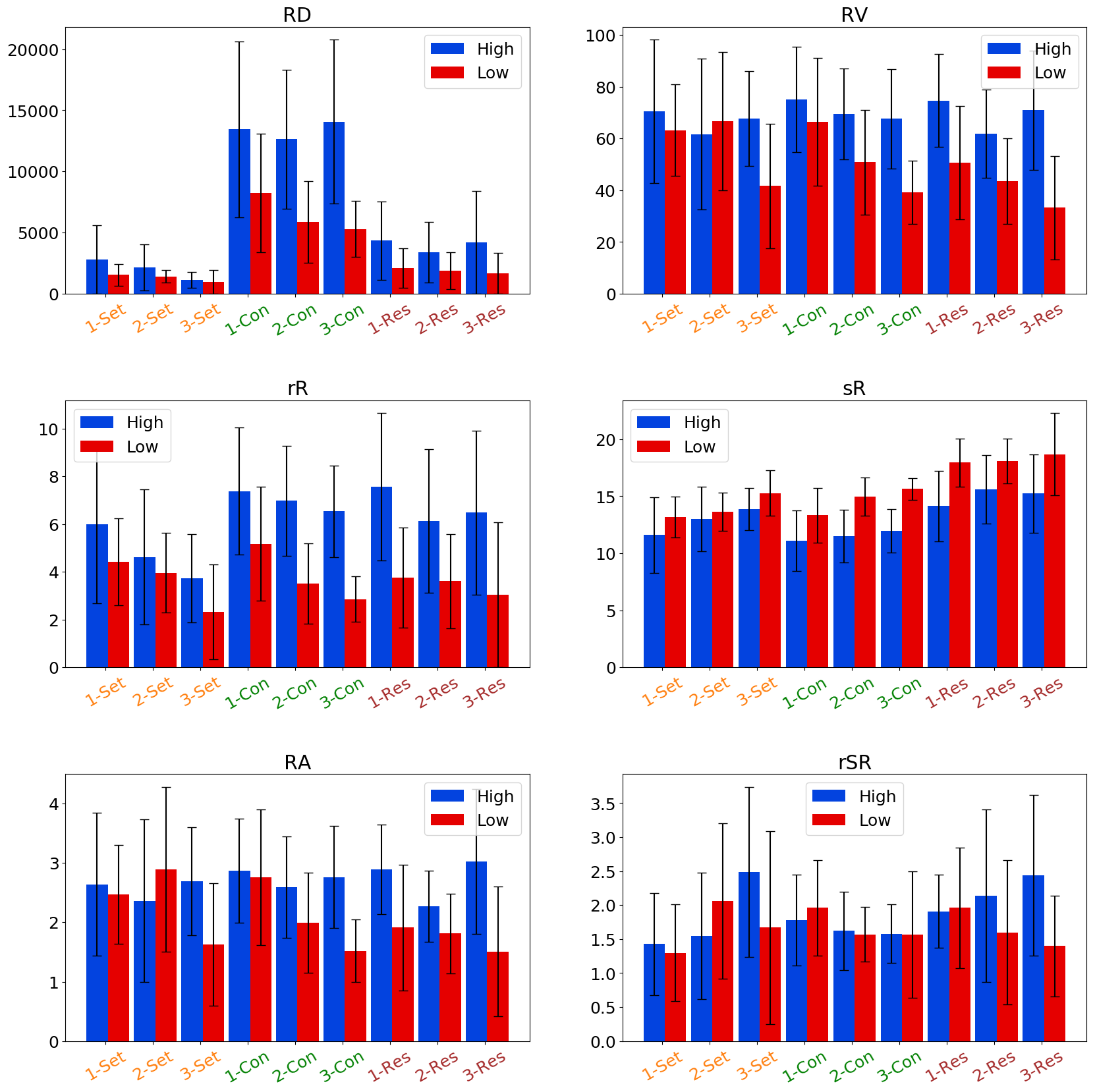}}
	\captionsetup{font={small}, labelfont=bf, skip=4pt}
	\caption{Regression feature on text-2 structure}
	\label{fig:Chapter-2fig10}
\end{figure}

As shown in figure- \ref{fig:Chapter-2fig10}, in RD graph, mean regression durations of both groups are according to respective section's size. 


The RV graph shows that low group's mean velocity decreases across days; whereas, the high group's mean velocity remains steady. The rR bar chat shows that participants read less words in a regression on setting as compared to the other sections. The sR bar chart shows the number of words skipped during a regression on the sections; `high' group skipped relatively more words. In RA graph, the mean regression amplitudes of both groups are plotted. In rSR graph, the ratio of read words in saccade vs those in regression is plotted.


\subsubsection{Comparison of Eye-movement Characteristics on Both Texts}

\begin{figure}[h!] 
	\centerline{
		\includegraphics[width=\columnwidth,height=90mm, scale=0.3]{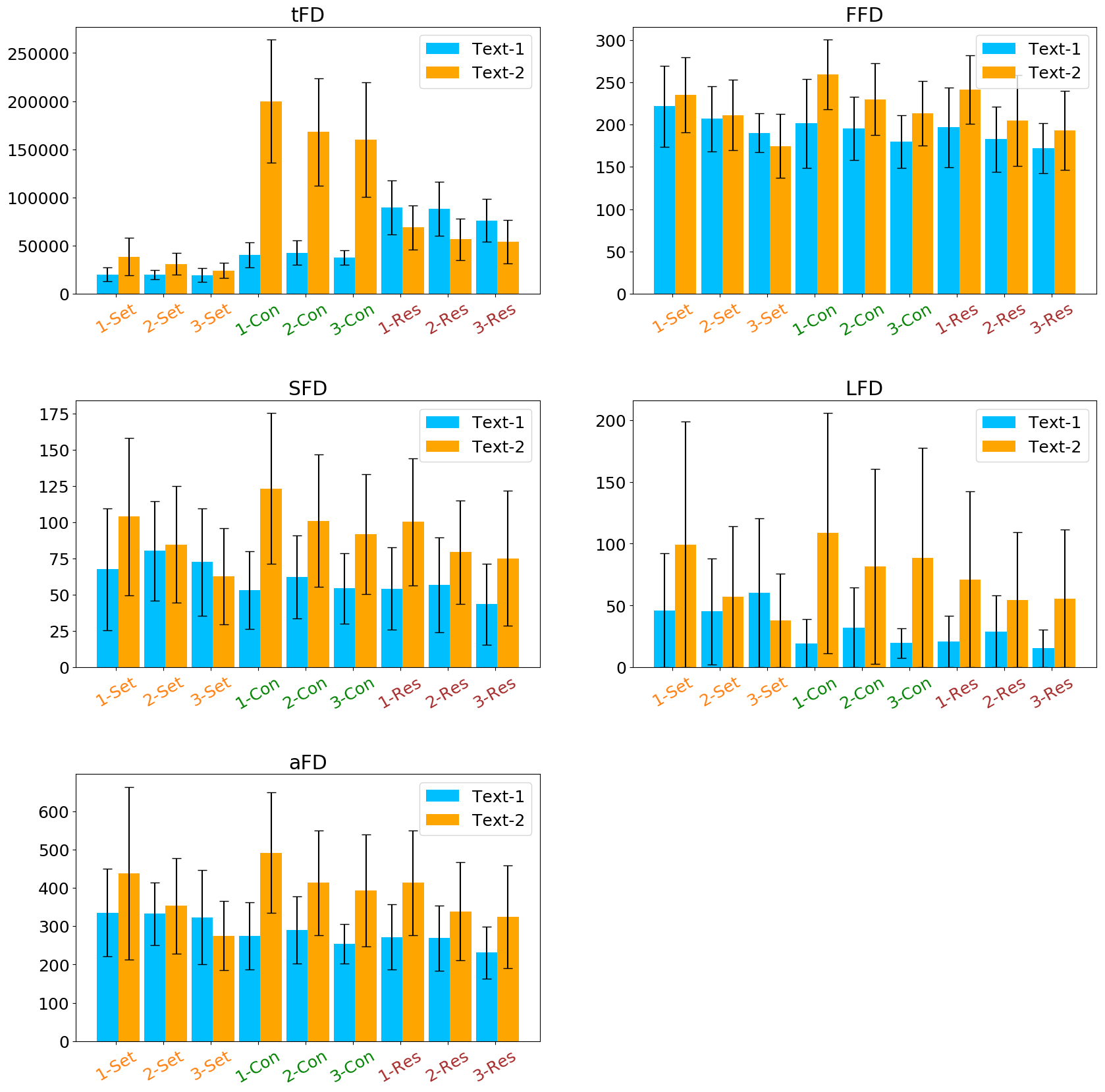}}
	\captionsetup{font={small}, labelfont=bf, skip=4pt}
	\caption{A comparison of fixation feature on text- 1 \& 2 structure}
	\label{fig:Chapter-2fig11}
\end{figure}

\begin{figure}[h!]
	\centerline{
		\includegraphics[width=\columnwidth,height=100mm, scale=0.35]{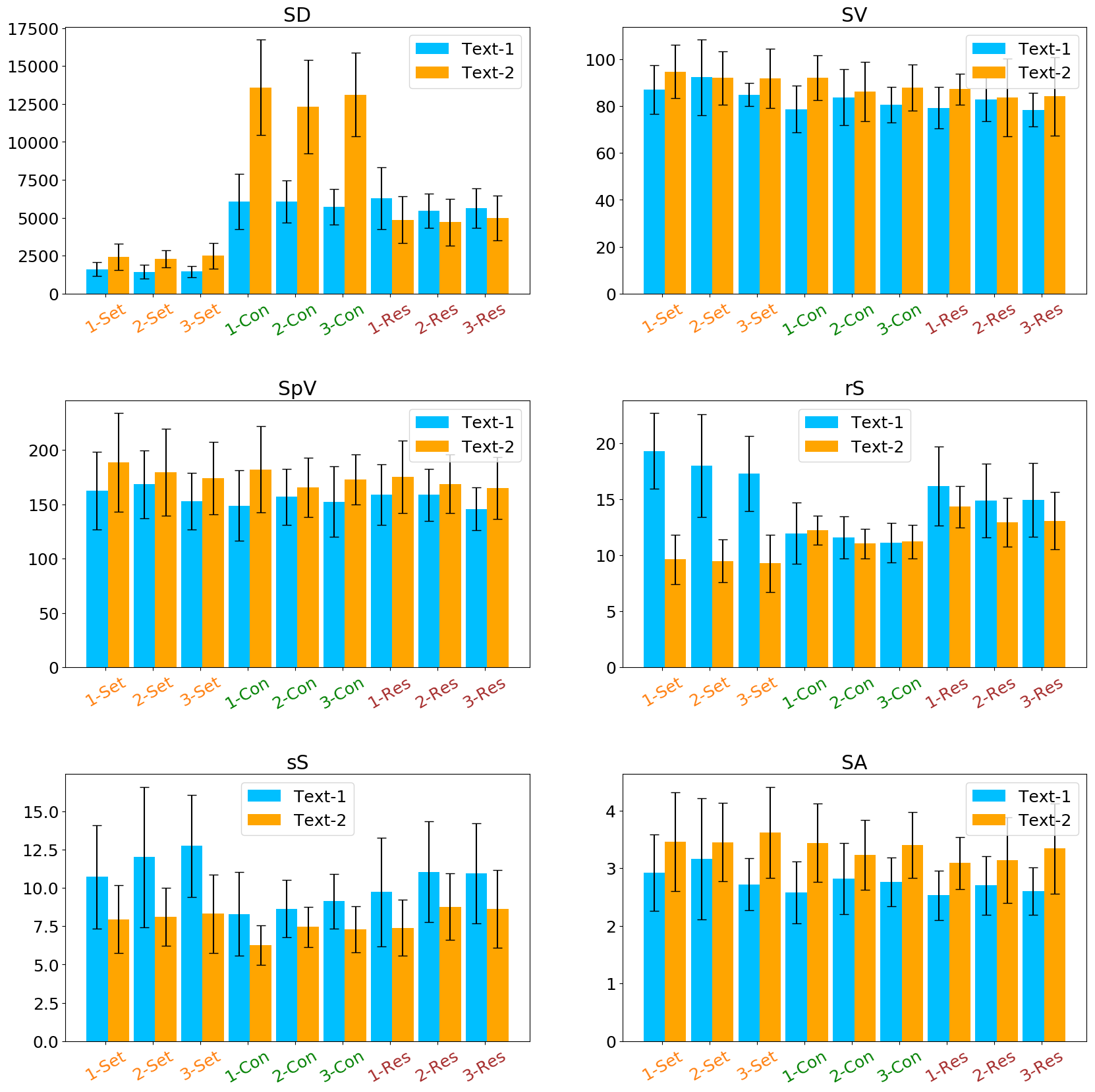}}
	\captionsetup{font={small}, labelfont=bf, skip=4pt}
	\caption{A comparison of saccade feature on text- 1 \& 2 structure}
	\label{fig:Chapter-2fig12}
\end{figure}

\begin{figure}[h!]
	\centerline{
		\includegraphics[width=\columnwidth,height=100mm,scale=0.35]{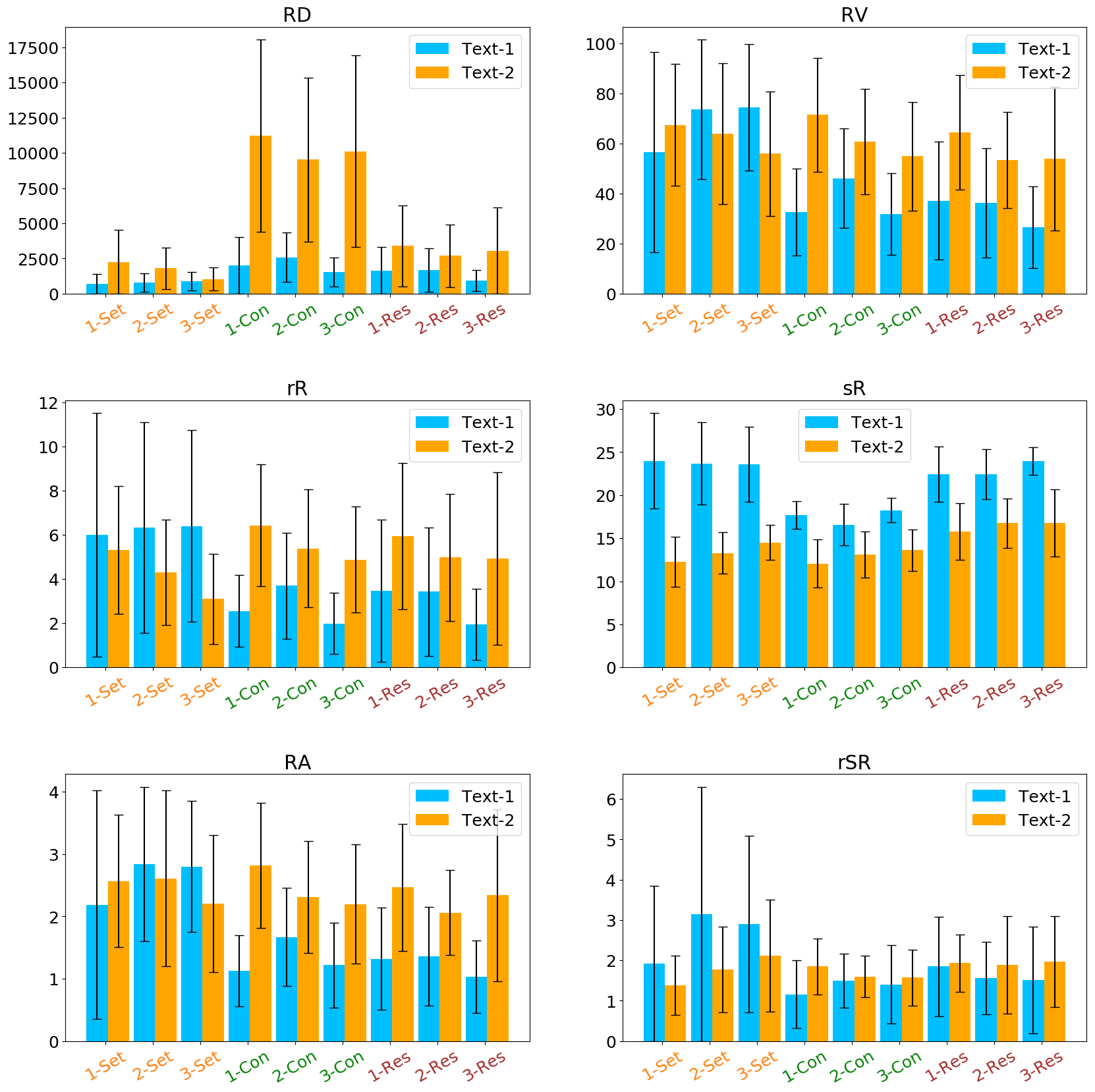}}
	\captionsetup{font={small}, labelfont=bf, skip=4pt}
	\caption{A comparison of regression feature on text- 1 \& 2 structure}
	\label{fig:Chapter-2fig13}
\end{figure}

Figures-- \ref{fig:Chapter-2fig11}, \ref{fig:Chapter-2fig12} and \ref{fig:Chapter-2fig13} show a comparison of fixation, saccade and regression features respectively on the common sections of both texts-- 1 and 2. In a bar chart, sky-blue and orange bars represent the mean (with the standard deviation) of all participants on sections of text 1 and 2 respectively. In all charts, a pair of sky-blue and orange bars displays feature values from left to right in the sequence-- day-1 setting, day-2 setting, day-3 setting, day-1 conflict, day-2 conflict, day-3 conflict, day-1 resolution, day-2 resolution, and day-3 resolution.

As shown in figure- \ref{fig:Chapter-2fig11}, the tFD bar chart shows that all participants' total fixation duration is very high on the conflict of text-2 with compared to text-1. 

The first fixation duration of all participants is more on text-2 sections than on text-1 sections with one exception-- day-3 setting; also it decreases on both text's sections across days. The second, last and average fixation-durations are more on text-2 sections than on text-1 sections in most cases.

As shown in figure- \ref{fig:Chapter-2fig12}, the SD bar chart shows that all participants' mean saccade duration is more on the conflict of text-2 with compared to text-1. Their mean saccade velocity, peak velocity and amplitude are more on text-2 sections than on text-1 sections. The rS and sS graphs show that, participants read and skip more words in saccade on text-1 than on text-2.

As shown in figure- \ref{fig:Chapter-2fig13}, in RD graph, mean regression duration of all participants is more on the conflict of text-2 with compared to text-1. The velocity and amplitude of the participants' regression are more on text-2 sections than text-1 in most cases. The rR bar chat shows that participants read more words in a regression on text-2 sections-- conflict and resolution than on those of text-1. They skipped more words on text-1 than on text-2. The last graph shows that, the ratio of read words in saccades vs those in regressions is higher on text-1 setting than on text-2 setting.

\subsection{LMER Analysis of Eye-movement Feature}

To study the effect of repeated reading (i.e., trial days were nested within the participants) on eye movements as well as other variables, even the sample size was low, we applied linear mixed-effects regression (LMER) models implemented with the lme4 package \cite{bates2014fitting} in the R environment (Version 3.6.2 R Core Team, 2019). Linear mixed-effects models are preferred over classical approaches, such as repeated measures ANOVA, as they provide insight into whether variance in a response variable is attributable to differences in ``between-individual variance" or ``within-individual variance" or both \cite{pinheiro2000linear}. 

We fitted four separate models for a) fixation, b) saccade, c) regression and d) gaze event (i.e., all the three as combined), on the two individual sessions data. In each model, the outcome variable- gaze value was regressed onto the predictor features of the feature-set. Thus, in a model, we specified gaze value as the outcome variable, whereas features name, group and trial day as fixed effects (also, independent variables), while student (student-id) was specified as random effects variable. Also, in a model trial day was nested within students; where within each student, there were 3 trial days. The nesting of days within student showed repeated measure in time and thus simulated the impact of repeated reading on students' eye-movement by statistically measure variations of feature values over time (day). The effect of a feature was considered to be statistically significant at $\alpha = 0.05$ level if the absolute t-value was more than 2.0. The t-value measures the size of the difference relative to the variation in the sample data. The greater the magnitude of t-value, the greater the evidence against the null hypothesis. This means there is greater evidence that there is a significant effect. If the t-value is closer to 0, which means there is not a significant effect.

Table \ref{tab:Chapter-2tab11} shows LMER results of participants' eye-movement features on sessions 1 and 2 data respectively. To study the effects of repeated reading on fixation in both sessions, the gaze of six fixation features– FFD, SFD, LFD, aFD, tFC and aFC were regressed with respect to the gaze of reference (baseline) fixation feature tFD and other baselines were day = 1 and group = high. The result shows the fixation intercept is more in session 2 than session 1, which indicates that participants fixated eyes for a longer time during repeated readings in session 2 than readings in session 1. Also, except aFD, all other fixation features were significantly important in both sessions. The fixation time was significantly decreased from day-1 to successor days only in session 2 readings. As expected, the fixation time was significantly low in the participants of group- low than those of the group- high in the both sessions' repeated readings.

To study the effects of repeated reading on saccade in both sessions, the gaze of six saccade features– SC, SV, SpV, rS, sS and SA were regressed with respect to the gaze of reference (baseline) feature SD and other baselines were day = 1 and group = high. The result shows the saccade intercept is nearly the same in both sessions, which indicates that participants had performed forward eye movement equally in both sessions. Also, all saccade features were significantly important in both sessions. However, in both sessions, saccade was not statistically effected from day-1 to successor days as well as in the low group participants than the high group.

To study the effects of repeated reading on regression in both sessions, the gaze of seven regression features– RC, RV, RpV, rR, sR, RA and rSR were regressed with respect to the gaze of reference (baseline) feature RD and other baselines were day = 1 and group = high. The result shows regression intercept is more than twice in session 2 than in session 1, which indicates that participants had performed backward eye movement more than double in session 2 than in session 1 readings. Also, all regression features were significantly important in both sessions. However, in both sessions, regression was not statistically effected from day-1 to successor days; but it was statistically lower in the low group participants than the high group.

\begingroup
\setstretch{1.3}
{\small	
	\begin{longtable}
		{>{\raggedright}m{3cm} L{2cm} L{3cm} L{2cm} L{2cm} }
		\captionsetup{font={small}, labelfont=bf, skip=4pt}
		\caption{LMER results of eye-movement features}
		\label{tab:Chapter-2tab11}\\
		\toprule\midrule
		\textbf{Eye-movement} & \multicolumn{2}{c}{\textbf{Session 1}} & \multicolumn{2}{c}{\textbf{Session 2}} \\
		
		\textbf{Features}& \textbf{Estimate} & \textbf{t value} & \textbf{Estimate} & \textbf{t value} \\
		\midrule
		
		\multicolumn{5}{c}{\textbf{Fixation}}\\ 
		(Intercept) & 285.7 & 38.54\textsuperscript{***} & 451.4 & 35.01\textsuperscript{***}\\ 
		FFD & -83.33 & -15.02\textsuperscript{***} & -180.7 & -16.79\textsuperscript{***}\\ 
		SFD & -216.7 & -39.06\textsuperscript{***} & -309.2 & -28.75\textsuperscript{***}\\ 
		LFD & -246.1 & -44.36\textsuperscript{***} & -326 & -30.31\textsuperscript{***}\\ 
		aFD & $\expnumber{-2.52}{-12}$ & 0 & $\expnumber{7.97}{-12}$ & 0 \\ 
		tFC & -272.1 & -49.04\textsuperscript{***} & -406.6 & -37.8\textsuperscript{***}\\ 
		aFC & -272.1 & -49.04\textsuperscript{***} & -406.6 & -37.8\textsuperscript{***}\\ 
		Day-2 & 3.72 & 0.59 & -32.83 & -3.29\textsuperscript{**}\\ 
		Day-3 & -9.41 & -1.36 & -38.63 & -3.59\textsuperscript{***}\\ 
		Group-Low & -22.9 & -3.4\textsuperscript{**} & -51.06 & -4.6\textsuperscript{***}\\ 
		\midrule
		\multicolumn{5}{c}{\textbf{Saccade}}\\ 
		(Intercept) & 595.22 & 70.93\textsuperscript{***} & 593.58 & 82.00\textsuperscript{***}\\ 
		SC & -572.22 & -61.3\textsuperscript{***} & -576.83 & -69.29\textsuperscript{***}\\ 
		SV & -508.08 & -54.43\textsuperscript{***} & -503.92 & -60.53\textsuperscript{***}\\ 
		SpV & -434.44 & -46.54\textsuperscript{***} & -418.53 & -50.27\textsuperscript{***}\\ 
		rS & -576.44 & -61.75\textsuperscript{***} & -580.93 & -69.78\textsuperscript{***}\\ 
		sS & -580.71 & -62.21\textsuperscript{***} & -585.13 & -70.29\textsuperscript{***}\\ 
		SA & -587.8 & -62.97\textsuperscript{***} & -589.15 & -70.77\textsuperscript{***}\\ 
		Day-2 & -5.33 & -0.89 & -10.26 & -1.92\\ 
		Day-3 & -7.35 & -1.13 & -4.50 & -0.79\\ 
		Group-Low & -1.28 & -0.21 & 7.12 & 1.35\\ 
		\midrule
		\multicolumn{5}{c}{\textbf{Regression}}\\ 
		(Intercept) & 200.98 & 27.85\textsuperscript{***} & 469.99 & 33.17\textsuperscript{***}\\ 
		RC & -188.96 & -22.75\textsuperscript{***} & -433.86 & -26.25\textsuperscript{***}\\ 
		RV & -154.65 & -18.62\textsuperscript{***} & -383.21 & -23.19\textsuperscript{***}\\ 
		RpV & -117.64 & -14.16\textsuperscript{***} & -321.30 & -19.44\textsuperscript{***}\\ 
		rR & -191.26 & -23.03\textsuperscript{***} & -439.91 & -26.62\textsuperscript{***}\\ 
		sR & -192.94 & -23.23\textsuperscript{***} & -443.56 & -26.84\textsuperscript{***}\\ 
		RA & -193.14 & -23.26\textsuperscript{***} & -442.84 & -26.8\textsuperscript{***}\\ 
		rSR & -174 & -20.95\textsuperscript{***} & -431.74 & -26.12\textsuperscript{***}\\ 
		Day-2 & 8.66 & 1.62 & -14.20 & -1.43 \\ 
		Day-3 & -0.84 & -0.14 & -14.07 & -1.33  \\ 
		Group-Low & -17.67 & -3.56\textsuperscript{***} & -35.86 & -3.60\textsuperscript{***}\\ 
		\midrule
		\multicolumn{5}{c}{\textbf{Gaze Event (fixation + saccade + regression)}}\\ 
		(Intercept)  & 125.97 & 13.93\textsuperscript{***} & 201.11 & 17.9\textsuperscript{***}\\ 
		Saccade  & 7.55 & 0.8 & -47.67 & -3.99\textsuperscript{***}\\ 
		Regression & -74.29 & -8.17\textsuperscript{***} & -92 & -7.95\textsuperscript{***}\\ 
		Day-2  & 2.57 & 0.29 & -18.23 & -1.61 \\ 
		Day-3  & -6.93 & -0.73 & -20.72 & -1.74 \\ 
		Group-Low & -15.45 & -2.05\textsuperscript{*} & -30.95 & -3.22\textsuperscript{**}\\ 
		\midrule
		\multicolumn{5}{l}{\textsuperscript{*}: $p<0.05$, \textsuperscript{**}: $p<0.01$, \textsuperscript{***}: $p<0.001$}\\
		
		\midrule\bottomrule
	\end{longtable}
}\endgroup
		
Finally, to study the effects of repeated reading on overall gaze event (fixation + saccade + regression) in both sessions, the features of saccade and regression were regressed with respect to the gaze of reference (baseline) fixation and other baselines were day = 1 and group = high. The result shows gaze event intercept is more in session 2 than in session 1, which indicates that participants had performed overall eye movement more in session 2 than in session 1 readings. In session 1, only regression was statistically effected than fixation, however, in session 2, both saccade and regression were statistically effected than fixation. Also, in both sessions, gaze event was not statistically effected from day-1 to successor days; but it was statistically lower in the low group participants than the high group.

\subsection{ANOVA Analysis of the LMER models}

Table \ref{tab:Chapter-2tab12} shows f-values of the ANOVA analysis of the LMER models discussed in the previous table \ref{tab:Chapter-2tab11} for variables of interest- eye-movement features, trial day and group of sessions 1 and 2 data. The ANOVA analysis of the LMER models give f-value of the gaze event (fixation, saccade and regression), group (high, low) and trial day (1, 2, 3) as a whole. Here, gaze value is dependent variable (DV), whereas feature-name, trial day and group are independent variables (IV). In the table \ref{tab:Chapter-2tab12}, f-values show that gaze features of all three events-- fixation, saccade and regression are significantly effected during repeated reading in both sessions. The f-value of fixation features is relatively high in session-1; which means in session-1, the variability of fixation features' mean is large relative to the within-feature variability. Thus, the analysis shows that among fixation features, their means were more distinct in session-1; whereas, among saccade features as well as regression features, their means were more distinct in session-2. As seen in the previous table \ref{tab:Chapter-2tab11}, the fixation was significantly effected during days in only session 2. We can also see that, fixation and regression were significantly effected the group in both sessions' readings. 

\begingroup
\setstretch{1.2}
{\small	
	\begin{longtable}
		{>{\raggedright}m{1.6cm} L{1.6cm} L{1.6cm} L{2.cm} L{1.6cm} L{1.6cm} L{2cm}}
		\captionsetup{font={small}, labelfont=bf, skip=4pt}
		\caption{ANOVA analysis of the LMER models of eye-movement features}
		\label{tab:Chapter-2tab12}\\
		\toprule\midrule
		\textbf{Variable} & \multicolumn{3}{c}{\textbf{Session 1}} & \multicolumn{3}{c}{\textbf{Session 2}} \\
		
		& \textbf{Fixation} & \textbf{Saccade} & \textbf{Regression} & \textbf{Fixation} & \textbf{Saccade} & \textbf{Regression} \\
		\midrule
		Features & 1004.12\textsuperscript{***} & 1038.24\textsuperscript{***} & 128.68\textsuperscript{***} & 536.03\textsuperscript{***} & 1326.18\textsuperscript{***} & 169.97\textsuperscript{***}\\ 
		Day & 1.73 & 0.74 & 1.73 & 8.24\textsuperscript{***} & 1.86 & 1.32 \\ 
		Group & 11.61\textsuperscript{**} & 0.04 & 12.69\textsuperscript{***} & 21.23\textsuperscript{***} & 1.82 & 13.01\textsuperscript{***}\\ 
		\midrule
		\multicolumn{5}{l}{\textsuperscript{*}: $p<0.05$, \textsuperscript{**}: $p<0.01$, \textsuperscript{***}: $p<0.001$}\\
		
		\midrule\bottomrule
	\end{longtable}
}\endgroup


\subsection{LMER Analysis of Fixation Features on Objective Score}

Table \ref{tab:Chapter-2tab14} shows LMER results for the effect of fixation features (independent variable) on participants' score (dependent variable) obtained from the answers of objective type questions asked during sessions 1 and 2.

\begin{table}[!h]
	\captionsetup{font={small}, labelfont=bf, skip=4pt}
	\caption{LMER result of fixation features for the score on sessions 1 \& 2 data} 
	\centering 
	{\small	
		\scalebox{1}{
			\renewcommand{\arraystretch}{1.4}
			\begin{tabular}{L{2.2cm} L{3.4cm} L{3.8cm} L{3.4cm}}
				\hline 
				\textbf{Fixation feature} & \textbf{Values} & \textbf{Session 1
					(objective-score)} & \textbf{Session 2
					(objective-score)}\\[0.5ex]
				\hline 
				
				\multirow{3}{*}{tFC} & Estimate & 0.94 & 0.82\\ 
				& t value & 1.60 \textit{(p = 0.11)} & 1.94 \textit{(p = 0.23)} \\[0.5ex]
				\hline 
				
				\multirow{3}{*}{tFD} & Estimate & 0.001 & 0.002 \\
				& t value & 0.96 \textit{(p = 0.33)} & 1.73 \textit{(p = 0.18)} \\[0.5ex]
				\hline 
				
				\multirow{3}{*}{FFD} & Estimate & -0.002 & -0.006 \\
				& t value & -0.50 \textit{(p = 0.61)} & -1.41 \textit{(p = 0.16)} \\[0.5ex]
				\hline 
				
				\multirow{3}{*}{LFD} & Estimate & 0.001 & 0.000 \\
				& t value & 1.09 \textit{(p = 0.27)} & 1.20 \textit{(p = 0.23)} \\[0.5ex]
				\hline 
				
				\multirow{3}{*}{SFD} & Estimate & -0.009 & -0.000 \\
				& t value & -1.89 \textit{(p = 0.16)} & -0.23 \textit{(p = 0.81)} \\[0.5ex]
				\hline 
				
				\multicolumn{4}{l}{\textsuperscript{*}: $p<0.05$, \textsuperscript{**}: $p<0.01$, \textsuperscript{***}: $p<0.001$ }\\[0.5ex]
				\hline 
			\end{tabular}}}
			\label{tab:Chapter-2tab14}
		\end{table}

In the table, t-value shows that fixation features were not a predictor of the objective score in the present study. T-value also shows that fixation features has no significant effects on the score. Thus, we got evidence that by measuring only eye-movement features of readers, their obtained score could not be predicted. However, some researchers have used eye movements during reading as a tool for second language proficiency assessment \cite{berzak2018assessing}.

\section{Conclusion}

This chapter presented an in-depth reading classification system based on eye movement feature data recorded during repeated reading sessions. The proposed method labelled the gaze data by classifying them into two labels-- high and low showing participants' in-depth reading efforts in terms of eye movements. The comparative performance of three machine learning techniques i.e. SVM, DT and MLP showed that, among these three classifiers SVM performed outstandingly.

The feature analysis showed that eye movement characteristics depend on text types (narrative and descriptive), their structures (setting, plot, conflict, and resolution) as well as reading frequency.

The LMER models showed, features of fixation, saccade and regression were statistically significant in repeated reading practice. However, among the three gaze events, only fixation was statistically effected from first day to successor days during repeated reading of complex and descriptive text. Also, only fixation and regression statistically showed the difference between the eye movement of participants groups labelled as high and low. We found that only regression was statistically effected than fixation in repeated reading of both texts. In both sessions, the gaze event was statistically lower in the low group participants than the high group. 

The analysis of the LMER models showed that features of all three events– fixation, saccade and regression are significantly effected during repeated reading in both sessions. We found that, fixation and regression were significantly effected the group in both sessions’ readings. At last, we found that there was no significant relation between participants' fixation features and their achieved scores.

The future work of this study could be to develop a regression-based machine learning system to predict the quantity of readers' reading efforts from reading gaze data. Also, more complex 2-way and 3-way interaction in LMER models could be analysed, such as the effect of specific trial day preference could enhance the preference of specific eye movement features  during specific type of readings (e.g., extensive and intensive) with various fixed variables such as gender, age-group, reading frequency, the gap between readings, various L1/L2 language, ethnicity, social status etc.

\chapter{Influence of Psycholinguistics Factors on Reading}
\label{Chapter3} 
\lhead{Chapter 3. \emph{Influence of Psycholinguistics Factors on Reading}} 

\section{Introduction}

In this chapter, we present the psycholinguistics analysis of EFL readers' eye movements during in-depth reading in repeated reading practice. In the field of eye-movement studies related to reading, two types of researches are progressing. In the first type, researchers are interested to understand the effects of relatively low level visual and linguistic factors on eye-movement control \cite{liao2017classification, mishra2018scanpath, reichle2003ez, rayner1998eye, hyona1990eye, kaakinen2007perspective, haikio2015role}, some of them are already discussed in the previous chapter-\ref{Chapter2}. In the second type, researchers are interested to make inferences about higher-order psycholinguistics processes underlying written language comprehension. These researchers are concerned with how psycholinguistics factors influence the reading behaviour of learners.

In this chapter, we explain some of these factors and also analyse their effects in eye movement trajectory generated during in-depth reading. The chapter first describes the context, i.e. some important psycholinguistics factors such as, word frequency, age-of-acquisition, familiarity or meaningfulness, imagery, concreteness and emotion. Then the correlation between these factors and the gaze features extracted from eye movements are explained. Next, we present the results of linear mixed-effects regression to analyse the effects of these factors on gaze features of repeated reading. At last, we conclude the outcomes.


\section{Role of Psycholinguistics Factors on Reading}

Psycholinguistics can be defined as the study of mental representations and processes involved in language use, including the production, comprehension, and storage of spoken and written language. It includes two disciplines-- cognitive psychology and linguistics. Cognitive psychology explores the workings of the human mind; whereas linguistics explores the nature of human language. Cognitive psychology views learning as an active and selective process. The process of language acquisition through reading is tied to the interaction of the human mind with the outside world. Linguistics, specifically transformational-generative grammar, classifies language into two major aspects-- surface structure and deep structure. The surface structure is the visual presentation of the text; whereas deep structure is the way of generating the meaning of the text. The bridge between the surface structure and the deep structure of language is syntax. A basic working model of the linguistic system applied to reading is explained here. The brain directs the eyes to pick up visual information from the configurations on the page; once the information starts coming into the brain, the brain processes it for meaning using its prior knowledge of the language (e.g., syntactic rules) and content. The outcome of the process is the identification of meaning. Psycholinguistics, then, combines cognitive psychology and linguistics in order to analyse and understand the language and thinking process, including reading, as it occurs in humans.

``The traditional psycholinguistics theories used the term mental lexicon to describe the mental store for words: the orthography (spelling), phonology (pronunciation) and semantic (meaning) information about words. Generally, orthography and phonology refer to the form of the word. These theories stated that words were represented as lexical entries containing that words orthographic and phonological information (stored separately) in the mental lexicon. The mental lexicon was the mental store of all the words a reader knows where there was one lexical entry for every word known by the reader'' \cite{shahid2014using}. The lexical processing of a text can be affected by several psycholinguistics factors such as word frequency, age-of-acquisition, imagery, familiarity, concreteness and emotion.

\subsection{Word Frequency}

Word frequency is one of the important factors that influence word processing during reading. It has been found to influence lexical decision and naming performance, fixation duration and eye movement measures. High-frequency words (e.g., music) are recognised more quickly than low-frequency words (e.g., waltz), and it is called the word frequency effect (FE). Word frequency (i.e., how often a word is encountered in natural language) has a direct effect on how long it takes a reader to process a word. For example, when readers read below two sentences (i) and (ii), they show longer fixation durations on low- (sent. ii) than high-frequency words (sent. i) during silent reading. \\
\textit{(i) The slow \textbf{music} captured her attention. (ii) The slow \textbf{waltz} captured her attention.}

In psycholinguistics models, The influence of word frequency on lexical access shows its fundamental relationship to the organisation of the mental lexicon. Therefore, the rereading of high-frequency words improves their lexical representations in mind. However, the relationship between word frequency and lexical access time is not straightforward. The rereading of low-frequency words leads to reduce their lexical access time more quickly as compared to the rereading of high-frequency words, where lexical access time is reduced slowly.

Some researchers \cite{reingold2010time} have reported that frequency effects in first fixation duration on a word, which means frequency influences early processing in children's reading, similar as in proficient readers. Also, the differences in eye-movements in school children are comparatively much larger than the differences of college students' eye movements.

Regarding the word frequency effect in second language acquisition, implicit learning models predict reduced the differences of the word frequency effects between L1 and L2 for proficient bilinguals; whereas poor bilinguals have larger the word frequency effects in L2, as compared with the word frequency effects in proficient bilinguals.

Whitford and Debra \cite{whitford2012second} reported ``Both, early-stage eye movement measures (first-fixation duration, gaze duration, skipping rate) as well as late-stage eye movement measures (proportion of regressions, total reading time) of reading revealed larger word frequency effects in L2 than L1, suggesting that word frequency had a more robust effect during initial L2 lexical access.''
\subsection{Age of Acquisition}

Age-of-acquisition (AoA) is the age at which a word was learned. The age-of-acquisition effect indicates that lexical learning is partially age-dependent, i.e., learners learn new words regularly, but there is an advantage for words learned earlier. The words learned first are easier to recall than the words learned later. Also, the order, in which words are learned, influences their activation-speed of forming mental representations, irrespective of their encountered frequency. Age-of-acquisition for words and pictures is an important factor for performing cognitive tasks such as picture naming, object recognition, word naming, lexical decision tasks, eye fixation, face recognition etc. 

The AoA effect can be understood through two theories. First, the semantic theory explains that when new concepts are learned, they are linked to the ones already in the mental network model; thus early-learned words become more central and better connected in the network for making them more easily accessible. The second (connectionist networks) explains, the word-items that were trained first always has an advantage over later-trained word-items; because the lexicon information, that enters a mental network first, gets benefits more from the plasticity of the network, and accordingly alters its connections, or adjusted weights. As new information keeps entering the network, the network reduces plasticity as well as reduces the flexibility of weight adjustment. Thus, early learned concepts have a larger impact on the mental network's final structure.
\subsection{Familiarity}

Juhasz and Barbara \cite{juhasz2003investigating} stated that ``word frequency estimates may mislead to false accuracy, especially for low-frequency words. Two words can have the same frequency in a lingual corpus yet vary in terms of subjective familiarity. Familiarity is a subjective rating measured by some participants to label how familiar (meaningful) they are with words. Several words, having the same word-frequency, can vary on subjective familiarity rating. Subjective familiarity has a significant effect on eye fixation durations for low-frequency words.'' Therefore, it is an important variable to investigate the effect of the word recognition process through reading eye movements data. 

Chaffin et al. \cite{chaffin2001learning} stated that ``during a reading task when readers' eye movements were monitored as they read pairs of sentences containing a target word from one of three subjective familiarity conditions-- high familiar, low familiar, or novel (unknown). The eye movements features including first fixation duration, gaze duration, and spillover effects disclosed that readers spent more time on less familiar words than on more familiar words. Also, when they encountered an unknown word, first fixation and gaze duration were similar to a known low familiar word." There is a strong relationship between the word frequency effect and subjective word familiarity rating on word processing in reading. Readers spent less time on high-frequency words than on low-frequency words; similarly, they spent less time on words that were rated as more familiar than on words that were rated as less familiar.

\subsection{Imagery}

Imagery (Imageability) refers to the ease with which a word gives rise to a sensory mental image. This factor is a type of semantic factor because it taps into the semantic aspect of meaning which varies from word to word. In imagery, words such as `tree' or `computer', are rated as high in imagery, whereas words such as `begin' or `trust' are rated as low. 

Previous studies have shown that high-imagery words are processed faster and more accurately than low-imagery words. This is called the imagery effect, and it influences the storage and processing of words in the mental lexicon, together with other psycholinguistics factors such as age-of-acquisition, frequency, familiarity and phonological properties. Several studies mention the imagery effects in various cognitive tasks including lexical decision, word production, and word recognition memory. Imagery effects have also been seen in tasks that use sentences. The theory of imagery effects states that, in readers, lexical items (words) for which a visual mental image is more easily formed are also more easily memorised and accessed than forms that are harder to visualise.

Holmes and Langford \cite{holmes1976comprehension} reported that readers can recall sentences having low-imagery words less accurately than sentences containing high-imagery words. Howard \& Franklin \cite{howard1988missing} found lower imagery ratings for verbs (e.g., `watch') than for nouns (e.g., `watch') in English, even when they shared the same word form. Gentner \cite{gentner1982nouns} argued that ``nouns are acquired early because they mostly denote concrete objects that are stable through time, allowing for a strong mapping between meanings and perceived form as seen in the world, whereas verbs are harder to memorise because they generally represent transient and abstract events". Similarly, Langacker \cite{langacker1987foundations} argued that ``nouns are prototypically objects, and thus conceptually autonomous and self-contained, whereas verbs, prototypically events, cannot be conceptualised independently". Hansen \cite{hansen2017makes} stated that ``Imagery is highly correlated with concreteness, but also depends on experience-- words with strong emotional connotations may be abstract, but quite imageable (e.g. `anger'), whereas words denoting rare objects are concrete, but may still be low on Imagery (e.g. `antitoxin')”.

\subsection{Concreteness}

Concreteness has been defined as the ability to see, hear and touch something. ``Words referring to objects are instructed to be given a high concreteness rating, and words referring to abstract concepts (which cannot be experienced by the senses) are instructed to be given a low concreteness rating. The concreteness effect refers to concrete words (e.g., `aeroplane') being more easily recognised (processed more quickly and accurately) than abstract words (e.g., `truth')." \cite{shahid2014using}

Schwanenflugel et al. \cite{schwanenflugel1988context} presented the context availability theory, according to which ``comprehension of a given word is dependent on verbal context which can be either from the preceding discourse or by the readers stored schemas in long-term semantic memory". Therefore, concrete words are easier to process because they are related to strongly supporting memory contexts, whereas abstract words are not.

Although often correlated with concreteness, imagery is not a redundant property. While most abstract things are hard to visualise, some call up images, e.g., `torture' calls up an emotional and even visual image. There are concrete things that are hard to visualise too, for example, `cloister' is harder to visualise than `house'.

Juhasz and Rayner \cite{juhasz2003investigating} reported the concreteness effect on fixation durations in natural reading. In their experiment, the materials were 72 nouns used as targets presented in single line sentences so that they were not the first or the last two words in the sentence. Results from this study showed that concreteness significantly predicted first fixation duration, gaze duration, and total time.
\subsection{Emotion}

Emotion has a significant effect on the cognitive processes in learners, including perception, attention, learning, memory, reasoning, and problem-solving. Grühn \cite{gruhn2016english} stated ``the influences of emotion on learning and memory have been studied and measured for remembering positive, negative, and neutral words. Negative words generally have a lower frequency in linguistic corpora including written and spoken text than positive words. Given that word frequency is a memory-relevant characteristic, ignoring frequency when selecting positive, negative, and neutral words for a memory task may produce spurious effects that might be attributed to valence rather than frequency." In the emotion domain, words are rated on mostly two primary dimensions-- valence and arousal. Kuperman et al. \cite{kuperman2014emotion} stated that ``Arousal is the extent to which a stimulus is calming or exciting, whereas valence is the extent to which a stimulus is negative or positive. These two dimensions are theoretically orthogonal: Negative stimuli can be either calming (e.g., `dirt') or exciting (e.g., `snake'), and positive stimuli can also be calming (e.g., `sleep') or exciting (e.g., `sex')." Emotion words are characterised as possessing high arousal and extreme valence. However, emotionality is not the same as valence or arousal. For example, `love' and `anger' are highly emotional words; but, they are on opposite ends of the valence dimension and `anger' is more arousing than `love'.

For measuring the impact of emotion on word recognition, Scott et al. \cite{scott2012emotion} found ``eye movements were monitored as participants read sentences containing an emotionally positive (e.g., `lucky'), negative (e.g., `angry'), or neutral (e.g., `plain') word. Fixation time demonstrated significant effects of emotion and word frequency. In detail, the fixation times on emotion words (positive or negative) were consistently more quickly than those on neutral words. However, high-frequency negative words were read slower than neutral words. These effects emerged in the earlier eye movement measures (first fixation duration), indicating that emotionality (in terms of arousal and valence) modulates lexical processing during reading comprehension."


\section{Problem}

As explained in the previous chapter- \ref{Chapter2}, we conducted an eye-tracking experiment, in which thirty EFL participants attended two repeated reading sessions for reading the same two texts in consecutive three days. The readings of text-1 (Appendix- \ref{AppendixA1}) and text-2 (Appendix- \ref{AppendixA2}) provide the experience of extensive reading and intensive reading respectively.

In this chapter, using the gaze data collected in the experiment, we analyse the impact of psycholinguistics factors on eye-movement features of in-depth reading. We propose a hypothesis that there is a correlation between fixation features and the ratings of words on different psycholinguistics factors. We also use a linear mixed-effects regression method for measuring the significance of these factors across the trials (days) of the repeated reading experiment.


\section{Words Rating on Psycholinguistics Factors}

Several researchers have proposed different rating-scores for the psycholinguistics factors including, word-frequency, age-of-acquisition, familiarity, imagery, concreteness, and emotion for English words as well as for other languages. The Medical Research Council (MRC) psycholinguistic database provided subjective rating scores for word-frequency, age-of-acquisition, familiarity, imagery and concreteness of English words \cite{wilson1988mrc}. Warriner et al. \cite{warriner2013norms} provided subjective emotional ratings of nearly 14,000 English words. For this, they used three components of emotions-- valence (the pleasantness of a stimulus), arousal (the intensity of emotion provoked by a stimulus), and dominance (the degree of control exerted by a stimulus). Rofes et al. \cite{rofes2018imageability} provided subjective imagery ratings of words of different languages, such as Basque, Catalan, Croatian, English, Greek, Hungarian, Italian, Norwegian, Serbian, Spanish, Swedish and Turkish. Miklashevsky \cite{miklashevsky2018perceptual} presented normative data on the modality (visual, auditory, haptic, olfactory, and gustatory) ratings, vertical spatial localization of the object, manipulability, imagery, age-of-acquisition, and subjective frequency for 506 Russian nouns. Using rating tasks, Khwaileh et al. \cite{khwaileh2018imageability} established norms for imagery, age-of-acquisition, and familiarity, as well as included linguistic factors such as syllable length and phoneme length for 330 Arabic words using subjective rating tasks. Juhasz et al. \cite{juhasz2015database} presented subjective ratings, which were collected on 629 English compound words for six factors including, familiarity, age-of-acquisition (AoA), semantic transparency, lexeme meaning dominance (LMD), imagery, and sensory experience ratings (SER). Desrochers and Thompson \cite{desrochers2009subjective} collected subjective frequency and imagery estimation for a sample of 3,600 French nouns from two independent groups of 72 young adults each. Brysbaert et al. \cite{brysbaert2014concreteness} presented concreteness ratings for 37,058 English words and 2,896 two-word expressions (such as `zebra crossing' and `zoom in'), obtained from over 4,000 participants by means of a norming study using internet crowdsourcing for data collection. Ljubešić et al. \cite{ljubevsic2018predicting} predicted concreteness and imagery of words within and across languages via supervised learning, using collections of cross-lingual word embeddings as explanatory variables and reported that the notions of concreteness and imagery were highly predictable both within and across languages, with a moderate loss of up to 20\% in correlation when predicting across languages. Grühn \cite{gruhn2016english} provided a database of English EMOtional TErms (EMOTE) containing subjective ratings for 1287 nouns and 985 adjectives. Nouns and adjectives were rated on valence, arousal, emotionality, concreteness, imagery, familiarity, and clarity of meaning; in addition, adjectives were rated on control, desirability, and likeableness.

\subsection{Psycholinguistics Ratings for Experiment Text}

The ratings of the content words (noun, adjective, verb, and adverb) found in the texts-- 1 and 2 for word frequency, age-of-acquisition, imagery, concreteness, familiarity, and emotion were calculated as below.

The ratings of the content words for word-frequency, concreteness, and emotion were obtained from three databases- the \textbf{MRC database} \cite{wilson1988mrc}, the database provided by \textbf{Brysbaert et al.} \cite{brysbaert2014concreteness} and the \textbf{EMOTE database} \cite{gruhn2016english} respectively. In all these cases, obtained ratings were rescaled on a 7-point scale and then were rounded to the nearest integer so that the ratings would be presented as integers on a scale from 1 to 7. 

In line with these ratings, we collected ratings of the content words for age-of-acquisition, imagery, and familiarity from EFL readers. For age-of-acquisition and imagery ratings, three PhD students participated to rate content words using a 7-point scale. For obtaining the age-of-acquisition ratings, the participants were instructed to rate words as to the age at which they had learned a word in school. Any word which was learned in early childhood was given a high AoA rating; any word that was learned later in higher classes was given a low AoA rating (from 7-rating for age 2–4 years to 1-rating for age 18 years or more). The imagery ratings were obtained following the instructions given by Paivio et al. \cite{paivio1968concreteness}. The participants were instructed to rate words as ``the ease or difficulty with which words aroused mental images". Any word which aroused a mental image (or, any sensory experience) very quickly and easily was given a high imagery rating; whereas a word that aroused a mental image with difficulty or not at all was given a low imagery rating. For example, a word- `ox' arouses an image easily and so was rated as high imagery (7-rating); another word- `whew' does hardly therefore, it was rated as low imagery (1-rating). For obtaining the familiarity ratings, three undergraduate students, enrolled in STEM discipline, participated to rate content words on a 7-point scale. These students were instructed to rate words as to the frequency at which a word is seen in daily life. Any word that they had seen very often was given a high familiarity rating; any word that they had never seen was given a low familiarity rating (from 1-rating for a word- `languish' to 7-rating for a word- `town'). In these three cases, the mean of the raters' ratings was calculated and rounded to the nearest integer (maximum value-7, minimum value-1).

The content words, which were not found in the rating database (i.e. `out of rating words') of a psycholinguistics factor were excluded. Table \ref{tab:Chapter-3tab1} show the distribution of content words (in percentage) of texts-- 1 and 2 on a 7-point rating scale for the six psycholinguistics factors. Since both texts contain a few words representing emotions, therefore more than 70\% of content words were excluded from the analysis of words' emotional impact on gaze features.

\begingroup
\setstretch{1.4}
{\small	
	\begin{longtable}
		{>{\raggedright}m{6.4cm} C{0.4cm} C{0.4cm} C{0.4cm} C{0.4cm} C{0.4cm} C{0.4cm} C{0.4cm} C{2.4cm}}
		\captionsetup{font={small}, labelfont=bf, skip=4pt}
		\caption{Distribution of content words on a 7-point scale}
		\label{tab:Chapter-3tab1}\\
		\toprule\midrule
			
		\textbf{7-point Rating Scale} & \textbf{1} & \textbf{2} & \textbf{3} & \textbf{4} & \textbf{5} & \textbf{6} & \textbf{7} & \textbf{out of rating words} \\
		\midrule
		\multicolumn{9}{c}{\textbf{Text-1 content words distribution (in \%)}} \\
		\midrule
		
		\textbf{Word Frequency}\\ {\footnotesize(least frequent:1 - most frequent:7)} & 38 & 23 & 15 & 7 & 4 & 6 & 3 & 4 \\
		\midrule
		\textbf{Age of Acquisition} \\{\footnotesize(higher classes:1 - early childhood:7)} & 2 & 2 & 5 & 11 & 24 & 30 & 25 & 1 \\
		\midrule
		\textbf{Familiarity}\\ {\footnotesize(least familiar:1 - most familiar:7)} & 7 & 14 & 9 & 25 & 26 & 10 & 8 & 1 \\
		\midrule
		\textbf{Imagery}\\ {\footnotesize(difficult imagination:1 - easy imagination:7)} & 2 & 15 & 18 & 21 & 24 & 11 & 8 & 1 \\
		\midrule
		\textbf{Concreteness} \\{\footnotesize(non-physical object:1 - physical object:7)} & 2 & 17 & 19 & 11 & 12 & 18 & 17 & 4 \\
		\midrule
		\textbf{Emotion}\\ {\footnotesize(mild:1 - intense:7)} & 1 & 8 & 8 & 3 & 1 & 1 & 1 & 77\\
	    \midrule
		
		\multicolumn{9}{c}{\textbf{Text-2 content words distribution (in \%)}} \\
		\midrule
		\textbf{Word Frequency}\\ {\footnotesize(least frequent:1 - most frequent:7)} & 55 & 25 & 10 & 2 & 1 & 1 & 3 & 3 \\
		\midrule
		\textbf{Age of Acquisition} \\{\footnotesize(higher classes:1 - early childhood:7)} & 12 & 18 & 9 & 3 & 13 & 16 & 27 & 2 \\
		\midrule
		\textbf{Familiarity}\\ {\footnotesize(least familiar:1 - most familiar:7)} & 28 & 12 & 13 & 10 & 16 & 12 & 7 & 2 \\
		\midrule
		\textbf{Imagery}\\ {\footnotesize(difficult imagination:1 - easy imagination:7)} & 29 & 16 & 11 & 9 & 20 & 5 & 8 & 2 \\
		\midrule
		\textbf{Concreteness} \\{\footnotesize(non-physical object:1 - physical object:7)} & 13 & 35 & 15 & 10 & 8 & 8 & 6 & 5 \\
		\midrule
		\textbf{Emotion}\\ {\footnotesize(mild:1 - intense:7)} & 1 & 4 & 3 & 3 & 6 & 7 & 3 & 73 \\
		\midrule\bottomrule
		\end{longtable}
		}\endgroup


\section{The Gaze Dataset}

To analyse the influence of psycholinguistics factors on eye-movement patterns, we used the gaze dataset collected in the repeated reading experiment, as already explained in chapter-\ref{Chapter2}. The dataset contained various features of gaze event patterns- fixations, saccades and regressions. Each participant's gaze data was labelled either as high or low by an expert, representing the participant's reading efforts in terms of eye movements.

The researchers of related domains agreed that ``among three gaze-patterns, only fixation features is closely linked with cognitive processing of words during in-depth reading". Therefore, we considered only the features related to fixation for analysis; these features are listed below:

\begin{enumerate}[(i)]
	\item \textbf{Total Fixation Duration (tFD):} the total of all fixation duration on a word.
	\item \textbf{First-Fixation Duration (FFD):} the duration of fixation during the first pass reading of a word.
	\item \textbf{Second-Fixation Duration (SFD):} the duration of fixation during the second pass reading of a word.
	\item \textbf{Last-Fixation Duration (LFD):} the duration of fixation during the third or more pass reading of a word.
	\item \textbf{Total Fixation Count (tFC):} the total count of all fixations during the multiple pass reading of a word.
\end{enumerate}


\section{Results and Discussion}

As described earlier, we analysed fixation data at the word level. To calculate the correlation between rated psycholinguistics factors and fixation features, the following steps were performed.
\begin{itemize}
	\item The content words of a text were labelled by using the rating scale of different psycholinguistics factors, as explained earlier. Unlabelled (out of rating) words were excluded from further analysis. 
	\item All labelled words were separated into seven clusters as per their ratings.
	\item As table \ref{tab:Chapter-3tab1} shows that the distribution of words on the rating scale was not uniform; therefore to reduce word distribution impact on results, the total of a fixation feature on the words of each cluster was divided by the number of words in the cluster. The output of this step gives the mean of each fixation feature on every rating point and thus, mapped a participant's fixation features with rated psycholinguistics factors.
	\item Now, for finding the final value of fixation features across the participants, the total of the previous step (the mean of a participant's fixation feature) was divided by the number of participants.
\end{itemize}

Thus, we found the value of fixation features mapped with the rating of different psycholinguistics factors. From the participants' gaze data, we observed that some words had hardly captured any fixations, whereas some other words captured enormous fixations. Therefore, the last step normalises the values of fixation features across the participants to improve the result. The algorithm to calculate mean and standard deviation of fixation features is given in algorithm-\ref{alg:Chapter-3alg1}.

\begin{algorithm}
	\caption{Mean and standard deviation of fixation features}\label{alg:Chapter-3alg1}
	\begin{algorithmic}[1]
		\Procedure{Mean \& Standard Deviation of fixation features}{}\\
		\textbf{Require:} Participants' gaze dataset and Psycholinguistics factors' word-rating sets\\
		\textbf{Output:} mean and standard deviation of five fixation features for six psycholinguistics factors
		\State \textit{PsycholingfactorDict =} $\{\}$
		\For {$ratingset \in \{Word Frequency, \dots, Emotion\}$}
		\State \textit{FixfeatDict =} $\{\}$
		\For {$fixfeat \in \{tFD, FFD, SFD, LFD, tFC\}$}:	
		\State \textit{FeatMeanList =} []
		\For {$fixations \in \{p_1gaze, \dots, p_{30}gaze\}$}:
		\State \textit{valueList =} []
		\For {$feat \in \{fixations\}$}:
		\If  {$feat.word$ not in $ratingset$}
		\State $continue$
		\Else
		\State \textit{valueList.append[feat.fixfeat]}
		\EndIf
		\EndFor
		\State \textit{FeatMeanList.append(mean(valueList))}
		\EndFor
		\State \textit{Feat-mean =  get\_mean(FeatMeanList)}
		\State \textit{Feat-sd =  get\_sd(FeatMeanList)}
		\State \textit{FixfeatDict[fixfeat]=[Feat-mean, Feat-sd]}
		\EndFor
		\State \textit{PsycholingfactorDict[ratingset] = FixfeatDict}
		\EndFor
		\State return \textit{PsycholingfactorDict}
		\EndProcedure
	\end{algorithmic}
\end{algorithm}

\subsection{Pearson's Product-moment Correlation Analysis}
It is common in statistical analysis to explore and to summarise the strength of association between two variables that can not be measured quantitatively. The most commonly used measure of association is Pearson's product-moment correlation coefficient \cite{gooch2011pearson}, often denoted by $r$. The value of $r$, always lies between -1 and +1, is a measure of any linear trend between two variables. The algorithm to calculate correlation coefficients of the five fixation features for six psycholinguistics factors is given in algorithm-\ref{alg:Chapter-3alg2}.

\begin{algorithm}
	\caption{Pearson's product-moment correlation}\label{alg:Chapter-3alg2}
	\begin{algorithmic}[1]
		\Procedure{Pearson's correlation coefficient}{}\\
		\textbf{Require:} Participants' gaze dataset and Psycholinguistics factors' word-rating sets\\
		\textbf{Output:} Correlation coefficients (\textit{r, p}) of five fixation features for six psycholinguistics factors
		\State \textit{PsycholingfactorDict =} $\{\}$
		\For {$ratingset \in \{Word Frequency, \dots, Emotion\}$}
		\State \textit{FixfeatDict =} $\{\}$
		\For {$fixfeat \in \{tFD, FFD, SFD, LFD, tFC\}$}:	
		\State \textit{ScoreMean =} []
		\For {$fixations \in \{p_1gaze, \dots, p_{30}gaze\}$}:
		\State $Scores = \{1:0, 2:0, 3:0, 4:0, 5;0, 6:0, 7:0\}$
		\For {$feat \in \{fixations\}$}:
		\If  {$feat.word$ not in $ratingset$}
		\State $continue$
		\Else
		\State \textit{rating = get\_ratingScore(feat.word, ratingset)}
		\State \textit{Scores[rating] = + feat.fixfeat}
		\EndIf
		
		\EndFor
		\State \textit{ScoreMean.append(mean(Scores))}
		\EndFor
		
		\State \textit{Score-meanList =  get\_mean(ScoreMean)}
		\State \textit{Sorted-Score-meanList =  ascending\_sort(Score-meanList)}
		\State $ratingScale =  [1, 2, 3, 4, 5, 6, 7]$
		\State \textit{r, p = get\_correlation(Sorted-Score-meanList, ratingScale)}
		\State \textit{FixfeatDict[fixfeat]=[r, p]}
		\EndFor
		\State \textit{PsycholingfactorDict[ratingset] = FixfeatDict}
		\EndFor
		\State return \textit{PsycholingfactorDict}
		\EndProcedure
	\end{algorithmic}
\end{algorithm}

\subsubsection{Correlations Between Fixation Features}
\label{sssec:3chap611}
We examined correlations among the fixation features of all six psycholinguistic factors on sessions 1 and 2 as shown in tables \ref{AppendixD1} and \ref{AppendixD2} respectively (in Appendix-\ref{AppendixD}). As shown in table-\ref{AppendixD1}, in the day-1 trial of session-1, tFC of age-of-acquisition and concreteness shows a highly significant correlation $(r>0.95, p<0.003)$ for all groups (Low, High \& Both); tFD of word-frequency and concreteness shows a significant correlation $(r>0.71, p<0.05)$ for all groups; FFD of word-frequency and familiarity shows a significant correlation $(r>0.82, p<0.01)$ for all groups; FFD of familiarity and imagery also shows a significant correlation $(r>0.7, p<0.05)$ for all groups. In day-2 trial gaze data, there is no correlation existed between the fixation features of any two psycholinguistics factors for all groups. However, in day-3 trial gaze data, there is a correlation existed between SFD of age-of-acquisition and imagery for all groups.

As shown in table-\ref{AppendixD2}, in the day-1 trial of session-2 gaze data, tFD of word-frequency and familiarity shows a highly significant correlation $(r>0.8, p<0.05)$ for all groups (Low, High \& Both); FFD of word-frequency and emotion shows a significant correlation $(r>0.79, p<0.05)$ for all groups; SFD of word-frequency and familiarity also shows a significant correlation $(r>0.82, p<0.05)$ for all groups. In days 2 and 3 trials gaze data, there is no correlation existed between the fixation features of any two psycholinguistics factors for all groups. 

Both correlation tables (\ref{AppendixD1} \& \ref{AppendixD2}) indicate that there is a significant correlation between the fixation features of word-frequency and familiarity for all groups. Thus, it supports researchers' findings that more-frequent words become more-familiar words.

\subsubsection{Correlations Between Fixations on Words and the Rating of Words under Psycholinguistics Factors}

Tables \ref{tab:Chapter-3tab2} and \ref{tab:Chapter-3tab3} show the correlation between fixation values of five fixation features on words and the ratings of those words under various psycholinguistics factors across participants' group (Low, High \& All) as well as across days (1, 2 \& 3) of sessions 1 and 2 data respectively. 

In both tables, the first column shows trial days, which were 1, 2, 3 and all three days; the second column shows fixation features' name; third, fifth and seventh columns show the mean and the standard deviation (SD) of fixation features for groups- Low, High and All (all participants of both groups-- Low \& High; it is used here as a reference) respectively; fourth, sixth and eighth columns show Pearson's product-moment correlation coefficients and significance stars for the groups- Low, High and All (all participants of both groups-- Low \& High; it is used here as a reference) respectively. The number of participants in the groups--Low \& High varied across days and sessions; details has been already given in table \ref{tab:Chapter-2tab3}. In tables \ref{tab:Chapter-3tab2} and \ref{tab:Chapter-3tab3}, correlation coefficients are mostly in negative; as the rating scale stepped up from one to seven and their fixation feature values stepped down from high to low, and thus produces a negative correlation between the words ratings and corresponding fixation feature value.

In table \ref{tab:Chapter-3tab2}, among the psycholinguistics factors, rated word-frequency is the most significant for all fixation features across groups and days. We have already found in the last subsection-\ref{sssec:3chap611} that the fixation features of word-frequency and familiarity are correlated; however, rated familiarity is comparatively less significant for all fixation features across groups and days. Rated emotion is least significant for the fixation features across groups and days because the text has very less emotional words. For rated concreteness, the fixation features have a negative correlation because the words representing non-physical objects captured more gaze during reading in compared to the words representing physical objects. For rated emotion, a few fixation features have a correlation, because only 23\% content words of text-1 found in emotion database. Most of the correlations are positive, means the words having mild emotion captured fewer gazes in compared to the words having intense emotion.

\begingroup
\setstretch{1.2}
{\footnotesize	
	\begin{longtable}
		{>{\raggedright}m{0.3cm} C{1.5cm} L{2.1cm} L{1.1cm} L{2.2cm} L{1.2cm} L{2.2cm} L{1.1cm}}
		\captionsetup{font={small}, labelfont=bf, skip=4pt}
		\caption{Pearson’s product-moment correlation on session-1 data}
		\label{tab:Chapter-3tab2}\\
		\toprule\midrule
		\textbf{\scriptsize Day} & \textbf{\scriptsize Fixation feature} & \textbf{\scriptsize Low (Mean,SD)} & \textbf{\scriptsize Low (r,p)} & \textbf{\scriptsize High (Mean,SD)} & \textbf{\scriptsize High (r,p)} & \textbf{\scriptsize All (Mean,SD)} & \textbf{\scriptsize All (r,p)}\\
		\bottomrule	
		\multicolumn{8}{l}{\textbf{1. Word Frequency:}} \\[0.5ex]
		\hline 
		\multirow{5}{*}{1} & tFD & 273.34(39.90) & -0.87\textsuperscript{**} & 344.65(72.27) & -0.77\textsuperscript{*} & 305.04(50.99) & -0.86\textsuperscript{**}\\ 
		& FFD & 207.07(31.10) & -0.70\textsuperscript{+} & 229.37(33.79) & -0.77\textsuperscript{*} & 216.98(28.80) & -0.82\textsuperscript{*}\\ 
		& SFD & 50.41(7.24) & -0.72\textsuperscript{*} & 74.65(21.71) & -0.65\textsuperscript{+} & 61.19(11.62) & -0.79\textsuperscript{*}\\ 
		& LFD & 15.86(9.99) & -0.76\textsuperscript{*} & 40.63(21.50) & -0.71\textsuperscript{+} & 26.87(12.53) & -0.87\textsuperscript{**}\\ 
		& tFC & 1.03(0.09) & -0.82\textsuperscript{*} & 1.22(0.22) & -0.79\textsuperscript{*} & 1.12(0.14) & -0.87\textsuperscript{**}\\[0.5ex]
		\hline 
		\multirow{5}{*}{2} & tFD & 283.91(38.39) & -0.91\textsuperscript{**} & 349.32(60.12) & -0.84\textsuperscript{**} & 319.59(49.30) & -0.88\textsuperscript{**}\\ 
		& FFD & 199.86(26.34) & -0.91\textsuperscript{**} & 205.80(29.17) & -0.85\textsuperscript{**} & 203.10(26.92) & -0.91\textsuperscript{**}\\ 
		& SFD & 53.89(11.69) & -0.62 & 88.53(20.15) & -0.82\textsuperscript{*} & 72.78(14.88) & -0.83\textsuperscript{*}\\ 
		& LFD & 30.16(9.49) & -0.37 & 54.99(17.36) & -0.53 & 43.70(10.91) & -0.61\\ 
		& tFC & 1.04(0.14) & -0.82\textsuperscript{*} & 1.33(0.17) & -0.78\textsuperscript{*} & 1.20(0.15) & -0.81\textsuperscript{*}\\[0.5ex]
		\hline 
		\multirow{5}{*}{3} & tFD & 252.84(50.92) & -0.92\textsuperscript{**} & 299.68(43.27) & -0.73\textsuperscript{*} & 277.64(43.58) & -0.89\textsuperscript{**}\\ 
		& FFD & 189.88(28.50) & -0.88\textsuperscript{**} & 190.26(27.98) & -0.87\textsuperscript{**} & 190.09(25.80) & -0.96\textsuperscript{***}\\ 
		& SFD & 46.89(15.41) & -0.82\textsuperscript{*} & 65.32(13.43) & -0.55 & 56.65(12.99) & -0.76\textsuperscript{*}\\ 
		& LFD & 16.06(11.46) & -0.78\textsuperscript{*} & 44.09(15.43) & -0.01 & 30.90(10.77) & -0.4\\ 
		& tFC & 0.95(0.13) & -0.83\textsuperscript{*} & 1.18(0.13) & -0.57 & 1.07(0.12) & -0.77\textsuperscript{*}\\ [0.5ex]
		\hline 
		\multirow{5}{0.3cm}{\raggedright All days} & tFD & 271.57(40.18) & -0.94\textsuperscript{**} & 334.08(54.92) & -0.86\textsuperscript{**} & 302.83(46.06) & -0.92\textsuperscript{**}\\ 
		& FFD & 200.71(26.42) & -0.88\textsuperscript{**} & 210.14(28.54) & -0.88\textsuperscript{**} & 205.43(26.50) & -0.91\textsuperscript{**}\\ 
		& SFD & 50.62(9.25) & -0.83\textsuperscript{*} & 77.15(15.90) & -0.83\textsuperscript{*} & 63.88(11.34) & -0.92\textsuperscript{**}\\ 
		& LFD & 20.24(6.96) & -0.96\textsuperscript{***} & 46.80(12.73) & -0.70\textsuperscript{+} & 33.52(9.24) & -0.84\textsuperscript{**}\\ 
		& tFC & 1.01(0.11) & -0.87\textsuperscript{**} & 1.25(0.16) & -0.81\textsuperscript{*} & 1.13(0.13) & -0.86\textsuperscript{**}\\ [0.5ex]
		\hline 
		\multicolumn{8}{l}{\textbf{2. Age of Acquisition:}} \\ [0.5ex]
		\hline 
		\multirow{5}{*}{1} & tFD & 410.89(196.70) & -0.80\textsuperscript{+} & 550.51(249.13) & -0.85\textsuperscript{*} & 472.94(219.57) & -0.83\textsuperscript{*}\\ 
		& FFD & 291.90(134.43) & -0.78\textsuperscript{+} & 369.59(193.03) & -0.81\textsuperscript{+} & 326.43(160.08) & -0.80\textsuperscript{+}\\ 
		& SFD & 82.05(36.20) & -0.89\textsuperscript{*} & 116.77(37.76) & -0.76\textsuperscript{+} & 97.48(34.83) & -0.88\textsuperscript{*}\\ 
		& LFD & 36.95(27.82) & -0.74\textsuperscript{+} & 64.14(31.44) & -0.84\textsuperscript{*} & 49.03(29.28) & -0.79\textsuperscript{+}\\ 
		& tFC & 1.19(0.14) & -0.94\textsuperscript{**} & 1.48(0.25) & -0.92\textsuperscript{**} & 1.32(0.19) & -0.94\textsuperscript{**}\\ [0.5ex]
		\hline 
		\multirow{5}{*}{2} & tFD & 427.80(171.55) & -0.86\textsuperscript{*} & 492.52(112.10) & -0.84\textsuperscript{*} & 463.11(131.30) & -0.90\textsuperscript{*}\\ 
		& FFD & 317.45(175.91) & -0.79\textsuperscript{+} & 279.56(74.10) & -0.91\textsuperscript{*} & 296.78(118.79) & -0.84\textsuperscript{*}\\ 
		& SFD & 79.07(29.28) & -0.36 & 120.26(27.87) & -0.56 & 101.54(28.26) & -0.47\\ 
		& LFD & 31.28(16.12) & 0.11 & 92.70(38.32) & -0.3 & 64.78(27.56) & -0.2\\ 
		& tFC & 1.21(0.15) & -0.72 & 1.53(0.17) & -0.61 & 1.38(0.16) & -0.67\\ [0.5ex]
		\hline 
		\multirow{5}{*}{3} & tFD & 403.61(141.49) & -0.83\textsuperscript{*} & 442.77(184.17) & -0.82\textsuperscript{*} & 424.34(155.79) & -0.87\textsuperscript{*}\\ 
		& FFD & 256.96(64.08) & -0.89\textsuperscript{*} & 293.92(170.93) & -0.76\textsuperscript{+} & 276.53(116.72) & -0.82\textsuperscript{*}\\ 
		& SFD & 99.63(49.09) & -0.88\textsuperscript{*} & 103.11(30.88) & -0.95\textsuperscript{**} & 101.48(39.08) & -0.91\textsuperscript{*}\\ 
		& LFD & 47.02(36.90) & -0.46 & 45.74(24.35) & 0.29 & 46.34(26.31) & -0.16\\ 
		& tFC & 1.16(0.20) & -0.68 & 1.31(0.12) & -0.63 & 1.24(0.16) & -0.65\\ [0.5ex]
		\hline 
		\multirow{5}{0.3cm}{\raggedright All days} & tFD & 414.25(169.56) & -0.85\textsuperscript{*} & 500.04(171.75) & -0.89\textsuperscript{*} & 457.14(170.37) & -0.87\textsuperscript{*}\\ 
		& FFD & 291.17(127.95) & -0.81\textsuperscript{+} & 316.21(142.25) & -0.82\textsuperscript{*} & 303.69(134.94) & -0.81\textsuperscript{*}\\ 
		& SFD & 85.40(32.11) & -0.88\textsuperscript{*} & 114.32(29.13) & -0.83\textsuperscript{*} & 99.86(30.32) & -0.86\textsuperscript{*}\\ 
		& LFD & 37.67(17.74) & -0.73\textsuperscript{+} & 69.51(20.98) & -0.56 & 53.59(18.67) & -0.66\\ 
		& tFC & 1.19(0.15) & -0.87\textsuperscript{*} & 1.46(0.16) & -0.90\textsuperscript{*} & 1.32(0.15) & -0.88\textsuperscript{*}\\ [0.5ex]
		\hline 
		\multicolumn{8}{l}{\textbf{3. Familiarity:}} \\ [0.5ex]
		\hline 
		\multirow{5}{*}{1} & tFD & 304.24(34.39) & -0.88\textsuperscript{**} & 395.39(51.87) & -0.86\textsuperscript{**} & 344.75(40.90) & -0.89\textsuperscript{**}\\ 
		& FFD & 223.53(21.30) & -0.85\textsuperscript{**} & 264.31(33.63) & -0.83\textsuperscript{**} & 241.65(24.86) & -0.91\textsuperscript{**}\\ 
		& SFD & 55.31(9.41) & -0.63\textsuperscript{+} & 86.21(13.65) & -0.76\textsuperscript{*} & 69.04(10.63) & -0.75\textsuperscript{*}\\ 
		& LFD & 25.40(10.80) & -0.58 & 44.87(11.35) & -0.53 & 34.05(10.51) & -0.58\\ 
		& tFC & 1.07(0.06) & -0.28 & 1.30(0.06) & -0.32 & 1.17(0.06) & -0.29\\ [0.5ex]
		\hline 
		\multirow{5}{*}{2} & tFD & 331.06(51.29) & -0.86\textsuperscript{**} & 418.94(68.31) & -0.82\textsuperscript{**} & 379.00(59.53) & -0.85\textsuperscript{**}\\ 
		& FFD & 232.07(30.77) & -0.91\textsuperscript{**} & 231.48(21.72) & -0.74\textsuperscript{*} & 231.75(22.39) & -0.96\textsuperscript{***}\\ 
		& SFD & 68.04(18.59) & -0.74\textsuperscript{*} & 106.18(24.61) & -0.70\textsuperscript{*} & 88.85(21.60) & -0.72\textsuperscript{*}\\ 
		& LFD & 30.96(15.87) & -0.14 & 81.28(25.85) & -0.87\textsuperscript{**} & 58.41(18.90) & -0.70\textsuperscript{*}\\ 
		& tFC & 1.13(0.11) & -0.33 & 1.46(0.13) & -0.62\textsuperscript{+} & 1.31(0.12) & -0.51\\ [0.5ex]
		\hline 
		\multirow{5}{*}{3} & tFD & 320.20(69.10) & -0.73\textsuperscript{*} & 347.73(52.79) & -0.88\textsuperscript{**} & 334.77(56.50) & -0.86\textsuperscript{**}\\ 
		& FFD & 216.95(24.52) & -0.63\textsuperscript{+} & 215.61(23.35) & -0.85\textsuperscript{**} & 216.24(19.91) & -0.89\textsuperscript{**}\\ 
		& SFD & 67.01(21.02) & -0.74\textsuperscript{*} & 79.65(15.78) & -0.72\textsuperscript{*} & 73.70(17.03) & -0.78\textsuperscript{*}\\ 
		& LFD & 36.24(30.24) & -0.65\textsuperscript{+} & 52.47(20.70) & -0.74\textsuperscript{*} & 44.83(22.28) & -0.78\textsuperscript{*}\\ 
		& tFC & 1.07(0.16) & -0.41 & 1.25(0.06) & -0.53 & 1.16(0.10) & -0.47\\ [0.5ex]
		\hline 
		\multirow{5}{0.3cm}{\raggedright All days} & tFD & 316.24(44.31) & -0.89\textsuperscript{**} & 390.95(53.47) & -0.92\textsuperscript{***} & 353.59(48.09) & -0.92\textsuperscript{***}\\ 
		& FFD & 224.52(21.70) & -0.95\textsuperscript{***} & 239.09(23.21) & -0.92\textsuperscript{***} & 231.80(21.45) & -0.98\textsuperscript{***}\\ 
		& SFD & 62.00(13.76) & -0.77\textsuperscript{*} & 91.68(16.52) & -0.79\textsuperscript{*} & 76.84(15.01) & -0.79\textsuperscript{*}\\ 
		& LFD & 29.71(13.07) & -0.63\textsuperscript{+} & 60.18(16.62) & -0.88\textsuperscript{**} & 44.95(14.34) & -0.80\textsuperscript{*}\\ 
		& tFC & 1.09(0.09) & -0.39 & 1.35(0.07) & -0.63\textsuperscript{+} & 1.22(0.08) & -0.52\\ [0.5ex]
		\hline 
		\multicolumn{8}{l}{\textbf{4. Imagery:}} \\ [0.5ex]
		\hline 
		\multirow{5}{*}{1} & tFD & 295.64(39.88) & -0.73\textsuperscript{*} & 381.75(68.07) & -0.70\textsuperscript{*} & 333.91(51.40) & -0.73\textsuperscript{*}\\ 
		& FFD & 212.26(21.95) & -0.90\textsuperscript{**} & 246.16(45.76) & -0.87\textsuperscript{**} & 227.33(31.29) & -0.92\textsuperscript{**}\\ 
		& SFD & 55.57(17.02) & -0.73\textsuperscript{*} & 86.84(19.06) & -0.45 & 69.47(16.74) & -0.64\textsuperscript{+}\\ 
		& LFD & 27.80(11.47) & 0.26 & 48.75(15.57) & 0.04 & 37.11(13.18) & 0.15\\ 
		& tFC & 1.11(0.12) & -0.18 & 1.31(0.14) & -0.35 & 1.20(0.13) & -0.27\\ [0.5ex]
		\hline 
		\multirow{5}{*}{2} & tFD & 288.26(61.94) & -0.84\textsuperscript{**} & 392.87(52.11) & -0.61\textsuperscript{+} & 345.32(51.33) & -0.80\textsuperscript{*}\\ 
		& FFD & 211.81(26.97) & -0.90\textsuperscript{**} & 224.25(26.96) & -0.66\textsuperscript{+} & 218.60(25.00) & -0.83\textsuperscript{**}\\ 
		& SFD & 52.53(23.18) & -0.80\textsuperscript{*} & 99.24(14.97) & -0.46 & 78.01(16.06) & -0.76\textsuperscript{*}\\ 
		& LFD & 23.92(14.29) & -0.64\textsuperscript{+} & 69.37(15.10) & -0.46 & 48.71(12.23) & -0.65\textsuperscript{+}\\ 
		& tFC & 1.05(0.19) & -0.71\textsuperscript{*} & 1.43(0.13) & -0.01 & 1.26(0.13) & -0.49\\ [0.5ex]
		\hline 
		\multirow{5}{*}{3} & tFD & 289.72(48.41) & -0.83\textsuperscript{**} & 328.35(48.79) & -0.55 & 310.17(44.40) & -0.75\textsuperscript{*}\\ 
		& FFD & 196.74(34.49) & -0.81\textsuperscript{**} & 208.36(23.52) & -0.64\textsuperscript{+} & 202.89(25.10) & -0.85\textsuperscript{**}\\ 
		& SFD & 59.77(17.60) & -0.81\textsuperscript{**} & 4.10(17.83) & -0.79\textsuperscript{*} & 67.36(16.83) & -0.84\textsuperscript{**}\\ 
		& LFD & 33.20(14.77) & 0.15 & 45.89(14.43) & 0.15 & 39.92(11.78) & 0.19\\ 
		& tFC & 1.07(0.12) & 0.09 & 1.27(0.14) & 0.1 & 1.18(0.12) & 0.1\\ [0.5ex]
		\hline 
		\multirow{5}{0.3cm}{\raggedright All days} & tFD & 291.96(44.58) & -0.87\textsuperscript{**} & 371.23(54.06) & -0.67\textsuperscript{*} & 331.60(47.81) & -0.78\textsuperscript{*}\\ 
		& FFD & 208.36(25.28) & -0.92\textsuperscript{**} & 227.88(30.87) & -0.81\textsuperscript{**} & 218.12(27.11) & -0.89\textsuperscript{**}\\ 
		& SFD & 55.67(17.26) & -0.85\textsuperscript{**} & 87.87(15.62) & -0.60\textsuperscript{+} & 71.77(15.51) & -0.78\textsuperscript{*}\\ 
		& LFD & 27.93(6.96) & -0.12 & 55.47(12.69) & -0.13 & 41.70(9.51) & -0.13\\ 
		& tFC & 1.08(0.11) & -0.44 & 1.34(0.13) & -0.1 & 1.21(0.11) & -0.26\\ [0.5ex]
		\hline 
		\multicolumn{8}{l}{\textbf{5. Concreteness:}}  \\ [0.5ex]
		\hline 
		\multirow{5}{*}{1} & tFD & 302.72(44.90) & -0.67\textsuperscript{+} & 374.51(38.19) & -0.90\textsuperscript{**} & 334.62(38.64) & -0.83\textsuperscript{**}\\ 
		& FFD & 221.30(29.12) & -0.72\textsuperscript{*} & 244.67(23.88) & -0.59\textsuperscript{+} & 231.69(25.21) & -0.71\textsuperscript{*}\\ 
		& SFD & 58.87(15.75) & -0.67\textsuperscript{+} & 85.47(18.00) & -0.75\textsuperscript{*} & 70.70(15.91) & -0.74\textsuperscript{*}\\ 
		& LFD & 22.54(11.87) & 0.15 & 44.36(14.87) & -0.47 & 32.24(10.66) & -0.2\\ 
		& tFC & 1.11(0.11) & -0.54 & 1.31(0.09) & -0.81\textsuperscript{**} & 1.19(0.09) & -0.74\textsuperscript{*}\\ [0.5ex]
		\hline 
		\multirow{5}{*}{2} & tFD & 327.42(48.79) & -0.66\textsuperscript{+} & 392.92(28.89) & -0.65\textsuperscript{+} & 363.15(36.75) & -0.67\textsuperscript{*}\\ 
		& FFD & 230.55(40.85) & -0.51 & 221.97(8.34) & -0.47 & 225.87(22.44) & -0.51\\ 
		& SFD & 64.57(10.92) & -0.61\textsuperscript{+} & 98.84(11.67) & -0.44 & 83.26(9.78) & -0.60\textsuperscript{+}\\ 
		& LFD & 32.31(11.24) & -0.43 & 72.11(17.64) & -0.54 & 54.02(12.49) & -0.59\textsuperscript{+}\\ 
		& tFC & 1.14(0.08) & -0.75\textsuperscript{*} & 1.44(0.14) & -0.54 & 1.30(0.11) & -0.65\textsuperscript{+}\\ [0.5ex]
		\hline 
		\multirow{5}{*}{3} & tFD & 284.57(25.25) & 0.21 & 326.05(30.77) & -0.86\textsuperscript{**} & 306.53(18.90) & -0.61\textsuperscript{+}\\ 
		& FFD & 206.56(22.80) & -0.31 & 202.09(17.95) & -0.52 & 204.19(14.28) & -0.58\\ 
		& SFD & 54.51(10.06) & 0.4 & 79.54(14.43) & -0.88\textsuperscript{**} & 67.76(7.72) & -0.63\textsuperscript{+}\\ 
		& LFD & 23.50(16.92) & 0.5 & 44.42(8.36) & -0.52 & 34.57(8.77) & 0.19\\ 
		& tFC & 1.01(0.09) & 0.41 & 1.27(0.10) & -0.91\textsuperscript{**} & 1.15(0.05) & -0.56\\ [0.5ex]
		\hline 
		\multirow{5}{0.3cm}{\raggedright All days} & tFD & 305.80(35.78) & -0.62\textsuperscript{+} & 367.99(29.64) & -0.90\textsuperscript{**} & 336.89(30.51) & -0.80\textsuperscript{*}\\ 
		& FFD & 220.53(29.65) & -0.59\textsuperscript{+} & 224.81(13.90) & -0.65\textsuperscript{+} & 222.67(20.38) & -0.65\textsuperscript{+}\\ 
		& SFD & 59.54(8.57) & -0.68\textsuperscript{*} & 88.72(13.41) & -0.76\textsuperscript{*} & 74.13(10.74) & -0.75\textsuperscript{*}\\ 
		& LFD & 25.73(10.31) & 0.13 & 54.46(11.97) & -0.60\textsuperscript{+} & 40.10(8.82) & -0.33\\ 
		& tFC & 1.09(0.07) & -0.51 & 1.34(0.09) & -0.82\textsuperscript{**} & 1.22(0.08) & -0.77\textsuperscript{*}\\ [0.5ex]
		\hline 
		\multicolumn{8}{l}{\textbf{6. Emotion:}} \\ [0.5ex]
		\hline 
		\multirow{5}{*}{1} & tFD & 271.79(110.56) & 0.18 & 299.21(135.40) & 0.22 & 283.98(119.45) & 0.21\\ 
		& FFD & 197.37(64.71) & 0.25 & 223.75(103.61) & 0.34 & 209.09(80.69) & 0.3\\ 
		& SFD & 55.14(39.24) & 0.12 & 51.56(30.35) & 0.03 & 53.55(31.80) & 0.1\\ 
		& LFD & 19.28(16.85) & -0.06 & 23.90(16.15) & -0.22 & 21.34(16.27) & -0.19\\ 
		& tFC & 0.97(0.32) & 0.09 & 1.09(0.37) & 0.33 & 1.02(0.32) & 0.21\\ [0.5ex]
		\hline 
		\multirow{5}{*}{2} & tFD & 260.72(137.77) & 0.05 & 355.48(174.57) & 0.36 & 312.41(153.90) & 0.24\\ 
		& FFD & 205.47(112.27) & 0.17 & 209.21(68.54) & 0.66 & 207.51(83.58) & 0.4\\ 
		& SFD & 42.37(28.40) & -0.2 & 85.51(63.28) & 0.15 & 65.90(44.35) & 0.06\\ 
		& LFD & 12.89(7.73) & -0.38 & 60.76(56.91) & 0.14 & 39.00(32.37) & 0.04\\ 
		& tFC & 0.92(0.36) & 0.05 & 1.21(0.45) & 0.35 & 1.08(0.39) & 0.24\\ [0.5ex]
		\hline 
		\multirow{5}{*}{3} & tFD & 339.22(129.70) & -0.31 & 267.46(129.43) & 0.22 & 301.23(102.10) & -0.04\\ 
		& FFD & 217.48(55.21) & 0.23 & 185.98(96.18) & 0.25 & 200.80(71.33) & 0.26\\ 
		& SFD & 61.85(38.20) & -0.61\textsuperscript{+} & 57.41(26.24) & 0.16 & 59.50(23.65) & -0.37\\ 
		& LFD & 59.89(48.19) & -0.44 & 24.08(21.90) & -0.01 & 40.93(30.90) & -0.46\\ 
		& tFC & 1.16(0.30) & -0.33 & 1.06(0.38) & 0.34 & 1.10(0.27) & 0.09\\ [0.5ex]
		\hline 
		\multirow{5}{0.3cm}{\raggedright All days} & tFD & 284.78(104.92) & 0.01 & 311.02(143.16) & 0.29 & 297.90(120.65) & 0.18\\ 
		& FFD & 204.70(71.73) & 0.22 & 208.16(84.94) & 0.42 & 206.43(76.96) & 0.34\\ 
		& SFD & 52.90(26.26) & -0.2 & 65.50(39.50) & 0.12 & 59.20(31.15) & 0.0\\ 
		& LFD & 27.19(18.10) & -0.54 & 37.35(26.07) & 0.03 & 32.27(18.59) & -0.24\\ 
		& tFC & 1.00(0.27) & -0.02 & 1.12(0.38) & 0.36 & 1.06(0.31) & 0.21\\ [0.5ex]
		\hline 
		\multicolumn{8}{l}{ \textsuperscript{+}: $p<0.1$, \textsuperscript{*}: $p<0.05$, \textsuperscript{**}: $p<0.01$, \textsuperscript{***}: $p<0.001$} \\
		\midrule\bottomrule
	\end{longtable}
}\endgroup

In table \ref{tab:Chapter-3tab3}, among the psycholinguistics factors, rated familiarity is the most significant for all fixation features across groups and days. As we see in the last subsection-\ref{sssec:3chap611} that the fixation features of word-frequency and familiarity are correlated; however, rated word-frequency is comparatively less significant for all fixation features across groups and days. Rated emotion is least significant for the fixation features across groups and days, because the text has very less emotional words. For rated concreteness, the fixation features have a positive correlation, because 73\% words of text- 2 represent non-physical objects, which captured lower gaze during reading in compared to the words of physical objects, making 22\% of the text. The distribution of words has been given in table \ref{tab:Chapter-3tab1}. For rated emotion, a few fixation features have a correlation because only 27\% content words of text-2 found in emotion database. Most of the correlations are negative, means the words showing mild emotion captured relatively more gaze in compared to the words showing intense emotion. 

After comparing both tables \ref{tab:Chapter-3tab2} and \ref{tab:Chapter-3tab3}, we conclude that rated imagery is more significant in session-1 as compared to session-2. One major reason can be that the imagery rating of words in text-1 is normally distributed than those of text-2 words. Rated emotion is least significant in both sessions- 1 and 2. All five fixation features are negatively correlated for rated word-frequency; which shows least frequent (low ratings) words capture more gaze during a reading with respect to most frequent (high ratings) words. For rated age-of-acquisition, the fixation features are negatively correlated, since the words, learned at higher age, captured more gaze during reading whereas, the words, learned at lower age, captured relatively lower gaze. For rated familiarity, the fixation features have also negative correlation, since least familiar words captured more gaze whereas, most familiar words captured fewer gaze. For rated imagery, the fixation features have a negative correlation because the words having difficult imagination captured more gaze as compared to the words having easy imagination.

In most of the cases, the values of the fixation features are highest in first day trial and are lowest in last day trial.

 
\begingroup
\setstretch{1.4}
 {\footnotesize	
 	\begin{longtable}
 		{>{\raggedright}m{0.3cm} C{1.5cm} L{2.1cm} L{1.1cm} L{2.2cm} L{1.2cm} L{2.2cm} L{1.cm}}
 		\captionsetup{font={small}, labelfont=bf, skip=4pt}
 		\caption{Pearson’s product-moment correlation on session-2 data}
 		\label{tab:Chapter-3tab3}\\
 		
 		\toprule\midrule
 		\textbf{\scriptsize Day} & \textbf{\scriptsize Fixation feature} & \textbf{\scriptsize Low (Mean,SD)} & \textbf{\scriptsize Low (r,p)} & \textbf{\scriptsize High (Mean,SD)} & \textbf{\scriptsize High (r,p)} & \textbf{\scriptsize All (Mean,SD)} & \textbf{\scriptsize All (r,p)}\\
 		\midrule
 		\multicolumn{8}{l}{\textbf{1. Word Frequency:}} \\[0.5ex]
 		\hline 
 		\multirow{5}{*}{1} & tFD & 395.15(77.41) & -0.74\textsuperscript{+} & 527.33(115.04) & -0.87\textsuperscript{*} & 471.41(97.12) & -0.84\textsuperscript{*}\\ 
 		& FFD & 255.98(33.07)  & -0.79\textsuperscript{*} & 259.89(25.66) & -0.84\textsuperscript{*} & 258.23(27.52) & -0.86\textsuperscript{*}\\ 
 		& SFD & 97.11(23.06) & -0.79\textsuperscript{*} & 138.22(44.05) & -0.58 & 120.83(34.17) & -0.66\\ 
 		& LFD & 42.06(25.50) & -0.5 & 129.22(57.48) & -0.92\textsuperscript{**} & 92.34(41.30) & -0.87\textsuperscript{*}\\ 
 		& tFC & 1.30(0.19) & -0.6 & 1.82(0.26) & -0.73\textsuperscript{+} & 1.60(0.23) & -0.70\textsuperscript{+}\\[0.5ex]
 		\hline 
 		\multirow{5}{*}{2} & tFD & 329.53(74.51) & -0.57 & 432.80(126.10) & -0.87\textsuperscript{*} & 385.47(97.83) & -0.81\textsuperscript{*}\\ 
 		& FFD & 217.83(27.65)  & -0.42 & 242.03(55.70) & -0.70\textsuperscript{+} & 230.93(40.51) & -0.65\\ 
 		& SFD & 77.82(34.02) & -0.57 & 97.88(37.38) & -0.81\textsuperscript{*} & 88.69(33.43) & -0.75\textsuperscript{+}\\ 
 		& LFD & 33.89(17.50) & -0.68\textsuperscript{+} & 92.89(50.57) & -0.81\textsuperscript{*} & 65.85(33.34) & -0.83\textsuperscript{*}\\ 
 		& tFC & 1.15(0.19) & -0.54 & 1.44(0.33) & -0.90\textsuperscript{**} & 1.31(0.25) & -0.82\textsuperscript{*}\\[0.5ex]
 		\hline 
 		\multirow{5}{*}{3} & tFD & 281.11(78.34) & -0.74\textsuperscript{+} & 456.08(86.63) & -0.73\textsuperscript{+} & 377.34(75.78) & -0.80\textsuperscript{*}\\
 		& FFD & 208.09(46.32)  & -0.46 & 211.92(40.87) & -0.73\textsuperscript{+} & 210.20(39.98) & -0.65\\ 
 		& SFD & 52.93(27.76) & -0.80\textsuperscript{*} & 124.18(33.15) & -0.19 & 92.12(26.19) & -0.52\\ 
 		& LFD & 20.09(15.04) & -0.93\textsuperscript{**} & 119.98(49.43) & -0.54 & 75.03(31.58) & -0.67\\ 
 		& tFC & 0.92(0.18) & -0.93\textsuperscript{**} & 1.65(0.23) & -0.71\textsuperscript{+} & 1.32(0.20) & -0.84\textsuperscript{*}\\[0.5ex]
 		\hline 
 		\multirow{5}{0.3cm}{\raggedright All days} & tFD & 338.76(73.75) & -0.71\textsuperscript{+} & 475.72(108.62) & -0.86\textsuperscript{*} & 415.07(89.97) & -0.83\textsuperscript{*}\\ 
 		& FFD & 228.53(31.61)  & -0.62 & 240.41(36.79) & -0.80\textsuperscript{*} & 235.15(33.28) & -0.76\textsuperscript{*}\\ 
 		& SFD & 77.44(27.58) & -0.72\textsuperscript{+} & 120.81(31.80) & -0.68\textsuperscript{+} & 101.61(29.81) & -0.70\textsuperscript{+}\\ 
 		& LFD & 32.79(17.36) & -0.74\textsuperscript{+} & 114.50(49.17) & -0.85\textsuperscript{*} & 78.31(33.39) & -0.86\textsuperscript{*}\\ 
 		& tFC & 1.14(0.18) & -0.71\textsuperscript{+} & 1.64(0.27) & -0.83\textsuperscript{*} & 1.42(0.22) & -0.80\textsuperscript{*}\\ [0.5ex]
 		\hline 
 		\multicolumn{8}{l}{\textbf{2. Age of Acquisition:}} \\ [0.5ex]
 		\hline 
 		\multirow{5}{*}{1} & tFD & 525.24(143.39)  & -0.74\textsuperscript{*} & 669.35(152.12) & -0.81\textsuperscript{**} & 608.38(147.05) & -0.79\textsuperscript{*}\\ 
 		& FFD & 324.88(80.66)  & -0.69\textsuperscript{*} & 304.62(58.92) & -0.75\textsuperscript{*} & 313.19(65.07) & -0.75\textsuperscript{*}\\ 
 		& SFD & 134.11(43.02)  & -0.72\textsuperscript{*} & 169.68(42.59) & -0.68\textsuperscript{*} & 154.63(41.60) & -0.72\textsuperscript{*}\\ 
 		& LFD & 66.25(30.12) & -0.66\textsuperscript{+} & 195.05(67.12) & -0.75\textsuperscript{*} & 140.55(49.15) & -0.76\textsuperscript{*}\\ 
 		& tFC & 1.50(0.23) & -0.68\textsuperscript{*} & 2.04(0.28) & -0.74\textsuperscript{*} & 1.81(0.25) & -0.73\textsuperscript{*}\\[0.5ex]
 		\hline 
 		\multirow{5}{*}{2} & tFD & 391.11(81.60) & -0.83\textsuperscript{**} & 595.19(172.25) & -0.71\textsuperscript{*} & 501.66(128.64) & -0.75\textsuperscript{*}\\ 
 		& FFD & 239.15(38.72)  & -0.77\textsuperscript{*} & 283.39(57.04) & -0.74\textsuperscript{*} & 263.12(47.79) & -0.76\textsuperscript{*}\\ 
 		& SFD & 96.59(21.95) & -0.82\textsuperscript{**} & 142.34(38.37) & -0.59\textsuperscript{+} & 121.37(29.28) & -0.70\textsuperscript{*}\\ 
 		& LFD & 55.37(26.22) & -0.75\textsuperscript{*} & 169.46(84.77) & -0.67\textsuperscript{*} & 117.17(57.05) & -0.70\textsuperscript{*}\\ 
 		& tFC & 1.28(0.19) & -0.75\textsuperscript{*} & 1.83(0.35) & -0.64\textsuperscript{+} & 1.58(0.27) & -0.69\textsuperscript{*}\\[0.5ex]
 		\hline 
 		\multirow{5}{*}{3} & tFD & 364.88(67.21) & -0.83\textsuperscript{**} & 504.06(90.92) & -0.75\textsuperscript{*} & 441.43(77.56) & -0.81\textsuperscript{**}\\ 
 		& FFD & 239.98(36.31)  & -0.86\textsuperscript{**} & 242.76(42.87) & -0.60\textsuperscript{+} & 241.51(38.72) & -0.73\textsuperscript{*}\\ 
 		& SFD & 82.87(24.81) & -0.57 & 130.06(24.24) & -0.62\textsuperscript{+} & 108.83(21.62) & -0.68\textsuperscript{*}\\ 
 		& LFD & 42.03(17.02) & -0.59\textsuperscript{+} & 131.24(42.99) & -0.64\textsuperscript{+} & 91.09(26.72) & -0.74\textsuperscript{*}\\ 
 		& tFC & 1.13(0.16) & -0.74\textsuperscript{*} & 1.74(0.22) & -0.63\textsuperscript{+} & 1.47(0.18) & -0.71\textsuperscript{*}\\[0.5ex]
 		\hline 
 		\multirow{5}{0.3cm}{\raggedright All days} & tFD & 431.09(98.21) & -0.79\textsuperscript{*} & 598.01(136.01) & -0.79\textsuperscript{*} & 524.09(119.09) & -0.79\textsuperscript{*}\\ 
 		& FFD & 269.81(51.38)  & -0.77\textsuperscript{*} & 280.09(51.85) & -0.74\textsuperscript{*} & 275.54(51.25) & -0.76\textsuperscript{*}\\ 
 		& SFD & 105.92(28.36)  & -0.76\textsuperscript{*} & 149.39(33.85) & -0.67\textsuperscript{*} & 130.14(30.94) & -0.72\textsuperscript{*}\\ 
 		& LFD & 55.36(22.91) & -0.74\textsuperscript{*} & 168.52(60.11) & -0.77\textsuperscript{*} & 118.41(43.24) & -0.77\textsuperscript{*}\\ 
 		& tFC & 1.32(0.19) & -0.73\textsuperscript{*} & 1.89(0.28) & -0.70\textsuperscript{*} & 1.64(0.24) & -0.72\textsuperscript{*}\\ 
 		[0.5ex]
 		\hline 
 		\multicolumn{8}{l}{\textbf{3. Familiarity:}} \\ [0.5ex]
 		\hline 
 		\multirow{5}{*}{1} & tFD & 424.20(123.71)  & -0.89\textsuperscript{**} & 560.06(152.78) & -0.85\textsuperscript{**} & 502.58(139.66) & -0.87\textsuperscript{**}\\ 
 		& FFD & 275.22(60.58)  & -0.89\textsuperscript{**} & 271.39(57.15) & -0.88\textsuperscript{**} & 273.01(58.39) & -0.89\textsuperscript{**}\\ 
 		& SFD & 102.67(45.56)  & -0.82\textsuperscript{**} & 138.09(36.51) & -0.83\textsuperscript{**} & 123.11(38.97) & -0.85\textsuperscript{**}\\ 
 		& LFD & 46.30(22.00) & -0.86\textsuperscript{**} & 150.59(61.40) & -0.80\textsuperscript{*} & 106.47(44.06) & -0.82\textsuperscript{**}\\ 
 		& tFC & 1.31(0.26) & -0.79\textsuperscript{*} & 1.78(0.37) & -0.70\textsuperscript{*} & 1.58(0.32) & -0.74\textsuperscript{*}\\ [0.5ex]
 		\hline 
 		\multirow{5}{*}{2} & tFD & 333.64(79.67) & -0.84\textsuperscript{**} & 465.66(123.11) & -0.64\textsuperscript{+} & 405.16(101.06) & -0.73\textsuperscript{*}\\ 
 		& FFD & 214.55(34.20)  & -0.73\textsuperscript{*} & 242.18(44.93) & -0.60\textsuperscript{+} & 229.51(37.87) & -0.69\textsuperscript{*}\\ 
 		& SFD & 76.27(29.19) & -0.71\textsuperscript{*} & 115.31(29.59) & -0.63\textsuperscript{+} & 97.42(28.27) & -0.69\textsuperscript{*}\\ 
 		& LFD & 42.83(22.56) & -0.95\textsuperscript{***} & 108.17(58.00) & -0.57 & 78.22(38.19) & -0.73\textsuperscript{*}\\ 
 		& tFC & 1.11(0.22) & -0.72\textsuperscript{*} & 1.56(0.26) & -0.65\textsuperscript{+} & 1.35(0.23) & -0.70\textsuperscript{*}\\ [0.5ex]
 		\hline 
 		\multirow{5}{*}{3} & tFD & 309.41(88.65) & -0.79\textsuperscript{*} & 434.38(132.16) & -0.70\textsuperscript{*} & 378.14(111.52) & -0.74\textsuperscript{*}\\ 
 		& FFD & 2206.59(44.83)  & -0.75\textsuperscript{*} & 216.17(35.82) & -0.65\textsuperscript{+} & 211.86(39.11) & -0.71\textsuperscript{*}\\ 
 		& SFD & 71.76(31.90) & -0.82\textsuperscript{**} & 108.76(39.71) & -0.58\textsuperscript{+} & 92.11(35.03) & -0.70\textsuperscript{*}\\ 
 		& LFD & 31.06(15.96) & -0.66\textsuperscript{+} & 109.45(62.98) & -0.73\textsuperscript{*} & 74.17(40.12) & -0.75\textsuperscript{*}\\ 
 		& tFC & 0.98(0.19) & -0.70\textsuperscript{*} & 1.50(0.33) & -0.60\textsuperscript{+} & 1.27(0.26) & -0.64\textsuperscript{+}\\ [0.5ex]
 		\hline 
 		\multirow{5}{0.3cm}{\raggedright All days} & tFD & 358.74(95.19) & -0.87\textsuperscript{**} & 493.15(131.29) & -0.78\textsuperscript{*} & 433.62(114.74) & -0.82\textsuperscript{**}\\ 
 		& FFD & 233.77(45.04)  & -0.84\textsuperscript{**} & 246.08(44.81) & -0.77\textsuperscript{*} & 240.63(44.51) & -0.81\textsuperscript{**}\\ 
 		& SFD & 84.33(32.31) & -0.87\textsuperscript{**} & 122.22(32.90) & -0.74\textsuperscript{*} & 105.44(31.87) & -0.82\textsuperscript{**}\\ 
 		& LFD & 40.64(19.05) & -0.91\textsuperscript{**} & 124.85(56.14) & -0.76\textsuperscript{*} & 87.56(39.15) & -0.81\textsuperscript{**}\\ 
 		& tFC & 1.14(0.22) & -0.77\textsuperscript{*} & 1.63(0.31) & -0.69\textsuperscript{*} & 1.41(0.27) & -0.73\textsuperscript{*}\\ 
 		[0.5ex]
 		\hline 
 		\multicolumn{8}{l}{\textbf{4. Imagery:}} \\ [0.5ex]
 		\hline 
 		\multirow{5}{*}{1} & tFD & 435.90(60.68) & -0.54 & 576.58(87.88) & -0.43 & 517.06(75.21) & -0.48\\ 
 		& FFD & 278.34(37.12)  & -0.53 & 271.00(34.28) & -0.51 & 274.10(34.43) & -0.54\\ 
 		& SFD & 106.82(16.46)  & -0.68\textsuperscript{*} & 145.02(24.58) & -0.41 & 128.86(19.77) & -0.53\\ 
 		& LFD & 50.75(14.67) & -0.15 & 160.56(35.18) & -0.3 & 114.10(24.92) & -0.28\\ 
 		& tFC & 1.37(0.12) & -0.45 & 1.88(0.19) & -0.4 & 1.67(0.16) & -0.43\\ [0.5ex]
 		\hline 
 		\multirow{5}{*}{2} & tFD & 345.74(42.75) & -0.42 & 491.50(81.59) & -0.60\textsuperscript{+} & 424.69(63.34) & -0.55\\ 
 		& FFD & 218.63(20.88)  & -0.44 & 255.68(29.97) & -0.58\textsuperscript{+} & 238.70(25.12) & -0.54\\ 
 		& SFD & 84.39(11.63) & -0.61\textsuperscript{+} & 118.98(25.01) & -0.4 & 103.12(18.38) & -0.47\\ 
 		& LFD & 42.72(15.42) & -0.11 & 116.84(32.71) & -0.66\textsuperscript{+} & 82.87(22.85) & -0.55\\ 
 		& tFC & 1.18(0.13) & -0.25 & 1.61(0.21) & -0.55 & 1.41(0.16) & -0.45\\ [0.5ex]
 		\hline 
 		\multirow{5}{*}{3} & tFD & 319.89(47.77) & -0.68\textsuperscript{*} & 457.12(54.72) & -0.73\textsuperscript{*} & 395.36(51.37) & -0.71\textsuperscript{*}\\ 
 		& FFD & 216.82(25.56)  & -0.79\textsuperscript{*} & 221.59(26.04) & -0.65\textsuperscript{+} & 219.44(25.56) & -0.72\textsuperscript{*}\\ 
 		& SFD & 71.17(17.45) & -0.52 & 115.79(17.90) & -0.62\textsuperscript{+} & 95.71(17.41) & -0.58\textsuperscript{+}\\ 
 		& LFD & 31.90(11.63) & -0.29 & 119.74(16.56) & -0.71\textsuperscript{*} & 80.21(13.40) & -0.60\textsuperscript{+}\\ 
 		& tFC & 1.03(0.12) & -0.53 & 1.64(0.12) & -0.60\textsuperscript{+} & 1.37(0.12) & -0.59\textsuperscript{+}\\ [0.5ex]
 		\hline 
 		\multirow{5}{0.3cm}{\raggedright All days} & tFD & 370.23(49.17) & -0.56 & 514.52(73.61) & -0.57 & 450.62(62.53) & -0.57\\ 
 		& FFD & 239.29(26.92)  & -0.60\textsuperscript{+} & 251.96(28.57) & -0.61\textsuperscript{+} & 246.35(27.73) & -0.61\textsuperscript{+}\\ 
 		& SFD & 88.51(14.30) & -0.64\textsuperscript{+} & 128.09(21.78) & -0.47 & 110.56(18.08) & -0.54\\ 
 		& LFD & 42.43(12.09) & -0.2 & 134.47(25.77) & -0.57 & 93.71(18.90) & -0.49\\ 
 		& tFC & 1.21(0.12) & -0.42 & 1.72(0.17) & -0.52 & 1.49(0.14) & -0.49\\ [0.5ex]
 		\hline 
 		\multicolumn{8}{l}{\textbf{5. Concreteness:}}  \\ [0.5ex]
 		\hline 
 		\multirow{5}{*}{1} & tFD & 454.08(60.48) & 0.15 & 598.01(136.00) & 0.41 & 537.12(102.38) & 0.35\\ 
 		& FFD & 289.23(28.03)  & -0.02 & 276.77(45.70) & 0.3 & 282.04(35.80) & 0.21\\ 
 		& SFD & 109.58(23.51)  & 0.24 & 153.67(27.47) & 0.29 & 35.01(25.26) & 0.28\\ 
 		& LFD & 55.28(17.24) & 0.22 & 167.58(66.17) & 0.51 & 120.06(43.63) & 0.49\\ 
 		& tFC & 1.40(0.15) & 0.4 & 1.93(0.31) & 0.48 & 1.70(0.24) & 0.46\\ [0.5ex]
 		\hline 
 		\multirow{5}{*}{2} & tFD & 349.95(98.74) & 0.52 & 519.06(99.76) & 0.32 & 441.55(97.80) & 0.42\\ 
 		& FFD & 221.30(48.72)  & 0.47 & 258.92(37.33) & 0.28 & 241.68(42.08) & 0.38\\ 
 		& SFD & 86.34(33.75)  & 0.58 & 131.93(16.61) & 0.06 & 111.04(22.22) & 0.43\\ 
 		& LFD & 42.31(18.56) & 0.47 & 128.20(50.74) & 0.4 & 88.84(34.46) & 0.44\\ 
 		& tFC & 1.18(0.29) & 0.55 & 1.68(0.24) & 0.35 & 1.45(0.26) & 0.46\\ [0.5ex]
 		\hline 
 		\multirow{5}{*}{3} & tFD & 347.94(46.66) & -0.08 & 503.96(99.57) & 0.19 & 433.75(73.30) & 0.12\\ 
 		& FFD & 219.90(46.55)  & 0.3 & 232.98(28.97) & 0.13 & 227.09(35.71) & 0.23\\ 
 		& SFD & 80.01(19.76) & 0.01 & 124.60(25.61) & 0.38 & 104.54(20.28) & 0.27\\ 
 		& LFD & 48.03(30.44) & -0.58\textsuperscript{+} & 146.37(47.30) & 0.12 & 102.12(26.00) & -0.19\\ 
 		& tFC & 1.09(0.10) & 0.03 & 1.77(0.25) & 0.21 & 1.47(0.18) & 0.17\\ [0.5ex]
 		\hline 
 		\multirow{5}{0.3cm}{\raggedright All days} & tFD & 386.32(63.95) & 0.32 & 545.17(111.71) & 0.33 & 474.82(90.23) & 0.33\\ 
 		& FFD & 245.00(38.66)  & 0.31 & 258.47(37.37) & 0.26 & 252.50(37.52) & 0.29\\ 
 		& SFD & 92.75(21.94) & 0.41 & 138.23(22.22) & 0.28 & 118.09(21.79) & 0.34\\ 
 		& LFD & 48.57(8.58) & -0.08 & 148.47(53.55) & 0.4 & 104.23(32.11) & 0.36\\ 
 		& tFC & 1.23(0.17) & 0.45 & 1.80(0.26) & 0.38 & 1.55(0.22) & 0.41\\ 
 		[0.5ex]
 		\hline 
 		\multicolumn{8}{l}{\textbf{6. Emotion:}} \\ [0.5ex]
 		\hline 
 		\multirow{5}{*}{1} & tFD & 424.89(104.42)  & -0.19 & 611.02(245.38) & -0.51 & 532.27(182.05) & -0.44\\ 
 		& FFD & 270.79(61.88)  & -0.34 & 275.59(73.27) & -0.45 & 273.56(67.62) & -0.42\\ 
 		& SFD & 111.99(37.85)  & -0.21 & 153.38(47.22) & -0.4 & 135.87(42.15) & -0.34\\ 
 		& LFD & 42.11(22.27) & 0.4 & 182.05(128.82) & -0.56 & 122.84(74.33) & -0.51\\ 
 		& tFC & 1.40(0.19) & 0.05 & 1.94(0.46) & -0.54 & 1.71(0.32) & -0.44\\ [0.5ex]
 		\hline 
 		\multirow{5}{*}{2} & tFD & 382.15(142.52)  & -0.29 & 518.70(183.51) & -0.26 & 456.12(159.21) & -0.28\\ 
 		& FFD & 223.83(42.75)  & -0.08 & 263.50(64.00) & -0.23 & 245.31(51.28) & -0.19\\ 
 		& SFD & 102.17(50.48)  & -0.1 & 132.43(51.64) & -0.1 & 118.56(45.01) & -0.11\\ 
 		& LFD & 56.16(67.16) & -0.48 & 122.78(76.58) & -0.36 & 92.25(69.70) & -0.43\\ 
 		& tFC & 1.30(0.42) & -0.39 & 1.70(0.39) & -0.13 & 1.52(0.37) & -0.27\\ [0.5ex]
 		\hline 
 		\multirow{5}{*}{3} & tFD & 361.41(146.46)  & -0.36 & 460.49(133.70) & -0.15 & 415.91(133.75) & -0.26\\ 
 		& FFD & 213.02(30.98)  & -0.18 & 212.23(23.93) & 0.16 & 212.58(24.36) & -0.02\\ 
 		& SFD & 78.28(43.58) & -0.29 & 127.71(46.65) & -0.14 & 105.47(44.22) & -0.21\\ 
 		& LFD & 70.12(81.48) & -0.43 & 120.55(70.30) & -0.25 & 97.85(69.53) & -0.37\\ 
 		& tFC & 1.19(0.41) & -0.44 & 1.66(0.32) & 0.04 & 1.45(0.32) & -0.23\\ [0.5ex]
 		\hline 
 		\multirow{5}{0.3cm}{\raggedright All days} & tFD & 391.30(125.73)  & -0.29 & 537.79(186.18) & -0.37 & 472.91(158.83) & -0.35\\ 
 		& FFD & 237.36(41.71)  & -0.25 & 253.69(52.58) & -0.32 & 246.45(47.27) & -0.29\\ 
 		& SFD & 98.72(40.79) & -0.2 & 139.16(44.96) & -0.24 & 121.25(42.36) & -0.23\\ 
 		& LFD & 55.23(46.83) & -0.39 & 144.95(90.87) & -0.46 & 105.21(70.68) & -0.45\\ 
 		& tFC & 1.30(0.31) & -0.34 & 1.78(0.36) & -0.3 & 1.57(0.33) & -0.32\\ [0.5ex]
 		\hline 
 		\multicolumn{8}{l}{ \textsuperscript{+}: $p<0.1$, \textsuperscript{*}: $p<0.05$, \textsuperscript{**}: $p<0.01$, \textsuperscript{***}: $p<0.001$}\\ \midrule\bottomrule
 	\end{longtable}
 }\endgroup

\subsection{Linear Mixed-Effects Regression Analysis}

Several researchers applied LMER approach to analyse eye movements in different aspects. Juhasz \cite{juhasz2018experience} focused on morphologically complex words during reading to examine the impact of the psycholinguistics variables such as familiarity, age-of-acquisition, transparency, lexeme meaning dominance, sensory experience rating, and imagery on fixation duration features including first-fixation duration, single-fixation duration, gaze duration, and total fixation duration in an LMER model. The result showed that most robust effects were found for familiarity and age-of-acquisition, indicating that a reader's experience with compound words significantly impacted during reading. Kumcu and Thompson \cite{kumcu2020less} investigated the words, that were more difficult to remember, led to more looks to blank locations (looking at nothing). Their results revealed that more frequent looks occurred to blank locations during retrieval of high-difficulty nouns compared to low-difficulty ones. Their linear mixed-effects models demonstrated that imagery and concreteness could predict looking at nothing during retrieval. Huang et al. \cite{huang2020effects} studied the impact of contextual diversity on reading. Contextual diversity (CD) was defined as the proportion of text in a corpus in which a word occurred. Their linear mixed-effects model suggested that CD affected characters and words in beginning Chinese readers. Mangen et al. \cite{magyari2020eye} focused on two text domination styles- enactive and descriptive, to study the contributions of mental imagery to reading behaviour using eye-tracking methodology. Their LMER models' results showed that descriptive text were experienced as more difficult to imagine, while enactive text, which providing readers to experience through multisensory manner, more strongly involved mental imaging. They also reported that ease of imagery correlated with slower reading speed and longer average fixation duration, first fixation duration, dwell time and gaze duration. Also, LMER models were generally preferred over repeated-measures ANOVA, due to their ability to deal with unbalanced and incomplete data sets \cite{hesselmann2018applying}.

We applied linear mixed-effects models to analyse the gaze data using the lme4 package \cite{bates2014fitting} in the R statistical computing tool (Version 3.6.2 R Core Team, 2019) \cite{team2013r}. 

Tables \ref{tab:Chapter-3tab4}, \ref{tab:Chapter-3tab5}, \ref{tab:Chapter-3tab6}, and \ref{tab:Chapter-3tab7} show LMER results of the fixed effects of four fixed variables --fixation features, rating of words, trial day and participants' group respectively on gaze under psycholinguistics factors of session-1 and 2 data. Thus, in all models, we specified gaze value as the dependant variable, whereas fixation features (tFD, FFD, SFD, tLFD, tFC), rating (1-7), trial day (1, 2, 3), and group (high, low) as fixed effects variables (also, independent variables) and student (student-id) was specified as random effects variable. Therefore, all four tables show same intercept value under a psycholinguistics factor in a session. In these tables, the intercepts were the predicted value of the dependent variables when all the independent variables were 0, where the baseline of fixation feature = FFD, the baseline of rating = 1, the baseline of trial day = 1, and the baseline of group = high.

The maximal random effects structure would converge justified by the data with by-participants and by-days random intercepts and slopes \cite{matuschek2017balancing}. We fitted a separate LMER model for each of the six factors on the fixed variables. In each model, the outcome variable- gaze value was regressed onto the predictor features of fixed variables. Also, in a model trial day was nested within students; where within each student, there were 3 trial days. The nesting of day within student showed repeated measure in time and thus simulating the impact of repeated reading on students’ gaze movement by statistically measure variations of feature values over time (day).

In each LMER model, t-value was computed for each fixed variable under all psycholinguistics factors. The effect of a fixed variable under psycholinguistics factors was considered to be statistically significant at $\alpha = 0.05$ level if the absolute t-value was more than 2. T-value measures the size of the difference relative to the variation in the sample data, so the greater the magnitude of t-value, the greater the evidence against the null hypothesis. This means there is greater evidence that there is a significant effect of the fixed variable on gaze. The absolute t-value below to 2 and closer to 0 may show there is not a significant effect of the fixed variable on gaze. 

\subsubsection{LMER Analysis of Gaze on Fixation Feature under Psycholinguistics Factors}

Table \ref{tab:Chapter-3tab4} shows LMER models’ results of fixation features on gaze under psycholinguistics factors on sessions 1 and 2 data. 
 
\begingroup
\setstretch{1.3}
{\small	
	\begin{longtable}
		{>{\raggedright}m{1.4cm} L{1.4cm} L{1.7cm} L{1.7cm} L{1.8cm} L{1.4cm} L{1.6cm} L{1.4cm}}
		\captionsetup{font={small}, labelfont=bf, skip=4pt}
		\caption{LMER result of gaze on fixation feature under psycholinguistics factors}
		\label{tab:Chapter-3tab4}\\
		\toprule\midrule
		
		{\scriptsize\textbf{Fixation feature}} & {\scriptsize\textbf{Values}} & {\scriptsize\textbf{Word Frequency}} & {\scriptsize\textbf{Age of Acquisition}} & {\scriptsize\textbf{Familiarity}} & {\scriptsize\textbf{Imagery}} & {\scriptsize\textbf{Concreteness}} & {\scriptsize\textbf{Emotion}}\\
		\midrule
		\multicolumn{8}{l}{\textbf{LMER result on session-1 data}}\\ 
		\midrule
		\multirow{2}{*}{(Intercept)} & Estimate & 240.15 & 463.25 & 266.68 & 270.19 & 248.29 & 133.91 \\
		& t value & 31.73\textsuperscript{***} & 30.94\textsuperscript{***} & 33.37\textsuperscript{***} & 32.47\textsuperscript{***} & 28.09\textsuperscript{***} & 11.15\textsuperscript{***}\\
		\midrule
		\multirow{2}{*}{tFD} & Estimate & 97.40 & 153.45 & 121.79 & 113.47 & 114.22 & 91.46 \\
		& t value & 24.26\textsuperscript{***} & 14.28\textsuperscript{***} & 32.4\textsuperscript{***} & 27.68\textsuperscript{***} & 28.97\textsuperscript{***} & 11.08\textsuperscript{***}\\
		\midrule
		\multirow{2}{*}{SFD} &  Estimate & -141.54 & -203.83 & -154.96 & -146.35 & -148.53 & -147.23\\
		& t value & -35.25\textsuperscript{***} & -18.96\textsuperscript{***} & -41.22\textsuperscript{***} & -35.70\textsuperscript{***} & -37.68\textsuperscript{***} & -17.84\textsuperscript{***}\\
		\midrule
		\multirow{2}{*}{LFD} &  Estimate & -171.90 & -250.10 & -186.85 & -176.42 & -182.56 & -174.16\\
		& t value & -42.81\textsuperscript{***} & -23.27\textsuperscript{***} & -49.71\textsuperscript{***} & -43.04\textsuperscript{***} & -46.31\textsuperscript{***} & -21.10\textsuperscript{***}\\
		\midrule
		\multirow{2}{*}{tFC} &  Estimate & -204.29 & -302.37 & -230.58 & -216.91 & -221.45 & -205.37\\
		& t value & -50.88\textsuperscript{***} & -28.13\textsuperscript{***} & -61.34\textsuperscript{***} & -52.92\textsuperscript{***} & -56.17\textsuperscript{***} & -24.89\textsuperscript{***}\\
		\midrule
		\multicolumn{8}{l}{\textbf{LMER result on session-2 data}}\\ 
		\midrule
		\multirow{2}{*}{(Intercept)} & Estimate & 329.34 & 377.37 & 324.04 & 336.91 & 227.76 & 451.84 \\
		& t value & 25.82\textsuperscript{***} & 23.82\textsuperscript{***} & 26.26\textsuperscript{***} & 28.01\textsuperscript{***} & 16.36\textsuperscript{***} & 25.03\textsuperscript{***}\\
		\midrule
		\multirow{2}{*}{tFD} & Estimate & 179.91 & 248.54 & 192.99 & 204.27 & 222.31 & 226.45 \\
		& t value & 23.97\textsuperscript{***} & 32.5\textsuperscript{***} & 32.5\textsuperscript{***} & 39.85\textsuperscript{***} & 36.19\textsuperscript{***} & 21.29\textsuperscript{***}\\
		\midrule
		\multirow{2}{*}{SFD} &  Estimate & -133.54 & -145.4 & -135.18 & -135.78 & -134.41 & -125.2\\
		& t value & -17.79\textsuperscript{***} & -19.01\textsuperscript{***} & -22.76\textsuperscript{***} & -26.49\textsuperscript{***} & -21.88\textsuperscript{***} & -11.77\textsuperscript{***}\\
		\midrule
		\multirow{2}{*}{LFD} &  Estimate & -156.83 & -157.13 & -153.07 & -152.63 & -148.27 & -141.24\\
		& t value & -20.89\textsuperscript{***} & -20.55\textsuperscript{***} & -25.77\textsuperscript{***} & -29.78\textsuperscript{***} & -24.13\textsuperscript{***} & -13.28\textsuperscript{***}\\
		\midrule
		\multirow{2}{*}{tFC} &  Estimate & -233.72 & -273.9 & -239.21 & -244.85 & -250.95 & -244.88\\
		& t value & -31.14\textsuperscript{***} & -35.82\textsuperscript{***} & -40.28\textsuperscript{***} & -47.77\textsuperscript{***} & -40.85\textsuperscript{***} & -23.02\textsuperscript{***}\\
		\midrule
		\multicolumn{8}{l}{\textsuperscript{*}: $p<0.05$, \textsuperscript{**}: $p<0.01$, \textsuperscript{***}: $p<0.001$}\\
		
		\midrule\bottomrule
	\end{longtable}
}\endgroup

In both sessions, the gaze of four fixation features-- tFD, SFD, LFD, and tFC were regressed with respect to the gaze of reference (baseline) fixation feature FFD and other baselines were rating = 1, day = 1, and group = high. The results indicate that in both sessions gaze value of tFD, showing total fixation duration, was significantly highest; whereas gaze value of tFC, showing fixation count, was the lowest. In session 1, intercept was highest for age-of-acquisition and was lowest for emotion, which indicate that gaze value was highest for age-of-acquisition and was lowest for emotion. However in session 2, the predicted gaze value was highest for emotion and was lowest for concreteness.

Absolute t-value shows that the gaze of the fixation features under familiarity was the most robust predictor of text-1 processing during session 1 of the RR experiment. Absolute t-value under familiarity shows the highest significant effects on all fixation features among all rated psycholinguistics factors. However, in session 2, t-value shows that the gaze of the fixation features under imagery was the most robust predictor of text-2 processing during session 2. It has the highest significant effect on all fixation features among all rated psycholinguistics factors.

We can see that all psycholinguistics factors had significant effects on the gaze of all fixation features in both sessions of repeated readings. 

\subsubsection{LMER Analysis of Gaze on the Rating of Words under Psycholinguistics Factors}
Table \ref{tab:Chapter-3tab5} shows LMER models’ results of gaze on the rating of words under psycholinguistics factors on sessions 1 and 2 data. In both sessions, the gaze of six ratings-- 2, 3, 4, 5, 6, and 7 were regressed with respect to the gaze of reference (baseline) rating 1 and other baselines were fixation feature = FFD, day = 1, and group = high. In session 1, under word-frequency, age-of-acquisition, imagery and concreteness rating 1 had most gaze value than other ratings; under familiarity, rating 2 had most gaze value; and under emotion, rating 6 had most gaze value. In session 2, under word-frequency rating 6 had most gaze value than other ratings; under age-of-acquisition, rating 2 had most gaze value; under familiarity and imagery rating 3 had most gaze value; and under concreteness and emotion rating 6 had most gaze value. 

Absolute t-value shows that the gaze of the ratings under age-of-acquisition was the most robust predictor of text-1 processing during session 1 of the RR experiment. Absolute t-value under age-of-acquisition shows the highest significant effects on almost all ratings among all rated psycholinguistics factors. 

\begingroup
\setstretch{1.2}
{\small	
	\begin{longtable}
		{>{\raggedright}m{1.4cm} L{1.4cm} L{1.8cm} L{1.8cm} L{1.4cm} L{1.4cm} L{1.6cm} L{1.4cm}}
		\captionsetup{font={small}, labelfont=bf, skip=4pt}
		\caption{LMER results of gaze on the rating of words under psycholinguistics factors}
		\label{tab:Chapter-3tab5}\\
		\toprule\midrule
		
		{\scriptsize\textbf{Rating}} & {\scriptsize\textbf{Values}} & {\scriptsize\textbf{Word Frequency}} & {\scriptsize\textbf{Age of Acquisition}} & {\scriptsize\textbf{Familiarity}} & {\scriptsize\textbf{Imagery}} & {\scriptsize\textbf{Concreteness}} & {\scriptsize\textbf{Emotion}}\\
		\midrule
		\multicolumn{8}{l}{\textbf{LMER result on session-1 data}}\\ 
		\midrule
				\multirow{2}{*}{(Intercept)} & Estimate & 240.15 & 463.25 & 266.68 & 270.19 & 248.29 & 133.91 \\
				& t value & 31.73\textsuperscript{***} & 30.94\textsuperscript{***} & 33.37\textsuperscript{***} & 32.47\textsuperscript{***} & 28.09\textsuperscript{***} & 11.15\textsuperscript{***}\\
				\midrule
		\multirow{2}{*}{2} & Estimate & -8.11 & -145.62 & 11.03 & -24.41 & -12.81 & 69.85\\ 
		& t value & -1.59 & -12.37\textsuperscript{***} & 2.18\textsuperscript{*} & -4.44\textsuperscript{***} & -2.42\textsuperscript{*} & 6.31\textsuperscript{***}\\ 
		\midrule
		\multirow{2}{*}{3} & Estimate & -2.68 & -110.44 & -12.04 & -25.69 & -4.67 & 79.59\\ 
		& t value & -0.52 & -9.38\textsuperscript{***} & -2.38\textsuperscript{*} & -4.67\textsuperscript{***} & -0.88 & 7.19\textsuperscript{***}\\ 
		\midrule
		\multirow{2}{*}{4} & Estimate & -21.35 & -174.46 & -29.22 & -11.37 & -24.21 & 90.31\\ 
		& t value & -4.20\textsuperscript{***} & -14.82\textsuperscript{***} & -5.79\textsuperscript{***} & -2.06\textsuperscript{*} & -4.57\textsuperscript{***} & 8.15\textsuperscript{***}\\ 
		\midrule
		\multirow{2}{*}{5} & Estimate & -8.32 & -197.96 & -30.58 & -49.22 & -25.93 & 69.71\\ 
		& t value & -1.63 & -16.81\textsuperscript{***} & -6.06\textsuperscript{***} & -8.95\textsuperscript{***} & -4.90\textsuperscript{***} & 6.29\textsuperscript{***}\\ 
		\midrule
		\multirow{2}{*}{6} & Estimate & -25.98 & -193.69 & -21.9 & -44.44 & -16.52 & 188.39\\ 
		& t value & -5.11\textsuperscript{***} & -16.45\textsuperscript{***} & -4.34\textsuperscript{***} & -8.08\textsuperscript{***} & -3.12\textsuperscript{**} & 17.01\textsuperscript{***}\\ 
		\midrule
		\multirow{2}{*}{7} & Estimate & -35.80 & -201.4 & -35.69 & -56.51 & -17.07 & 40.55\\ 
		& t value & -7.05\textsuperscript{***} & 17.17\textsuperscript{***} & -7.07\textsuperscript{***} & -10.27\textsuperscript{***} & -3.22\textsuperscript{**} & 3.66\textsuperscript{***}\\ 
		\midrule
		\multicolumn{8}{l}{\textbf{LMER result on session-2 data}}\\ 
		\midrule
		\multirow{2}{*}{(Intercept)} & Estimate & 329.34 & 377.37 & 324.04 & 336.91 & 227.76 & 451.84 \\
		& t value & 25.82\textsuperscript{***} & 23.82\textsuperscript{***} & 26.26\textsuperscript{***} & 28.01\textsuperscript{***} & 16.36\textsuperscript{***} & 25.03\textsuperscript{***}\\
		\midrule
		\multirow{2}{*}{2} & Estimate & -42.27 & 44.31 & -22.21 & -40.07 & 84.87 & -197.53\\ 
		& t value & -4.76\textsuperscript{***} & 4.32\textsuperscript{***} & -2.78\textsuperscript{**} & -5.82\textsuperscript{***} & 10.29\textsuperscript{***} & -13.84\textsuperscript{***}\\ 
		\midrule
		\multirow{2}{*}{3} & Estimate & -49.69 & -53.77 & 35.18 & -14.29 & 110.89 & -207.53\\ 
		& t value & -5.59\textsuperscript{***} & -5.24\textsuperscript{***} & 4.41\textsuperscript{***} & -2.07\textsuperscript{*} & 13.45\textsuperscript{***} & -14.54\textsuperscript{***}\\ 
		\midrule
		\multirow{2}{*}{4} & Estimate & -55.44 & 8.72 & -15.12 & -45.58 & 82.02 & -199.78\\ 
		& t value & -6.24\textsuperscript{***} & 0.85 & -1.89 & -6.62\textsuperscript{***} & 9.95\textsuperscript{***} & -14.00\textsuperscript{***}\\ 
		\midrule
		\multirow{2}{*}{5} & Estimate & -27.78 & -32.35 & -20.1 & -61.6 & 57.4 & -185.46\\ 
		& t value & -3.12\textsuperscript{**} & -3.15\textsuperscript{**} & -2.52\textsuperscript{*} & -8.95\textsuperscript{***} & 6.96\textsuperscript{***} & -12.99\textsuperscript{***}\\ 
		\midrule
		\multirow{2}{*}{6} & Estimate & 46.02 & -82.59 & -28.1 & -26.83 & 129.18 & -106.06\\ 
		& t value & -5.18\textsuperscript{***} & -8.05\textsuperscript{***} & -3.52\textsuperscript{***} & -3.90\textsuperscript{***} & 15.67\textsuperscript{***} & -7.43\textsuperscript{***}\\ 
		\midrule
		\multirow{2}{*}{7} & Estimate & -127.38 & -102.28 & -36.3 & -59.11 & 42.45 & -142.83\\ 
		& t value & -14.34\textsuperscript{***} & -9.97\textsuperscript{***} & -4.55\textsuperscript{***} & -8.59\textsuperscript{***} & 5.15\textsuperscript{***} & -10.01\textsuperscript{***}\\ 
		\midrule
		\multicolumn{8}{l}{\textsuperscript{*}: $p<0.05$, \textsuperscript{**}: $p<0.01$, \textsuperscript{***}: $p<0.001$}\\
		
		\midrule\bottomrule
	\end{longtable}
}\endgroup

However, in session 2, t-value shows that the gaze of the ratings under emotion was the most robust predictor of text-2 processing during session 2. It has the highest significant effect on most of all fixation features among all rated psycholinguistics factors. We can see that all psycholinguistics factors had significant effects on the gaze of almost all ratings in both sessions of repeated readings. 

\subsubsection{LMER Analysis of Gaze on Trial Day under Psycholinguistics Factors}
Table \ref{tab:Chapter-3tab6} shows LMER models’ results of gaze on trial day under psycholinguistics factors on sessions 1 and 2 data. In both sessions, the gaze of two days-- 2 and 3 were regressed with respect to the gaze of reference (baseline) day 1 and other baselines were fixation feature = FFD, rating = 1 and group = high. In session 1, neither of the psycholinguistics factors had significant effects on the gaze of day 2 or 3 with respect to the gaze of the baseline day 1. However, in session 2, three psycholinguistics factors-- word-frequency, familiarity and imagery had significant effects on the gaze of day 2 with respect to the gaze of day 1; also, all six psycholinguistics factors had significant effects on the gaze of day 3 with respect to the gaze of day 1. The table also shows that the gaze value of successor day was decreased than that of predecessor day under all psycholinguistics factors in session-2. However, in session 1, it did not happen.

\begingroup
\setstretch{1.2}
{\small	
	\begin{longtable}
		{>{\raggedright}m{1.2cm} L{1.4cm} L{1.6cm} L{1.6cm} L{1.7cm} L{1.4cm} L{2cm} L{1.4cm}}
		\captionsetup{font={small}, labelfont=bf, skip=4pt}
		\caption{LMER results of gaze on trial day under psycholinguistics factors}
		\label{tab:Chapter-3tab6}\\
		\toprule\midrule
		
		{\scriptsize\textbf{Day}} & {\scriptsize\textbf{Values}} & {\scriptsize\textbf{Word Frequency}} & {\scriptsize\textbf{Age of Acquisition}} & {\scriptsize\textbf{Familiarity}} & {\scriptsize\textbf{Imagery}} & {\scriptsize\textbf{Concreteness}} & {\scriptsize\textbf{Emotion}}\\
		\midrule
		\multicolumn{8}{l}{\textbf{LMER result on session-1 data}}\\ 
		\midrule
				\multirow{2}{*}{(Intercept)} & Estimate & 240.15 & 463.25 & 266.68 & 270.19 & 248.29 & 133.91 \\
				& t value & 31.73\textsuperscript{***} & 30.94\textsuperscript{***} & 33.37\textsuperscript{***} & 32.47\textsuperscript{***} & 28.09\textsuperscript{***} & 11.15\textsuperscript{***}\\
				\midrule
		\multirow{2}{*}{2} & Estimate & 4.32 & -3.72 & 13.18 & 3.63 & 11.11 & 9.46 \\
		& t value & 0.48 & -0.24 & 1.31 & 0.35 & 0.98 & 0.86\\
		\midrule
		\multirow{2}{*}{3} & Estimate & -12.24 & -20.82 & -4.56 & -11.21 & -12.08 & 6.86 \\
		& t value & -1.249 & -1.25 & -0.41 & -1 & -0.98 & 0.58\\
		\midrule
		\multicolumn{8}{l}{\textbf{LMER result on session-2 data}}\\ 
		\midrule
		\multirow{2}{*}{(Intercept)} & Estimate & 329.34 & 377.37 & 324.04 & 336.91 & 227.76 & 451.84 \\
		& t value & 25.82\textsuperscript{***} & 23.82\textsuperscript{***} & 26.26\textsuperscript{***} & 28.01\textsuperscript{***} & 16.36\textsuperscript{***} & 25.03\textsuperscript{***}\\
		\midrule
		\multirow{2}{*}{2} & Estimate & -31.23 & -39.02 & -36.83 & -34.05 & -35.75 & -28.83\\
		& t value & -2.10\textsuperscript{*} & -1.8 & -2.43\textsuperscript{*} & -2.23\textsuperscript{*} & -1.9 & -1.45\\
		\midrule
		\multirow{2}{*}{3} & Estimate & -34.3 & -63.64 & -48.68 & -47.01 & -39.08 & -43.83\\
		& t value & -2.20\textsuperscript{*} & -3.11\textsuperscript{**} & -3.06\textsuperscript{**} & -2.94\textsuperscript{**} & -2.1\textsuperscript{*} & -2.11\textsuperscript{*}\\
		\midrule
		\multicolumn{8}{l}{\textsuperscript{*}: $p<0.05$, \textsuperscript{**}: $p<0.01$, \textsuperscript{***}: $p<0.001$}\\
		
		\midrule\bottomrule
	\end{longtable}
}\endgroup

\subsubsection{LMER Analysis of Gaze on Group under Psycholinguistics Factors}

Table \ref{tab:Chapter-3tab7} shows LMER models’ results of gaze on participants' group under psycholinguistics factors on sessions 1 and 2 data. In both sessions, the gaze of group-- low was regressed with respect to the gaze of reference (baseline) group-- high and other baselines were fixation feature = FFD, rating = 1 and day = 1. In session 1, except emotion all other psycholinguistics factors had significant effects on the gaze of low group with respect to the gaze of the baseline group-- high. In session 2, all six psycholinguistics factors had significant effects on the gaze of low group with respect to the gaze of high group. 

Absolute t-value shows that the gaze of the groups under imagery was the most robust predictor of texts 1 and 2 processing during sessions 1 and 2 respectively in the RR experiment. Absolute t-value under imagery shows the highest significant effects among all rated psycholinguistics factors.

We can see that all psycholinguistics factors had significant effects on the gaze of the group in both sessions of repeated readings.
 
\begingroup
\setstretch{1.2}
{\small	
	\begin{longtable}
		{>{\raggedright}m{1.2cm} L{1.4cm} L{1.6cm} L{1.6cm} L{1.7cm} L{1.4cm} L{2cm} L{1.4cm}}
		\captionsetup{font={small}, labelfont=bf, skip=4pt}
		\caption{LMER results of gaze on group under psycholinguistics factors}
		\label{tab:Chapter-3tab7}\\
		\toprule\midrule
		
		{\scriptsize\textbf{Group}} & {\scriptsize\textbf{Values}} & {\scriptsize\textbf{Word Frequency}} & {\scriptsize\textbf{Age of Acquisition}} & {\scriptsize\textbf{Familiarity}} & {\scriptsize\textbf{Imagery}} & {\scriptsize\textbf{Concreteness}} & {\scriptsize\textbf{Emotion}}\\
		\midrule
		\multicolumn{8}{l}{\textbf{LMER result on session-1 data}}\\ 
		\midrule
				\multirow{2}{*}{(Intercept)} & Estimate & 240.15 & 463.25 & 266.68 & 270.19 & 248.29 & 133.91 \\
				& t value & 31.73\textsuperscript{***} & 30.94\textsuperscript{***} & 33.37\textsuperscript{***} & 32.47\textsuperscript{***} & 28.09\textsuperscript{***} & 11.15\textsuperscript{***}\\
				\midrule
		\multirow{2}{*}{Low} & Estimate & -24.05 & -32.89 & -27.04 & -29.95 & -22.52 & -9.86\\
		& t value & -9.1\textsuperscript{***} & -4.69\textsuperscript{***} & -10.9\textsuperscript{***} & -11.08\textsuperscript{***} & -8.65\textsuperscript{***} & -1.83\\
		\midrule
		\multicolumn{8}{l}{\textbf{LMER result on session-2 data}}\\ 
		\midrule
		\multirow{2}{*}{(Intercept)} & Estimate & 329.34 & 377.37 & 324.04 & 336.91 & 227.76 & 451.84 \\
		& t value & 25.82\textsuperscript{***} & 23.82\textsuperscript{***} & 26.26\textsuperscript{***} & 28.01\textsuperscript{***} & 16.36\textsuperscript{***} & 25.03\textsuperscript{***}\\
		\midrule
		\multirow{2}{*}{Low} & Estimate & -59.72 & -76.57 & -60.55 & -65.18 & -72.18 & -65.65\\
		& t value & -11.98\textsuperscript{***} & -15.01\textsuperscript{***} & -15.28\textsuperscript{***} & -19.0\textsuperscript{***} & -17.58\textsuperscript{***} & -9.31\textsuperscript{***}\\
		\midrule
		\multicolumn{8}{l}{\textsuperscript{*}: $p<0.05$, \textsuperscript{**}: $p<0.01$, \textsuperscript{***}: $p<0.001$}\\
		
		\midrule\bottomrule
	\end{longtable}
}\endgroup

\subsection{ANOVA Analysis of the LMER models}

Table \ref{tab:Chapter-3tab8} shows f-values of the ANOVA analysis of the LMER models discussed in tables \ref{tab:Chapter-3tab4}, \ref{tab:Chapter-3tab5}, \ref{tab:Chapter-3tab6} and \ref{tab:Chapter-3tab7} for variables of interest- fixation features, ratings of words, trial day and group respectively on sessions 1 and 2 data. The ANOVA analysis of the LMER models give f-value of the fixation features, rating, group, and trial day as a whole. Here, fixation gaze value is dependent variable (DV), whereas fixation feature name, group, day and rating are independent variables (IV). The table \ref{tab:Chapter-3tab8} shows that, in both sessions 1 and 2 there were significant effects of all variables except trial day on gaze value under all psycholinguistics factors. However, in session 2, there were low significant effects of trial day on gaze value under age-of-acquisition, familiarity and imagery.  
 
In session 1, the f-values given in the first row show that the fixation features most significantly influenced gaze value under familiarity; and it was least influenced under emotion. In the next row, f-values show that the ratings most significantly influenced gaze value under age-of-acquisition; and it was least influenced under concreteness. In the third row, f-values show that no trial day significantly influenced gaze value. The f-values of the last row show that the group most significantly influenced gaze value under imagery.

Similarly, in session 2, the f-values given in the first row show that the fixation features most significantly influenced gaze value under imagery; and it was least influenced under emotion. In the next row, f-values show that the ratings most significantly influenced gaze value under familiarity; and it was least influenced under imagery. In the third row, f-values show that trial day significantly influenced gaze value under familiarity. The f-values of the last row show that the group most significantly influenced gaze value under imagery and it was least under emotion.

\begingroup
\setstretch{1.4}
{\small	
	\begin{longtable}
		{>{\raggedright}m{1.4cm} L{1.8cm} L{1.8cm} L{1.9cm} L{1.6cm} L{2.cm} L{1.5cm}}
		\captionsetup{font={small}, labelfont=bf, skip=4pt}
		\caption{ANOVA analysis of fixed effect variables in the LMER models}
		\label{tab:Chapter-3tab8}\\
		\toprule\midrule

		{\scriptsize\textbf{Variable}} & {\scriptsize\textbf{Word Frequency}} & {\scriptsize\textbf{Age of Acquisition}} & {\scriptsize\textbf{Familiarity}} & {\scriptsize\textbf{Imagery}} & {\scriptsize\textbf{Concreteness}} & {\scriptsize\textbf{Emotion}}\\
		\midrule
		\multicolumn{7}{l}{\textbf{Analysis on session-1 data}}\\ 
		\midrule
		
		Fixation & 2030.9\textsuperscript{***} & 633.82\textsuperscript{***} & 3054.56\textsuperscript{***} & 2266.43\textsuperscript{***} & 2543.75\textsuperscript{***} & 474.84\textsuperscript{***}\\
		Rating  & 30.17\textsuperscript{***} & 80.48\textsuperscript{***} & 32.77\textsuperscript{***} & 27.26\textsuperscript{***} & 12.0\textsuperscript{***} & 42.86\textsuperscript{***}\\
		Day & 1.37 & 0.83 & 1.41 & 0.86 & 1.62 & 0.4\\
		Group & 82.8\textsuperscript{***} & 22.03\textsuperscript{***} & 118.94\textsuperscript{***} & 122.94\textsuperscript{***} & 74.94\textsuperscript{***} & 3.37\\
		
		\midrule
		\multicolumn{7}{l}{\textbf{Analysis on session-2 data}}\\ 
		\midrule
		Fixation & 938.56\textsuperscript{***} & 1378.31\textsuperscript{***} & 1613.69\textsuperscript{***} & 2317.55\textsuperscript{***} & 1763.3\textsuperscript{***} & 577.67\textsuperscript{***}\\ 
		Rating  & 38.41\textsuperscript{***} & 48.61\textsuperscript{***} & 74.83\textsuperscript{***} & 29.82\textsuperscript{***} & 43.25\textsuperscript{***} & 44.70\textsuperscript{***}\\ 
		Day & 3.22 & 5.08\textsuperscript{*} & 5.42\textsuperscript{*} & 4.86\textsuperscript{*} & 2.94 & 2.4\\ 
		Group & 143.59\textsuperscript{***} & 225.44\textsuperscript{***} & 233.68\textsuperscript{***} & 361.94\textsuperscript{***} & 309.34\textsuperscript{***} & 86.67\textsuperscript{***}\\
		\midrule
		\multicolumn{7}{l}{\textsuperscript{*}: $p<0.05$, \textsuperscript{**}: $p<0.01$, \textsuperscript{***}: $p<0.001$ }\\
		\midrule\bottomrule
	\end{longtable}
}\endgroup
		

\section{Conclusion}

This study examined the influence of word-rating under six psycholinguistics factors on gaze value during English word processing in intensive and extensive readings in three-day repeated reading practice.

In each text, the distribution of content words on a 7 point rating-scale is different for all psycholinguistics factors. Since both texts have low number of words having emotion rating; therefore, only one-fourth of content words were used for analysing the impact of emotion on fixations.

During readings of either narrative text-1 or informative text-2, participants' eyes were fixated on content words according to the given rating for a psycholinguistics factor, i.e., there is a correlation between ratings and fixations (gaze) throughout trials. In session-1 reading, a significant correlation between ratings of words and fixations on the words is highest for word-frequency, lowest for emotion and medium for others. While in session-2 reading, a significant correlation between ratings of words and fixations on the words is highest for familiarity, lowest for concreteness, absent for emotion and medium for others. In both session, as trial day stepped up, fixation on words gradually decreased; which indicates improvements in participants' comprehension on psycholinguistics factors. However, the improvement rate in both sessions are different and dependents on the complexity of text; therefore, the fixations are lower for text-1 as compared to text-2. 

Our LMER results showed that all psycholinguistics factors had significant effects on the gaze of all fixation features in the repeated reading of two different style of texts-- one was narrative and the other was informative. The results also showed that these factors had significant effects on the gaze of almost all
ratings of words in the sessions of repeated readings. The significance of these factors was least on the gaze of trial days. however, all six psycholinguistics factors had significant effects on the gaze of the groups in both sessions of repeated reading.

ANOVA analysis of these LMER models provided sufficient evidence that rated psycholinguistics factors have influence on the gaze of fixation features, word-rating, trial day and group during repeated readings, and we could infer that eye movement features of the reading of text could be important to evaluate mental process happening during reading comprehension.

The future work of this study could be to analyse the 2-way and 3-way interaction in LMER models, such as the effect of specific trial day preference could enhance the preference of specific psycholinguistics factor on L2 word processing in mind during specific type of readings (e.g., extensive and intensive) with various fixed variables such as gender, age-group, reading frequency, the gap between readings, various L1/L2 language, ethnicity, social status etc.

\chapter{Linguistic Analysis of English as Foreign Language Learners' Writings}
\label{Chapter4} 
\lhead{Chapter 4. \emph{Linguistic Analysis of EFL Learners' Writings}} 

\section{Introduction}

Writing and speaking, the forms of text production, are important for a student’s educational and professional careers. The major component of high-stakes tests and interviews require higher-order writing and speaking skills \cite{jenkins2004reading}. Students who fail to develop appropriate text production skills in school may not be able to articulate ideas, argue opinions, and synthesise multiple perspectives, all essential to communicating with others, including peers, teachers, co-workers, and the community at large \cite{crossley2011understanding}. 

Researches focused on L2 writing proficiency investigated linguistic features such as Lexical complexity \cite{laufer1995vocabulary, lu2019examining}, word repetition, word frequency \cite{crossley2014frequency}, and cohesive resources \cite{jung2019predicting, crossley2016development} to distinguish differences among L2 writing proficiency levels. Based on their outcomes, several computational tools have been proposed as well as developed in recent decades.

In this chapter, we explain different linguistics measures and their role in measuring L2 writing proficiency in repeated reading (RR) intervention. The chapter first describes the context, i.e. relationship between comprehension and language production, individual differences in L2 writing, linguistic complexity indices, and computational tools for text analysis. Then we provide the details of the experiment and different feature-sets we used to prove our hypothesis. So, machine learning approaches for differentiating L2 writing proficiency levels are explained. Next, we present the results of linear mixed-effects regression (LMER) to analyse the significant impact of RR on linguistics measures. At last, we discuss the results and conclude the outcomes.


\section{Context}

Only a few research \cite{gorsuch2010developing} studied repeated reading intervention for improving writing fluency. We explain here some theories, concepts and the findings of previous researches proposed in related fields.

\subsection{Relationship Between Text Comprehension and Text Production}

In chapters \ref{Chapter2} and \ref{Chapter3}, we explained the relationship between in-depth eye movement properties and comprehension during in-depth text reading. Here, we mention theories and concepts showing the relationship between text comprehension and text production. Crossley and McNamara \cite{crossley2011understanding} explained the process of text comprehension in brief. They stated, ``in a reader, text comprehension starts with word decoding, therefore, text having less frequent words, diverse words, less familiar words, and complexed syntactic and/or grammatical structures are more difficult to process, especially for readers having poor comprehension. Thus such text increase learning challenges for the reader. Once the reader has identified words and its meaning, the reader linked that word to previous words in order to develop deeper level concepts. The process of linking words together into meaningful phrases and constituents is known as syntactic parsing. Once words are parsed together into units, the reader can begin to determine relations between the words and organise the words into concepts. A complexed syntactic sentence requires higher demands of working memory processing because poor readers cannot immediately construct the appropriate syntactic structures in their mind. However, proficient readers process less frequent words and complex structures immediately than poor skilled readers, because they are more familiar with a greater range of syntactic structures and varieties of words. Once the reader has developed meaningful phrases; connections between discourse constituents at various levels (sentence, paragraph, and text) have made for developing a coherent mental representation of the text. These connections are strengthened by cohesive resources (e.g., reference, substitution, ellipsis, conjunction, repetition, anaphora etc.) that maintain links among textual elements and develop coherence in the mind of the readers. Therefore, writing quality is highly related to cohesion. These cohesive resources and other linguistic features are important in connecting ideas for developing a continuous theme as well as linking ideas with topics."

Linguistic studies have demonstrated that the presence of such cohesive resources in text improve text comprehension, especially for low-knowledge readers.

\subsection{Individual Differences in L2 Writing}

Kormos \cite{kormos2012role} stated “writing is a complex process that requires the skilful coordination of a large number of cognitive and linguistic processes and resources, and therefore, individuals with different cognitive abilities can be expected to execute and orchestrate these processes with varying degrees of efficiency and differ in how they learn to write in L2 language.” In L1 acquisition, the importance of individual differences, such as working memory capacity, motivation and self-efficacy beliefs, has been explored. Gardner \cite{gardner1985role} stated, “individual difference variables were traditionally divided into cognitive, affective and personality-related factors”. 

Carroll \cite{carroll1981twenty} identified four components of language aptitude in learners: i) phonetic coding ability- the ability to identify distinct sounds, and to form associations between those sounds; ii) grammatical sensitivity- the ability to recognize the grammatical functions of words in sentence structures; iii) rote learning ability- the ability to learn associations between sounds, images and meanings efficiently, and to retain these associations; and iv) deductive learning ability- the ability to infer the rules governing a set of language materials.

McCutchen \cite{mccutchen2011novice} explained the role of short-term working memory (STWM) and long-term working memory (LTWM) in language production transiting L2 learners from novice to expert.

\subsection{Linguistics Complexity Indices of Written L2 Text}

Several researchers in the domain of language learning have conducted experiments to examine learners' writing behaviours and linguistic complexity. Text (writing or speech) produced by the participants were analyzed for a range of linguistic complexity measures. Some of these experiments were \cite{crossley2014linguistic, zupanc2014automated, palma2018coherence, yang2019automated, janda2019syntactic, bhatt2020graph, revesz2017effects, kim2014predicting, mostafa2020verb, jung2019predicting, berger2019using, bulte2020investigating, yoon2017linguistic, qin2020beyond, chon2021comparing, okinina2020ctap, delic2017linguistic, jin2020syntactic, matthews2018exploring, kyle2017assessing}. 

Lu et al. \cite{lu2019analyzing} studied the relationship between published scientific articles and their scientific impacts, and for this, they selected 12 variables of linguistic complexity as a proxy for depicting scientific writing in Biology and Psychology disciplines. Their results suggested no practical significant relationship between linguistic complexity and citation strata in either discipline. This suggested that textual complexity plays little role in scientific impact in their data sets.

\subsubsection{Lexical Richness Measures}

In the study of L2 vocabulary acquisition, lexical richness measures are used to evaluate the lexical proficiency level of a learner by comparing their lexical richness with an external reference point. Good writing should have low-frequency words that are suitable to the topic, rather than just general, everyday vocabulary; and it allows writers to express their meanings in an appropriate and knowledgeable manner. Read \cite{o2000assessing} proposed that there were four aspects of lexical richness in content compositions- lexical density, lexical sophistication, lexical variation, and the number of errors. Kyle and Scott \cite{kyle2015automatically} proposed a Tool for the Automatic Analysis of LExical Sophistication (TAALES), which calculates text scores for 135 classic and newly developed lexical indices related to word frequency, range, bigram and trigram frequency, academic language, and psycholinguistics word information.

Lu \cite{lu2012relationship} examined the relationship of lexical richness to the quality of L2 English learners’ oral narratives. He stated that “many L2 development studies have examined the extent to which these measures, along with measures of accuracy, fluency, and grammatical complexity, can be used as reliable and valid indices of the learner's developmental level or overall proficiency in an L2.” He designed a computational tool to automate the measurement of mentioned components of lexical richness using 25 different metrics. In the present work, we used these metrics to measure the improvements in L2 learners’ text production through writing proficiency.

\subsubsection{Syntactic Complexity Measures}

Two popular computational tools to measure the structural complexity of written text are: ‘L2 Syntactical Complexity Analyzer’ (L2SCA) \cite{lu2010automatic} and ‘Tool for the Automatic Analysis of Syntactic Sophistication and Complexity’ (TAASSC) \cite{kyle2016measuring}. L2SCA produces 23 syntactic complexity indices of written English text. It has been used in numerous studies in the field of L2 writing development to compute indices of syntactic complexity. TAASSC 1.0 calculates 377 indices in five categories- clause complexity (32 indices), phrase complexity (132 indices), syntactic sophistication (190 indices), syntactic component scores (9 indices), and classic syntactic complexity indices (14 indices).

\subsubsection{Cohesion Measures}

Cohesion becomes an important element for understanding complex text that demand depth knowledge to the reader. Several studies examining differences in L2 writing proficiency demonstrate that linguistic variables related to cohesion and linguistic sophistication can be used to distinguish high and low proficiency essays. The cohesion indices are grouped in local cohesion, global cohesion, and overall text cohesion. Local cohesion refers to cohesion at the sentence level, global cohesion refers to cohesion between paragraphs of text (e.g., noun overlap between paragraphs in a text), and overall text cohesion refers to the incidence of cohesion features in an entire text \cite{crossley2016tool}.

\subsubsection{Readability Measures}

Readability of a text refers to how well a reader is able to comprehend the content of a text, through reading. Studies have shown that easy to read text improve comprehension, retention, reading speed and reading persistence. Readability roughly indicates the cognitive load of a text for a reader, which depends on the characteristics of the text like lexical richness, syntactic and semantic complexity, discourse-level complexity as well as on the knowledge proficiency of the user. Traditional readability indices incorporate the easy to compute syntactic features of a text like the Average Sentence Length (ASL), the Average Word Length (AWL), the Average number of Syllables per Word (ASW) etc. The examples of such measures are the Flesch Reading Ease score \cite{flesch1949art}, the Kincaid readability formula \cite{kincaid1975derivation}, the Fog Index \cite{gunning1952technique}, and SMOG grading \cite{mc1969smog} etc.

Sinha et al. \cite{sinha2012new} presented computational models to compute readability of two Indian languages- Bangla and Hindi text documents.

\subsection{Impact of Repeated Reading on Second Language Acquisition}

Repeated reading or multiple reading, as discussed in previous chapters, involves multiple, successive encounters with the same visual material, the key being repetition—whether of the same words, sentences, or connected discourse. Repeated reading has been practised with both disabled and non-disabled students in a variety of fashions, ranging from having the learner read aloud, to listening to and simultaneously or subsequently reading aloud, and to silently reading, the same material multiple times. Han and Chen \cite{han2010repeated} reported, “a consistent RR practice can facilitate both intentional and incidental vocabulary gains that would not otherwise have been possible through conventional reading or vocabulary instruction.” Chang and Millett’s study \cite{chang2013improving} shed additional light on the effect of RR on improving reading rates, comprehension levels and potential transfer effects (i.e., transferring RR reading rate to reading an unpracticed passage). Kim's \cite{kim2017effect} findings indicated that the effect of RR on improvements in both speaking and writing of adult EFL learners were significant. These were highlighted especially in the aspects of speaking fluency, interaction and writing organization using difficult sentence structure. Serrano and Huang \cite{serrano2018learning} examined the effects of time distribution in RR—how different encounters with the same text should be spaced for repeated reading to have the strongest impact on L2 vocabulary acquisition. Their results revealed that intensive practice led to more immediate vocabulary gains but spaced practise led to greater long-term retention.

\subsection{Computational Tools for free-Text Analysis}

Several free-text (e.g., essay) writing assessments computing tools have been proposed and developed for L2 tasks; some of them were- automatic essay feedback generation \cite{liu2016automated}, TAACO- Tool for the Automatic Analysis of Cohesion \cite{crossley2016tool, crossley2019tool}, ETS e-rater\textsuperscript{\textregistered} scoring \cite{enright2010complementing},  Coh-Metrix \cite{graesser2004coh}, GAMET- Grammar And Mechanics Error Tool \cite{crossley2019using}, ReaderBench- a multi-lingual framework for analyzing text complexity \cite{dascalu2017readerbench, dascalu2017areaderbench}, ACT CRASE+\textsuperscript{\textregistered} \cite{team2019actnext}, CNN based AES \cite{dong2016automatic}, and CRNN based AES \cite{dasgupta2018augmenting}.


\section{Problem}

In this chapter, we propose a hypothesis that using significant linguistics features derived from L2 learners’ text written during repeated reading intervention, learners’ writing proficiency could be automatically classified by applying machine learning methods. We also propose another hypothesis that repeated reading should have an impact on linguistics features and show improvements in learners’ writing skills.

In order to test these hypotheses, we conducted a reading and writing experiment, in which participants attended two repeated reading sessions for reading and writing. In a session, they read one text and then wrote a summary of the text once in a day for consecutive three days. The reading of text-1 and text-2 had given the experience of intensive and extensive reading respectively. After finishing the reading of a text, they wrote a summary of the text, as much detail as they could. 

The summary text collected in the repeated reading experiment are analysed in the current chapter as well as in the next chapter. In this chapter, we test the proposed hypotheses.


\section{Experiment}
The present experiment was similar to the experiment discussed in chapter-\ref{Chapter2}, with three major exceptions, i) no participant was common in both experiments, ii) in the present study, eye-tracker was not required during reading, and iii) in each trial day, participants wrote a summary of the read text. A brief description of the participants, materials, and procedure that we used in this study is described here.

\subsection{Participants}

There were 50 EFL bilingual students (12 females and 38 males, age-range = 20-24 years, mean age = 22.3 years) from Indian Institute of Information Technology Allahabad participating in the present study. The participants were enrolled in a bachelor/master program of STEM discipline. They were compensated with some course credits for their participation in the study. None of the participants reported having any language or reading impairments. They all performed regular academic activities (e.g., listening to class lectures, writing assignments, watching lecture videos) in English (L2) only, whereas their primary language (L1) was different from each other. Also, they reported that they could carry on a conversation, read and comprehend instructions, read articles, books, as well as watch TV shows and movies in English (L2) also. 

\subsection{Materials}
As discussed in chapter-\ref{Chapter2}, to fulfil the purpose of the experiment, participants needed to read completely strange text. Therefore, the same two texts-- text-1 (\ref{AppendixA1}) and text-2 (\ref{AppendixA2}), used in the previous experiment, were chosen for the present study. As stated earlier the texts-- 1 and 2 were used for simulating extensive reading and intensive reading respectively. These texts were unread in the participants’ lifespan until the experiment began. A summary of the properties of text is already given in table \ref{tab:Chapter-2tab2}.

\subsection{Procedure}

All experimental sessions were held in the SILP research laboratory in a group of 10 students. Upon their first arrival in the research laboratory, the participants read and signed a consent form prior to starting the experimental procedure. They were aware that they had to read two texts followed by summary writing in each of the consecutive three days, but they had no prior information that day-1 text would be repeated on the next two days. The format of the data collection experiment was inspired by Sukhram et al. \cite{sukhram2017effects} repeated reading experiment. 

In the experiment, the participants attended two reading sessions in a day for three consecutive days. In a session, they individually sat in front of a monitor on which the slide of a text was displaying and then they started to read the text. There was no time limit for finishing their reading to provide them a natural reading condition. So, participants were able to read at their own pace. They were instructed to read the text silently. After finishing reading, the text was removed from the display and they were told to solve some puzzles (Appendix \ref{pdf:puzzle}) on a paper sheet in two minutes. After two minutes, they started to write a summary of the read text on the same computer. In the experiment, puzzles were used to clear their rote/working/short-term memory to ensure that summaries would come from their long-term memory where their comprehension skills (understanding of the text) stored. Title, heading, bullets, images etc. of the original text were excluded and only core sentences were displayed, which made the text anonymous to the participants.

In session-1, participants read text-1 and then by typing on a keyboard, wrote its summary on a given computer without looking at the text. The same had happened in session-2 for text-2. The participants took 20—40 minutes to finish a session. Between the two sessions, they were given a break of 15-minute for refreshment.


\section{Feature Analysis}

\subsection{Summary Rating}

The typed summary files were examined for correcting errors in text including grammar, spelling, punctuation, white space, and repetition errors. After manual correction, the files were given to two Ph.D. scholars for rating- one labelled each text either high or low, whereas the other gave each text a score from 1 up to 5. The former one rated summary text based on overall L2 linguistic quality and gave a label accordingly, whereas the latter one rated based on content enrichment (how much part of the text covered in a summary) and gave a score accordingly. The distribution of the ratings (label \& score) of summaries is reported in table \ref{tab:Chapter-4tab1}.


\begingroup
\setstretch{1.3}
{\small	
	\begin{longtable}
		{C{1.5cm} C{1cm} C{2.5cm} C{1cm} C{1cm} C{0.7cm} C{0.7cm} C{0.7cm} C{0.7cm} C{0.7cm}}
		\captionsetup{font={small}, labelfont=bf, skip=4pt}
		\caption{Details of rating distribution among summary text}
		\label{tab:Chapter-4tab1}\\
		\toprule\midrule

				\textbf{Session} & \textbf{Day} & \textbf{Total participants} & \multicolumn{2}{c}{\textbf{Label distribution}} & \multicolumn{5}{c}{\textbf{Score distribution}} \\
				\midrule
				& & & \textbf{Low} & \textbf{High} & \textbf{1} & \textbf{2} & \textbf{3} & \textbf{4} & \textbf{5}\\
				\midrule
					
				\multirow{3}{*}{1} & 1 & 50 & 26 & 24 & 16 & 13 & 19 & 2 & 0\\ 
				                   & 2 & 50 & 26 & 24 & 11 & 14 & 12 & 10 & 3\\
				                   & 3 & 50 & 26 & 24 & 4  & 20  & 8 & 12 & 6\\
				\midrule
				\multirow{3}{*}{2} & 1 & 50 & 26 & 24 & 17 & 16 & 15 & 2 & 0\\
				                   & 2 & 50 & 26 & 24 & 15 & 12 & 17 & 6 & 0\\
				                   & 3 & 50 & 24 & 26 & 12 & 12 & 14 & 10 & 2\\
				
		\midrule\bottomrule
	\end{longtable}}
	\endgroup

\subsection{Feature Extraction}
For measuring overall linguistics quality, we targeted to extract various kinds of features automatically through lingual measure computing tools proposed and applied by researchers in related domains. Major researchers of second language acquisition (SLA) domain focused mainly five kinds of lingual measures- i) lexical richness, ii) syntactic complexity, iii) cohesion, iv) readability and v) psycholinguistics factors.

We applied the following process for extracting these five feature-sets.

\subsubsection{Vocabulary Knowledge Features}
In the study of L2 vocabulary acquisition, lexical richness measures are used to evaluate the lexical proficiency level of a learner, comparing their lexical richness with an external reference point. Lu \cite{lu2012relationship} designed a computational tool named `Lexical Complexity Analyzer' (LCA)\footnote{\url{http://www.personal.psu.edu/xxl13/downloads/lca.html}} to automate the measurement of three important components- lexical density, lexical sophistication, and lexical variation of lexical richness using 33 different features. In the present work, we used these features to measure the improvements in L2 student’s text production. In the calculation of these features, `British National Corpus'\footnote{\url{http://hdl.handle.net/20.500.12024/2554}} dataset was used as an external reference point.
\begin{enumerate}[(i)]
	\item \textbf{Lexical density:} It refers to the ratio of the number of lexical words to the total number of words in a text. Lexical words refer as nouns, adjectives, verbs (excluding modal verbs and auxiliary verbs), and adverbs with an adjectival base (e.g., particularly). Good writing is expected to have a high percentage of lexical words. These indices are: lexical density (LD), proportion of word types, sophisticated word types, lexical types, sophisticated lexical types, word tokens, sophisticated word tokens, lexical tokens, sophisticated lexical tokens.
	
	\item \textbf{Lexical sophistication:} It refers to the ratio of relatively un-usual or advanced words in learner’s text. These indices are:  Lexical Sophistication-I (LS1), Lexical Sophistication-II (LS2), Verb Sophistication-I (VS1), Corrected VS1 (CVS1), Verb Sophistication-II (VS2).
	
	\item \textbf{Lexical variation:} It refers to the range of a learner’s vocabulary as displayed in his or her language use. These indices are: Number of Different Words (NDW), NDW (first 50 words) (NDW-50), NDW (expected random 50) (NDW–ER50), NDW (expected sequence 50) (NDW–ES50), Type–Token Ratio (TTR), Mean Segmental TTR (50) (MSTTR–50), Corrected TTR (CTTR), Root TTR (RTTR), Bi-logarithmic TTR (LogTTR), Uber Index (Uber), Lexical Word Variation (LV), Verb Variation-I (VV1), Squared VV1 (SVV1), Corrected VV1 (CVV1), Verb Variation-II (VV2), Noun Variation (NV), Adjective Variation (AdjV), Adverb Variation (AdvV), Modifier Variation (ModV).

\end{enumerate}

\subsubsection{Text Structure Complexity Features}

Lu \cite{lu2010automatic} developed `Syntactical Complexity Analyzer' (L2SCA)\footnote{\url{http://www.personal.psu.edu/xxl13/downloads/l2sca.html}}, which produces 23 syntactic complexity indices of written English language text. It has been used in numerous studies in the field of L2 writing development to compute indices of syntactic complexity. The syntactic complexity analyzer first retrieves and counts all the occurrences of nine relevant production units and syntactic structures in a text, i.e. Words (W), Sentences (S), Clauses (C), Dependent clauses (DC), T-units (T), Complex T-units (CT), Coordinate phrases (CP), Complex nominals (CN), and Verb phrases (VP) and then calculates the fourteen indices, i.e. Mean length of clause (MLC), Mean length of sentence (MLS), Mean length of T-unit (MLT), Sentence complexity ratio (C/S), T-unit complexity ratio (C/T), Complex T-unit ratio (CT/T), Dependent clause ratio (DC/C), Dependent clauses per T-unit (DC/T), Coordinate phrases per clause (CP/C), Coordinate phrases per T-unit (CP/T), Sentence coordination ratio (T/S), Complex nominals per clause (CN/C), Complex nominals per T-unit (CN/T), and Verb phrases per T-unit (VP/T) using those counts. 

\subsubsection{Cohesion Measures}

Coh-Metrix 3.0\footnote{\url{http://141.225.61.35/cohmetrix2017}} \cite{graesser2004coh} is a web-based tool that has been used to investigate the mental representation of content and the relationship between learning outcomes and the cognitive and psycholinguistics variables captured in text, e.g. analyzing the overall quality of writing as well as speech transcripts \cite{barnwal2017using}. The Coh-Metrix measures a number of linguistic features related to lexical sophistication, syntactic complexity, and multi-levels of cohesion included co-referential cohesion, causal cohesion, the density of connectives, temporal cohesion, spatial cohesion, and latent semantic analysis (LSA). The publicly available version of the tool includes around 108 indices (Appendix \ref{AppendixE1}).

\subsubsection{Readability Measures}

We calculated six traditional readability indices of each summary text including, the Flesch Reading Ease score \cite{flesch1949art}, the SMOG grading \cite{mc1969smog}, the Coleman-Liau Index \cite{coleman1975computer}, Automated readability index \cite{smithautomated}, the Dale-Chall Readability Formula \cite{dale1948formula}, the Fog Index \cite{gunning1952technique}, as well as also calculated the number of complex words (having at-least 3 syllables), and the total number of syllables of each text. 

\subsubsection{Psycholinguistics Factors' Rating}

As we already discussed in chapter-\ref{Chapter3}, six rated psycholinguistics factors played cognitive roles in text comprehension and therefore influenced eye-movements during reading. Here, we used these rated psycholinguistics factors for measuring L2 learners’ writing proficiency. The words of a summary text were rated for i) word frequency (using the database given by BrysBaert and New \cite{brysbaert2009moving}), ii) age-of-acquisition (Kuperman et al. \cite{kuperman2012age}), iii) imagery \& iv) concreteness (Ljubešić et al. \cite{ljubevsic2018predicting}), v) familiarity/meaningfulness (using the EMOTE database \cite{gruhn2016english} and the MRC database \cite{wilson1988mrc}), and vi) emotion (using the EMOTE database \cite{gruhn2016english}). The obtained ratings were rescaled on a 9-point scale and then were rounded to the nearest integer so that the ratings could be presented as integers on a scale from 1 to 9. Now, rated words were grouped into low (having rating 1 to 3), mid (having rating 4 to 6) and high (having rating 7 to 9). Thus, for each factor, the total number of words in the three groups were calculated. This data having 18 features showed the percentage of all words of a summary fell in low, mid and high rating-group under a psycholinguistics factor. 

\subsection{Feature Selection}

The data of all five feature-sets were normalized using the Z-score normalization method. We employed the Welch's t-test technique \cite{welch1947generalization, delacre2017psychologists} to select discriminating features from all feature-sets. This approach is based on the concept that whether the mean of two sample groups (here, high and low) of the population were similar or not. This test was used to detect the significant differences between the features of high-labelled and low-labelled summary in a session. Only the features with a p-value lower than 0.05 were selected as statistically significant features for classification and regression. 

From session-1 summary text, the number of significant features in the five feature-sets were [25, 22, 19, 27], [16, 18, 9, 18], [17, 13, 18, 25], [2, 3, 2, 5], [17, 18, 16, 18] for vocabulary, text structure complexity, cohesion, readability and psycholinguistics factors’ respectively in the sequence of trial days- 1, 2, 3 and all days. Similarly, from session-2 summary text, the number of significant features in the five feature-sets were [21, 18, 20, 25], [16, 18, 9, 18], [13, 16, 34, 21], [5, 3, 5, 6], [14, 14, 17, 18] for vocabulary, text structure complexity, cohesion, readability and psycholinguistics factors’ respectively in the sequence of trial day- 1, 2, 3 and all days.

\section{Automatic Machine Learning Method}

To test the first hypothesis “using significant linguistic features, participants’ writing proficiency can be automatically labelled as well as can be determined a score”, we made classifications and regressions models of the statistically significant lingual features by using the Scikit-learn machine learning library (Version 0.20.4, 2019) \cite{pedregosa2011scikit} for the Python programming language. For classification, three different classifiers were used to predict labels- the decision tree (DT), the multilayer perceptron (MLP) and the support vector machine (SVM); whereas for regression, the Support Vector Regression (SVR) was used to determine scores. In addition, the 5-fold cross-validation technique was used to create another separate training and validation set. 

The configuration of the classifiers and the regression are given here. For DT classifier, the quality of a split was set to Gini impurity and the strategy used to choose the split at each node = best. For MLP, the learning rate = 0.001, Number of epochs = 200, Activation function for the hidden layer = relu, and hidden layer size was set to 100. For SVM, the C parameter was set to 1.0 using RBF kernel and the tolerance parameter was set to 0.001. For SVR, the C parameter was set to 1.0 using RBF kernel, gamma was 0.1, and the tolerance parameter was set to 0.001. 

To analyze the performance of the classifications, we used classification accuracy (C. rate), Unweighted Average Recall (UAR) and mean cross-validation (CV) accuracy. Similarly, to analyze the performance of the regression, we used root-mean-square error (RMSE), mean cross-validation error (CV RMSE) and quadratic weighted kappa (QWK). The Kappa coefficient \cite{cohen1968weighted} is a chance-adjusted index of agreement. In machine learning, it can be used to quantify the amount of agreement between a regression's predictions and a human rater’s label of the same objects. A perfect score of 1.0 is granted when both the predictions and actuals are the same. Whereas, the least possible score is -1 which is given when the predictions are furthest away from actuals.

\section{Results and Discussion on Machine Learning Outcomes}

Tables \ref{tab:Chapter-4tab2} and \ref{tab:Chapter-4tab3} show the classification results in the task of identifying the label of a summary (High or Low) using statistically significant features of individual feature-set extracted from the summary text collected during sessions 1 and 2 respectively. Similarly, tables \ref{tab:Chapter-4tab4} and \ref{tab:Chapter-4tab5} show the regression results in the task of determining the score of a summary (1, 2, 3, 4, or 5) using the same statistically significant features of individual feature-set extracted from the summary text collected during sessions 1 and 2 respectively. The purpose of reporting these values is to represent them as the classifiers’ and regression’s baseline accuracy and these tables \ref{tab:Chapter-4tab2} \& \ref{tab:Chapter-4tab3} and \ref{tab:Chapter-4tab4} \& \ref{tab:Chapter-4tab5} are compared with final results reporting in tables \ref{tab:Chapter-4tab6} (for classification) and \ref{tab:Chapter-4tab8} (for regression) respectively.
\subsection{Classification Results using Individual Feature-set}

In table \ref{tab:Chapter-4tab2}, the average accuracy were [0.84, 0.82, 0.81], [0.8, 0.8, 0.78], and [0.81, 0.78, 0.77] for SVM, DT, and MLP respectively in the sequence of mean cross-validation accuracy, UAR and classification rate. SVM showed better average performance in all three accuracy measures. The classifiers combinedly showed slightly better performance for three feature-sets- Vocabulary Knowledge, Readability, Psycholinguistics factors rating; whereas, their combined accuracy was lowest for Cohesion. The day column reported that the classifiers combinedly showed better performance for day-2 and worst performance for all-day. 

In terms of accuracy measures, across classifiers, the minimum CV accuracy was 0.67 and was found once, belonged to day-3 and was given by MLP; whereas, the maximum CV accuracy was 0.99 and was found two times, all belonged to day-2 and was given by SVM. The minimum UAR and classification rate across classifiers were 0.64 and the value was found five times, all belonged to all-day and was given by DT and MLP. The maximum UAR and  classification rate across classifiers was 1.0 and was found two times, all belonged to day-3 and was given by DT. 

In table \ref{tab:Chapter-4tab3}, the average accuracy were [0.77, 0.77, 0.76], [0.75, 0.71, 0.7], and [0.78, 0.79, 0.78] for SVM, DT, and MLP respectively in the sequence of mean cross-validation accuracy, UAR and classification rate. MLP showed better average performance in all three accuracy measures. The classifiers combinedly showed slightly better performance for one feature-set- Psycholinguistics Factors Rating; whereas, their combined accuracy was lowest for the Readability. The day column reported that the classifiers combinedly showed better performance for day-1 and worst performance for all-day. 

In terms of accuracy measures, across classifiers, the minimum CV accuracy was 0.61 and was found once, belonged to day-2 and was given by DT; whereas, the maximum CV accuracy was 0.92 and was found once, belonged to day-2 and was given by MLP. The minimum UAR and classification rate across classifiers were 0.4 and the value was found two times, all belonged to all-day and was given by DT. The maximum UAR was 0.95 and was found once, belonged to day-2 and was given by DT. The classification rate across classifiers was 0.93 and was found two times, belonged to days- 1 \& 2 and was given by DT \& MLP. 

As compared to the previous accuracy table \ref{tab:Chapter-4tab2}, all classifiers in the present table \ref{tab:Chapter-4tab3} performed below accuracy. We can say that the classifiers perform better performance on the significant features extracted from session-1 summary than those of session-2.

\begingroup
\setstretch{1.3}
{\small	
	\begin{longtable}
		{>{\raggedright}m{1.8cm} L{0.8cm} L{1cm} L{1.2cm} L{0.8cm} L{1cm} L{1.2cm} L{0.8cm} L{1cm} L{0.8cm}}
		\captionsetup{font={small}, labelfont=bf, skip=4pt}
		\caption{Classification results on individual linguistic feature-set of session-1}
		\label{tab:Chapter-4tab2}\\
		\toprule\midrule
		\textbf{Day} & \multicolumn{3}{c}{\textbf{SVM}} & \multicolumn{3}{c}{\textbf{DT}} & \multicolumn{3}{c}{\textbf{MLP}}\\
		\midrule
		& \textbf{CV acc.} & \textbf{UAR} & \textbf{C. rate} & \textbf{CV acc.} & \textbf{UAR} & \textbf{C. rate} & \textbf{CV acc.} & \textbf{UAR} & \textbf{C. rate}\\
		\midrule
		\multicolumn{10}{l}{\textbf{1. Vocabulary Knowledge Feature-set:} all features- 33} \\	
		\midrule
		1 (\# 25) & 0.76 & 0.87 & 0.87 & 0.72 & 0.75 & 0.73 & 0.75 & 0.86 & 0.87\\ 
		2 (\# 22) & 0.99 & 0.94 & 0.93 & 0.99 & 0.92 & 0.93 & 0.96 & 0.86 & 0.87\\ 
		3 (\# 19) & 0.87 & 0.81 & 0.8 & 0.77 & 0.69 & 0.67 & 0.79 & 0.86 & 0.87\\ 
		All (\# 27) & 0.79 & 0.66 & 0.67 & 0.71 & 0.71 & 0.71 & 0.76 & 0.69 & 0.69\\
		\midrule 
		\multicolumn{10}{l}{\textbf{2. Text Structure Complexity Features:} all features- 23}\\ 
		\midrule
		1 (\# 16) & 0.81 & 0.89 & 0.87 & 0.75 & 0.86 & 0.87 & 0.77 & 0.69 & 0.67\\ 
		2 (\#18) & 0.96 & 0.89 & 0.87 & 0.89 & 0.75 & 0.73 & 0.92 & 0.89 & 0.87\\ 
		3 (\#9) & 0.85 & 0.81 & 0.8 & 0.84 & 0.75 & 0.73 & 0.67 & 0.78 & 0.73\\ 
		All (\#18) & 0.79 & 0.78 & 0.78 & 0.71 & 0.8 & 0.8 & 0.75 & 0.83 & 0.82\\
		\midrule 
		\multicolumn{10}{l}{\textbf{3. Cohesion Feature-set:} all features- 100}\\ 
		\midrule
		1 (\#17) & 0.8 & 0.7 & 0.67 & 0.73 & 0.8 & 0.73 & 0.73 & 0.75 & 0.73\\ 
		2 (\#13) & 0.91 & 0.75 & 0.73 & 0.89 & 0.9 & 0.87 & 0.8 & 0.75 & 0.73\\ 
		3 (\#18) & 0.73 & 0.9 & 0.87 & 0.8 & 0.75 & 0.67 & 0.75 & 0.8 & 0.73\\ 
		All (\#25) & 0.84 & 0.82 & 0.82 & 0.78 & 0.78 & 0.78 & 0.81 & 0.64 & 0.64\\
		\midrule 
		\multicolumn{10}{l}{\textbf{4. Readability Feature-set:} all features- 8}\\ 
		\midrule
		1 (\#2) & 0.76 & 0.81 & 0.8 & 0.71 & 0.94 & 0.93 & 0.79 & 0.78 & 0.73\\ 
		2 (\#3) & 0.92 & 0.81 & 0.8 & 0.92 & 0.86 & 0.87 & 0.89 & 0.86 & 0.87\\ 
		3 (\#2) & 0.87 & 0.89 & 0.87 & 0.76 & 1.0 & 1.0 & 0.85 & 0.78 & 0.73\\ 
		All (\#5) & 0.78 & 0.72 & 0.73 & 0.72 & 0.64 & 0.64 & 0.72 & 0.71 & 0.73\\ 
		\midrule
		\multicolumn{10}{l}{\textbf{5. Psycholinguistics Factors’ Rating:} all features- 18}\\ 
		\midrule
		1 (\#17) & 0.8 & 0.87 & 0.87 & 0.76 & 0.87 & 0.87 & 0.84 & 0.66 & 0.67\\ 
		2 (\#18) & 0.96 & 0.8 & 0.8 & 0.92 & 0.75 & 0.73 & 0.96 & 0.8 & 0.8\\ 
		3 (\#16) & 0.88 & 0.87 & 0.87 & 0.91 & 0.73 & 0.73 & 0.87 & 0.87 & 0.87\\ 
		All (\#18) & 0.77 & 0.8 & 0.8 & 0.71 & 0.65 & 0.64 & 0.72 & 0.82 & 0.82\\
				
		\midrule\bottomrule
	\end{longtable}}
	\endgroup

\begingroup
\setstretch{1.3}
{\small	
	\begin{longtable}
		{>{\raggedright}m{1.8cm} L{0.8cm} L{1cm} L{1.2cm} L{0.8cm} L{1cm} L{1.2cm} L{0.8cm} L{1cm} L{0.8cm}}
		\captionsetup{font={small}, labelfont=bf, skip=4pt}
		\caption{Classification results on individual linguistic feature-set of session-2}
		\label{tab:Chapter-4tab3}\\
		\toprule\midrule
		\textbf{Day} & \multicolumn{3}{c}{\textbf{SVM}} & \multicolumn{3}{c}{\textbf{DT}} & \multicolumn{3}{c}{\textbf{MLP}}\\
		\midrule
		& \textbf{CV acc.} & \textbf{UAR} & \textbf{C. rate} & \textbf{CV acc.} & \textbf{UAR} & \textbf{C. rate} & \textbf{CV acc.} & \textbf{UAR} & \textbf{C. rate}\\
		\midrule
		\multicolumn{10}{l}{\textbf{1. Vocabulary Knowledge Feature-set:} all features- 33} \\	
		\midrule
		1 (\#21) & 0.84 & 0.81 & 0.8 & 0.79 & 0.75 & 0.73 & 0.77 & 0.89 & 0.87\\ 
		2 (\#18) & 0.8 & 0.75 & 0.73 & 0.79 & 0.81 & 0.8 & 0.75 & 0.75 & 0.73\\ 
		3 (\#20) & 0.69 & 0.77 & 0.77 & 0.66 & 0.77 & 0.77 & 0.74 & 0.63 & 0.62\\ 
		All (\#25) & 0.75 & 0.82 & 0.81 & 0.73 & 0.73 & 0.72 & 0.76 & 0.85 & 0.86\\ 
		\midrule 
		\multicolumn{10}{l}{\textbf{2. Text Structure Complexity Features:} all features- 23}\\ 
		\midrule
		1 (\#16) & 0.88 & 0.86 & 0.87 & 0.85 & 0.81 & 0.8 & 0.85 & 0.75 & 0.73\\ 
		2 (\#18) & 0.81 & 0.72 & 0.73 & 0.81 & 0.67 & 0.6 & 0.73 & 0.89 & 0.87\\ 
		3 (\#9) & 0.68 & 0.7 & 0.69 & 0.65 & 0.63 & 0.62 & 0.74 & 0.86 & 0.85\\ 
		All (\#18) & 0.66 & 0.75 & 0.74 & 0.69 & 0.58 & 0.58 & 0.67 & 0.69 & 0.7\\ 
		\midrule 
		\multicolumn{10}{l}{\textbf{3. Cohesion Feature-set:} all features- 100}\\ 
		\midrule
		1 (\#13) & 0.77 & 0.65 & 0.6 & 0.84 & 0.8 & 0.73 & 0.75 & 0.65 & 0.6\\ 
		2 (\#16) & 0.83 & 0.8 & 0.8 & 0.76 & 0.95 & 0.93 & 0.76 & 0.85 & 0.87\\ 
		3 (\#34) & 0.86 & 0.94 & 0.92 & 0.78 & 0.78 & 0.77 & 0.88 & 0.94 & 0.92\\ 
		All (\#21) & 0.74 & 0.71 & 0.7 & 0.66 & 0.4 & 0.4 & 0.69 & 0.6 & 0.63\\ 
		\midrule 
		\multicolumn{10}{l}{\textbf{4. Readability Feature-set:} all features- 8}\\ 
		\midrule
		1 (\#5) & 0.76 & 0.69 & 0.67 & 0.77 & 0.58 & 0.53 & 0.77 & 0.81 & 0.8\\ 
		2 (\#3) & 0.71 & 0.72 & 0.73 & 0.61 & 0.72 & 0.73 & 0.72 & 0.61 & 0.6\\ 
		3 (\#5) & 0.72 & 0.7 & 0.69 & 0.65 & 0.54 & 0.54 & 0.75 & 0.93 & 0.92\\ 
		All (\#6) & 0.7 & 0.73 & 0.72 & 0.7 & 0.63 & 0.63 & 0.71 & 0.73 & 0.72\\  
		\midrule
		\multicolumn{10}{l}{\textbf{5. Psycholinguistics Factors’ Rating:} all features- 18}\\ 
		\midrule
		1 (\#14) & 0.84 & 0.79 & 0.8 & 0.79 & 0.87 & 0.87 & 0.87 & 0.93 & 0.93\\ 
		2 (\#14) & 0.81 & 0.86 & 0.87 & 0.84 & 0.74 & 0.73 & 0.92 & 0.79 & 0.8\\ 
		3 (\#17) & 0.82 & 0.86 & 0.85 & 0.86 & 0.77 & 0.77 & 0.88 & 0.93 & 0.92\\ 
		All (\#18) & 0.72 & 0.79 & 0.79 & 0.77 & 0.76 & 0.77 & 0.81 & 0.65 & 0.65\\
		
		\midrule\bottomrule
	\end{longtable}}
	\endgroup
\subsection{Regression Results using Individual Feature-set}
In tables \ref{tab:Chapter-4tab4}, \ref{tab:Chapter-4tab5} and \ref{tab:Chapter-4tab8}, the second column shows the root-mean-square error of SVR regression model on test data; whereas the fifth column shows the mean of root-mean-square errors in 5-fold cross-validation. The output of SVR was floating-point numbers, so these were converted to nearest integers. The regression's prediction and the rater’s score of the same summary text were used to calculate Pearson's correlation ($r$) and QWK agreement. These values are shown in columns 3 and 4 respectively. 

The third column of all three tables shows strong correlations between the regression's prediction and the rater’s score on the significant features of each feature-set as well as all significant features of all feature-sets across the days. Similarly, the fourth column of these tables shows better agreement between the regression's prediction and the rater’s score on the significant features of each feature-set as well as all significant features of all feature-sets across the days. Generally, the value of QWK greater than 0.6 is considered to be a good agreement.

In table \ref{tab:Chapter-4tab4}, the feature-set Vocabulary knowledge had the minimum mean of RMSE (SVR) of all days (0.59), whereas the maximum mean of RMSE (SVR) of all days was given by  Psycholinguistics factors rating (0.67). Readability had the maximum mean of both Pearson’ $r$ and QWK of all days, 0.9 and 0.81 respectively. In day-wise analysis, we found that the mean of RMSE of all feature-sets was lowest for day-1 and highest for day-3, whereas the mean of Pearson’s $r$ and QWK of all feature-sets was highest for day-all, and the mean of CV-RMSE of all feature-sets was lowest for day-1.

\begingroup
\setstretch{1.2}
{\small	
	\begin{longtable}
		{>{\raggedright}m{1.8cm} C{1.8cm} C{6cm} C{1.6cm} C{1.6cm}}
		\captionsetup{font={small}, labelfont=bf, skip=4pt}
		\caption{Regression results on individual linguistic feature-set of session-1}
		\label{tab:Chapter-4tab4}\\
		\toprule\midrule
		\textbf{Day} & \textbf{RMSE (SVR)} & \textbf{Pearson correlation coefficient (\textit{r})} & \textbf{QWK} & \textbf{CV-RMSE}\\
		\midrule
		\multicolumn{5}{l}{\textbf{1. Vocabulary Knowledge Feature-set:} all features- 33} \\	
		\midrule
		1 (\#25) & 0.37 & 0.91\textsuperscript{***} & 0.88 & 0.5\\ 
		2 (\#22) & 0.52 & 0.92\textsuperscript{***} & 0.83 & 0.65\\ 
		3 (\#19) & 0.77 & 0.80\textsuperscript{***} & 0.67 & 0.73\\ 
		All (\#27) & 0.71 & 0.85\textsuperscript{***} & 0.79 & 0.64\\ 
		\midrule 
		\multicolumn{5}{l}{\textbf{2. Text Structure Complexity Features:} all features- 23}\\ 
		\midrule
		1 (\#16) & 0.37 & 0.88\textsuperscript{***} & 0.87 & 0.57\\ 
		2 (\#18) & 0.63 & 0.84\textsuperscript{***} & 0.81 & 0.61\\ 
		3 (\#9) & 0.77 & 0.80\textsuperscript{***} & 0.67 & 0.74\\ 
		All (\#18) & 0.71 & 0.90\textsuperscript{***} & 0.8 & 0.68\\ 
		\midrule 
		\multicolumn{5}{l}{\textbf{3. Cohesion Feature-set:} all features- 100}\\ 
		\midrule
		1 (\#17) & 0.52 & 0.92\textsuperscript{***} & 0.65 & 0.49\\ 
		2 (\#13) & 0.63 & 0.80\textsuperscript{***} & 0.77 & 0.72\\ 
		3 (\#18) & 0.73 & 0.86\textsuperscript{***} & 0.76 & 0.66\\ 
		All (\#25) & 0.58 & 0.94\textsuperscript{***} & 0.86 & 0.54\\ 
		\midrule 
		\multicolumn{5}{l}{\textbf{4. Readability Feature-set:} all features- 8}\\ 
		\midrule
		1 (\#2) & 0.45 & 0.91\textsuperscript{***} & 0.84 & 0.62\\ 
		2 (\#3) & 0.68 & 0.90\textsuperscript{***} & 0.77 & 0.61\\ 
		3 (\#2) & 0.58 & 0.90\textsuperscript{***} & 0.84 & 0.63\\ 
		All (\#5) & 0.73 & 0.89\textsuperscript{***} & 0.78 & 0.62\\ 
		\midrule
		\multicolumn{5}{l}{\textbf{5. Psycholinguistics Factors’ Rating:} all features- 18}\\ 
		\midrule
		1 (\#17) & 0.63 & 0.76\textsuperscript{***} & 0.65 & 0.6\\ 
		2 (\#18) & 0.63 & 0.78\textsuperscript{***} & 0.78 & 0.56\\ 
		3 (\#16) & 0.82 & 0.72\textsuperscript{***} & 0.64 & 0.65\\ 
		All (\#18) & 0.58 & 0.91\textsuperscript{***} & 0.89 & 0.47\\
		\midrule
		\multicolumn{5}{l}{\textsuperscript{*}: $p<0.05$, \textsuperscript{**}: $p<0.01$, \textsuperscript{***}: $p<0.001$} \\
		\midrule\bottomrule
	\end{longtable}}
\endgroup

In table \ref{tab:Chapter-4tab5}, the feature-set Psycholinguistics factors rating had the minimum mean of RMSE (SVR) of all days (0.44), whereas the maximum mean of RMSE (SVR) of all days was given by Text structure complexity (0.62). Psycholinguistics factors rating had the maximum mean of both Pearson’ $r$ and QWK of all days, 0.9 and 0.83 respectively. In day-wise analysis, we found that the mean of RMSE of all feature-sets was lowest for day-1 and highest for day-all, whereas the mean of Pearson’s $r$ and QWK of all feature-sets was highest for day-1, and the mean of CV-RMSE of all feature-sets was lowest for day-1.

\begingroup
\setstretch{1.2}
{\small	
	\begin{longtable}
		{>{\raggedright}m{1.8cm} C{1.8cm} C{6cm} C{1.6cm} C{1.6cm}}
		\captionsetup{font={small}, labelfont=bf, skip=4pt}
		\caption{Regression results on individual linguistic feature-set of session-2}
		\label{tab:Chapter-4tab5}\\
		\toprule\midrule
		\textbf{Day} & \textbf{RMSE (SVR)} & \textbf{Pearson correlation coefficient (\textit{r})} & \textbf{QWK} & \textbf{CV-RMSE}\\
		\midrule
		\multicolumn{5}{l}{\textbf{1. Vocabulary Knowledge Feature-set:} all features- 33} \\	
		\midrule
		1 (\#25) & 0.52 & 0.95\textsuperscript{***} & 0.86 & 0.3\\ 
		2 (\#22) & 0.37 & 0.94\textsuperscript{***} & 0.88 & 0.5\\ 
		3 (\#19) & 0.62 & 0.85\textsuperscript{***} & 0.78 & 0.53\\ 
		All (\#27) & 0.61 & 0.80\textsuperscript{***} & 0.76 & 0.6\\ 
		\midrule 
		\multicolumn{5}{l}{\textbf{2. Text Structure Complexity Features:} all features- 23}\\ 
		\midrule
		1 (\#16) & 0.63 & 0.91\textsuperscript{***} & 0.73 & 0.36\\ 
		2 (\#18) & 0.73 & 0.60\textsuperscript{*} & 0.53 & 0.66\\ 
		3 (\#9) & 0.55 & 0.77\textsuperscript{**} & 0.78 & 0.52\\ 
		All (\#18) & 0.55 & 0.84\textsuperscript{***} & 0.78 & 0.57\\ 
		\midrule 
		\multicolumn{5}{l}{\textbf{3. Cohesion Feature-set:} all features- 100}\\ 
		\midrule
		1 (\#17) & 0.45 & 0.91\textsuperscript{***} & 0.83 & 0.41\\ 
		2 (\#13) & 0.45 & 0.88\textsuperscript{***} & 0.84 & 0.54\\ 
		3 (\#18) & 0.62 & 0.89\textsuperscript{***} & 0.74 & 0.66\\ 
		All (\#25) & 0.78 & 0.83\textsuperscript{***} & 0.59 & 0.57\\ 
		\midrule 
		\multicolumn{5}{l}{\textbf{4. Readability Feature-set:} all features- 8}\\ 
		\midrule
		1 (\#2) & 0.58 & 0.92\textsuperscript{***} & 0.81 & 0.41\\ 
		2 (\#3) & 0.52 & 0.84\textsuperscript{***} & 0.78 & 0.58\\ 
		3 (\#2) & 0.68 & 0.76\textsuperscript{**} & 0.63 & 0.6\\ 
		All (\#5) & 0.53 & 0.83\textsuperscript{***} & 0.81 & 0.56\\ 
		\midrule
		\multicolumn{5}{l}{\textbf{5. Psycholinguistics Factors’ Rating:} all features- 18}\\ 
		\midrule
		1 (\#17) & 0.37 & 0.96\textsuperscript{***} & 0.91 & 0.3\\ 
		2 (\#18) & 0.52 & 0.86\textsuperscript{***} & 0.65 & 0.54\\ 
		3 (\#16) & 0.39 & 0.96\textsuperscript{***} & 0.93 & 0.42\\ 
		All (\#18) & 0.48 & 0.82\textsuperscript{***} & 0.82 & 0.5\\
		\midrule
		\multicolumn{5}{l}{\textsuperscript{*}: $p<0.05$, \textsuperscript{**}: $p<0.01$, \textsuperscript{***}: $p<0.001$} \\
		\midrule\bottomrule
	\end{longtable}}
	\endgroup

\subsection{Classification Results on all Significant Linguistic Features}

Table \ref{tab:Chapter-4tab6} shows the classification results in the task of identifying the label of a summary (High or Low) using integrated (early fusion of) significant features of all feature-sets for sessions 1 and 2. 

After comparing with the baseline accuracy tables \ref{tab:Chapter-4tab2} \& \ref{tab:Chapter-4tab3}, best results were highlighted in the present table- \ref{tab:Chapter-4tab6}. The results reported in both tables indicated improvements of the classifiers’ overall performance also the results were comparatively more stable. 


\begingroup
\setstretch{1.4}
{\small	
	\begin{longtable}
		{>{\raggedright}m{1.5cm} C{1cm} C{1cm} C{1.2cm} C{1cm} C{1cm} C{1.2cm} C{1cm} C{1cm} C{0.7cm}}
		\captionsetup{font={small}, labelfont=bf, skip=4pt}
		\caption{Classification results on all significant linguistic features}
		\label{tab:Chapter-4tab6}\\
		\toprule\midrule

				\textbf{Day} & \multicolumn{3}{c}{\textbf{SVM}} & \multicolumn{3}{c}{\textbf{DT}} & \multicolumn{3}{c}{\textbf{MLP}}\\
				\midrule
				& {\footnotesize\textbf{CV acc.}} & {\footnotesize\textbf{UAR}} & {\footnotesize\textbf{C. rate}} & {\footnotesize\textbf{CV acc.}} & {\footnotesize\textbf{UAR}} & {\footnotesize\textbf{C. rate}} & {\footnotesize\textbf{CV acc.}} & {\footnotesize\textbf{UAR}} & {\footnotesize\textbf{C. rate}}\\
				\midrule
				\multicolumn{10}{l}{\textbf{Classification results on all significant linguistic features of session-1}}\\
				\midrule
				
				1 (\#77) & \textbf{0.83} & 0.81 & 0.8 & 0.72 & 0.81 & 0.8 & 0.79 & 0.78 & 0.8\\
				2 (\#74) & 0.95 & \textbf{1.0} & \textbf{1.0} & 0.92 & \textbf{1.0} & \textbf{1.0} & 0.92 & \textbf{1.0} & \textbf{1.0}\\
				3 (\#64) & 0.85 & 0.89 & 0.87 & 0.83 & 0.83 & 0.8 & 0.81 & \textbf{0.89} & 0.87\\
				All (\#94) & \textbf{0.85} & \textbf{0.91} & \textbf{0.91} & 0.75 & 0.75 & 0.76 & \textbf{0.84} & \textbf{0.83} & 0.82\\ 
				\midrule
				\multicolumn{10}{l}{\textbf{Classification results on all significant linguistic features of session-2}}\\
				\midrule
				1 (\#71) & 0.87 & 0.81 & 0.8 & \textbf{0.93} & 0.81 & 0.8 & 0.84 & 0.75 & 0.73\\ 
				2 (\#65) & \textbf{0.87} & \textbf{1.0} & \textbf{1.0} & 0.73 & 0.86 & 0.87 & 0.87 & \textbf{1.0} & \textbf{1.0}\\ 
				3 (\#94) & \textbf{0.89} & 0.94 & 0.92 & 0.78 & \textbf{0.81} & 0.77 & 0.82 & 0.88 & 0.85\\ 
				All (\#90) & 0.71 & 0.72 & 0.7 & 0.75 & 0.69 & 0.7 & 0.71 & 0.7 & 0.7\\ 
				\midrule\bottomrule
			\end{longtable}}
			\endgroup
\subsection{Regression Results on all Significant Linguistic Features}
Table \ref{tab:Chapter-4tab8} shows the regression results in the task of determining a score of a summary (1, 2, 3, 4, or 5) using integrated (early fusion of) significant features of all feature-sets for sessions 1 and 2.

After comparing with the baseline accuracy tables \ref{tab:Chapter-4tab4} \& \ref{tab:Chapter-4tab5}, we found that table \ref{tab:Chapter-4tab8} indicates improvements in the SVR regression model’s overall performance.

\vspace{40mm}

\begingroup
\setstretch{1.2}
{\small	
	\begin{longtable}
		{>{\raggedright}m{1.8cm} C{1.8cm} C{6cm} C{1.6cm} C{1.4cm}}
		\captionsetup{font={small}, labelfont=bf, skip=4pt}
		\caption{Regression results on all significant linguistic features}
		\label{tab:Chapter-4tab8}\\
		\toprule\midrule

				\textbf{Day} & \textbf{RMSE (SVR)} & \textbf{Pearson correlation coefficient (\textit{r})} & \textbf{QWK} & \textbf{CV-RMSE}\\
				\midrule
				\multicolumn{5}{l}{\textbf{Regression results on all significant linguistic features of session-1}}\\
				\midrule
				
				1 (\#77) & 0.37 & 0.88\textsuperscript{***} & 0.85 & 0.57\\
				2 (\#74) & 0.58 & 0.90\textsuperscript{***} & 0.85 & 0.69\\
				3 (\#64) & 0.73 & 0.82\textsuperscript{***} & 0.74 & 0.71\\
				All (\#94) & 0.49 & 0.93\textsuperscript{***} & 0.9 & 0.6\\ 
				\midrule
				\midrule
				\multicolumn{5}{l}{\textbf{Regression results on all significant linguistic features of session-2}}\\
				\midrule
				
				1 (\#71) & 0.58 & 0.89\textsuperscript{***} & 0.74 & 0.29\\
				2 (\#65) & 0.63 & 0.83\textsuperscript{***} & 0.63 & 0.44\\
				3 (\#94) & 0.68 & 0.96\textsuperscript{***} & 0.7 & 0.56\\
				All (\#90) & 0.78 & 0.85\textsuperscript{***} & 0.55 & 0.57\\ 
				\midrule
				\multicolumn{5}{l}{\textsuperscript{*}: $p<0.05$, \textsuperscript{**}: $p<0.01$, \textsuperscript{***}: $p<0.001$} \\
				\midrule\bottomrule
			\end{longtable}}
			\endgroup
		
\section{Linear Mixed-Effects Regression Analysis}

Several researchers used LMER approach to analyse linguistic measures in different aspects. Ströbel et al. \cite{strobel2020relationship} applied linear mixed-effects models to study significant relationships between L1 complexity and L2 complexity for lexical and syntactic measures. Dowell et al. \cite{dowell2014works} investigated the linguistic patterns of students group chats, within an online collaborative learning exercise, on five discourse dimensions (narrativity, deep cohesion, referential cohesion, syntactic simplicity, word concreteness) extracted using Coh-Metrix. The results of linear mixed-effects models indicated that students who engaged in deeper cohesive integration and generated more complicated syntactic structures performed significantly better.
 
We applied linear mixed-effects regression models implemented with the lme4 package \cite{bates2014fitting} in the R environment (Version 3.6.2 R Core Team, 2019). We fitted separate models for each of the 6 linguistic feature-sets. In each model, the outcome variable- feature value (linguistic complexity) was regressed onto the predictor features of the feature-set. Thus, in a model of a feature-set, we specified feature value as the outcome variable, whereas score (1, 2, 3, 4, 5), group (high, low), and trial day (1, 2, 3) as fixed effects as well as student (student-id) was specified as random effects. Day was nested within student; where within each student, there were 3 trial days. The nesting of day within student showed repeated measure in time and thus simulating the impact of repeated reading on students’ performance by statistically measure variations of feature values over time (day).

\subsection{Results of LMER Model}
The results of LMER models, configured for studying the significance of score, indicated that scores 2 and higher (3, 4 \& 5 in session-1; 3 \& 4 in session-2) were significantly achieved than score-1, where day baseline was 1. The significance of score was also achieved for other days 2 \& 3. Similarly, the results of LMER models, configured for studying the significance of group-label, indicated that label high was significantly higher than the label low, where day baseline was 1. This was also true for other days 2 \& 3. The significance of label was also achieved for other days 2 \& 3. The models also predicted significance in performance improvement (score and label) from day- 1 to 2, day-1 to 3 as well as day-2 to 3.
 
In the Vocabulary Knowledge feature set, 15 features (wordtokens, wordtypes, vs2, uber, rttr, slextokens, slextypes, swordtokens, swordtypes, ndw, ndwerz, ndwesz, ndwz, lextokens, \& lextypes) showed significance over the repeated measure. 

In Text Structure feature set, all 23 features showed significance over the repeated measure. 
In Cohesion feature set, 88 features (CNCAdd, CNCAll, CNCCaus, CNCLogic, CRFAO1, CRFAOa, CRFCWO1, CRFCWO1d, CRFCWOa, CRFCWOad, CRFNO1, CRFNOa, CRFSO1, CRFSOa, DESSL, DESWC, DESWLlt, DESWLltd, DESWLsy, DESWLsyd, DRAP, DRINF, DRNEG, DRNP, DRPP, DRVP, LDMTLD, LDTTRa, LDTTRc, LDVOCD, LSAGN, LSAGNd, LSASS1, LSASS1d, LSASSp, LSASSpd, PCCNCp, PCCNCz, PCCONNp, PCCONNz, PCDCp, PCDCz, PCNARp, PCNARz, PCREFp, PCREFz, PCSYNp, PCSYNz, PCTEMPp, PCTEMPz, PCVERBp, PCVERBz, RDFRE, RDL2, SMCAUSlsa, SMCAUSr, SMCAUSv, SMCAUSvp, SMCAUSwn, SMINTEp, SMINTEr, SMTEMP, SYNLE, SYNMEDlem, SYNMEDpos, SYNMEDwrd, SYNNP, SYNSTRUTa, SYNSTRUTt, WRDADJ, WRDADV, WRDAOAc, WRDCNCc, WRDFAMc, WRDFRQa, WRDFRQc, WRDFRQmc, WRDHYPn, WRDHYPnv, WRDHYPv, WRDIMGc, WRDMEAc, WRDNOUN, WRDPOLc, WRDPRO, WRDPRP1p, WRDPRP1s, WRDVERB) showed significance over the repeated measure. 

In Readability feature set, 2 features (Flesch-Reading-Ease, No-of-syllables) showed significance over the repeated measure. In Psycholinguistics factors’ rating feature set, 15 features (AoaLow, AoaMid, ConHigh, ConMid, EmoHigh, EmoLow, EmoMid, FamHigh, FamLow, FamMid, FrqHigh, FrqLow, FrqMid, ImgHigh, ImgMid) showed significance over the repeated measure. 

All of the said features had absolute t value more than 2.0 and so, the effect of these features was considered to be statistically significant at the $\alpha = 0.05$.

\subsection{ANOVA Analysis of LMER Model}

Table \ref{tab:Chapter-4tab10} shows f-values of ANOVA analysis of the LMER models on each feature-set of sessions 1 and 2. ANOVA analysis of the LMER models gave f-value of the linguistic features, group (high, low), trial day (1, 2, 3) and scores (1, 2, 3, 4, 5) as a whole. Here, linguistic feature value was dependent variable (DV), whereas linguistic feature name, group, day and score were independent variables (IV). In the table, f-values show that feature, group label and score were significantly influenced during repeated reading. We also found that linguistics feature-sets of session-1 showed relativity high f-value, which meant the variability of group means was large relative to the within-group variability. In order to reject the null hypothesis that the group means were equal, we needed a high f-value. Thus, the feature-sets of session-1 discriminated the group (i.e., null hypothesis rejection) more effectively than those of session-2. 
 
\begingroup
\setstretch{1.4}
{\small	
	\begin{longtable}
		{>{\raggedright}m{2cm} C{2.2cm} C{2.2cm} C{1.8cm} C{1.9cm} C{2.8cm}}
		\captionsetup{font={small}, labelfont=bf, skip=4pt}
		\caption{ANOVA analysis of linguistic feature-sets}
		\label{tab:Chapter-4tab10}\\
		\toprule\midrule

			{\footnotesize\textbf{Variable}} & {\footnotesize\textbf{Vocabulary Knowledge}} & {\footnotesize\textbf{Text Structure Complexity}} & {\footnotesize\textbf{Cohesion}}  & {\footnotesize\textbf{Readability}} & {\footnotesize\textbf{Psycholinguistics Factors’ Rating}}\\
				\midrule
				\multicolumn{6}{l}{\textbf{ANOVA analysis of the session-1 linguistic feature-sets}} \\
				\midrule
				Feature & 863.87\textsuperscript{***} & 822.74\textsuperscript{***} & 6137.0\textsuperscript{***} & 834.59\textsuperscript{***} & 689.86\textsuperscript{***}\\
				Group\\ (High, Low) & 111.46\textsuperscript{***} & 109.92\textsuperscript{***} & 25.73\textsuperscript{***} & 70.18\textsuperscript{***} & 105.05\textsuperscript{***}\\
				Day\\ (1, 2, 3) & 147.0\textsuperscript{***} & 52.48\textsuperscript{***} & 25.73\textsuperscript{***} & 70.18\textsuperscript{***} & 37.0\textsuperscript{***}\\
				Score\\ (1, 2, 3, 4, 5) & 157.16\textsuperscript{***} & 82.61\textsuperscript{***} & 16.44\textsuperscript{***} & 36.02\textsuperscript{***} & 189.04\textsuperscript{***}\\ 
				\midrule
				\multicolumn{6}{l}{\textbf{ANOVA analysis of the session-2 linguistic feature-sets}} \\
				\midrule			
				Feature & 668.06\textsuperscript{***} & 616.28\textsuperscript{***} & 5216.9\textsuperscript{***} & 590.27\textsuperscript{***} & 540.2\textsuperscript{***}\\
				Group\\ (High, Low) & 33.03\textsuperscript{***} & 33.27\textsuperscript{***} & 7.19\textsuperscript{*} & 32.33\textsuperscript{***} & 34.93\textsuperscript{***}\\
				Day\\ (1, 2, 3) & 52.0\textsuperscript{***} & 34.7\textsuperscript{***} & 6.38\textsuperscript{***} & 10.56\textsuperscript{***} & 22.21\textsuperscript{***}\\
				Score\\ (1, 2, 3, 4, 5) & 119.98\textsuperscript{***} & 89.68\textsuperscript{***} & 6.59\textsuperscript{***} & 45.99\textsuperscript{***} & 125.78\textsuperscript{***}\\ 
				\midrule
				\multicolumn{6}{l}{\textsuperscript{*}: $p<0.05$, \textsuperscript{**}: $p<0.01$, \textsuperscript{***}: $p<0.001$} \\
				\midrule\bottomrule
			\end{longtable}}
			\endgroup

We can also see that the variability of the means of high and low was large relative to the within variability of each label in session-1 as compare to those in session-2. Thus, we can infer that the labels represented participants' writing level more significantly in session-1 summary than in session-2 summary. The same inference can be drawn for day and score also.
\section{Conclusion}

This chapter presented an approach to label as well as to score students’ writings based on different linguistics complexity indices in L2 repeated reading intervention, where reading two texts simulated extensive and intensive readings. We used five kinds of feature-sets-- Lexical richness, Syntactic complexity, Cohesion, Readability, Psycholinguistics factors' rating for measuring L2 participants' text production (writing) skills. For predicting labels of summary text, we applied three classifiers-- SVM, DT \& MLP and to score these texts, SVR regression model was used. Among the classifiers, SVM performed with highest accuracy to predict the label of participants. Also, Pearson correlation coefficient and QWK agreement between the regression model's prediction and human rater's score were higher on significant features.

The LMER models showed that the fixed-effect variables– score, label and day significantly influenced L2 linguistics features in repeated reading intervention. The results of LMER models indicated that score and label were significantly achieved for the both sessions of the trial days. The models also predicted significance in performance improvement (score and label) from day- 1 to 2, day-1 to 3 as well as day-2 to 3. The ANOVA analysis of the LMER models showed that the fixed-effect variables– score, label and day significantly influenced overall L2 writing measures. By comparing the ANOVA results of both sessions' LMER models, we inferred that labels, score, day and feature-sets could represent participants' writing more significantly in session-1 summary than in session-2 summary.

The future work of this study could be to analyse the 2-way and 3-way interaction in LMER models, such as the effect of specific trial day preference could enhance the preference of specific L2 writing measures on L2 linguistics proficiency in specific type of readings (e.g., extensive and intensive) and writings (e.g., narrative and descriptive) with various fixed variables such as gender, age-group, reading frequency, the gap between readings, various L1/L2 language, ethnicity, social status etc.

\chapter{Content Enrichment Analysis of English as Foreign Language Learners' Writings}
\label{Chapter5} 
\lhead{Chapter 5. \emph{Content Enrichment Analysis of EFL Learners' Writings}} 

\section{Introduction}

Writings reveal learners' ability to integrate, synthesize, design, and communicate their ideas in natural language. Therefore, it is an important part of assessing learners' acquired knowledge. Assessment of writings (essays, summaries, descriptive answers etc.) is more laborious and subjective for humans as well as much harder to automate. Several researchers developed some techniques to automate the assessment of writings (short free-text such as an essay) \cite{fauzi2017automatic, contreras2018automated, tashu2018layered, tashu2018pair, wang2018automatic, horbach2019influence, darwish2019automated, hendre2020efficacy, abimanyu2020automatic, peng2021research}. Developing a general solution to this is a hard problem owing to multiple reasons viz. linguistic variations in student answers (multiple ways of expressing the same answer), subjectivity of questions (multiple correct answers) and topical variations (science vs literature).

In this chapter, we explain different techniques to score learners' free-text writing and to analyse the effect of repeated reading on their writing score. The chapter first describe the context, i.e. description of available techniques and tools for calculating content similarity. Then we propose our methodology for measuring content similarity between texts. Next, we present the results of linear mixed-effects regression (LMER) models to analyse the significant impact of RR on the content score in writings. At last, we discuss the results and conclude the outcomes. 


\section{Techniques and Tools}

Similarity between textual contents becomes important in text-related research and applications such as information retrieval, text classification, document clustering, topic detection, topic tracking, questions generation, question answering, essay scoring, short answer scoring, machine translation, text summarization etc. The task of understanding whether two texts are on the same topic or are somehow related still remains difficult for NLP program. The main difficulties behind automatic text comparison are semantic ambiguity of words and lexical and syntactic differences between sentences. 

In this section, we review techniques and tools which had been used for the automatic essay scoring (AES) in the prior art. All such techniques are briefly described under key themes and are presented in an organized manner.

\subsection{Word Similarity Measures}

Approaches to determine the semantic similarity of sentences use measures of semantic similarity between individual words. Three types of approaches were introduced- corpus-based, knowledge-based and hybrid-based approaches \cite{gomaa2013survey}. Corpus-Based similarity measure determines the similarity between words according to information gained from large corpora. Knowledge-Based similarity measure determines the degree of similarity between words using information derived from semantic networks. Hybrid-based similarity measure uses both corpus-based and knowledge-based measures of similarity. Ferreira et al. \cite{ferreira2016assessing} proposed a measure to assess the degree of sentence similarity based on lexical, syntactic and semantic analysis. Neculoiu et al. \cite{neculoiu2016learning} presented a deep neural network architecture for learning a similarity metric on variable-length character sequences. Tsekouras et al. \cite{tsekouras2017graph} introduced a text-similarity measure which employs named-entities' information extracted from the text and the n-gram graphs' model for representing documents.


\subsection{Semantic Similarity with Word Embeddings}

For semantic features, many approaches use external sources of structured semantic knowledge such as Wikipedia \cite{gabrilovich2007computing} or WordNet \cite{ferreira2016assessing}. Wikipedia is structured around entities and undergoes constant development so its breadth and depth steadily increase over time. A drawback of using dictionaries or WordNet is that proper names, domain-specific technical terms and slang tend to be underrepresented. In distributional semantics, a large amount of unlabelled text data is used to create a semantic space. Words/terms are represented in this semantic space as vectors that are called word embeddings. The geometric properties of this space prove that words that are semantically or syntactically similar tend to be close in the semantic space. Several researchers (e.g., \cite{nguyen2019learning, kenter2015short}) used word embeddings models such as Word2vec- word vector \cite{mikolov2013distributed}, Glove \cite{pennington2014glove} or ELMo- Embeddings from Language Models \cite{peters2018deep} for determining similarity from word-level to text-level semantics. 


\subsection{Sentence Similarity Measures}

The syntactic structure of the compared sentences influences the calculation of semantic similarity. The meaning of a sentence is made up of not only the meanings of its individual words, but also the structural way the words are combined. Therefore, in these approaches, multiple sentences or a text as a whole were processed to obtain the semantic vector and calculate the similarity of two short texts by performing algebraic operations on the obtained vectors. Under this approach, several methods have been developed including, Latent Semantic Analysis (LSA) \cite{rus2013semilar}, Probabilistic Latent Semantic Analysis (PLSA), Latent Dirichlet Analysis (LDA), Bag Of Words (BOW), Explicit Semantic Analysis (ESA)\cite{gabrilovich2007computing} etc.


\subsection{Semantic Similarity between Concepts}

Each word carries one or multiple concepts. For example, a word - `bank' can represent a financial organization, the edge of a river, or storage facility. Therefore, the meanings of a word in two different sentences may denote two very different meanings. One popular solution of such problem is obtaining semantic information from WordNet \cite{miller1998wordnet}. WordNet is a lexical database of semantic relations between words. It links words into semantic relations including synonyms, hyponyms, and meronyms. Using its syntactic structure, several researchers already proposed different types of measures of semantic similarity between concepts, including, path-based measures \cite{hirst1998lexical}, information content measures \cite{resnik1995using}, and gloss-based measures \cite{banerjee2003extended}. 


\subsection{Graph-based Semantic Similarity}

Knowledge graphs encode semantics that describe resources in terms of several aspects, e.g., hierarchies, neighbours, and node degrees. Recently, the impact of aspects on the problem of determining relatedness between entities in a knowledge graph has been shown, and semantic similarity measures for knowledge graphs have been proposed \cite{paul2016efficient, pilehvar2015senses}. Traverso et al. \cite{traverso2016gades} proposed GADES (Graph-bAseD Entity Similarity), a semantic similarity measure for comparing entities in knowledge graphs. GADES considers the knowledge encoded in aspects, e.g., hierarchies, neighbourhoods, and to determine relatedness between entities in a knowledge graph. Publicly available knowledge graphs like DBpedia\footnote{\url{http://dbpedia.org}}, Yago\footnote{\url{http://yago-knowledge.org}}, or ConceptNet\footnote{\url{https://conceptnet.io/}} represent general domain concepts such as films, politicians, or sports, using RDF vocabularies.


\subsection{Content Similarity using Concept Maps}

In academics, as compared to using multiple-choice questions concept maps can be utilized more effectively to determine the depth of knowledge possessed by a student. Concept maps can provide visual data on students misconceptions and their level of understanding. Jain et al. \cite{jain2013artificial, jain2014artificial} proposed a tool coined as Artificial Intelligence Based Student Learning Evaluation Tool (AISLE) to apply artificial intelligence techniques in evaluating a student's understanding of a particular topic of study using concept maps. Similarly, Gurupur et al. \cite{gurupur2015evaluating} described a tool that could be effectively used to evaluate student learning outcomes using concept maps and Markov chain analysis.


\section{Problem}

In this chapter, we propose a hypothesis that by computing content similarity between the referred text and L2 learners' text written during repeated reading, their content writing proficiency could be automatically scored as human rater did. We also propose another hypothesis that repeated reading has an impact on similarity scores and show improvement in learners' writing skills in terms of content enrichment.

In order to test these hypotheses, we used the data, described already in the chapter- 4, collected by conducting a reading and writing experiment, in which participants attended two repeated reading sessions. In a session, they read one text and after finishing the reading of the text, they wrote a summary of the text, as much detail as they could. This practice was done by all fifty participants once in a day for consecutive three days. The reading of text-1 and text-2 had given the experience of extensive and intensive reading respectively. 

Black and Bower \cite{black1980story} investigated how people understand and recall simple stories. They proposed a hierarchical state transition (HST) network, to predict the recall probabilities of different statements in stories. They stated, ``the best remembered part of a story was the critical path that provided the transition from the beginning state to the ending state of the story. If the story described the critical path at various levels of detail, then the higher (i.e., more general, less detailed) level of a statement would be better remembered." 

As shown in table \ref{tab:Chapter-2tab2}, both texts-- 1 and 2 contained more than 650 words in 30+ sentences. According to HST network theory, It was quite impossible for participants to write a summary of the text exactly the same as it was. In the first day, they wrote more general statements and in later days, they included fine details. Also, they had not written all events of the text as well as had not followed the same sequence as were in the text. Therefore, determining the degree of conceptual similarity between the text and the summaries become a challenging task.


\section{Method}

In this section, we present our proposed method to determine the content similarity between the read text and participants' written summaries. The overall working process contains two main modules: preprocessing the texts and determining similarity. 


\subsection{Preprocessing}
A text may contain one or more sentences. Each sentence carries some unique information. Therefore, the first step of the preprocessing was to split the text into sentences without changing the syntactic structure. Second, the tokeniser split the text into simple tokens, such as numbers, punctuation, symbols, and words of different types (e.g. with an initial capital, all upper case, etc). Punctuation marks- the period, exclamation point, comma, semicolon, colon, dash, apostrophe etc. and stop words- a, an, the etc. did not contribute much information in measuring the similarity. Therefore, in the next step, such punctuation marks, symbols, stop words were removed, but still maintained the structure of sentences. Now named entity recognition (NER) and NE coreference resolution were performed. In text-2, `Andrew Jackson', `Jackson', `president', `the head of Democratic party'-- they represented the same entity though they were different in words. To resolve such problems, we trained SpaCy\footnote{\url{https://spacy.io/}} models and then models were used for NER and coreference resolution. The models achieved low performance as shown in table \ref{tab:Chapter-5tab4}. Therefore, we manually resolved issues regarding coreference resolution. After that, we performed word segmentation to group tokens of the same phrasal verbs or idioms. For example, when considering `paper money aristocracy' as three distinct tokens, it would not express the real meaning since `paper money aristocracy' was related to `bankers' in the text-2. Therefore, word segmentation to recognize phrases helped maintain the meaning of the text. In the last step of the preprocessing, each of the tokens was translated into its basic form using SpaCy. For example, noun words in plural form were made singular and all verb forms (v2, v3, v4 and v5) were replaced by basic form (v1). 

For example, below two lines (1 \& 2) were preprocessed as stated above i.e. i) splitting text into sentences, ii) splitting sentences into tokens, iii) removing noise (punctuation marks, symbols, and stop words), iv) named entity coreference resolution, v) word segmentation, and vi) converting word into the basic form. \\
\textbf{line-1 from text-1:} “Presently the horse came to him on Monday morning, with a saddle on his back. The horse said, Camel, O Camel, come out and trot like the rest of us.”\\
\textbf{line-2 from a summary text:} “Initially, a horse came to the camel and he said to do some work.”
\begin{enumerate}[(i)]
\item \textbf{splitting text into sentences:}\\
line1\_1 = Presently the horse came to him on Monday morning, with a saddle on his back.\\
line1\_2 = The horse said, Camel, O Camel, come out and trot like the rest of us.\\
line2\_1 = Initially, a horse came to the camel.\\
line2\_2 = he said to do some work.
\item \textbf{splitting sentences into tokens:}\\
line1\_1 = [presently, the, horse, came, to, him, on, monday, morning, ,, with, a, saddle, on, his, back, .]\\
line1\_2 = [the, horse, said, ,, camel, ,, o, camel, , come, out, and, trot, like, the, rest, of, us, .]\\
line2\_1 = [initially, ,, a, horse, came, to, the, camel, .]\\
line2\_2 = [he, said, to, do, some, work, .]
\item \textbf{removing noise:}\\
line1\_1 = [presently, horse, came, him, monday, morning, saddle, his, back]\\
line1\_2 = [horse, said, camel, o, camel, come, out, trot, like, rest, of, us]\\
line2\_1 = [initially, horse, came, camel]\\
line2\_2 = [he, said, do, work]
\item \textbf{named entity coreference resolution:}\\
line1\_1 = [presently, horse, came, camel, monday, morning, saddle, horse, back]\\
line1\_2 = [horse, said, camel, o, camel, come, out, work, like, animal]\\
line2\_1 = [initially, horse, came, camel]\\
line2\_2 = [horse, said, do, work]
\item \textbf{word segmentation:}\\
line1\_1 = [presently, horse, came, camel, monday, morning, {saddle, horse, back}]\\
line1\_2 = [horse, said, {camel, o, camel}, come, out, work, like, animals]\\
line2\_1 = [initially, horse, came, camel]\\
line2\_2 = [horse, said, do, work]
\item \textbf{converting word into the basic form:}\\
line1\_1 = [presently, horse, come, camel, monday, morning, {saddle, horse, back}]\\
line1\_2 = [horse, say, {camel, o, camel}, come, out, work, like, animal]\\
line2\_1 = [initially, horse, come, camel]\\
line2\_2 = [horse, say, do, work]
\end{enumerate}	


\subsection{Determining Similarity}

We proposed a method to determine concept similarity between the read text and the summaries using three levels of text- lexical, syntactic and semantic, as explained by Ferreira et al. \cite{ferreira2016assessing}. 


\subsubsection{Lexical Similarity}

For calculating lexical similarity among the words of two sentences, we used four external knowledge sources- WordNet\footnote{\url{https://wordnet.princeton.edu/download}},  Word2vec\footnote{\url{https://code.google.com/archive/p/word2vec}}, GloVe\footnote{\url{https://nlp.stanford.edu/projects/glove}}, and SpaCy web-large-vector model\footnote{\url{https://spacy.io/models/en-starters\#en\_vectors\_web\_lg}}.\\ 
In WordNet based calculation, the distance or similarity between the concepts of two words in WordNet Graph was calculated. There were six measures used to calculate the similarity between words- path measure, Resnik measure, Lin measure, Wu and Palmer measure, and Leacock and Chodorow measure. In our approach, we relied on the first one. \\
In word embedding based calculation, we used three pre-trained word embeddings- Word2vec, GloVe and a SpaCy word-vector. Word2vec is 300-dimensional vectors trained on the 100 billion-word Google News dataset. The model covers more than 3 million words and phrases. GloVe embedding is also 300-dimensional vectors. The word vectors are trained on a very large corpus of 840 billion tokens corpus. A SpaCy model for English is also 300-dimensional vectors having 1.1 million keys i.e., 1.1 million unique vectors. Gensim\footnote{\url{https://radimrehurek.com/gensim}}, an open-source library for unsupervised topic modelling and natural language processing, has a Python\footnote{\url{https://www.python.org}} implementation of Word2vec which provides an in-built utility for finding similarity between two given words. 

We observed that several words were not found in these sources. These out of vocabulary (OOV) words might include the name of a person and organisation, specific domain words etc. To handle OOV words, we applied the Levenshtein distance metric to calculate the character-based similarity. The Levenshtein distance is a string metric for measuring the difference between two words, which is the minimum number of single-character edits (i.e. insertions, deletions or substitutions) required to change one word into the other.

For measuring similarity between numbers (e.g., year in figures), if two numbers had the same digits were given the highest score 1.0, otherwise the lowest score (0.0). 
 
The relations between words that achieved similarity score lower than a threshold (here, 0.7) using any source were discarded. To obtain a similarity for two sentences, maximum similarity score between their word pairs were added, and then the total was divided by the number of word pairs to normalise the similarity value. All unique word (token) pairs of sentences were used to calculate their syntactic similarity in the next step. The lexical similarity between words of lines 1 \& 2 are given in table \ref{tab:Chapter-5tab1}.

\begingroup
\setstretch{1.4}
	{\small 
	\begin{longtable}
		{>{\raggedright}m{3.2cm} C{1.8cm} C{1.8cm} C{1.2cm} C{1.2cm} C{2.6cm}}
		\captionsetup{font={small}, labelfont=bf, skip=4pt}
		\caption{Lexical similarity of word-pairs}
		\label{tab:Chapter-5tab1}\\
		\toprule\midrule
		\textbf{Word pairs\\ (lines 1 and 2)} & \textbf{WordNet} & \textbf{Word2vec} & \textbf{GloVe} & \textbf{SpaCy} & \textbf{Levenshtein distance}\\
		\midrule
		
		\textbf{presently, initially} & 0.0 & 0.19 & 0.16 & \textbf{0.78} & 0.22\\ 
		\textbf{horse, horse} & \textbf{1.0} & \textbf{1.0} & \textbf{1.0} & \textbf{1.0} & \textbf{1.0}\\ 
		\textbf{come, come} & \textbf{1.0} & \textbf{1.0} & \textbf{1.0} & \textbf{1.0} & \textbf{1.0}\\ 
		\textbf{camel, camel} & \textbf{1.0} & \textbf{1.0} & \textbf{1.0} & \textbf{1.0} & \textbf{1.0}\\ 
		monday, say & 0.04 & 0.11 & 0.36 & 0.45 & 0.33\\ 
		morning, horse & 0.05 & 0.14 & 0.17 & 0.57 & 0.29\\ 
		saddle, horse & 0.06 & 0.54 & 0.46 & 0.65 & 0.17\\ 
		back, horse & 0.0 & 0.19 & 0.33 & 0.39 & 0.0\\ 
		\textbf{say, say} & \textbf{1.0} & \textbf{1.0} & \textbf{1.0} & \textbf{1.0} & \textbf{1.0}\\ 
		o, work & 0.09 & 0.05 & 0.08 & 0.24 & 0.25\\ 
		out, work & 0.0 & 0.15 & 0.53 & 0.02 & 0.25\\ 
		\textbf{work, work} & \textbf{1.0} & \textbf{1.0} & \textbf{1.0} & \textbf{1.0} & \textbf{1.0}\\ 
		like, come & 0.0 & 0.35 & 0.68 & 0.38 & 0.25\\ 
		animal, horse & 0.11 & 0.45 & 0.4 & 0.54 & 0.0\\ 
		\midrule\bottomrule
	\end{longtable}}
\endgroup

\begingroup
\small 
\begin{multline}
\label{eqn:Chap-5eqn1} 
total\_lexical\_similarity\; (TLS) = \\\frac{\sum_{i=1}^{n} \{_{lexical\_similarity\; +=\;0,\; otherwise}^{lexical\_similarity\; +=\;score,\; if\; score\; >=\; threshold}\}}{1 + number\; of\; matched\; word-pairs}\\
where,\; n\; is\; the\; number\; of\; all\; word-pairs.
\end{multline}
\begin{flalign}
\label{eqn:Chap-5eqn2} 
&TLS_{(line1\&2)} = (0.78 + 1.0 + 1.0 + 1.0 + 1.0 + 1.0) / (1 + 6.0) = 0.82 & 
\end{flalign}
\endgroup
\subsubsection{Syntactic Similarity}
\begin{figure}[h!]
	\centering
	\includegraphics[scale=0.4]{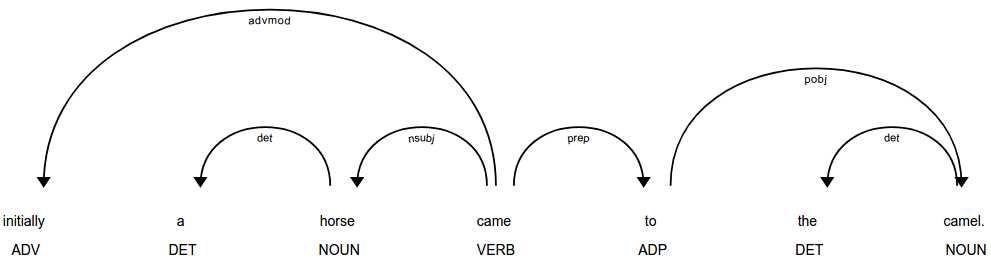}
	\captionsetup{font={small}, labelfont=bf, skip=4pt}
	\caption{A Dependency tree}
	\label{fig:Chapter-5fig1}
\end{figure}

In this step, dependency relations (root, auxiliary verb, subject, direct object etc.) and dependency distance were used to calculate the syntactic similarity between two sentences. Dependency distance (DD) was measured by the linear distance between two syntactically related words in a sentence. Linguistician agreed that DD was an important index of memory burden and an indicator of syntactic difficulty \cite{haitao15probability}. SpaCy provides a deep syntactic parser tree having syntactic relation and showing dependency among words of a sentence. Figure \ref{fig:Chapter-5fig1} shows a dependency tree generated by SpaCy for line2\_1 sentence. For calculating the syntactic similarity between two sentences, their token (word) pairs, found in the previous step, were used. 

Each token contained quartet- base form (lemma), part-of-speech tag as well as dependency relation and dependency distance from other peers. Table \ref{tab:Chapter-5tab2} shows aligned tokens of lines 1 and 2; and each token displays its base word, POS tag, dependency relation (DR) and dependency distance (DD). Using this data, we calculated the mean dependency distance (MDD) of sentences and the mean dependency relation similarity (MDRS) between two sentences.

\begingroup
\setstretch{1.4}
{\small 
	\begin{longtable}
		{>{\raggedright}m{1.5cm} C{1.5cm} C{1.5cm} C{1.3cm} L{1.5cm} C{1.6cm} C{1.5cm} C{1.3cm}}
		\captionsetup{font={small}, labelfont=bf, skip=4pt}
		\caption{Token-quartet}
		\label{tab:Chapter-5tab2}\\
		\toprule\midrule
		\multicolumn{4}{c}{\textbf{Line-1}} & \multicolumn{4}{c}{\textbf{Line-2}}\\
		\midrule
		\textbf{Base word} & \textbf{POS tag} & \textbf{DR} & \textbf{DD} & \textbf{Base word} & \textbf{POS tag} & \textbf{DR} & \textbf{DD}\\
		\midrule
		presently & ADV & advmod & 1 & initially & ADV & advmod & 1\\
		horse & NOUN & nsubj & 1 & horse & NOUN & nsubj & 1\\
		come & VERB & ROOT & 0 & come & VERB & ROOT & 0\\
		camel & NOUN & pobj & 2 & camel & NOUN & pobj & 2\\
		say & VERB & ROOT & 0 & say & VERB & ROOT & 0\\
		work & VERB & conj & 2 & work & NOUN & dobj & 2\\
		\midrule\bottomrule
	\end{longtable}}
	\endgroup
We calculated the mean dependency distance (MDD), using the following formula:
\begingroup
\small 
\begin{multline}
\label{eqn:Chap-5eqn3} 
mean\; dependency\; distance \; (MDD) = \frac{\sum_{i=1}^{n} \{{DD_i}\}}{n}\\
where,\; n\; is\; the\; number\; of\; words\; in\; a\; set.
\end{multline}
\endgroup
Similarly, we calculated the mean dependency relation similarity (MDRS) of word-pairs,
\begingroup
\small 
\begin{multline}
\label{eqn:Chap-5eqn4} 
mean\; dependency\; relation\; similarity \; (MDRS) = \\\frac{\sum_{i=1}^{n} \{_{DR\_similarity\; +=\; 0,\; otherwise}^{DR\_similarity\; +=\; 1,\; if\; DR\_word1,i\; ==\; DR\_word2,i}\}}{n}\\
where,\; n\; is\; the\; number\; of\; word-pairs.
\end{multline}
\begin{flalign}
\label{eqn:Chap-5eqn5} 
&total\_syntactic\_similarity\: (TSS) = \frac{1}{1 + |MDD_1 - MDD_2|} + MDRS &
\end{flalign}
\endgroup
Therefore, two sentences having lower differences in their dependency distance as well as having more number of same dependency relation were given a higher syntactic similarity score.
\begingroup
\small 
\begin{flalign} 
\label{eqn:Chap-5eqn6} 
&MDD\textsubscript{(line1)}= MDD\textsubscript{(line2)} = 6/6 = 1.0 & \\
\label{eqn:Chap-5eqn7} 
&MDRS\textsubscript{(line1\&2)} = 5.0/(6-1) = 0.8 & \\
\label{eqn:Chap-5eqn8} 
&TSS\textsubscript{(line1\&2)} = 1.0 + 0.8 = 1.8 &
\end{flalign}
\endgroup

\subsubsection{Concept Similarity}
Words represent events and concepts. In a text, words are written such that readers can perceive events and concepts in an order (e.g., chronicle order). We observed that participants had not written all events and concepts of the read text in their summaries; as well as, the events and concepts, written in summaries, loosely followed the sequence of the original text.  
In this step, we determined how many events and concepts of the original text were mentioned in a summary using the following formula:\\
\begin{equation}
\small
\label{eqn:Chap-5eqn9} 
concept\_score = \frac{total\_events\_\&\_concepts\_found\_in\_summary}{total\_events\_\&\_concepts\_in\_original\_text}
\end{equation}
We also compared the sequence of events and concepts written in a summary with that of the original text using pairwise sequence alignment technique. In sequence alignment, the two sequences were given a score on how similar (or different) they were to each other. Three basic aspects were considered when assigning scores. They were,\\
\textbf{Match-value —} Value assigned for matching events and concepts; we set it as 2.0.\\
\textbf{Mismatch-value —} Value assigned for mismatching events and concepts; we set it as -1.0.\\
\textbf{Gap-penalty —} Value assigned for missed events and concepts; we set it as -0.5.
\begingroup
\small 
\begin{multline}
\label{eqn:Chap-5eqn10} 
alignment\_score = no.\; of\; match\; \times\; match\_value\; +\; no.\; of\; mismatch\; \times\\
 mismatch\_value\; +\; no.\; of\; gap\; \times\; gap\_penalty
\end{multline}
\begin{equation}
\label{eqn:Chap-5eqn11} 
normalized\_alignment\_score = \frac{alignment\_score}{total\_events\_\&\_concepts\_in\_original\_text}
\end{equation}
\begin{flalign}
\label{eqn:Chap-5eqn12} 
&total\_concept\_similarity\; (TCS) = concept\_score + normalized\_alignment\_score &
\end{flalign}
\endgroup
In the above example, line2\_1 had- four matched concept, zero mismatched concepts, and three missed (gap) concepts. line2\_2 had- three matched concept, one mismatched concept, and three missed (gap) concepts.
\begingroup
\small 
\begin{equation}
\label{eqn:Chap-5eqn13} 
\begin{split}
concept\_score_{(line1\&2)} & = \frac{total\_events\_\&\_concepts\_in\_line2\_1\_\&\_line2\_2 }{total\_events\_\&\_concepts\_in\_line1\_1\_\&\_line1\_2} \\
& = \frac{4 + 3 }{7 + 6} = 0.54
\end{split}
\end{equation}
\begin{equation}
\label{eqn:Chap-5eqn14} 
alignment\_score_{(line1\&2)} = (4 \times 2.0-0 \times 1.0-3 \times 0.5) + (3 \times 2.0-1 \times 1.0-3 \times 0.5) = 10.0 
\end{equation}
\begin{flalign}
\label{eqn:Chap-5eqn15} 
&normalized\_alignment\_score_{(line1\&2)} = \frac{10.0} {7 + 6} = 0.77 & \\
\label{eqn:Chap-5eqn16} 
&TCS_{(line1\&2)} = 0.54 + 0.77 = 1.31 &
\end{flalign}
\endgroup

\subsubsection{Overall Semantic Similarity}

To calculate overall semantic similarity between the read text and a summary, we combined the three similarity measures in order to have a global view of the degree of similarity between the text. Below formula presented the combination of the lexical, syntactic and concept measures to provide the overall measure of sentence similarity.

\begingroup
\small 
\begin{flalign}
\label{eqn:Chap-5eqn17} 
&Semantic\_Similarity = TLS + TSS + TCS &
\end{flalign}
\endgroup
Therefore, the semantic similarity between example lines 1 and 2 was calculated as:
\begingroup
\small 
\begin{flalign}
\label{eqn:Chap-5eqn18} 
&Semantic\_Similarity\textsubscript{(line1\&2)} = 0.82+1.8+1.31 = 3.93 &
\end{flalign}
\endgroup

\section{Results and Discussion}
Table \ref{tab:Chapter-5tab3} shows the average of semantic similarity score of the participants grouped into high and low. The average score of both groups increased as trials proceeded from day-1 to day-3. The table also shows that the semantic similarity, calculated using WordNet, was worst among all the four knowledge sources, because of relatively more concept words were absent in WordNet.

\begingroup
\setstretch{1.4}
{\small 
	\begin{longtable}
		{>{\raggedright}m{3.5cm} C{1.3cm} C{1.3cm} C{1.5cm} C{1.3cm} C{1.3cm} C{1.5cm}}
		\captionsetup{font={small}, labelfont=bf, skip=4pt}
		\caption{Average of Semantic similarity at the group level}
		\label{tab:Chapter-5tab3}\\
		\toprule\midrule

		\multirow{3}{3.5cm}{\raggedright \textbf{Knowledge source}} & \multicolumn{3}{c}{Group: \textbf{High}} & \multicolumn{3}{c}{Group: \textbf{Low}}\\\cline{2-7}

		& \textbf{Day-1} & \textbf{Day-2} & \textbf{Day-3} & \textbf{Day-1} & \textbf{Day-2} & \textbf{Day-3}\\
		
		\midrule
		\multicolumn{7}{l}{\textbf{Session-1 Summary}}\\
		\midrule
		WordNet & 23.42 & 32.66 & 36.94 & 15.56 & 21.34 & 27.26\\
		Word2Vec & 160.06 & 233.78 & 281.62 & 100.22 & 146.98 & 195.4\\
		GloVe & 174.67 & 249.7 & 296.55 & 109.69 & 156.58 & 206.31\\
		SpaCy & 172.14 & 244.31 & 292.08 & 105.81 & 153.77 & 202.95\\
		
		\midrule
		\multicolumn{7}{l}{\textbf{Session-2 Summary}}\\
		\midrule
		WordNet & 9.51 & 13.85 & 17.54 & 5.9 & 10.77 & 13.05\\
		Word2Vec & 56.59 & 83.95 & 106.73 & 33.26 & 63.99 & 77.1\\
		GloVe & 72.42 & 108.83 & 133.25 & 43.75 & 83.51 & 99.33\\
		SpaCy & 63.77 & 96.87 & 119.18 & 39.29 & 75.4 & 88.16\\
		\midrule\bottomrule
	\end{longtable}}
	\endgroup
	
In tables \ref{tab:Chapter-5tab4}, \ref{tab:Chapter-5tab5}, and \ref{tab:Chapter-5tab6}, the semantic similarity between a summary text and the read text and raters' score given to the same summary text were used to calculate Pearson's correlation ($r$), Spearman's rank correlation coefficient ($\rho$), and QWK agreement. Their values are shown in columns 2, 3 and 4 respectively. The results, shown in table \ref{tab:Chapter-5tab4} were given when SpaCy models were used for NER and co-reference resolution. In the table, most correlation coefficients were below 0.5 for both sessions' summaries and thus the table shows weak correlations. Similarly, all QWK coefficients were below 0.5 for both sessions' summaries and thus the coefficients show slight to moderate agreements. Generally, the value of QWK greater than 0.6 is considered to be a good agreement.

\begingroup
\setstretch{1.4}
{\small	
	\begin{longtable}{>{\raggedright}m{1cm} C{5cm} C{6cm} C{1cm}}
	\captionsetup{font={small}, labelfont=bf, skip=4pt}
	\caption{Semantic similarity for summaries (NER and Coref. resolution using SpaCy model)} 
	\label{tab:Chapter-5tab4}\\
	\toprule\midrule
	\textbf{Day} & \textbf{Pearson correlation coefficient (\textit{r})} & \textbf{Spearman's rank correlation coefficient ($\rho$)} & \textbf{QWK}\\
	\midrule
	\multicolumn{3}{l}{\textbf{Semantic similarity for session-1 summaries (Knowledge source: SpaCy)}}\\
	\midrule
	1 & 0.29\textsuperscript{*} & 0.28 & 0.22\\
	2  & 0.4\textsuperscript{**} & 0.4\textsuperscript{**} & 0.33\\
	3 & 0.29\textsuperscript{*} & 0.27 & 0.34\\
	\midrule
	\multicolumn{3}{l}{\textbf{Semantic similarity for session-2 summaries (Knowledge source: SpaCy)}}\\
	\midrule
	1 & 0.43\textsuperscript{**} & 0.53\textsuperscript{***} & 0.34\\
	2 & 0.49\textsuperscript{***} & 0.5\textsuperscript{***} & 0.45\\
	3 & 0.48\textsuperscript{**} & 0.51\textsuperscript{**} & 0.48\\
	\midrule
	\multicolumn{4}{l}{\textsuperscript{*}: $p<0.05$, \textsuperscript{**}: $p<0.01$, \textsuperscript{***}: $p<0.001$} \\
	\midrule\bottomrule
	\end{longtable}
}\endgroup

The results, shown in tables \ref{tab:Chapter-5tab5} and \ref{tab:Chapter-5tab6} were given by the proposed method when all NER and coreference resolution issues were resolved manually. The second and third columns of both tables show strong correlations between the semantic similarity and the rater's score on each knowledge source across the days. The fourth column of both tables also shows better agreement between the similarity score and the rater's score on each knowledge source across the days. 

\begingroup
\setstretch{1.4}
{\small	
	\begin{longtable}{>{\raggedright}m{1cm} C{4cm} C{5cm} C{1cm}}
		\captionsetup{font={small}, labelfont=bf, skip=4pt}
		\caption{Semantic similarity for session-1 summaries (NER and Coreference resolution  by manual)} 
		\label{tab:Chapter-5tab5}\\
		\toprule\midrule
		\textbf{Day} & \textbf{Pearson correlation coefficient (\textit{r})} & \textbf{Spearman's rank correlation coefficient ($\rho$)} & \textbf{QWK}\\
		\midrule
		\multicolumn{3}{l}{\textbf{Knowledge source: WordNet}}\\
		\midrule
		1  & 0.86\textsuperscript{***} & 0.85\textsuperscript{***} & 0.78\\ 
		2  & 0.87\textsuperscript{***} & 0.87\textsuperscript{***} & 0.86\\ 
		3 & 0.82\textsuperscript{***} & 0.83\textsuperscript{***} & 0.7\\ 
		\midrule
		\multicolumn{3}{l}{\textbf{Knowledge source: Word2Vec}}\\
		\midrule
		1 & 0.8\textsuperscript{***} & 0.84\textsuperscript{***} & 0.54\\ 
		2 & 0.84\textsuperscript{***} & 0.83\textsuperscript{***} & 0.76\\ 
		3 & 0.89\textsuperscript{***} & 0.9\textsuperscript{***} & 0.83\\ 
		\midrule
		\multicolumn{3}{l}{\textbf{Knowledge source: GloVe}}\\
		\midrule
		1 & 0.82\textsuperscript{***} & 0.87\textsuperscript{***} & 0.56\\ 
		2 & 0.86\textsuperscript{***} & 0.86\textsuperscript{***} & 0.77\\ 
		3 & 0.9\textsuperscript{***} & 0.91\textsuperscript{***} & 0.83\\ 
		\midrule
		\multicolumn{3}{l}{\textbf{Knowledge source: SpaCy}}\\
		\midrule
		1 & 0.8\textsuperscript{***} & 0.84\textsuperscript{***} & 0.5\\ 
		2 & 0.85\textsuperscript{***} & 0.85\textsuperscript{***} & 0.77\\ 
		3 & 0.91\textsuperscript{***} & 0.92\textsuperscript{***} & 0.85\\ 
		\midrule
		\multicolumn{4}{l}{\textsuperscript{*}: $p<0.05$, \textsuperscript{**}: $p<0.01$, \textsuperscript{***}: $p<0.001$} \\
	\midrule\bottomrule
\end{longtable}
}\endgroup


\begingroup
\setstretch{1.4}
{\small	
	\begin{longtable}{>{\raggedright}m{1cm} C{4cm} C{5cm} C{1cm}}
		\captionsetup{font={small}, labelfont=bf, skip=4pt}
		\caption{Semantic similarity for session-2 summaries (NER and Coreference resolution  by manual)} 
		\label{tab:Chapter-5tab6}\\
		\toprule\midrule
		\textbf{Day} & \textbf{Pearson correlation coefficient (\textit{r})} & \textbf{Spearman's rank correlation coefficient ($\rho$)} & \textbf{QWK}\\
		\midrule
		\multicolumn{3}{l}{\textbf{Knowledge source: WordNet}}\\
		\midrule
		1 & 0.9\textsuperscript{***} & 0.89\textsuperscript{***} & 0.76\\ 
		2 & 0.88\textsuperscript{***} & 0.88\textsuperscript{***} & 0.83\\ 
		3 & 0.84\textsuperscript{***} & 0.85\textsuperscript{***} & 0.78\\ 
		\midrule
		\multicolumn{3}{l}{\textbf{Knowledge source: Word2Vec}}\\
		\midrule
		1 & 0.81\textsuperscript{***} & 0.8\textsuperscript{***} & 0.73\\ 
		2 & 0.82\textsuperscript{***} & 0.84\textsuperscript{***} & 0.75\\ 
		3 & 0.75\textsuperscript{***} & 0.81\textsuperscript{***} & 0.5\\ 
		\midrule
		\multicolumn{3}{l}{\textbf{Knowledge source: GloVe}}\\
		\midrule
		1 & 0.83\textsuperscript{***} & 0.8\textsuperscript{***} & 0.73\\ 
		2 & 0.81\textsuperscript{***} & 0.83\textsuperscript{***} & 0.77\\ 
		3 & 0.8\textsuperscript{***} & 0.85\textsuperscript{***} & 0.61\\ 
		\midrule
		\multicolumn{3}{l}{\textbf{Knowledge source: SpaCy}}\\
		\midrule
		1 & 0.81\textsuperscript{***} & 0.76\textsuperscript{***} & 0.67\\ 
		2 & 0.79\textsuperscript{***} & 0.83\textsuperscript{***} & 0.82\\ 
		3 & 0.76\textsuperscript{***} & 0.81\textsuperscript{***} & 0.5\\ 
		\midrule
		\multicolumn{4}{l}{\textsuperscript{*}: $p<0.05$, \textsuperscript{**}: $p<0.01$, \textsuperscript{***}: $p<0.001$} \\
		\midrule\bottomrule
	\end{longtable}
}\endgroup
	

\section{Linear Mixed-Effects Regression Analysis}

We applied linear mixed-effects regression models implemented with the lme4 package \cite{bates2014fitting} in the R environment (Version 3.6.2 R Core Team, 2019). We fitted separate models for each of the four knowledge source based similarity. In each model, the outcome variable- semantic similarity value was regressed onto the predictor features. Thus, in a model, we specified feature value as the outcome variable, whereas score (1, 2, 3, 4, 5) and trial day (day-id) as fixed effects, while student (student-id) was specified as random effects. Day was nested within student; where within each student, there were 3 trial days. The nesting of day within student showed repeated measure in time and thus simulating the impact of repeated reading on students' performance by statistically measure variations of feature values over time.

Table \ref{tab:Chapter-5tab7} shows LMER models' results using text-element wise semantic similarity as the dependent variable reported significant effects for score, day and text-element on the summaries collected during sessions 1 and 2. In the table, the intercept is the mean of the dependent variable in the three base levels-- human rater's score (1, 2, 3, 4, 5), trial-days (1, 2, 3) and text-element (setting, plot, conflict, resolution). In both sessions, results indicate that scores 2 and higher (3, 4 \& 5 in session-1; 3 \& 4 in session-2) are significantly achieved than score-1, where other baseline is text-element = setting and day = 1. We can see that, in contrast to similarity over text-2 setting, higher score over text-1 setting is achieved. Also, plot, conflict \& resolution wise similarities in session-1 and conflict \& resolution wise similarities in session-2 are significantly effected with respect to baseline setting text-element, where other baselines are day = 1 and score = 1. However, similarity scores achieved in days-2 and 3 are significantly higher than day-1 score with respect to text-element = setting and score = 1 on session-2 summaries only. Thus, the table indicates that repeated reading intervention has significant effects on content enrichment with respect to text-elements.


\begingroup
\setstretch{1.4}
{\small 
	\begin{longtable}
		{>{\raggedright}m{5cm} C{2cm} C{2cm} C{2cm} C{2cm}}
		\captionsetup{font={small}, labelfont=bf, skip=4pt}
		\caption{LMER results for predicting text-element wise similarity score}
		\label{tab:Chapter-5tab7}\\
		\toprule\midrule
		
		\textbf{Fixed Effect (baseline)} & \multicolumn{2}{c}{\textbf{Session 1}} & \multicolumn{2}{c}{\textbf{Session 2}} \\
		\textbf{}& \textbf{Estimate} & \textbf{t value} & \textbf{Estimate} & \textbf{t value} \\
		\midrule
		
		(Intercept)  & 0.91  &  13.82\textsuperscript{***} & 0.26  &    14.78\textsuperscript{***}\\
		Score-2 (baseline Score-1) & 0.47 &   6.01\textsuperscript{***}& 0.14 &    6.72\textsuperscript{***}\\
		Score-3 (baseline Score-1) & 0.96 &     10.66\textsuperscript{***} & 0.30 &    13.61\textsuperscript{***}\\
		Score-4 (baseline Score-1) & 1.39 &   13.26\textsuperscript{***} & 0.53 &    13.63\textsuperscript{***}\\
		Day-2 (baseline Day-1)  & 0.09 &   1.41 & 0.11 &     6.31\textsuperscript{***}\\
		Day-3 (baseline Day-1)  & 0.08 &     1.15 & 0.16  &   8.16\textsuperscript{***}\\
		Plot (baseline Setting) & -0.23 &    -16.35\textsuperscript{***} & & \\
		Conflict (baseline Setting) & 0.07  &    5.35\textsuperscript{***} &  0.02  &   2.33\textsuperscript{*}\\
		Resolution (baseline Setting) & 0.08 &     6.11\textsuperscript{***} & -0.03 &     -2.94\textsuperscript{**}\\
		
		\midrule
		\multicolumn{4}{l}{\textsuperscript{*}: $p<0.05$, \textsuperscript{**}: $p<0.01$, \textsuperscript{***}: $p<0.001$} \\
		\midrule\bottomrule
	\end{longtable}}
	\endgroup


\section{ANOVA Analysis of LMER Model}

Table \ref{tab:Chapter-5tab8} shows ANOVA analysis of LMER models on all knowledge sources applied to determine the semantic similarity between the referred text and summaries collected during sessions 1 and 2. The ANOVA analysis of the LMER models give f-value of semantic similarity, trial day (1, 2, 3) and score (1, 2, 3, 4, 5) as a whole. Here, semantic similarity value was dependent variable (DV), whereas day and score were independent variables (IV). In the table, f-values show that overall semantic scores, calculated using any external sources-- WordNet, Word2Vec, GloVe \& SpaCy, were significantly influenced during repeated reading. By comparing both sessions, we find that f-values were relatively high at session-1, which meant the variability of group means was large relative to the within-group variability. In order to reject the null hypothesis that the group means were equal, we needed a high f-value. Thus, the similarity score of session-1 discriminated participants' writing (i.e., null hypothesis rejection) more effectively than that of session-2.

\begingroup
\setstretch{1.4}
{\small 
	\begin{longtable}
		{>{\raggedright}m{5cm} C{2cm} C{2cm} C{2cm} C{1.8cm}}
		\captionsetup{font={small}, labelfont=bf, skip=4pt}
		\caption{ANOVA analysis of LMER models on semantic similarity}
		\label{tab:Chapter-5tab8}\\
		\toprule\midrule

				\textbf{Variable} & \textbf{WordNet} & \textbf{Word2Vec} & \textbf{GloVe} & \textbf{SpaCy}\\
				\midrule
				\multicolumn{5}{l}{\textbf{Session-1 semantic similarity}}\\
				\midrule
				
				Score (1, 2, 3, 4, 5) & 15.8\textsuperscript{***} & 27.3\textsuperscript{***} & 32.15\textsuperscript{***} & 31.04\textsuperscript{***}\\
				
				Day (1, 2, 3) & 5.8\textsuperscript{**} & 10.0\textsuperscript{***} & 8.41\textsuperscript{***} & 8.32\textsuperscript{***}\\
				
				\midrule
				\multicolumn{5}{l}{\textbf{Session-2 semantic similarity}}\\
				\midrule
				
				Score (1, 2, 3, 4, 5) & 20.46\textsuperscript{***} & 22.93\textsuperscript{***} & 27.46\textsuperscript{***} & 24.75\textsuperscript{***}\\
				
				Day (1, 2, 3) & 27.34\textsuperscript{***} & 26.47\textsuperscript{***} & 29.95\textsuperscript{***} & 28.63\textsuperscript{***}\\
				
				\midrule
				\multicolumn{5}{l}{\textsuperscript{*}: $p<0.05$, \textsuperscript{**}: $p<0.01$, \textsuperscript{***}: $p<0.001$} \\
				\midrule\bottomrule
			\end{longtable}}
			\endgroup


\section{Conclusion}

This chapter presented an approach to score students' writings based on content enrichment in repeated reading intervention, where reading two texts simulated extensive and intensive readings. We used four different knowledge sources-- WordNet, Word2Vec, GloVe and SpaCy models to calculate semantic similarity among words. We first applied text preprocessing and then measured similarity at four levels-- lexical, syntactic, concept and overall between two texts. The results showed that, the semantic similarity, calculated using WordNet, was worst among all the four knowledge sources. The average semantic similarity scores of both groups showed that contents enrichment were increased in the summaries as repeated reading trials proceeded from day-1 to day-3. To evaluate the performance of the proposed method, we used three approaches-- Pearson correlation coefficient, Spearman's rank correlation coefficient and QWK. Their results showed a strong agreement between expert's ratings and the calculated scores.  

The LMER models showed that the fixed-effect variables– text-element, score and day significantly influenced text-element level semantic similarity in repeated reading intervention. The ANOVA analysis of the LMER models showed that the fixed-effect variables– score and day significantly influenced overall content enrichment in writings. 

We have not explored several other methods which could be used for measuring content enrichment of learners' text, such as ontology based measurement framework \cite{liu2012ontology, harispe2014framework, konys2018ontology, ma2014generic}, contextualized word embeddings (e.g., BERT model  \cite{li2020sentence, reimers2019sentence, zhang2020semantics}), and artificial general intelligence (e.g., OpenNARS for Applications \cite{hammer2020opennars}). Implementation of these methods for measuring learners' text production skill can be an extension of our work.

Another future work of this study could be to analyse the 2-way and 3-way interaction in LMER models, such as the effect of specific trial day preference could enhance the preference of specific semantic level on content enrichment in writings in specific type of readings (extensive and intensive) and writings (e.g., narrative and descriptive) with various fixed variables such as gender, age-group, reading frequency, the gap between readings, various L1/L2 language, ethnicity, social status etc.

\chapter{Speech Fluency Analysis of English as Foreign Language Learners' Oral Proficiency}
\label{Chapter6} 
\lhead{Chapter 6. \emph{Speech Fluency Analysis of EFL Learners' Oral Proficiency}} 

\section{Introduction}

Acceptable pronunciation is one of the key characteristics of a fluent speaker since speech, which is grammatically accurate but full of disfluencies, is likely to be unintelligible for listeners. Second-language speakers, whose speech is not easily understood, are sometimes perceived as less intelligent and such speakers may achieve less score in some tasks such as in academic and professional interviews. Conversely, if speech fluency is good, then such non-native speakers may have more opportunities for socialization with others.

In an attempt to address this gap, the present study attempts to explore whether Repeated Reading, a pedagogical procedure based on reading text, is an adequate technique to help learners (i) enhance their English fluency, and (ii) transfer their reading gains to their spontaneous speech production. In order to answer these questions, in this chapter, we explain different techniques to score learners’ speech fluency and to analyse the effect of repeated reading on their speech fluency score. The chapter first describe the context, i.e. description of available techniques and tools for determining speech fluency. Then we propose our machine learning based methodology to classify automatically L2 learners spontaneous speech fluency. Next, we present the results of linear mixed-effects regression (LMER) to analyse the significant impact of RR on speech fluency scores. At last, we discuss the results and conclude the outcomes. 


\section{Context}

Fluency is one of the most important components of oral proficiency and can be used to represent general oral proficiency. Most people rarely reflect on the complications of speech fluency that enable them to use language unless confronted by their absence. Measuring oral proficiency has been limited due to difficulties in collecting and analyzing speech samples. Recent developments in computer technology have aided efforts to effectively test speaking rates in a language test, and large scale collection and analysis of speech samples to investigate the components of oral proficiency has become easier for researchers.

In this section, we explain some theories, concepts, techniques and tools which had been applied for the automatic fluency assessment in the prior art. All these are briefly described under key themes and present in an organized manner.

\subsection{Fluency}

In the narrow and focused definition, fluency can be defined as the speed and smoothness of oral delivery. It is often characterized as the ability to speak smoothly, accurately, confidently, and at a rate consistent with native-speaker norms. 

The speed of oral delivery can be represented by temporal variables, and research on fluency has focused on the speed of oral delivery. Also, fluency measurement as represented by pausing patterns and temporal variables in L2 English speech samples are examined together. Recently, phonological fluency is also examined by researchers and were used to predict L2 speech fluency score \cite{liu2020dolphin}.


\subsection{Repeated Reading method}
One of the most relevant methodological procedures, used both in L1 and L2 contexts to help learners develop their reading skills, is the Repeated Reading (RR) method. RR consists of re-reading a short passage both silently and aloud until the reader is able to do so with ease, effortlessly and fluently. Most studies have explored RR is effective in enhancing learners’ fluency when reading \cite{stevens2017effects}. Gil et al. \cite{gil2017effect} found that RR can also be used to improve L2 learners’ pronunciation skills and may help learners to articulate consonants and vowels or to use appropriate rhythm and stress. Within RR, two varieties have been proposed: (i) Assisted Repeated Reading- learners read text loudly or silently while are guided by the teacher or an audio-tape model; and (ii) Unassisted Repeating Reading- learners read selected text loudly or silently and independently \cite{xin2020improving, taguchi2004developing, webb2012vocabulary}.


\subsection{L2 Speech Production}

To study L2 speech production, several models were proposed by linguistics. For example, Levelt \cite{levelt1993speaking, levelt1999producing} proposed a model, which postulates three primary stages of production. Lambert et al. \cite{lambert2020task} explained as ``the first is a conceptualization stage in which speakers select information from world knowledge (getting through reading and listening) to include in a message and organize this information into a structure representing a pre-verbal plan. The second is a formulation stage in which this preverbal plan is encoded in the L2. The third stage is the articulation stage in which the phonetic plan is buffered and parsed as syllables at the motor level. Parallel processing plays an important role in this model of L2 speech production. It occurs in speech production when work in one stage is sufficiently automated for work in another stage to progress unrestricted. Like L1 speakers, L2 higher proficient speakers can conceptualize a message and formulate it at the same time because lexical retrieval and syntactic encoding tend to be largely automatized processes that accompany emergent concepts. Whereas, L2 lower proficient speakers might not be able to parallel process these two stages of speech production because lexical retrieval and syntactic encoding are often not sufficiently automatic for them. Lower proficiency L2 speakers are thus often required to employ serial processing in which encoding must occur as a separate stage after conceptualization. This naturally results in frequent pausing and other dysfluencies while messages are encoded in real-time. However, effective preparation techniques such as planning \& rehearsal, multiple reading and repeated reading can support pre-conceptualization and pre-formulation of messages for lower proficiency learners. These preparations allow such learners to rely on previously conceptualized content and recently activated linguistic resources during oral text production."


\subsection{Temporal Features of Fluency}

There are several components in oral proficiency such as grammar, vocabulary, pronunciation, smoothness, coherence, fluency that play important roles in language use for speaking as well as listening, reading and writing. However, one important feature of fluency is that it is relatively easy to measure as compared to other components in oral language proficiency. Researchers measured the speed of oral delivery by measuring temporal variables such as speech rate and mean syllables per run \cite{park2016measuring, ginther2010conceptual}. 
Temporal information from speech samples are categorized into length and number variables, such as the length of spoken and silent time periods, and the number of syllables and pauses. The syllable is the basic unit of production and the average number of syllables with a given time period has been recognized as a good measure of oral proficiency. Pauses are silent parts that occur between runs and denote hesitation or breathing, and long silent pauses are regarded as basic evidence of non-fluency. However, not every pause is silent and pauses vocalizations such as ‘uh’ are called filled pauses. Researchers have already calculated various temporal variables of quantity and rate of production (e.g., speech time ratio, speech rate, and mean syllable per run), and frequency and length of pauses (e.g., number of silent pauses per second, silent pause total response ratio) \cite{ginther2010conceptual, park2016measuring}. 


\subsection{Acoustics-Prosody Feature of Fluency}

Prosody describes variation in intonation, duration, rhythm, and intensity, is a critical component of perceived fluency in spoken language, as prosodic variation signals syntactic and semantic structure of sentences. For example, speakers often cue syntactic phrase boundaries through the employment of intonational phrase boundaries, the presence of silence between words and a pitch excursion, which can be rising in interrogative sentences or falling in declarative sentences. Kuhn et al. \cite{kuhn2010aligning} stated ‘in addition to the role of rate and accuracy, prosodic fluency requires appropriate expression or intonation coupled with phrasing that allows for the maintenance of meaning’. Deng et al. \cite{deng2020integrating} described an automatic fluency evaluation of spontaneous speech using integrated diverse features of acoustics, prosody, and disfluency-based ones. Verkhodanova et al.\cite{verkhodanova2017hesitations} presented the acoustic analysis of frequent disfluencies - voiced hesitations (filled pauses and lengthenings) across different speaking styles in spontaneous Russian speech, as well as results of experiments on their detection using SVM classifier on a joint Russian and English spontaneous speech corpus.


\section{Problem}

In this chapter, we propose a hypothesis that using significant acoustics-prosodic and temporal features of spontaneous speech derived from L2 learners’ summary speech during a repeated reading session, learners’ oral proficiency (in terms of fluency) can be automatically classified by applying machine learning methods. We also propose another hypothesis that repeated reading has an impact on oral features and shows improvements in learners’ speaking skills.

In order to test these hypotheses, we conducted an experiment in which participants attended two repeated reading sessions for reading and then retelling. In a session, they read one text and then recited a summary of the text once in a day for consecutive three days. The reading of text-1 and text-2 had given the experience of intensive and extensive reading respectively. After finishing the reading of a text, they recited a summary of the text, as much detail as they could. 

The oral summary speech in the form of audios, collected in the repeated reading experiment, are analysed in the current chapter to test the proposed hypotheses.


\section{Experiment}

The present experiment was similar to the experiment discussed in chapter-4, with two exceptions, i) no participant was common in both experiments, and ii) in each trial day, participants recited a summary of the read text. A brief description of the participants, materials, and procedure that we used in this study is described here.


\subsection{Participants}

There were 20 EFL bilingual students (7 females and 13 males, age-range = 18-22 years, mean age = 20.29 years) from Indian Institute of Information Technology Allahabad participating in the present study. The participants were enrolled in a bachelor program of STEM discipline. They were compensated with some course credits for their participation in the study. None of the participants reported having any language or reading impairments. They all performed regular academic activities (e.g., listening to class lectures, writing assignments, watching lecture videos) in English (L2) only, whereas their primary language (L1) was different from each other. Also, they reported that they could carry on a conversation, read and comprehend instructions, read articles, books, as well as watch TV shows and movies in English (L2) also. 


\subsection{Materials}

As discussed in previous chapters- \ref{Chapter2} and \ref{Chapter4}, to fulfil the purpose of the experiment, participants needed to read completely strange text. Therefore, the same two texts-- text-1 (\ref{AppendixA1}) and text-2 (\ref{AppendixA2}), used in both previous experiments, were chosen for the present study. As stated in earlier chapters, the text-1 was used for simulating enjoyable extensive reading and the text-2 was for intensive reading. These texts were unread in the participants’ lifespan until the experiment began. A summary of the properties of text was already given in table \ref{tab:Chapter-2tab2}.


\subsection{Procedure}

All experimental sessions were held in the SILP research laboratory in a group of 5 students. Upon their first arrival in the research laboratory, the participants read and signed a consent form prior to starting the experimental procedure. They were aware that they had to read two texts followed by a summary retelling in each of the consecutive three days, but they had no prior information that day-1 text would be repeated on the next two days. The design of the data collection experiment was inspired by Sukhram et al. \cite{sukhram2017effects} experiment.

In the experiment, the participants attended two reading sessions in a day for three consecutive days. In a session, they individually sat in front of a monitor on which the slide of a text was displaying and then they started reading independently the text loudly or silently as they wished (i.e. unassisted repeating reading). There was no time limit for finishing their reading to provide them a natural reading condition. So, participants were able to read at their own pace. After finishing reading, the text was removed from the display and they were told to solve some puzzles (Appendix \ref{pdf:puzzle}) on a paper sheet in two minutes. After two minutes, while the room was completely silent; they started to tell a summary of the read text and the audio was recorded using a noise-cancelling microphone connected to the same computer. In the experiment, puzzles were used to clear their rote/working/short-term memory to ensure that summaries would come from their long-term memory where their comprehension skills (understanding of the text) stored. Title, heading, bullets, images etc. of the original text were excluded and only core sentences were displayed, which made the text anonymous to the participants.

In session-1, participants read text-1 and then told its summary without looking at the text, and the audio was recorded on a given computer. The same had happened in session-2 for text-2. The participants took 15—25 minutes to finish a session. Between the two sessions, they were given a break of 15-minute for refreshment.


\section{Feature Analysis}


\subsection{Oral Proficiency Labelling}

The recorded audio summary files were given to one PhD student for labelling each summary audio either high or low based on overall oral proficiency. The distribution of labels of summaries is reported in table \ref{tab:Chapter-6tab1}.

\begingroup
\setstretch{1.4}
{\small 
	\begin{longtable}
		{>{\raggedright}m{2.5cm} C{2cm} C{4cm} C{2cm} C{2cm}}
		\captionsetup{font={small}, labelfont=bf, skip=4pt}
		\caption{Details of rating distribution among summary audio}
		\label{tab:Chapter-6tab1}\\
		\toprule\midrule

		\textbf{Session} &  \textbf{Day} &  \textbf{Total participants} & \multicolumn{2}{c}{\textbf{Label distribution}}\\\cline{4-5}
		& & & \textbf{Low} & \textbf{High}\\
		\midrule

		\multirow{3}{2.5cm}{1}  & 1 & 20 & 10 & 10\\  
		& 2 & 20 & 9 & 11\\ 
		& 3 & 20 & 7 & 13\\
		\midrule
		
		\multirow{3}{2.5cm}{2} & 1 & 20 & 11 & 9\\
		& 2 & 20 & 11 & 9\\ 
		& 3 & 20 & 10 & 10\\
		\midrule\bottomrule
	\end{longtable}}
	\endgroup

\subsection{Feature Extraction}

For measuring overall oral fluency quality, we extracted temporal as well as acoustics-prosody features from the summary audio files. 


\subsubsection{Acoustics Feature Set}

The process used to extract acoustic features did not differentiate between silences and unvoiced regions, which could produce errors in the functions applied to each feature. Therefore, all silences from each summary audio were extracted using intensity-threshold and pitch-threshold and these silences were excluded from the acoustic analysis process. In order to increase the number of audio segments for acoustic-prosodic analyses, we segmented the silence-free summary audio into 5 seconds chunks (termed fragments). Acoustic low-level descriptors (LLD) were automatically extracted from each audio fragment using the openSmile toolkit \cite{eyben2013recent}. Several researchers have used this toolkit with different feature sets for detecting disfluency in speech \cite{moniz2015disfluency, tian2015recognizing, medeiros2013disfluency, deng2020integrating}. In our experiment, the Continuous Audio/Visual Emotion and Depression Recognition Challenge 2013 (avec2013) feature set of this toolkit was used \cite{valstar2013avec}. The avec2013 is an acoustic feature set consists of 76 features (frame-based), composed of 32 energy and spectral related low-level descriptors (LLD), 6 voicing related LLD, 32 delta coefficients of the energy/spectral LLD, and 6 delta coefficients of the voicing related LLD. 

The features extracted from each recording are sorted into three groups:
\begin{enumerate}[(i)]
\item \textbf{Energy-related LLD:} Loudness; Sum of RASTA-style filtered auditory spectrum; RMS energy, Zero-crossing rate.
\item \textbf{Spectral-related LLD:} RASTA-style auditory spectrum, Bands 1–26 (0–8 kHz); MFCC 1–14; Spectral energy 250–650 Hz, 1 k–4 kHz; Spectral roll off point 0.25, 0.50, 0.75, 0.90; Spectral flux, Centroid, Entropy, Slope; Psychoacoustic sharpness, Harmonicity; Spectral variance, Skewness, Kurtosis.
\item \textbf{Voicing-related LLD:} F0 (SHS and viterbi smoothing); Probability of voice; Logarithmic HNR, Jitter (local, delta), Shimmer (local).
\end{enumerate}	

\subsubsection{Temporal Feature Set}
\begin{figure}[h!] 
	\centering
	\includegraphics[scale=0.33]{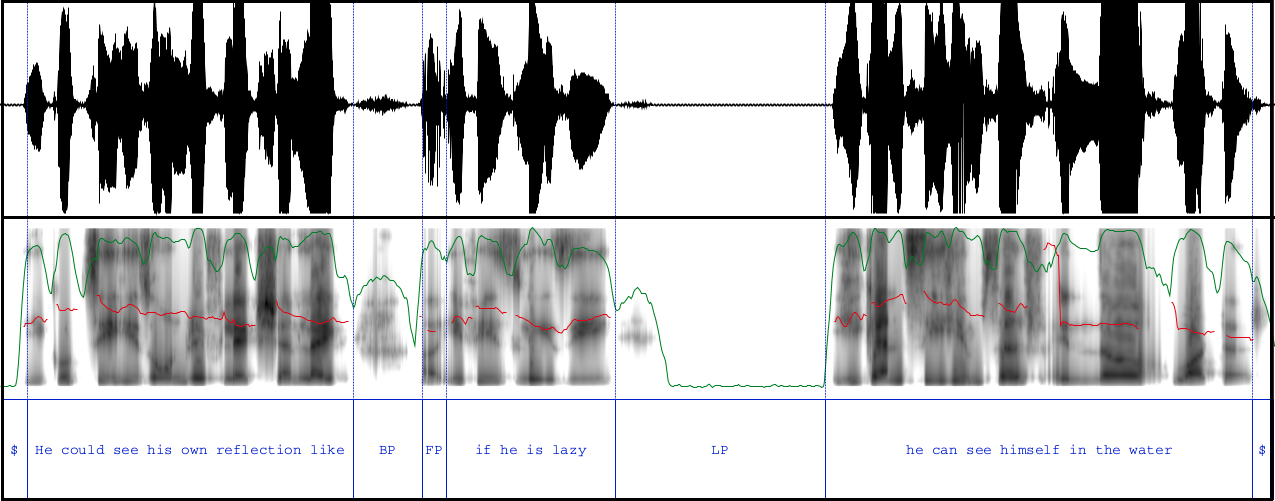}
	\captionsetup{font={small}, labelfont=bf, skip=4pt}
	\caption{Boundary of speech and pause}
	\label{fig:Chapter-6fig1}
\end{figure}

In order to obtain temporal features we needed to transcribe the original summary audio files. Therefore, the recordings were transcribed in English at word-level by manual correction of transcripts generated automatically using the Google Speech API recognizer\footnote{\url{https://cloud.google.com/speech/docs/languages}}, where brief and long unfilled pauses (BP \& LP) were marked by separate symbols. We also marked filled pauses (FP)  (e.g., ‘uh’ and ‘um’) using a different symbol. We analysed the sequence of pauses and speech in summary audio using PRAAT \cite{boersma2001praat} in order to extract 15 temporal features listed in Table \ref{tab:Chapter-6tab2}. The boundaries of speech and pause are shown in figure \ref{fig:Chapter-6fig1}.

In the figure, the red and green lines show pitch and intensity respectively. These temporal features were suggested by Ginther et al. \cite{ginther2010conceptual}. These features were added to improve the information about the temporal characterization of the oral summary recordings. In this case, the initial and final silence (marked as \$ in fig. \ref{fig:Chapter-6fig1}) of each recording were excluded from the analysis process because their lengths were different due to the recording process.


\begingroup
\setstretch{1.4}
{\small 
	\begin{longtable}
		{>{\raggedright}m{5cm} L{9cm}}
		\captionsetup{font={small}, labelfont=bf, skip=4pt}
		\caption{List of temporal features}
		\label{tab:Chapter-6tab2}\\
		\toprule\midrule

		\textbf{Feature name} &  \textbf{Definition}\\
		\midrule
		Total Response Time (TRT) & speaking + silent pause + filled pause time.\\
		\midrule
		Speech Time (ST) & speaking time, excluding silent and filled pauses.\\ 
		\midrule
		Speech Time Ratio (STR) & speech time/total response time.\\ 
		\midrule
		Number of Syllables (NumSyl) & total number of syllables in a given speech sample.\\ 
		\midrule
		Speech Rate (SR) & total number of syllables divided by the total response time in seconds. This figure was multiplied by 60 to obtain the speech rate per minute.\\ 
		\midrule
		Articulation Rate (AR) & total number of syllables divided by the sum of speech time and total filled pause time. This figure was multiplied by 60 to obtain the articulation rate per minute.\\ 
		\midrule
		Mean Syllables per Run (MSR) & number of syllables divided by number of runs in a given speech sample. Runs were defined as number of syllables produced between two silent pauses. Silent pauses were considered pauses equal to or longer than 0.25 seconds.\\ 
		\midrule
		Silent Pause Time (SPT) & total time in seconds of all silent pauses in a given speech sample.\\ 
		\midrule
		Number of Silent Pauses (NumSP) & total number of silent pauses per speech sample. Silent pauses were considered pauses of 0.25 seconds or longer.\\ 
		\midrule
		Mean Silent Pause Time (MSP) & silent pause time/number of silent pauses.\\ 
		\midrule
		Silent Pause Ratio (SPR) & silent pause time as a decimal percent of total response time.\\ 
		\midrule
		Filled Pause Time (FPT) & total time in seconds of all filled pauses in a given speech sample.\\ 
		\midrule
		Number of Filled Pauses (NumFP) & total number of filled pauses.\\ 
		\midrule
		Mean Filled Pause Time (MFP) & filled pause time/number of filled pauses.\\ 
		\midrule
		Filled Pause Ratio (FPR) & filled pause time as a decimal percent of total response time.\\
		\midrule\bottomrule
	\end{longtable}}
	\endgroup

\subsection{Temporal Feature Analysis}
Tables \ref{tab:Chapter-6tab6} and \ref{tab:Chapter-6tab7} show mean and standard deviation of temporal features across participants' group (Low \& High) as well as across days (1, 2 \& 3) of sessions 1 and 2 respectively.  

In table \ref{tab:Chapter-6tab6}, the value of four features– TRT, ST, STR, NumSyl, SR \& MSR increased across trial-days in both groups– low \& high. As compare to high-group, low-group had comparatively less value across days in features-- ST, STR, NUmSyl, SR, AR \& MSR; whereas the group had comparatively higher value across days in features-- TRT, SPT, NumSP, MSP, SPR, FPT, NumFP, MFP \& FPR. Thus, the table shows that low-group spent comparatively more time in pause than speaking and vice versa for high-group.
  
\begingroup
\setstretch{1.4}
{\footnotesize 
	\begin{longtable}
		{>{\raggedright}m{1.2cm} C{1.9cm} C{1.9cm} C{1.9cm} C{1.9cm} C{1.9cm} C{1.9cm}}
		\captionsetup{font={small}, labelfont=bf, skip=4pt}
		\caption{Temporal features details of session-1}
		\label{tab:Chapter-6tab6}\\
		\toprule\midrule

		\textbf{Feature} & \multicolumn{3}{c}{\textbf{Low (Mean (SD))}} & \multicolumn{3}{c}{\textbf{High (Mean (SD))}}\\
		\midrule
		& \textbf{Day-1} & \textbf{Day-2} & \textbf{Day-3} &  \textbf{Day-1} & \textbf{Day-2} & \textbf{Day-3}\\
		\midrule
				
		TRT & 148.79 (55.96) & 165.03 (65.84) & 243.05 (36.54) & 118.13 (27.64) & 160.62 (49.04) & 232.43 (76.84)\\
		ST & 75.84 (39.54) & 118.33 (46.49) & 160.78 (19.82) & 99.90 (20.46) & 128.70 (37.57) & 185.82 (68.67)\\
		STR & 0.68 (0.13) & 0.73 (0.10) & 0.81 (0.02)  & 0.78 (0.05) & 0.81 (0.03) & 0.89 (0.03)\\
		NumSyl & 387.44 (147.67) & 459.11 (199.67) & 695.00 (67.00) & 402.38 (88.21) & 485.50 (152.85) & 711.80 (143.00)\\
		SR & 163.13 (41.60) & 169.62 (36.08) & 172.99 (9.47)  & 184.80 (12.12) & 193.10 (14.83) & 198.08 (20.81)\\
		AR & 236.35 (24.43) & 230.89 (29.52) & 260.23 (7.08)  & 237.99 (13.15) & 242.66 (20.64) & 275.64 (32.04)\\
		MSR & 4.72 (1.09) & 7.44 (1.39) & 8.48 (0.37)  & 6.47 (1.26) & 8.58 (1.75) & 13.9 (2.89)\\
		SPT & 39.60 (26.55) & 35.63 (20.15) & 73.04 (13.22) & 18.07 (10.70) & 32.68 (15.88) & 27.58 (7.06)\\
		NumSP & 61.00 (33.41) & 61.33 (25.87) & 107.50 (21.50) & 32.62 (17.70) & 56.75 (23.52) & 52.20 (14.05)\\
		MSP & 0.61 (0.13) & 0.56 (0.11) & 0.68 (0.01) & 0.48 (0.19) & 0.55 (0.07) & 0.53 (0.03)\\
		SPR & 0.25 (0.12) & 0.22 (0.09) & 0.30 (0.01)  & 0.18 (0.08) & 0.20 (0.05) & 0.16 (0.04)\\
		FPT & 21.58 (16.99) & 19.81 (14.36) & 29.57 (7.44)  & 5.40 (1.99) & 3.32 (2.94) & 9.76 (5.97)\\
		NumFP & 23.56 (16.35) & 21.67 (14.29) & 32.00 (11.00)  & 9.75 (4.87) & 5.50 (3.61) & 10.00 (4.90)\\
		MFP & 0.81 (0.40) & 0.81 (0.24) & 0.96 (0.10)  & 0.60 (0.11) & 0.55 (0.17) & 0.94 (0.27)\\
		FPR & 0.13 (0.11) & 0.11 (0.07) & 0.12 (0.01) & 0.06 (0.02) & 0.02 (0.02) & 0.05 (0.02)\\ 
		\midrule\bottomrule
	\end{longtable}}
	\endgroup

In table \ref{tab:Chapter-6tab7}, the value of four features– TRT, ST, NumSyl, SR, MSR \& MFP increased across trial-days in both groups– low \& high. As compare to high-group, low-group had comparatively less value across days in features-- ST, STR, NumSyl, SR, AR \& MSR; whereas the group had comparatively higher value across days in features-- TRT, SPT, NumSP, MSP, SPR, FPT, NumFP, MFP \& FPR. Thus, the table shows that low-group spent comparatively more time in pause than speaking and vice versa for high-group.


\begingroup
\setstretch{1.4}
{\footnotesize 
	\begin{longtable}
		{>{\raggedright}m{1.2cm} C{1.9cm} C{1.9cm} C{1.9cm} C{1.9cm} C{1.9cm} C{1.9cm}}
		\captionsetup{font={small}, labelfont=bf, skip=4pt}
		\caption{Temporal features details of session-2}
		\label{tab:Chapter-6tab7}\\
		\toprule\midrule

		\textbf{Feature} & \multicolumn{3}{c}{\textbf{Low (Mean (SD))}} & \multicolumn{3}{c}{\textbf{High (Mean (SD))}}\\
		\midrule
		& \textbf{Day-1} & \textbf{Day-2} & \textbf{Day-3} &  \textbf{Day-1} & \textbf{Day-2} & \textbf{Day-3}\\
		\midrule
			
		TRT & 110.39 (57.95) & 154.56 (63.48) & 204.67 (71.67) &  103.15 (49.10) & 142.46 (54.70) & 146.98 (84.56)\\
		ST & 69.28 (35.42) & 91.64 (25.84) & 145.97 (34.60) & 79.93 (36.81) & 113.85 (45.01) & 168.50 (77.26)\\
		STR & 0.65 (0.12) & 0.64 (0.14) & 0.74 (0.09)  & 0.79 (0.06) & 0.80 (0.06) & 0.78 (0.06)\\
		NumSyl & 286.80 (154.13) & 369.00 (125.44) & 611.75 (170.39) & 350.43 (154.52) & 479.00 (204.46) & 656.67 (272.96)\\
		SR & 139.28 (34.23) & 154.66 (40.06) & 167.50 (24.50) &  168.31 (20.78) & 174.85 (21.70) & 187.12 (19.07)\\
		AR & 251.24 (27.82) & 240.86 (31.51) & 251.67 (35.62) &  266.93 (24.07) & 251.81 (19.40) & 255.05 (25.93)\\
		MSR & 4.93 (1.21) & 6.98 (1.60) & 13.84 (3.49)  & 6.71 (0.80) & 10.62 (2.77) & 13.49 (2.41)\\
		SPT & 32.63 (28.81) & 48.04 (37.22) & 47.16 (36.77)  & 17.00 (14.41) & 26.24 (15.25) & 25.27 (8.75)\\
		NumSP & 41.30 (31.40) & 59.25 (31.88) & 69.00 (44.23) & 30.43 (23.25) & 46.33 (24.53) & 44.67 (20.95)\\
		MSP & 0.58 (0.33) & 0.71 (0.23) & 0.63 (0.11)  & 0.45 (0.20) & 0.50 (0.19) & 0.60 (0.08)\\
		SPR & 0.25 (0.15) & 0.27 (0.14) & 0.21 (0.09) & 0.15 (0.08) & 0.17 (0.08) & 0.19 (0.04)\\
		FPT & 14.03 (10.97) & 31.18 (24.57) & 24.28 (16.43)  & 7.46 (4.06) & 7.47 (5.73) & 9.24 (1.72)\\
		NumFP & 17.00 (10.64) & 25.38 (17.84) & 20.75 (14.31)  & 12.86 (6.71) & 9.89 (6.44) & 10.33 (3.09) \\
		MFP & 0.69 (0.30) & 0.96 (0.43) & 1.26 (0.26)  & 0.58 (0.08) & 0.73 (0.21) & 0.96 (0.23)\\
		FPR & 0.12 (0.08) & 0.16 (0.12) & 0.11 (0.04)  & 0.07 (0.02) & 0.06 (0.04) & 0.08 (0.04)\\ 
		\midrule\bottomrule
\end{longtable}}
\endgroup

Thus, both tables \ref{tab:Chapter-6tab6} and \ref{tab:Chapter-6tab7} show that low-group had spent comparatively more time in pause than speaking and for high-group vice versa was true. By analysing both tables, we could infer that participants had spent less time in reciting of the complex text-2 as compare to the simple text-1. This was because of less words (NumSyl) in the summary of the complex text than those of simple text. 

\subsection{Feature Selection}

The features of all three feature-sets were normalized using the Z-score normalization method. We employed the Welch's t-test technique \cite{welch1947generalization, delacre2017psychologists} to select discriminating features from the acoustic feature-set. This approach is based on the concept that whether the mean of two sample groups (here, high and low) of the population were similar or not. This test was used to detect the significant differences between the features of high-labelled and low-labelled oral summary in a session. Only the features with a p-value lower than 0.05 were selected as statistically significant features for classification. Since temporal feature-set had only 15 features, therefore, the Welch’s t-test was not applied for getting any significant temporal features.

From session-1 oral summary, the number of significant features in the acoustic feature-set were [21, 20, 21] in the sequence of trial days- 1, 2 and 3. Similarly, from session-2 oral summary, the number of significant features in the acoustic feature-set were [22, 21, 18] in the sequence of trial day- 1, 2 and 3.

\section{Automatic Machine Learning Method}

To test the first hypothesis “using significant acoustics-prosodic and temporal features, participants’ oral proficiency can be automatically labelled”, we made classification models of the statistically significant features by using the Scikit-learn machine learning library (Version 0.20.4, 2019) \cite{pedregosa2011scikit} for the Python programming language. Three different classifiers were used to predict labels- the decision tree (DT), the multilayer perceptron (MLP) and the support vector machine (SVM). In addition, the 5-fold cross-validation technique was used to create another separate training and validation set. 

The configuration of the classifiers and the regression are given here. For DT classifier, the quality of a split was set to Gini impurity and the strategy used to choose the split at each node = best. For MLP, the learning rate = 0.001, Number of epochs = 200, Activation function for the hidden layer = relu, and hidden layer size was set to 100. For SVM, the C parameter was set to 1.0 using RBF kernel and the tolerance parameter was set to 0.001.

To analyze the performance of the classifiers, we used classification accuracy (C. rate), Unweighted Average 
Recall (UAR) and mean cross-validation (CV) accuracy.


\section{Result and Discussion on Machine Learning Outcomes}

Tables \ref{tab:Chapter-6tab3} and \ref{tab:Chapter-6tab4} show the classification results in the task of identifying oral proficiency label of a summary speech using acoustic LLD feature-set and temporal feature-set extracted from the summary audio collected during sessions 1 and 2 respectively. In both tables, the results of classification were achieved using a) all features of these two feature-sets, b) using statistically significant acoustic LLD features and c) using combined significant LLD and all temporal features. The purpose of reporting classification accuracy using all features of the two (LLD \& temporal) feature-sets is to represent them as the classifiers’ baseline accuracy and these are compared with the results of the same classifiers’ accuracy with significant features.

In table \ref{tab:Chapter-6tab3}, SVM classifier showed better average performance in all three accuracy measures. Among both feature-sets, all three classifiers produced better results on acoustic LLD feature-set. SVM raised its accuracy on the significant features of the acoustic feature-set across days in all three accuracy measures. But the other two classifiers could not achieve better performance on significant features across the days and across the accuracy measures. Among trial days, the classifiers produced better results mostly on day-2 trial features. The accuracy of all classifiers raised highest on combined feature-set having significant LLD features \& all temporal features. 

\vspace{40mm}

\begingroup
\setstretch{1.4}
{\small	
	\begin{longtable}
		{>{\raggedright}m{1.8cm} C{0.8cm} C{1cm} C{1.2cm} C{0.8cm} C{1cm} C{1.2cm} C{0.8cm} C{1cm} C{0.8cm}}
		\captionsetup{font={small}, labelfont=bf, skip=4pt}
		\caption{Classification results on audio feature-set of session-1}
		\label{tab:Chapter-6tab3}\\
		\toprule\midrule
		\textbf{Day} & \multicolumn{3}{c}{\textbf{SVM}} & \multicolumn{3}{c}{\textbf{DT}} & \multicolumn{3}{c}{\textbf{MLP}}\\
		\midrule
		& \textbf{CV acc.} & \textbf{UAR} & \textbf{C. rate} & \textbf{CV acc.} & \textbf{UAR} & \textbf{C. rate} & \textbf{CV acc.} & \textbf{UAR} & \textbf{C. rate}\\
		\midrule
		\multicolumn{10}{c}{\textbf{Classification results based on all features}} \\
		\midrule
		\multicolumn{10}{l}{\textbf{Acoustic LLD Feature-set:} all features- 76} \\	
		\midrule
		1  & 0.82 & 0.69 & 0.86 & 0.81 & 0.78 & 0.83 & 0.73 & 0.52 & 0.75\\
		2 & 0.9 & 0.91 & 0.91 & 0.86 & 0.81 & 0.82 & 0.88 & 0.91 & 0.91\\
		3  & 0.88 & 0.77 & 0.86 & 0.82 & 0.8 & 0.85 & 0.86 & 0.88 & 0.9\\
		\midrule 
		\multicolumn{10}{l}{\textbf{Temporal Feature-set:} all features- 15}\\ 
		\midrule
		1 & 0.73 & 0.83 & 0.83 & 0.67 & 0.67 & 0.67 & 0.53 & 0.33 & 0.33\\ 
		2 & 0.77 & 0.67 & 0.67 & 0.67 & 0.5 & 0.5 & 0.77 & 0.5 & 0.5\\ 
		3 & 0.75 & 0.7 & 0.7 & 0.65 & 0.65 & 0.65 & 0.6 & 0.6 & 0.6\\ 
		\midrule
		\multicolumn{10}{c}{\textbf{Classification results based on only statistically significant features}} \\
		\midrule 
		\multicolumn{10}{l}{\textbf{Acoustic LLD Feature-set}}\\ 
		\midrule
		1 (\#21) & 0.89 & 0.92 & 0.92 & 0.78 & 0.66 & 0.68 & 0.89 & 0.83 & 0.83\\ 
		2 (\#20) & 0.96 & 0.96 & 0.96 & 0.86 & 0.88 & 0.88 & 0.9 & 0.88 & 0.87\\ 
		3 (\#21) & 0.93 & 0.91 & 0.93 & 0.82 & 0.75 & 0.78 & 0.91 & 0.89 & 0.9\\ 
		\midrule
		\multicolumn{10}{c}{\textbf{Classification results based on acoustics significant features}}\\
		\multicolumn{10}{c}{\textbf{and all temporal features}} \\
		\midrule 
		1 (\#36) & 0.99 & 0.98 & 0.98 & 1.0 & 1.0 & 1.0 & 0.99 & 0.98 & 0.98\\ 
		2 (\#35) & 1.0 & 1.0 & 1.0 & 1.0 & 1.0 & 1.0 & 0.99 & 0.95 & 0.96\\ 
		3 (\#36) & 1.0 & 1.0 & 1.0 & 1.0 & 1.0 & 1.0 & 1.0 & 1.0 & 1.0 \\ 
		\midrule\bottomrule
	\end{longtable}
}\endgroup

In table \ref{tab:Chapter-6tab4}, SVM classifier showed better average performance in most of the accuracy measures across days. All three classifiers raised their accuracy on significant features corresponding to the accuracy on all features of the LLD feature-set across days in all three accuracy measures. Also, the accuracy of all classifiers raised highest on combined feature-set having significant LLD features \& all temporal features.

\vspace{40mm}

\begingroup
\setstretch{1.4}
{\small	
	\begin{longtable}
		{>{\raggedright}m{1.8cm} C{0.8cm} C{1cm} C{1.2cm} C{0.8cm} C{1cm} C{1.2cm} C{0.8cm} C{1cm} C{0.8cm}}
		\captionsetup{font={small}, labelfont=bf, skip=4pt}
		\caption{Classification results on audio  feature-set of session-2}
		\label{tab:Chapter-6tab4}\\
		\toprule\midrule
		\textbf{Day} & \multicolumn{3}{c}{\textbf{SVM}} & \multicolumn{3}{c}{\textbf{DT}} & \multicolumn{3}{c}{\textbf{MLP}}\\
		\midrule
		& \textbf{CV acc.} & \textbf{UAR} & \textbf{C. rate} & \textbf{CV acc.} & \textbf{UAR} & \textbf{C. rate} & \textbf{CV acc.} & \textbf{UAR} & \textbf{C. rate}\\
		\midrule
		\multicolumn{10}{c}{\textbf{Classification results based on all features}} \\
		\midrule
		\multicolumn{10}{l}{\textbf{Acoustic LLD Feature-set:} all features- 76} \\	
		\midrule
		1  & 0.83 & 0.58 & 0.8 & 0.69 & 0.51 & 0.67 & 0.74 & 0.57 & 0.79\\ 
		2 & 0.89 & 0.89 & 0.91 & 0.77 & 0.82 & 0.8 & 0.83 & 0.62 & 0.79\\ 
		3  & 0.85 & 0.7 & 0.81 & 0.77 & 0.72 & 0.73 & 0.78 & 0.67 & 0.77\\ 
		\midrule 
		\multicolumn{10}{l}{\textbf{Temporal Feature-set:} all features- 15}\\ 
		\midrule
		1 & 0.57 & 0.5 & 0.5 & 0.77 & 0.67 & 0.67 & 0.67 & 0.33 & 0.33\\ 
		2 & 0.67 & 0.38 & 0.5 & 0.77 & 0.5 & 0.5 & 0.77 & 0.62 & 0.67\\ 
		3 & 0.7 & 0.65 & 0.68 & 0.68 & 0.58 & 0.59 & 0.7 & 0.68 & 0.65\\ 
		\midrule
		\multicolumn{10}{c}{\textbf{Classification results based on only statistically significant features}} \\
		\midrule 
		\multicolumn{10}{l}{\textbf{Acoustic LLD Feature-set}}\\ 
		\midrule
		1 (\#22) & 0.89 & 0.9 & 0.9 & 0.8 & 0.9 & 0.9 & 0.87 & 0.85 & 0.87\\ 
		2 (\#21) & 0.92 & 0.94 & 0.94 & 0.84 & 0.82 & 0.83 & 0.89 & 0.93 & 0.93\\ 
		3 (\#18) & 0.95 & 0.92 & 0.92 & 0.81 & 0.73 & 0.73 & 0.92 & 0.92 & 0.92\\  
		\midrule
		\multicolumn{10}{c}{\textbf{Classification results based on acoustics significant features}}\\
		\multicolumn{10}{c}{\textbf{and all temporal features}} \\
		\midrule 
		1 (\#37) & 1.0 & 1.0 & 1.0 & 1.0 & 1.0 & 1.0 & 0.98 & 0.99 & 0.99\\ 
		2 (\#36) & 1.0 & 1.0 & 1.0 & 1.0 & 1.0 & 1.0 & 0.98 & 0.98 & 0.98\\ 
		3 (\#33) & 0.99 & 0.98 & 0.98 & 1.0 & 1.0 & 1.0 & 1.0 & 0.95 & 0.95 \\ 

		\midrule\bottomrule
	\end{longtable}
}\endgroup

\section{ANOVA Analysis of LMER Model}

We applied linear mixed-effect regression models implemented with the lme4 package \cite{bates2014fitting} in the R environment (Version 3.6.2 R Core Team, 2019). We fitted separate models for each of the two audio feature-sets of each session data. In each model, the outcome variable- feature value (dependent variable) was regressed onto the predictor features of the feature-set. Thus, in a LMER model of a feature-set, we specified feature value as the outcome variable, whereas group (high, low), features name (of the feature-set) and trial day (1, 2, 3) as fixed effects, while participants (participant-id) was specified as random effects. Trial day was nested within participant; since within each participant, there were 3 trial days. The nesting of days within participants showed repeated measure in time and thus simulating the impact of repeated reading on participants’ performance by statistically measure variations of feature values over time. 

The t-value of LMER models showed that the effect of all fifteen features of temporal feature-set in each session data were statistically significant at $\alpha = 0.05$ level during repeated reading. However, the models also showed that only a few features of acoustic LLD feature-set in each session were statistically significant at $\alpha = 0.05$ level during repeated reading, and not all significant features were common for both sessions. 

Table \ref{tab:Chapter-6tab5} shows ANOVA analysis of LMER models on all feature-sets of sessions 1 and 2 respectively. F-value of feature sets shows that, in both sessions both feature-sets were significantly influenced during repeated reading. In any session, f-value of group was not statistically significant during repeated reading on any feature-set. However, f-value of day was statistically significant during repeated reading on only temporal feature-set. By comparing both tables, we found that acoustic feature-sets of session-2 showed relativity high F value, which meant the variability of the feature-group, means was large relative to the within-group variability. In order to reject the null hypothesis that the LLD features' group means across the days were equal, we needed a high F-value. Thus, the LLD features of session-2 discriminated participants’ performance across days (i.e., null hypothesis rejection) more effectively than those of session-1. Similarly, the temporal feature-set of session-1 discriminated participants’ performance across days (i.e., null hypothesis rejection) more effectively than those of session-2. Also, the day of session-1 discriminated participants’ performance across group more effectively than those of session-2.

\begingroup
\setstretch{1.4}
{\small 
	\begin{longtable}
		{>{\raggedright}m{4cm} C{4.5cm} C{4.5cm}}
		\captionsetup{font={small}, labelfont=bf, skip=4pt}
		\caption{ANOVA analysis of the audio feature-sets}
		\label{tab:Chapter-6tab5}\\
		\toprule\midrule

				\textbf{Variable} & \parbox[t]{5cm}{\centering \textbf{Acoustic Feature-set}\\ (76 features)} & \parbox[t]{5cm}{\centering \textbf{Temporal Feature-set}\\ (15 features)}\\
	
				\midrule
				\multicolumn{3}{l}{\textbf{Session-1 audio feature-sets}}\\
				\midrule
				Feature & 820.96\textsuperscript{***} &  227.93\textsuperscript{***}\\
		    	
				Group (High, Low) & 2.22 & 0.08\\
				Day (1, 2, 3) & 0.14 & 9.5\textsuperscript{***}\\
				\midrule
				\multicolumn{3}{l}{\textbf{Session-2 audio feature-sets}}\\
				\midrule
				Feature & 945.58\textsuperscript{***} & 185.68\textsuperscript{***}\\
		        
				Group (High, Low) & 0.97 & 0.65\\
				Day (1, 2, 3) & 0.73 & 8.84\textsuperscript{***}\\
				\midrule
				\multicolumn{3}{l}{\textsuperscript{*}: $p<0.05$, \textsuperscript{**}: $p<0.01$, \textsuperscript{***}: $p<0.001$} \\
			\midrule\bottomrule
		\end{longtable}}
		\endgroup


\section{Conclusion}

This chapter presents an approach for classifying spontaneous speech based on oral fluency in repeated reading intervention, where reading two texts simulated extensive and intensive readings. We extracted two different feature-sets from speech audio, one was acoustics LLD and the other was temporal. To predict the label of oral fluency, we applied three machine learning classifiers, SVM, DT and MLP which got their best performance on a combined set of statistically significant LLDs and all temporal features.

The temporal features showed that low fluent participants had spent comparatively more time in pause than speaking. The feature set also showed that all participants took less speaking time in telling a summary of the complex text than the simple text. This was because of fewer words in the summary of the complex text than those of the simple text.

The LMER models showed that all fifteen features of temporal feature-set in each session data were statistically significant during repeated reading. So, we could say that, temporal features were important to study the impact of repeated reading intervention on oral fluency. 

The ANOVA analysis of the LMER models showed that the fixed-effect variables-- feature-sets (acoustics LLD and temporal features) and day were significantly influence oral fluency during repeated reading.

The future work of this study could be to analyse the 2-way and 3-way interaction in LMER models, such as the effect of specific trial day preference could enhance the preference of specific temporal and acoustics features on oral fluency in specific type of readings (extensive and intensive) with various fixed variables such as gender, age-group, reading frequency, the gap between readings, various L1/L2 language, ethnicity, social status etc.

\chapter{General Discussions}
\label{Chapter7} 
\lhead{Chapter 7. \emph{General Discussions}} 


\section{Learning Assessment Framework}
\begin{figure}[h!]
	\centering
	\includegraphics[height=0.55\textheight, width=\textwidth]{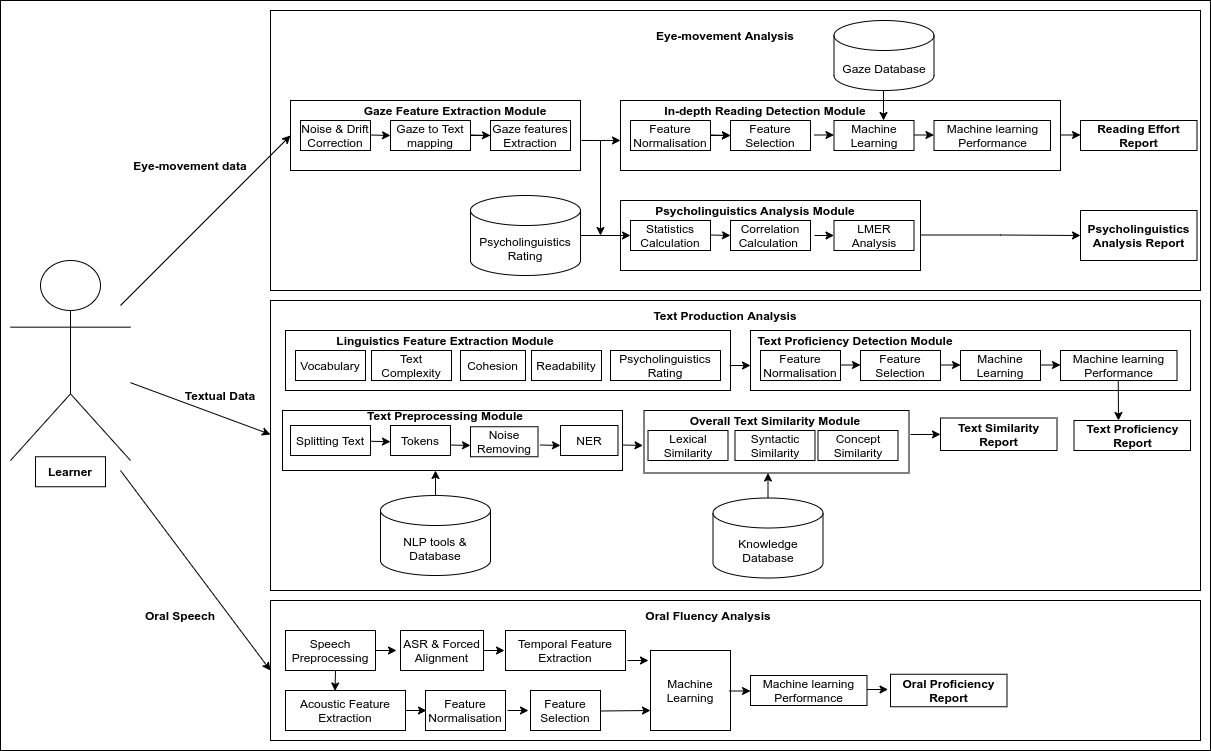}
	\captionsetup{font={small}, labelfont=bf, skip=4pt}
	\caption{Computational framework for Learning Assessment}
	\label{fig:Chapter-7fig1}
\end{figure}

We propose a learning assessment framework, as shown in figure \ref{fig:Chapter-7fig1}, which can be used to develop a computational tool for multimodal analysis of learners' reading behaviour as well as language comprehension and text production. The framework is divided into three segments-- eye movement analysis, text production analysis, and oral fluency analysis. The first segment, \textbf{eye-movement analysis} describes the process to evaluate eye-movement recorded during readings and provides results in form of reading effort report and psycholinguistics analysis report. For generating the former report, the steps are involved-- gaze feature extraction and in-depth reading detection using machine learning methods. In the gaze feature extraction module, noises in the eye data are removed and drifts are corrected to align with text lines. The gaze coordinates are mapped with displayed text to identify the read word of the moment. Now, AoI and gaze based features are extracted and processed in reading detection module. In this module, the gaze features are normalised using a normalisation method, such as scaling, clipping, log scaling and Z-score. To improve the performance of machine learning methods as well as to reduce computational cost of machine learning models, only important features are selected using a feature selection method, such as ANOVA, Chi-Square, T-test etc. The selected features are used to train machine learning models including classifiers and regressions. The trained model is further used to output reading effort report of unlabelled eye data. The test performance shows the accuracy of the model. 

To analyse the effect of psycholinguistics factors rating on eye-movement, statistics (mean \& standard deviation) of gaze data is calculated. The mean of gaze features and word-ratings are used to calculate correlation coefficients between them to identify the impact of word rating on eye-movement. The LMER analysis is also performed to get significant features, showing learners performance over time.
     
The second segment, \textbf{text production analysis} describes the process to predict learners' text proficiency using linguistic analysis and content similarity from a reference text. From learners' text, five linguistics feature-sets are extracted and these features are used to train a machine learning model to generate a proficiency report. For calculating similarity score, the two texts-- a learner's text and a reference text, are first preprocessed by following steps including, splitting text into sentences, converting into tokens, noisy words removal, named-entity recognition (NER) and co-reference resolution. In overall text similarity module, the preprocessed text is used to calculate lexical, syntactic and conceptual similarity and the output of the module shows content enrichment in leaners' text with respect to the reference text. The LMER analysis is used to get significant features, showing learners' text production performance over time.  

The last segment of the framework, \textbf{oral fluency analysis} describes the process to predict learners' speech fluency using acoustics and temporal features. Machine learning models are used to predict the fluency proficiency label of the speech.

\section{Scalability}

The results of different experiments encourage to scale up multi-modality response based students’ comprehension evaluation process from the laboratory experiments to the academic curriculum in form of computer-assisted assessments (CAAs). One way to scale up the results is to find common variables in both the lab experiments and in the academic assessments, such as essay-writings, facts-based descriptive answers etc. 


\section{Roadmap of results}

We have conducted several studies to implement a multi-modality response based L2 learners' comprehension evaluation system. Following are the summary of main results:
\begin{enumerate}[(i)]
	\item \textbf{Detection of in-depth reading effort:} We find that gaze can be used to label a learner’s in-depth reading efforts. We present results of different classifiers on gaze features composed of fixation, saccade, and regression at different textual area-of-interests including, word, sub-sentence, sentence, paragraph, page, and whole-text. Machine learning classifier SVM shows significantly improved performance compared to other classifiers like DT and MLP. The classification system labels the gaze data of reading over one and more days (repeated reading). Also, the system can accurately label the gaze data of two approaches of readings- extensive and intensive.
	\item \textbf{Measuring the influence of psycholinguistics factors on reading:} We find that some psycholinguistics factors viz. familiarity, imagery and concreteness have a clear effect on eye-movement during repeated readings, whereas word-frequency and age-of-acquisition have a modest impact and imagery has the least impact on eye-movement while word processing. Also, word processing during repeated reading is affected by EFL learners’ psycholinguistics experience with the words; and word processing improved when more repeated reading practices are performed.
	\item \textbf{Measuring descriptive writing proficiency:} We find that linguistics and psycholinguistics have clear effects on English text writing productions of EFL learners during repeated readings. The results show that the labels and scores predicted by a machine learning based writing proficiency tool can be very near to the labels and scores, given by human raters.
	\item \textbf{Measuring content improvement in writings:} We find that lexical, syntactic and content based overall similarity scores show relevant semantic similarity between referred text and summary text written by learners. Also, there is a clear effect on content enrichment improvement in the summary text due to repeated reading intervention.
	\item \textbf{Measuring oral fluency in spontaneous speech:} We find that classifiers can predict more accurate fluency label of speech audio on acoustics-prosody feature-sets with respect to temporal features. Also, the feature-sets can catch the impact of repeated reading intervention on speech fluency.
\end{enumerate}	


\section{Contributions}

In this section, we discuss the contributions of this dissertation within the relevant research areas.

\subsection{Multimodal Response based Automatic Assessment Framework}

In prior research, single modality based comprehension skill assessment was discussed. However, determining comprehension skills using multi-modality measures can provide more robust and reliable results. Therefore, we propose multi-modality measures, including, eye-tracking, psycholinguistics, linguistics, semantics and fluency based learning assessment framework (figure \ref{fig:Chapter-7fig1}). We show the impact of psycholinguistics on eye-movement during reading as well as writing proficiency. Similarly, linguistics and semantic measures applied on summary writing and machine learning based writing proficiency evaluation are also discussed.


\subsection{Measuring Repeated Reading Intervention}

In prior research, the impact of repeated reading intervention was studied majorly on reading fluency improvement. We study the impact of repeated reading intervention on different skills including reading, writing, language production and oral proficiency. Our results show improvements in all these skills due to repeated reading intervention.


\subsection{Effect of Intensive and Extensive Readings}

In prior research, intensive and extensive approach based repeated readings were not discussed in details. However, researchers agreed on the positive influence of both types of readings on reading comprehension. Therefore, we study the influence of both readings parallelly on different skills through learners’ multi-modality response. Our results show extensive repeated reading was more effective in skills improvements with the comparison of intensive repeated reading.


\section{Limitations and Future Work}

In this section, we discuss some of the limitations of the methods, which we used in our research. The research was done on a small number of participants. Though they all belong to the same academic discipline (STEM), still their skill levels were predicted by our methods with higher accuracy. \\
In our experiments, only two texts were used to stimulate readers’ skills. However, both text has different textual complexity level (easy vs. harder) and also they belong to different categories - one is narrative, story type and the other is history, descriptive type. In general, a student deals with these type of text in their study.\\
Previous research included higher reading attempts (e.g., 10 or more readings \cite{chang2013improving}) in a repeated reading experiment to study the impact on reading fluency skill improvements in school-going children. But in our experiments, only three reading trials were included in a repeated reading. Since our participants were adults and they had already matured second language skills; therefore, the major objective of our experiments was to how much contents from the reading text were grasped by participants in the trials of repeated reading.\\

We have not explored the relationship between different modality responses of learners. The impact of one modality of response on other responses can be the next level of analysis to explain the comprehension skills in more depth. This can be another branch of investigation stemming out from this dissertation. 


\section{Final Words}

This dissertation presented the outcome of a few years of research during which, 1) an eye-tracking for second language comprehension was conducted. Based upon the findings of which, 2) we focused our investigation to the triumvirate of linguistics, psycholinguistics and NLP. Finally, 3) adding speech fluency in our study, we presented a framework of AI-enabled students assessment system, which could be used to investigate long term effects of learning (in particular, reading) on comprehension skills improvement. This dissertation could inspire a few different directions to focus the investigation upon.


\addtocontents{toc}{\vspace{2em}} 

\appendix 


\chapter{Selected Text for Reading Trials} 

\label{AppendixA} 

\lhead{Appendix A. \emph{Text-1 \& 2}} 
\textbf{Text: 1} is a simple narrative story having title- `How the Camel got his Hump' by Rudyard Kipling, and is taken from the NCERT class 8 English textbook `It So Happened'\footnote{\url{http://ncertbooks.prashanthellina.com/class_8.English.ItSoHappend/index.html}} chapter-1. The text has four basic and important elements-- the setting, the plot, the conflict, and the resolution.

\noindent\textbf{Text: 2} is a relatively more complex informative article `Nineteenth-century politics in the United States'\footnote{\url{http://toeflpreparationsources.weebly.com/uploads/1/2/5/1/12516531/reaing_test_-_text_1.pdf}} taken from the reading section of TOEFL-iBT test guide book. The text has three basic elements-- the setting, the conflict, and the resolution.

\section{Text: 1}
\label{AppendixA1} 
 \textbf{SETTING:}\\
In the beginning, when the world was new and the Animals were just beginning to work for Man, there was a Camel, and he lived in the middle of a Howling Desert because he did not want to work. He ate sticks and thorns and prickles, and when anybody spoke to him he said "Humph!" Just "Humph!" and no more.

\noindent\textbf{PLOT:}\\
Presently the Horse came to him on Monday morning, with a saddle on his back and said, "Camel, O Camel, come out and trot like the rest of us."
"Humph!" said the Camel, and the Horse went away and told the Man.

Presently the Dog came to him, with a stick in his mouth, and said, "Camel, O Camel, come and fetch and carry like the rest of us."
"Humph!" said the Camel, and the Dog went away and told the Man.

Presently the Ox came to him, with the yoke on his neck, and said, "Camel, O Camel, come and plough like the rest of us."
"Humph!" said the Camel, and the Ox went away and told the Man.

At the end of the day the Man called the Horse and the Dog and the Ox together, and said, "Three, O Three, I am very sorry for you; but that Humph thing in the Desert can not work, or he would have been here by now, so I am going to leave him alone, and you must work double time to make up for it."

\noindent\textbf{CONFLICT:}\\
That made the Three very angry, and they held a panchayat on the edge of the Desert; and the Camel came chewing cud and laughed at them. Then he said "Humph!" and went away again.

Presently there came along the Djinn who was in charge of All Deserts, rolling in a cloud of dust.
"Djinn of All Deserts," said the Horse, "is it right for anyone to be idle?"
"Certainly not," said the Djinn.
"Well," said the Horse, "there is a thing in the middle of your Desert with a long neck and long legs, and he has not done a stroke of work since Monday morning. He will not trot."
"Whew!" said the Djinn whistling, "That is my Camel. What does he say about it?"
"He says 'Humph!', and he will not plough," said the Ox.
"Very good," said the Djinn. "I will humph him if you will kindly wait a minute."

The Djinn rolled himself up in his dust cloak, and took a walk across the Desert, and found the Camel looking at his own reflection in a pool of water.
"My friend," said the Djinn, "what is this I hear of your doing no work?"
The Djinn sat down, with his chin in his hand, while the Camel looked at his own reflection in the pool of water.
"You have given the three extra work ever since Monday morning, all on account of your idleness," said the Djinn. And he went on thinking with his chin in his hand.
"Humph!" said the Camel.
"I should not say that again if I were you," said the Djinn; "you might say it once too often. I want you to work."

And the Camel said "Humph!" again; but no sooner had he said it than he saw his back, that he was so proud of, puffing up and puffing up into a great big hump.
"Do you see that?" said the Djinn. "That is your very own humph that you have brought upon your very own self by not working. 
Today is Thursday, and you have done no work since Monday, when the work began. Now you are going to work."
"How can I," said the Camel, "with this humph on my back?"

\noindent\textbf{RESOLUTION:}\\
"That has a purpose," said the Djinn, "all because you missed those three days. You will be able to work now for three days without eating, because you can live on your humph; and do not you ever say I never did anything for you. Come out of the Desert and go to the Three, and behave."
And the Camel went away to join the Three. 

And from that day to this the Camel always wears a humph; but he has never yet caught up with the three days that he missed at the beginning of the world, and he has never yet learned how to behave.

\section{Text: 2}
\label{AppendixA2}
 \textbf{SETTING:}\\
The development of the modern presidency in the United States began with Andrew Jackson who swept to power in 1829 at the head of the Democratic Party and served until 1837. During his administration he immeasurably enlarged the power of the presidency. "The President is the direct representative of the American people," he lectured the Senate when it opposed him. "He was elected by the people, and is responsible to them." With this declaration, Jackson redefined the character of the presidential office and its relationship to the people.\\

\noindent\textbf{CONFLICT:}\\
During Jackson's second term, his opponents had gradually come together to form the Whig party. Whigs and Democrats held different attitudes toward the changes brought about by the market, banks, and commerce. The Democrats tended to view society as a continuing conflict between "the people"-- farmers, planters, and workers - and a set of greedy aristocrats. This "paper money aristocracy" of bankers and investors manipulated the banking system for their own profit, Democrats claimed, and sapped the nation's virtue by encouraging speculation and the desire for sudden, unearned wealth. The Democrats wanted the rewards of the market without sacrificing the features of a simple agrarian republic. They wanted the wealth that the market offered without the competitive, changing society; the complex dealing; the dominance of urban centres; and the loss of independence that came with it.

Whigs, on the other hand, were more comfortable with the market. For them, commerce and economic development were agents of civilization. Nor did the Whigs envision any conflict in society between farmers and workers on the one hand and businesspeople and bankers on the other. Economic growth would benefit everyone by raising national income and expanding opportunity. The government's responsibility was to provide a well-regulated economy that guaranteed opportunity for citizens of ability.

Whigs and Democrats differed not only in their attitudes toward the market but also about how active the central government should be in people's lives. Despite Andrew Jackson's inclination to be a strong President, Democrats as a rule believed in limited government. Government's role in the economy was to promote competition by destroying monopolies and special privileges. In keeping with this philosophy of limited government, Democrats also rejected the idea that moral beliefs were the proper sphere of government action. Religion and politics, they believed, should be kept clearly separate, and they generally opposed humanitarian legislation.

The Whigs, in contrast, viewed government power positively. They believed that it should be used to protect individual rights and public liberty, and that it had a special role where individual effort was ineffective. By regulating the economy and competition, the government could ensure equal opportunity. Indeed, for Whigs the concept of government promoting the general welfare went beyond the economy. In particular, Whigs in the northern sections of the United States also believed that government power should be used to foster the moral welfare of the country. They were much more likely to favour social-reform legislation and aid to education.

\noindent\textbf{RESOLUTION:}\\
In some ways the social makeup of the two parties was similar. To be competitive in winning votes, Whigs and Democrats both had to have significant support among farmers, the largest group in society, and workers. Neither party could win an election by appealing exclusively to the rich or the poor. The Whigs, however, enjoyed disproportionate strength among the business and commercial classes. Whigs appealed to planters who needed credit to finance their cotton and rice trade in the world market, to farmers who were eager to sell their surpluses, and to workers who wished to improve themselves. Democrats attracted farmers isolated from the market or uncomfortable with it, workers alienated from the emerging industrial system, and rising entrepreneurs who wanted to break monopolies and open the economy to newcomers like themselves. The Whigs were strongest in the towns, cities, and those rural areas that were fully integrated into the market economy, whereas Democrats dominated areas of semi subsistence farming that were more isolated and languishing economically.

\chapter{Questionnaire}
\label{pdf:questionnaire} 

\lhead{Appendix B. \emph{Questionnaire}}

\section{Objective questions on text-1 for trial days-- 1, 2 \& 3}
\begin{enumerate}[(I)]
	\item \textbf{Set 1: Please fill up the blank with suitable word:}
	\begin{enumerate}
		\item The Camel lived in the middle of a \underline{\hspace{3cm}} Desert.
		\item Horse came to the Camel on \underline{\hspace{3cm}} morning.
		\item The Camel ate \underline{\hspace{3cm}} and \underline{\hspace{3cm}} and \underline{\hspace{3cm}}.
		\item When anybody spoke to the Camel, he said \underline{\hspace{3cm}} Just \underline{\hspace{3cm}} and no more.
		\item The Camel missed \underline{\hspace{3cm}} days, which were: \underline{\hspace{3cm}}, \underline{\hspace{3cm}}, \underline{\hspace{3cm}}.

	\end{enumerate}
	\item \textbf{Set 2: Please choose a perfect option from multiple choices:}
	\begin{enumerate}
		\item And the Camel went away to join the Three. And from that day to this the Camel always wears a humph (we call it ‘hump' now, not to hurt his feelings); but he has never yet caught up with the three days that he missed at the beginning of the world, and he has never yet learned how to \textbf{behave}.\\
		The word \textbf{behave} in the passage is closest in meaning to\\
		i) act,     ii) live,     iii) learn,    iv) say
		\item The Ox came to The Camel, with the yoke on his neck on \underline{\hspace{3cm}}.\\
		i) Monday,   ii) Tuesday,   iii) Wednesday,    iv) Thursday
		\item Djinn was \underline{\hspace{3cm}} of All Deserts.\\
		i) king,     ii) in charge,      iii) owner,      iv) worker
		\item Select mismatched word in following options\\
		i) plough,     ii) prickle,      iii) trot,     iv) fetch and carry
		\item The Camel looked at his \underline{\hspace{3cm}} in the pool of water. \\
		i) own reflection,  ii) owner’s face reflection,   iii) friends’ reflection,   iv) owner’s dust-cloak
	\end{enumerate}
\end{enumerate}

\section{Objective questions on text-2 for trial days-- 1, 2 \& 3}
\begin{enumerate}[(I)]
	\item \textbf{Set 1: Please fill up the blank with suitable word:}
	\begin{enumerate}
		\item During his administration \underline{\hspace{3cm}} immeasurably enlarged the power of the presidency.
		\item \underline{\hspace{3cm}} held different attitudes toward the changes brought about by the market, banks, and commerce.
		\item The Whigs, in contrast, viewed government power \underline{\hspace{3cm}}.
		\item \underline{\hspace{3cm}} attracted farmers isolated from the market or uncomfortable with it.
		\item This “\underline{\hspace{3cm}}” of bankers and investors manipulated the banking system for their own profit.
		
	\end{enumerate}
	\item \textbf{Set 2: Please choose a perfect option from multiple choices:}
	\begin{enumerate}
		\item During his administration he \textbf{immeasurably} enlarged the power of the presidency. “The President is the direct representative of the American people,” he lectured the Senate when it opposed him.\\
		The word \textbf{immeasurably} in the passage is closest in meaning to\\
		i. frequently,   ii. greatly,   iii. rapidly,   iv. reportedly 
		\item The author mentions bankers and investors in the passage as an example of which of the following?\\
		i. The Democratic Party's main source of support,\\
		ii. The people that Democrats claimed were unfairly becoming rich,\\
		iii. The people most interested in a return to a simple agrarian republic,\\
		iv. One of the groups in favour of Andrew Jackson's presidency
		\item The development of the modern presidency in the United States began with Andrew Jackson who swept to power in \underline{\hspace{3cm}} at the head of the Democratic Party and served until \underline{\hspace{3cm}}. \\
		i. 1829,	ii. 1830,		 iii. 1836,	iv. 1837
		\item Select mismatched word in following options\\
		i. votes,		ii. farmers,	iii. society,	iv. history 
		\item The Whigs were supported by which of the following groups: \\
		i. workers unhappy with the new industrial system,	ii. planters involved in international trade,	
		iii. rising entrepreneurs,		iv. individuals seeking to open the economy to newcomers
	\end{enumerate}
\end{enumerate}

\chapter{Puzzle}
\label{pdf:puzzle} 
\lhead{Appendix C. \emph{Puzzle}}

\includepdf[pages=-,scale=1.2,trim=-5cm 0 0 0, clip]{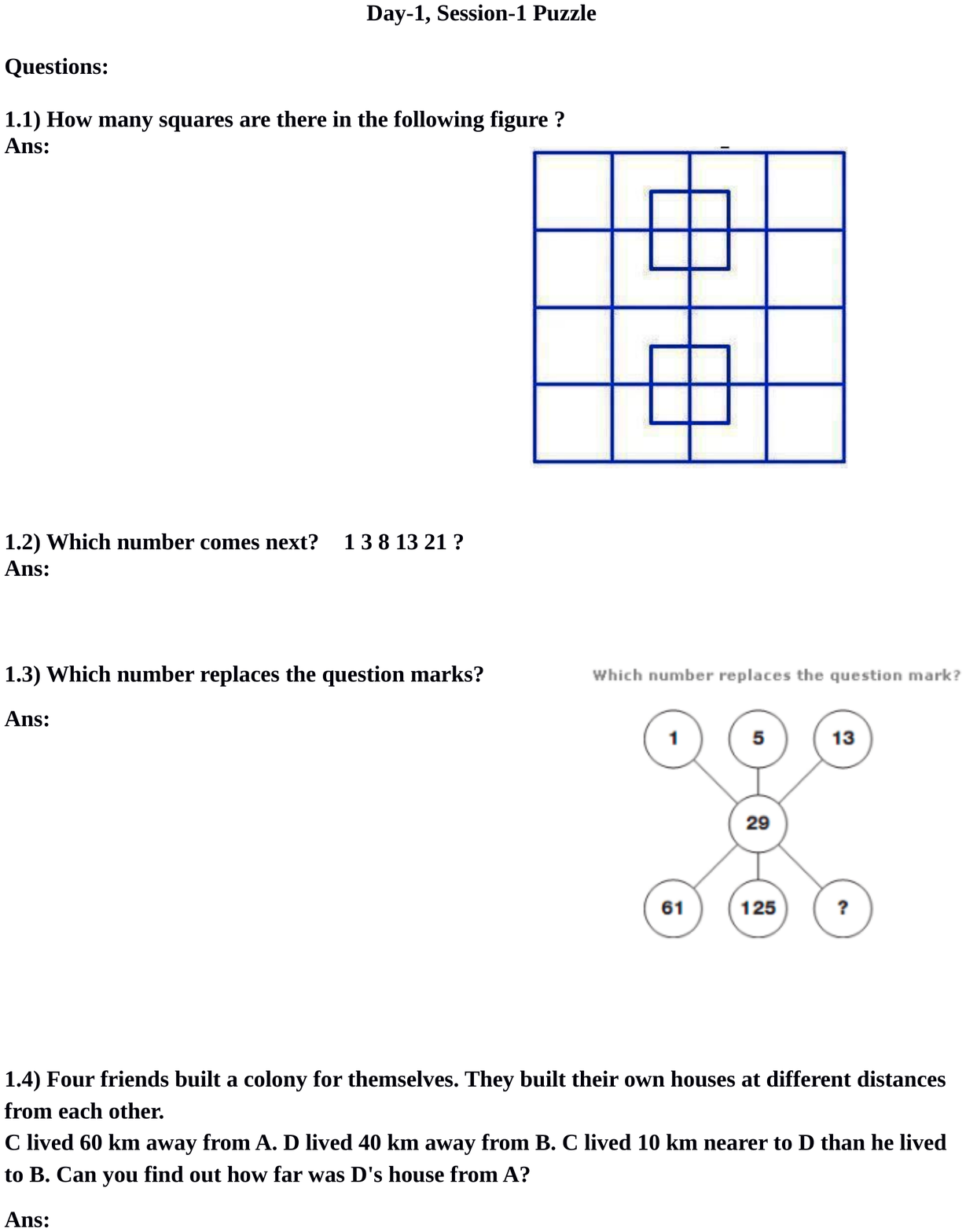}

\chapter{Pearson's Correlation among Fixation Features of Psycholinguistics Factors} 

\label{AppendixD} 

\lhead{Appendix D. \emph{Pearson's Correlation among Fixation Features of Psycholinguistics Factors}} 

\section{Pearson's Correlation among Fixation Features of Psycholinguistics Factors on Session 1 data}
\label{AppendixD1} 

\begingroup
\setstretch{1.2}
{\small	
	\setlength{\tabcolsep}{3pt}
	\renewcommand{\arraystretch}{1.}
	\begin{longtable}
		{>{\raggedright}m{2cm} L{5cm} C{2cm} C{2cm} C{2cm}}
		\caption{Correlation among psycholinguistics factor pairs' fixation features on session 1 data}
		\label{tab:AppendixDtab1}\\
		
		\toprule\midrule

		\textbf{Fixation Features} & \textbf{Psycholinguistics factors pair} & \textbf{Low(r,p)} & \textbf{High(r,p)} & \textbf{Both(r,p)}\\
		\midrule
		\multicolumn{5}{c}{Day-1}\\
		\midrule
		\multirow{5}{1cm}{tFC} & WF \& AoA &  0.85\textsuperscript{*} & 0.51 & 0.67 \\
		& WF \& Concreteness & 0.93\textsuperscript{**} & 0.61 & 0.78\textsuperscript{*} \\
		& AoA \& Concreteness & 0.96\textsuperscript{**} & 0.97\textsuperscript{**} & 0.99\textsuperscript{***} \\
		& Familiarity \& Emotion & 0.77\textsuperscript{*} & 0.38 & 0.7\textsuperscript{*} \\
		& Imagery \& Emotion & -0.72\textsuperscript{*} & -0.41 & -0.57 \\
		\midrule
		\multirow{6}{1cm}{tFD} & WF \& Familiarity &  0.62 & 0.73\textsuperscript{*} & 0.78\textsuperscript{*} \\
		& WF \& Concreteness & 0.72\textsuperscript{*} & 0.72\textsuperscript{*} & 0.73\textsuperscript{*} \\
		& AoA \& Familiarity & 0.62 & 0.88\textsuperscript{*} & 0.79\textsuperscript{+} \\
		& AoA \& Imagery & 0.8\textsuperscript{+} & 0.87\textsuperscript{*} & 0.86\textsuperscript{*} \\
		& AoA \& Concreteness & 0.59 & 0.85\textsuperscript{*} & 0.77\textsuperscript{+} \\
		& Familiarity \& Concreteness & 0.5 & 0.91\textsuperscript{**} & 0.73\textsuperscript{*} \\
		
		\midrule
		\multirow{6}{1cm}{FFD} & WF \& Familiarity &  0.83\textsuperscript{*} & 0.83\textsuperscript{*} & 0.89\textsuperscript{**} \\
		& AoA \& Familiarity & 0.45 & 0.93\textsuperscript{**} & 0.78\textsuperscript{+} \\
		& AoA \& Imagery & 0.61 & 0.85\textsuperscript{*} & 0.8\textsuperscript{+} \\
		& Familiarity \& Imagery & 0.76\textsuperscript{*} & 0.7\textsuperscript{*} & 0.73\textsuperscript{*} \\
		& Imagery \& Concreteness & 0.81\textsuperscript{**} & 0.4 & 0.7\textsuperscript{*} \\
		
		\midrule
		\multirow{3}{1cm}{LFD} & WF \& AoA &  0.83\textsuperscript{*} & 0.22 & 0.42 \\
		& WF \& Concreteness & 0.36 & 0.41 & 0.75\textsuperscript{*} \\
		& AoA \& Concreteness & -0.27 & 0.94\textsuperscript{**} & 0.71 \\
		
		\midrule
		\multirow{5}{1cm}{SFD} & WF \& AoA &  0.83\textsuperscript{*} & 0.27 & 0.55 \\
		& WF \& Concreteness & 0.79\textsuperscript{*} & 0.52 & 0.63\textsuperscript{+} \\
		& AoA \& Imagery & 0.96\textsuperscript{**} & 0.36 & 0.82\textsuperscript{*} \\
		& AoA \& Concreteness & 0.95\textsuperscript{**} & 0.61 & 0.87\textsuperscript{*} \\
		& Familiarity \& Concreteness & 0.24 & 0.84\textsuperscript{**} & 0.67\textsuperscript{*} \\
		\midrule
		\multicolumn{5}{c}{Day-2}\\
		\midrule
		\multirow{2}{1cm}{tFC} & Imagery \& Concreteness &  0.51 & 0.5 & 0.74\textsuperscript{*} \\
		& Concreteness \& Emotion & -0.48 & -0.73\textsuperscript{*} & -0.7\textsuperscript{*} \\
		
		\midrule
		\multirow{4}{1cm}{tFD} & WF \& Familiarity &  0.72\textsuperscript{*} & 0.65\textsuperscript{+} & 0.7\textsuperscript{+} \\
		& WF \& Imagery  & 0.8\textsuperscript{*} & 0.16 & 0.55 \\
		& AoA \& Imagery  & 0.8\textsuperscript{+} & 0.76\textsuperscript{+} & 0.95\textsuperscript{**} \\
		& Familiarity \& Concreteness & 0.75\textsuperscript{*} & 0.53 & 0.69\textsuperscript{*} \\
		
		\midrule
		\multirow{7}{1cm}{FFD} & WF \& Familiarity &  0.68\textsuperscript{+} & 0.64\textsuperscript{+} & 0.87\textsuperscript{**} \\
		& WF \& Imagery & 0.8\textsuperscript{*} & 0.25 & 0.56 \\
		& AoA \& Imagery & 0.72 & 0.89\textsuperscript{*} & 0.91\textsuperscript{*} \\
		& AoA \& Emotion & -0.58 & -0.85\textsuperscript{*} & -0.72 \\
		& Familiarity \& Imagery & 0.84\textsuperscript{**} & 0.2 & 0.73\textsuperscript{*} \\
		& Familiarity \& Concreteness & 0.33 & 0.73\textsuperscript{*} & 0.56 \\
		& Imagery \& Emotion & -0.19 & -0.75\textsuperscript{*} & -0.55 \\
		\midrule
		\multirow{2}{1cm}{LFD} & AoA \& Emotion &  0.87\textsuperscript{*} & 0.4 & 0.6 \\
		& Imagery \& Concreteness & 0.41 & 0.69\textsuperscript{*} & 0.69\textsuperscript{*} \\
		
		\midrule
		\multirow{2}{1cm}{SFD} & WF \& Familiarity &  0.62\textsuperscript{+} & 0.65\textsuperscript{+} & 0.76\textsuperscript{*} \\
		& Familiarity \& Concreteness & 0.29 & 0.71\textsuperscript{*} & 0.6\textsuperscript{+} \\
		\midrule
		\multicolumn{5}{c}{Day-3}\\
		\midrule
		\multirow{2}{1cm}{tFC} & WF \& AoA  &  0.49 & 0.87\textsuperscript{*} & 0.75\textsuperscript{+} \\
		& AoA \& Concreteness & 0.52 & 0.73 & 0.87\textsuperscript{*} \\
		
		\midrule
		\multirow{4}{1cm}{tFD} & WF \& AoA &  0.78\textsuperscript{+} & 0.67 & 0.92\textsuperscript{*} \\
		& WF \& Imagery  & 0.82\textsuperscript{*} & 0.28 & 0.62\textsuperscript{+} \\
		& AoA \& Imagery  & 0.96\textsuperscript{**} & 0.53 & 0.87\textsuperscript{*} \\
		& Imagery \& Concreteness  & -0.03 & 0.8\textsuperscript{*} & 0.72\textsuperscript{*} \\
		
		\midrule
		\multirow{4}{1cm}{FFD} & WF \& Familiarity &  0.52 & 0.62 & 0.92\textsuperscript{**} \\
		& WF \& Imagery & 0.67\textsuperscript{+} & 0.38 & 0.77\textsuperscript{*} \\
		& AoA \& Imagery & 0.94\textsuperscript{**} & 0.67 & 0.81\textsuperscript{+} \\
		& Familiarity \& Concreteness & 0.84\textsuperscript{**} & 0.36 & 0.65\textsuperscript{+} \\
		
		\midrule
		LFD & WF \& Emotion  &  0.71\textsuperscript{*} & 0.78\textsuperscript{*} & 0.61 \\
		
		\midrule
		\multirow{7}{1cm}{SFD} & WF \& AoA  &  0.83\textsuperscript{*} & 0.78\textsuperscript{+} & 0.92\textsuperscript{**} \\
		& WF \& Familiarity  & 0.79\textsuperscript{*} & -0.11 & 0.39 \\
		& WF \& Imagery   & 0.84\textsuperscript{**} & 0.38 & 0.66\textsuperscript{+} \\
		& AoA \& Imagery   & 0.94\textsuperscript{**} & 0.87\textsuperscript{*} & 0.98\textsuperscript{***} \\
		& AoA \& Emotion   & 0.84\textsuperscript{*} & -0.11 & 0.61 \\
		& Familiarity \& Concreteness  & -0.53 & 0.73\textsuperscript{*} & 0.41 \\
		& Imagery \& Concreteness   & -0.24 & 0.74\textsuperscript{*} & 0.63\textsuperscript{+} \\[0.5ex]
		\hline 
		\multicolumn{5}{l}{ \textsuperscript{+}: $p<0.1$, \textsuperscript{*}: $p<0.05$, \textsuperscript{**}: $p<0.01$, \textsuperscript{***}: $p<0.001$} \\
		\midrule
		\multicolumn{5}{l}{WF: Word-Frequency, AoA: Age-of-Acquisition}\\
		\midrule\bottomrule	
	\end{longtable}}
\endgroup

\section{Pearson's Correlation among Fixation Features of Psycholinguistics Factors on Session 2 data}
\label{AppendixD2} 
\begingroup
\setstretch{1.2}
{\small	
	\setlength{\tabcolsep}{3pt}
	\renewcommand{\arraystretch}{1.}
	\begin{longtable}
		{>{\raggedright}m{2cm} L{5cm} C{2cm} C{2cm} C{2cm}}
		\caption{Correlation among psycholinguistics factor pairs' fixation features on session 2 data}
		\label{tab:AppendixDtab2}\\
		
		\toprule\midrule
		
		\textbf{Fixation Features} & \textbf{Psycholinguistics factors pair} & \textbf{Low(r,p)} & \textbf{High(r,p)} & \textbf{Both(r,p)}\\
		\midrule
		\multicolumn{5}{c}{Day-1}\\
		\midrule
		tFC & WF \& Familiarity &  0.7\textsuperscript{+} & 0.91\textsuperscript{**} & 0.86\textsuperscript{*} \\
		\midrule
		\multirow{2}{1cm}{tFD} & WF \& Familiarity &  0.81\textsuperscript{*} & 0.96\textsuperscript{**} & 0.94\textsuperscript{**} \\
		& Concreteness \& Emotion & -0.58 & -0.75\textsuperscript{*} & -0.71\textsuperscript{*} \\

		\midrule
		\multirow{4}{1cm}{FFD} & WF \& Familiarity &  0.79\textsuperscript{*} & 0.73\textsuperscript{+} & 0.79\textsuperscript{*} \\
		& WF \& Concreteness & -0.15 & -0.86\textsuperscript{*} & -0.7\textsuperscript{+} \\
		& WF \& Emotion & 0.79\textsuperscript{*} & 0.8\textsuperscript{*} & 0.82\textsuperscript{*} \\
		& Concreteness \& Emotion & -0.31 & -0.82\textsuperscript{**} & -0.72\textsuperscript{*} \\
		
		\midrule
		\multirow{2}{1cm}{LFD} & WF \& Familiarity &  0.52 & 0.93\textsuperscript{**} & 0.94\textsuperscript{**} \\
		& Concreteness \& Emotion & -0.04 & -0.77\textsuperscript{*} & -0.75\textsuperscript{*} \\
		
		\midrule
		SFD & WF \& Familiarity &  0.83\textsuperscript{*} & 0.9\textsuperscript{**} & 0.89\textsuperscript{**} \\

		\midrule
		\multicolumn{5}{c}{Day-2}\\
		\midrule
		tFC & Concreteness \& Emotion &  -0.78\textsuperscript{*} & -0.24 & -0.58 \\
		
		\midrule
		tFD & Concreteness \& Emotion &  -0.73\textsuperscript{*} & -0.43 & -0.61\textsuperscript{+} \\
		
		\midrule
		FFD & WF \& Familiarity &  0.46 & 0.78\textsuperscript{*} & 0.71\textsuperscript{+} \\

		\midrule
		\multirow{2}{1cm}{LFD} & WF \& Familiarity &  0.8\textsuperscript{*} & 0.28 & 0.52 \\
		& Concreteness \& Emotion & -0.86\textsuperscript{**} & -0.41 & -0.67\textsuperscript{*} \\
		
		\midrule
		SFD & --- &  --- & --- & --- \\
		\midrule
		\multicolumn{5}{c}{Day-3}\\
		\midrule
		tFC & WF \& AoA  &  0.63 & 0.85\textsuperscript{*} & 0.9\textsuperscript{**} \\
		
		\midrule
		tFD & --- &  --- & --- & --- \\
		
		\midrule
		FFD & AoA \& Imagery &  0.69\textsuperscript{*} & 0.39 & 0.53 \\
		
		\midrule
		\multirow{2}{1cm}{LFD} & AoA \& Imagery  &  -0.1 & 0.71\textsuperscript{*} & 0.63\textsuperscript{+} \\
		 & Concreteness \& Emotion  &  0.95\textsuperscript{***} & -0.15 & 0.18 \\
		\midrule
		SFD & --- &  --- & --- & --- \\[0.5ex]
		\hline 
		\multicolumn{5}{l}{ \textsuperscript{+}: $p<0.1$, \textsuperscript{*}: $p<0.05$, \textsuperscript{**}: $p<0.01$, \textsuperscript{***}: $p<0.001$} \\
		\midrule
		\multicolumn{5}{l}{WF: Word-Frequency, AoA: Age-of-Acquisition}\\
		\midrule\bottomrule

	\end{longtable}}
\endgroup

\chapter{L2 Linguistics Features} 

\label{AppendixE} 

\lhead{Appendix E. \emph{L2 Linguistics Features}} 

\section{Coh-Metrix version 3.0 indices}
\label{AppendixE1} 

Coh-Metrix is a computational tool that produces indices of the linguistic and discourse representations of a text. Coh-Metrix version 3.0 contains 108 indices under eleven groups: (1) Descriptive, (2) Text Easability Principal Component Scores, (3) Referential Cohesion, (4) LSA, (5) Lexical Diversity, (6) Connectives, (7) Situation Model, (8) Syntactic Complexity, (9) Syntactic Pattern Density, (10) Word Information, and (11) Readability. The details of these indices are available online at: \url{http://cohmetrix.memphis.edu/cohmetrixhome/documentation_indices.html}

\begingroup
\setstretch{1.1}
{\small	
	\setlength{\tabcolsep}{3pt}
	\renewcommand{\arraystretch}{1.}
	\begin{longtable}
		{>{\raggedright}m{1cm} L{3cm} L{10cm}}
		\caption{Coh-Metrix version 3.0 indices}
		\label{tab:AppendixEtab1}\\
		\toprule\midrule

		\textbf{No.} & \textbf{Abbreviation} & \textbf{Description}\\
		\midrule
		\multicolumn{3}{l}{1. Descriptive}\\
		\midrule
		1 & DESPC & Paragraph count, number of paragraphs\\ 
		2 & DESSC & Sentence count, number of sentences\\ 
		3 & DESWC & Word count, number of words\\ 
		4 & DESPL & Paragraph length, number of sentences, mean\\ 
		5 & DESPLd & Paragraph length, number of sentences, standard deviation\\ 
		6 & DESSL & Sentence length, number of words, mean\\ 
		7 & DESSLd & Sentence length, number of words, standard deviation\\ 
		8 & DESWLsy & Word length, number of syllables, mean\\ 
		9 & DESWLsyd & Word length, number of syllables, standard deviation\\ 
		10 & DESWLlt  & Word length, number of letters, mean\\ 
		11 & DESWLltd & Word length, number of letters, standard deviation\\ 
		\midrule
		\multicolumn{3}{l}{2. Text Easability Principal Component Scores}\\
		\midrule
		12 & PCNARz & Text Easability PC Narrativity, z score\\ 
		13 & PCNARp & Text Easability PC Narrativity, percentile\\ 
		14 & PCSYNz & Text Easability PC Syntactic simplicity, z score\\ 
		15 & PCSYNp & Text Easability PC Syntactic simplicity, percentile\\ 
		16 & PCCNCz & Text Easability PC Word concreteness, z score\\ 
		17 & PCCNCp & Text Easability PC Word concreteness, percentile\\ 
		18 & PCREFz & Text Easability PC Referential cohesion, z score\\ 
		19 & PCREFp & Text Easability PC Referential cohesion, percentile\\ 
		20 & PCDCz & Text Easability PC Deep cohesion, z score\\ 
		21 & PCDCp & Text Easability PC Deep cohesion, percentile\\ 
		22 & PCVERBz & Text Easability PC Verb cohesion, z score\\ 
		23 & PCVERBp & Text Easability PC Verb cohesion, percentile\\ 
		24 & PCCONNz & Text Easability PC Connectivity, z score\\ 
		25 & PCCONNp & Text Easability PC Connectivity, percentile\\ 
		26 & PCTEMPz & Text Easability PC Temporality, z score\\ 
		27 & PCTEMPp & Text Easability PC Temporality, percentile\\ 
		\midrule
		\multicolumn{3}{l}{3. Referential Cohesion}\\
		\midrule
		28 & CRFNO1 & Noun overlap, adjacent sentences, binary, mean\\ 
		29 & CRFAO1 & Argument overlap, adjacent sentences, binary, mean\\ 
		30 & CRFSO1 & Stem overlap, adjacent sentences, binary, mean\\ 
		31 & CRFNOa & Noun overlap, all sentences, binary, mean\\ 
		32 & CRFAOa & Argument overlap, all sentences, binary, mean\\ 
		33 & CRFSOa & Stem overlap, all sentences, binary, mean\\ 
		34 & CRFCWO1 & Content word overlap, adjacent sentences, proportional, mean\\ 
		35 & CRFCWO1d & Content word overlap, adjacent sentences, proportional, standard deviation\\ 
		36 & CRFCWOa & Content word overlap, all sentences, proportional, mean\\ 
		37 & CRFCWOad & Content word overlap, all sentences, proportional, standard deviation\\ 
		38 & CRFANP1 & Anaphor overlap, adjacent sentences\\ 
		39 & CRFANPa & Anaphor overlap, all sentences\\ 
		\midrule
		\multicolumn{3}{l}{4. LSA}\\
		\midrule
		40 & LSASS1 & LSA overlap, adjacent sentences, mean\\ 
		41 & LSASS1d & LSA overlap, adjacent sentences, standard deviation\\ 
		42 & LSASSp & LSA overlap, all sentences in paragraph, mean\\ 
		43 & LSASSpd & LSA overlap, all sentences in paragraph, standard deviation\\ 
		44 & LSAPP1 & LSA overlap, adjacent paragraphs, mean\\ 
		45 & LSAPP1d & LSA overlap, adjacent paragraphs, standard deviation\\ 
		46 & LSAGN & LSA given/new, sentences, mean\\ 
		47 & LSAGNd & LSA given/new, sentences, standard deviation\\ 
		\midrule
		\multicolumn{3}{l}{5. Lexical Diversity}\\
		\midrule
		48 & LDTTRc & Lexical diversity, type-token ratio, content word lemmas\\ 
		49 & LDTTRa & Lexical diversity, type-token ratio, all words\\ 
		50 & LDMTLDa & Lexical diversity, MTLD, all words\\ 
		51 & LDVOCDa & Lexical diversity, VOCD, all words\\ 
		\midrule
		\multicolumn{3}{l}{6. Connectives}\\
		\midrule
		52 & CNCAll & All connectives incidence\\ 
		53 & CNCCaus & Causal connectives incidence\\ 
		54 & CNCLogic & Logical connectives incidence\\ 
		55 & CNCADC & Adversative and contrastive connectives incidence\\ 
		56 & CNCTemp & Temporal connectives incidence\\ 
		57 & CNCTempx & Expanded temporal connectives incidence\\ 
		58 & CNCAdd & Additive connectives incidence\\ 
		59 & CNCPos & Positive connectives incidence\\ 
		60 & CNCNeg & Negative connectives incidence\\ 
		\midrule
		\multicolumn{3}{l}{7. Situation Model}\\
		\midrule
		61  & SMCAUSv  &  Causal verb incidence\\ 
		62  & SMCAUSvp  & Causal verbs and causal particles incidence\\ 
		63  & SMINTEp  &  Intentional verbs incidence\\ 
		64  & SMCAUSr  &  Ratio of casual particles to causal verbs\\ 
		65  & SMINTEr  &  Ratio of intentional particles to intentional verbs\\ 
		66  & SMCAUSlsa  &  LSA verb overlap\\ 
		67  & SMCAUSwn  &  WordNet verb overlap\\ 
		68  & SMTEMP  &  Temporal cohesion, tense and aspect repetition, mean\\ 
		\midrule
		\multicolumn{3}{l}{8. Syntactic Complexity}\\
		\midrule

		69 & SYNLE & Left embeddedness, words before main verb, mean\\ 
		70 & SYNNP & Number of modifiers per noun phrase, mean\\ 
		71 & SYNMEDpos & Minimal Edit Distance, part of speech \\ 
		72 & SYNMEDwrd & Minimal Edit Distance, all words\\ 
		73 & SYNMEDlem & Minimal Edit Distance, lemmas \\ 
		74 & SYNSTRUTa & Sentence syntax similarity, adjacent sentences, mean\\ 
		75 & SYNSTRUTt & Sentence syntax similarity, all combinations, across paragraphs, mean\\ 
		\midrule
		\multicolumn{3}{l}{9. Syntactic Pattern Density}\\
		\midrule
		76  & DRNP  & Noun phrase density, incidence\\ 
		77  & DRVP  & Verb phrase density, incidence\\ 
		78  & DRAP  & Adverbial phrase density, incidence\\ 
		79  & DRPP  & Preposition phrase density, incidence\\ 
		80  & DRPVAL  & Agentless passive voice density, incidence\\ 
		81  & DRNEG  & Negation density, incidence\\ 
		82  & DRGERUND & Gerund density, incidence\\ 
		83  & DRINF  & Infinitive density, incidence\\
		\midrule
		\multicolumn{3}{l}{10. Word Information}\\
		\midrule
		84 & WRDNOUN & Noun incidence\\ 
		85 & WRDVERB & Verb incidence\\ 
		86 & WRDADJ & Adjective incidence\\ 
		87 & WRDADV & Adverb incidence\\ 
		88 & WRDPRO & Pronoun incidence\\ 
		89 & WRDPRP1s & First person singular pronoun incidence\\ 
		90 & WRDPRP1p & First person plural pronoun incidence\\ 
		91 & WRDPRP2 & Second person pronoun incidence\\ 
		92 & WRDPRP3s & Third person singular pronoun incidence\\ 
		93 & WRDPRP3p & Third person plural pronoun incidence\\ 
		94 & WRDFRQc & CELEX word frequency for content words, mean\\ 
		95 & WRDFRQa & CELEX Log frequency for all words, mean\\ 
		96 & WRDFRQmc & CELEX Log minimum frequency for content words, mean\\ 
		97 & WRDAOAc & Age of acquisition for content words, mean\\ 
		98 & WRDFAMc & Familiarity for content words, mean\\ 
		99 & WRDCNCc & Concreteness for content words, mean\\ 
		100 & WRDIMGc & Imagability for content words, mean\\ 
		101 & WRDMEAc & Meaningfulness, Colorado norms, content words, mean\\ 
		102 & WRDPOLc & Polysemy for content words, mean\\ 
		103 & WRDHYPn & Hypernymy for nouns, mean\\ 
		104 & WRDHYPv & Hypernymy for verbs, mean\\ 
		105 & WRDHYPnv & Hypernymy for nouns and verbs, mean\\ 
		\midrule
		\multicolumn{3}{l}{11. Readability}\\
		\midrule
		106  & RDFRE   & Flesch Reading Ease\\ 
		107  & RDFKGL   & Flesch-Kincaid Grade Level\\ 
		108  & RDL2 & Coh-Metrix L2 Readability\\ 

		\midrule\bottomrule	
	\end{longtable}}
\endgroup

\chapter{Publications} 

\label{AppendixJ} 

\lhead{Appendix J. \emph{List of Publications}} 

\textbf{Publications in Conferences:}
\begin{enumerate}[(i)]
	\item \textbf{Santosh Kumar Barnwal}, and Uma Shanker Tiwary; ``Using Psycholinguistic Features for the Classification of Comprehenders from Summary Speech Transcripts."; IHCI; Evry, France; December 11-13, Springer LNCS, 2017.
	
	\item Rohit Mishra, \textbf{Santosh Kumar Barnwal}, Shrikant Malviya, Prasoon Mishra, Uma Shanker Tiwary; ``Prosodic Feature Selection of Personality Traits for Job Interview Performance."; December 6-8, 2018 Springer AISC, volume 940
	
	\item Rohit Mishra, \textbf{Santosh Kumar Barnwal}, Shrikant Malviya, Varsha Singh, Punit Singh, Sumit Singh, and Uma Shanker Tiwary; ``Computing with Words Through Interval Type-2 Fuzzy Sets for Decision Making Environment"; IHCI 2019, LNCS 11886, 2020.
	
	\item Sumit Singh, Shrikant Malviya, Rohit Mishra, \textbf{Santosh Kumar Barnwal}, and Uma Shanker Tiwary; ``RNN Based Language Generation Models for a Hindi Dialogue System"; IHCI 2019, LNCS 11886, 2020.
\end{enumerate}

\noindent\textbf{Publications in SCI Journals:}
\begin{enumerate}[(i)]
\item Shrikant Malviya, Rohit Mishra, \textbf{Santosh Kumar Barnwal}, Uma Shanker Tiwary; ``HDRS: Hindi Dialogue Restaurant Search Corpus for Dialogue State Tracking in Task-Oriented Environment"; IEEE/ACM Transactions on Audio Speech and Language Processing, 2021\\
\end{enumerate}

\vspace{40mm}
\noindent\textbf{Patent Pending in India:}
\begin{enumerate}[(i)]
\item Inventors: \textbf{Santosh Kumar Barnwal}, U S Tiwary; title: An artificial intelligence-based learning assessment system for determining improvements using collected data from various equipment; applicants: Indian Institute of Information of Technology Allahabad, \textbf{Santosh Kumar Barnwal}, and Uma Shanker Tiwary; application no: 202111034867; filing date: 3\textsuperscript{rd} August 2021.
\end{enumerate}
\vfill

\addtocontents{toc}{\vspace{2em}} 

\backmatter

\nocite{*}
\label{Bibliography}
\lhead{\emph{Bibliography}} 


\bibliographystyle{./AllPackages/MyIEEEtranN} 

\bibliography{./AllPackages/MyBibliography} 

\clearpage 
\addtotoc{Research to Practice}
\thispagestyle{empty}
\begin{center}{\huge\bf Research to Practice \par}\end{center}
\vspace{1em}
To serve feasible solutions of known and hidden problems through CAA to the growing population of online students, I ended my doctoral research on founding a startup in Education domain.
\begin{tabbing}
	\textbf{Company Name:} \textit{COGNISTAR TECHNOLOGIES PRIVATE LIMITED}\\
	\textbf{CIN:} \textit{U72900DL2019PTC356992}\\
	\textbf{RoC:} \textit{RoC-Delhi}\\
	\textbf{Registration Number:} \textit{356992}\\
	\textbf{Date of Incorporation:} \textit{04 November 2019}\\
	\textbf{Founding Directors:} \=(i) \textit{Smt Shakuntala Devi} (mother)\\
	\>(ii) \textit{Santosh Kumar Barnwal} (myself).
\end{tabbing}

\begin{figure}[!htb]
	\centering
	\includegraphics[scale=1]{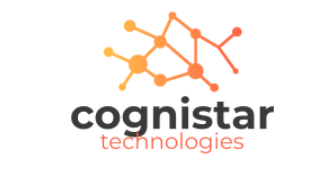}
	\label{fig:Cognistar-Logo}
\end{figure}
\begin{figure}[!htb]
	\minipage{0.3\textwidth}
	\includegraphics[scale=0.5, width=\linewidth]{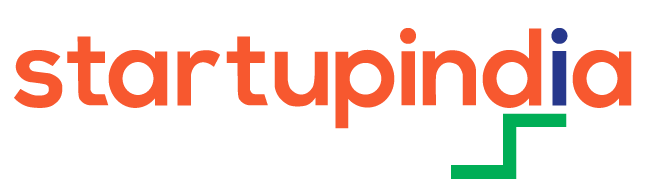}
	\label{fig:Startup-India}
	\endminipage\hfill
	\minipage{0.3\textwidth}
	\includegraphics[scale=0.5, width=\linewidth]{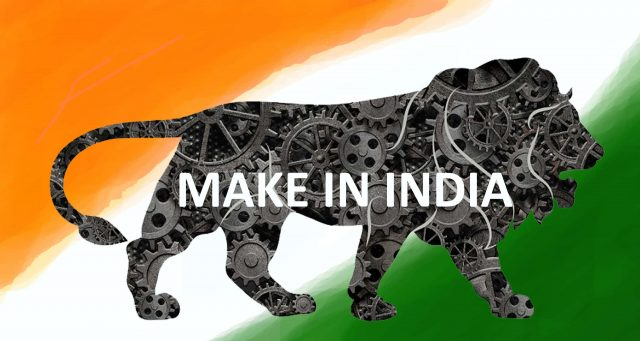}
	\label{fig:Make-In-India}
	\endminipage\hfill
	\minipage{0.3\textwidth}%
	\includegraphics[scale=0.5, width=\linewidth]{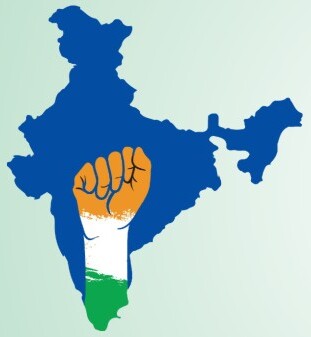}
	\label{fig:Aatma-Nirbhar-Bharat}
	\endminipage
\end{figure}

\end{document}